\documentclass[runningheads]{llncs}
\usepackage{graphicx}
\usepackage{comment}
\usepackage{amsmath,amssymb} 
\usepackage{color}
\usepackage{multirow}
\usepackage[table]{xcolor}
\usepackage[export]{adjustbox}
\usepackage{subcaption}
\usepackage[moderate,mathdisplays=normal,wordspacing=normal]{savetrees}

\newlength{\mytabcolsep}%
\newcommand{\calL}{\mathcal{L}}

\def\S{\mathbf{S}}%
\def\p{\mathbf{p}}%
\def\E{\mathbb{E}}
\def\p{\mathbf{p}}%
\def\z{\mathbf{z}}%
\def\m{\mathbf{m}}%
\def\z{\mathbf{z}}%

\newlength{\imwidth}%
\newcommand{\ignorethis}[1]{}
\usepackage[outercaption]{sidecap}  
\begin{document}
\pagestyle{headings}
\mainmatter
\def\ECCVSubNumber{4322}  

\title{Look here! A parametric learning based approach\\ to redirect visual attention} 

\newcommand\blfootnote[1]{%
  \begingroup
  \renewcommand\thefootnote{}\footnote{#1}%
  \addtocounter{footnote}{-1}%
  \endgroup
}

\setlength{\textfloatsep}{11pt}
\setlength{\intextsep}{13pt}
\titlerunning{Look here! A parametric learning based approach to redirect visual attention}
%
\author{Youssef A. Mejjati \inst{1}* \and
Celso F. Gomez\inst{2} \and
Kwang In Kim\inst{3} \and \\ Eli Shechtman\inst{2} \and Zoya Bylinskii\inst{2}}
\authorrunning{Mejjati et al.}
%
\institute{University of Bath, UK \and
Adobe Research \and UNIST 
}
\maketitle

\begin{abstract}
Across photography, marketing, and website design, being able to direct the viewer's attention is a powerful tool. 
Motivated by professional workflows, we introduce an automatic method to make an image region more attention-capturing via subtle image edits that maintain realism and fidelity to the original. From an input image and a user-provided mask, our GazeShiftNet model predicts a distinct set of global parametric transformations to be applied to the foreground and background image regions separately. 
We present the results of quantitative and qualitative experiments that demonstrate improvements over prior state-of-the-art. In contrast to existing attention shifting algorithms, our global parametric approach better preserves image semantics and avoids typical generative artifacts. Our edits enable inference at interactive rates on any image size, and easily generalize to videos. Extensions of our model allow for multi-style edits and the ability to both increase and attenuate attention in an image region. Furthermore, users can customize the edited images by dialing the edits up or down via interpolations in parameter space. This paper presents a practical tool that can simplify future image editing pipelines.

\keywords{automatic image editing, visual attention, adversarial networks}
\end{abstract}
\begin{figure*}[!ht]
\centering
\includegraphics[width=0.90\linewidth]{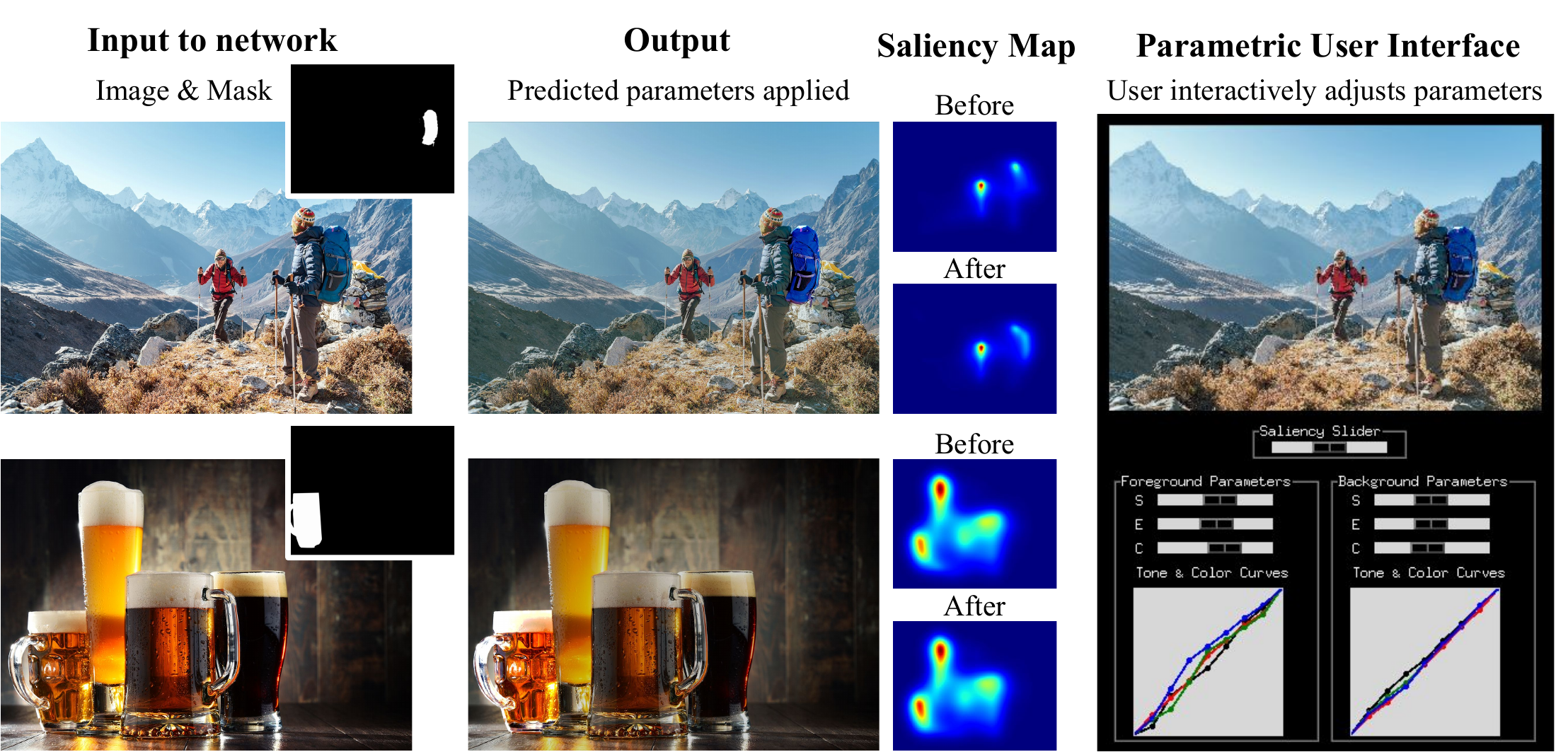} 
\caption{GazeShiftNet takes an image and binary mask as input and predicts a set of parameters (sharpening, exposure, contrast, tone, and color curves) that are sequentially applied to the image to produce the output. The transformed image \emph{subtly} redirects visual attention towards the mask region, seen from the saliency maps. A user can then tune the edits up or down (as shown on the right)  at interactive rates, using the saliency slider.}
\label{fig:teaser}
\end{figure*}

\section{Introduction}
\blfootnote{\noindent* Work done while Youssef was interning at Adobe Research.}
Photographers, advertisers, and educators seek to control the attention of their audiences, redirecting it to the content that matters most. Professionals working with images accomplish this via subtle adjustments to the contrast,
tone, color, etc., of the relevant image regions to make them ``pop-out". Motivated by professional workflows, we propose an automated learning based approach that predicts a set of global parametric edits to apply to an image to redirect a viewer's attention towards (or away from) a specific image region. Importantly, our approach is constrained, via an adversarial module, to produce realistic edits that 
remain faithful to the photographer's intentions and original image semantics. To ensure that the edited image successfully redirects attention, we use a state of the art saliency model during training. 

While a glowing red arrow in an image would surely attract attention, this would not be practical for many use cases. Our GazeShiftNet model is specifically designed to be a practical solution according to the following criteria: (1) the model predictably redirects attention towards (or away from) image pixels denoted by a user-provided binary mask, (2) the proposed edits conserve the image semantics to maintain realism
, (3) the model performs consistently and robustly across a variety of image content (e.g., objects and people), and (4)~image edits are predicted at interactive rates, for use within applications.   



Our solution involves applying global parametric transformations - sharpening, exposure, contrast, tone, and color - to the image, and employing an adversarial training strategy~\cite{goodfellow2014generative}, rather than modifying each pixel separately~\cite{chenguide,gatys2017guiding}. 
Our network predicts two sets of parameters to apply to the foreground and background of the image, demarcated by the input mask. The goal is to create a ``pop-out" effect, which cannot be achieved by a single global transformation to the entire image. In sum, our model takes the form of a parametric generator followed by a cascade of differentiable layers that apply common photo editing transformations, mimicking a professional workflow.
The choice and implementation of these transformations is motivated by related work on image enhancement~\cite{bianco2019learning,hu2018exposure}.
We train GazeShiftNet on a subset of the MS-COCO dataset~\cite{lin2014microsoft}, and present quantitative and qualitative results, including from user studies, on three different datasets. Compared to existing attention shifting methods~\cite{chenguide,gatys2017guiding,hagiwara2011saliency,mateescu2014attention,mechrez2019saliency}, 
our approach successfully shifts viewers' attention, while achieving more realistic results that conserve the original image semantics and do not suffer from local color and texture artifacts.

We demonstrate the advantages of our global parametric approach for future applications. In particular, inference and editing take place at interactive rates, independent of image resolution, as the global transformations are predicted on a low resolution image using a feed-forward pass (rather than an optimization procedure~\cite{hagiwara2011saliency,mateescu2014attention,mechrez2019saliency}) and can be seamlessly applied to a high resolution version. Users can tweak the final result using 
photo editing sliders that control the parameters applied, since the interpolation takes place in parameter space. Our method also works for video data by applying the parameters predicted for the first frame to the rest of the video, thereby ensuring temporal continuity. Finally, we present extensions of our model to produce image edits of different styles, and the option to both increase and attenuate visual attention in an image region.

\section{Related Work}

While image editing has been a popular research topic for many years~\cite{barnes2009patchmatch,chen1993view,hertzmann2001image,perez2003poisson}, recent breakthroughs have been made possible by generative adversarial networks~\cite{goodfellow2014generative}, including in image generation~\cite{karras2019style,park2019gaugan}, image-to-image translation~\cite{mejjati2018unsupervised,zhu2017unpaired}, and style transfer~\cite{gatys2016image,huang2017arbitrary}. However, very few image editing methods exist for redirecting the viewer's attention towards a specific image region. 

Among such methods, Gatys et al.~\cite{gatys2017guiding} use an encoder-decoder model that takes an image and target saliency map as inputs, and generates a new image satisfying the target saliency map. This approach has a number of weaknesses:  (1) operating directly in pixel space often produces artifacts in the final image, 
(2) a target saliency map is required as input, which is not straightforward for a user to provide, and (3) the edits from the generator are limited to a fixed image resolution. Chen et al.~\cite{chenguide} similarly use an encoder-decoder model, but with an additional cyclic loss to stabilize the training procedure and reduce artifacts. However, this approach still suffers from the last two issues above, making it inconvenient for practical editing scenarios. 

Su et al.~\cite{su2005emphasis} use smoothened power maps with steerable pyramids to equalize texture. Mechrez et al.~\cite{mechrez2019saliency} use patches from the same image to increase the saliency of a given region. The space of possible edits is naturally limited to appearances of other pixels in the image. 
Furthermore, this approach is very time consuming and requires an optimization per image, with compute time scaling with image resolution. 

Related to our approach Hagiwara et al.~\cite{hagiwara2011saliency} predict color and intensity changes in RGB space and add them pixel-wise to the original image. Mateescu et al.~\cite{mateescu2014attention} use LAB space and adjust 
the hue within the object mask to maximize the KL divergence between the color distributions inside and outside the mask. Both methods require computationally intensive optimization at test time. Additionally, while these algorithms are based on heuristics that successfully redirect human attention, they do not preserve the semantics of the image, and can generate color anomalies affecting image realism~\cite{mateescu2014attention}. 

Wong et al.~\cite{wong2011saliency} also use an optimization strategy to predict parameters such as saturation, sharpness and brightness to apply on different segments of an image, which require significant human supervision to generate. 
Compared to other parametric methods~\cite{hagiwara2011saliency,mateescu2014attention}, this method struggles to effectively redirect the viewer's attention~\cite{mechrez2019saliency}.
In contrast, our method uses a deep neural network to predict global parametric transformations like exposure and contrast to apply to the image, all while preserving the semantics of the image, avoiding artifacts, and significantly reducing computation time. 

Similar parametric transformations as ours are used for other image processing tasks, including for photo enhancement~\cite{chandakkar2016structured} and for recovering an image from its raw counterpart by emulating the image processing pipeline~\cite{hu2018exposure}.
Tsai et al.~\cite{tsai2017deep} tackle a related problem, to make the composition of image foreground and background as natural as possible, by enhancing the foreground with respect to its background using an encoder-decoder network. We rather use an encoder-decoder network to enhance the foreground with the goal of redirecting the viewer's attention. 

Attention modeling has also seen significant progress in the past few years thanks to deep neural networks~\cite{borji2019saliency,bylinskii2016should}. 
Attention models are becoming increasingly more accurate and practical for applications including object detection~\cite{Wang_2019_CVPR,Wang_2019_cv}, object recognition~\cite{li2016objects,moosmann2006learning}, content aware image re-targeting~\cite{achanta2009saliency,zund2013content}, graphic design~\cite{bylinskii2017learning,shen2014webpage,zheng2018task}, image captioning~\cite{cornia2019show,zhou2020}, and action recognition~\cite{koutras2019susinet,mathe2012dynamic}. 
While the application of such models has proven prolific, relatively little effort has been dedicated towards automatically manipulating attention, i.e. creating new image content that satisfies a given attention objective.



\section{Method}\label{sec:method}

\hspace{\parindent}\textbf{Motivation:} Given an image and region of interest, our goal is to automatically edit the image to make the selected region more attention-capturing. 
To learn how professionals complete this task, we ran an exploratory study on \url{www.usertesting.com}, providing participants with image-mask pairs, and asking them to edit the images in Adobe's Photoshop to make the masked regions more attention-capturing. We collected edits from 5 different participants on 30 high-resolution stock photographs (see Supplemental Material). 
Lessons learned: (1) image semantics are preserved, (2) edits are mostly restricted to parametric transformations, and (3) different operations are applied to image pixels inside and outside the mask. We designed our computational approach accordingly. 
Our approach (1) strives to maintain fidelity/faithfulness to the input image, while (2) applying global parametric transformations to the image, such that, (3) one set of transformations is applied to the mask (foreground), and a separate set is applied to the background. 

\textbf{Computational approach:} Given an input image $I$ 
and a binary mask $\m$, 
our goal is to generate a new image $I'$ 
redirecting viewer attention towards the image region specified by $\m$. 
We generate $I'$ by sequentially applying a set of parametric transformations commonly used in photo editing software and by computational photography algorithms~\cite{chandakkar2016structured,hu2018exposure,kaufman2012content,yan2016automatic}. 
We apply the following ordered sequence of parameters: 
\emph{sharpening}, \emph{exposure}, \emph{contrast}, \emph{tone adjustment}, and \emph{color adjustment}~\cite{hu2018exposure}. 
We discuss the order of operations in Sec.~\ref{ssec:ablations}, and provide mathematical definitions in the Supplemental Material.

One possible approach is to train an encoder to predict a set of parameters to apply to the foreground region within a mask $\mathbf{m}$. However, we found that predicting a set of parameters both for the foreground $\p_f$ and for the background $\p_b$, is especially effective in cluttered scenes. Where multiple regions in the input image may have high saliency, attenuating the saliency of the background can be easier than increasing the saliency of the foreground. 
Formally, our encoder $G_E$ is a neural network with a shared feature extractor and two specialized heads, predicting two sets of parameters:  $G_E(I,  \m)=(\p_f, \p_b)$.

\begin{figure}[ht!]
\centering
\includegraphics[width=\linewidth]{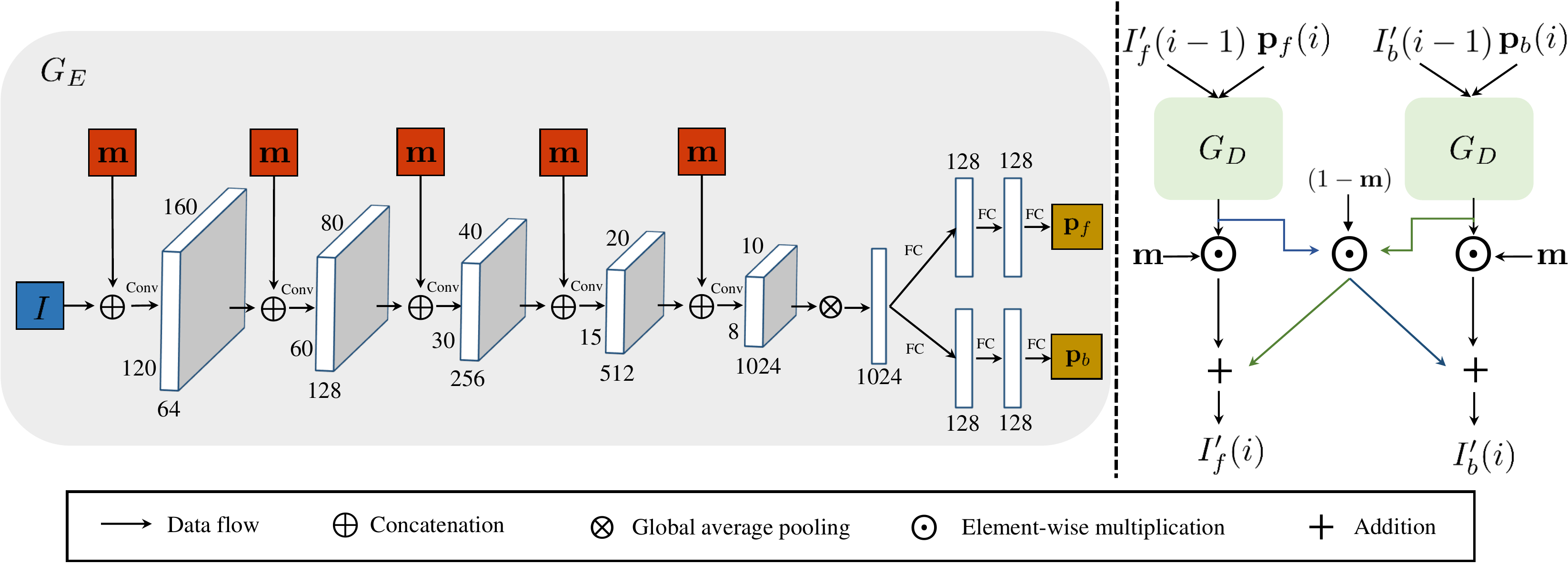} 
\caption{GazeShiftNet architecture. Left: $G_E$ takes the image $I$ and the mask $\m$ as inputs and encodes them through a series of convolutional layers, then predicts the foreground and background parameters using fully connected network heads. Right: $G_D$ applies a series of differentiable functions sequentially using the predicted parameters from $G_E$ and creates the intermediate images $I'_f(i)$ and $I'_b(i)$ at iteration $i$.}
\label{fig:graph}
\end{figure}

\textbf{Model architecture:}
Our generator $G$ is composed of an encoder $G_E$ and a decoder $G_D$, where $G_E$ predicts global parametric transformations conditioned on $I$ and $\m$, and $G_D$ applies these transformations to the image to produce the final edited image. 
The pipeline is as follows (Fig.~\ref{fig:graph}): we concatenate $I$ with the mask $\m$ and feed this to a down-sampling convolutional neural network. Providing the mask as input allows the network to focus on the region to be enhanced. Furthermore, we reinforce the mask conditioning by applying the concatenation layer-wise throughout the convolutional part of the network. We bilinearly down-sample the mask before concatenating on each hidden layer to fit the corresponding input dimensions. The resulting representation is then fed to a series of fully connected layers via global average pooling. Global average pooling provides global information about high level features in the image, which is useful for predicting global parametric transformations. The foreground and background parameters are then predicted by two fully connected network heads. 

The convolutional part of $G_E$ encodes the semantics of the image and the region specified by the mask, and is the shared part of our network. The subsequent fully connected networks are the specialized heads that leverage this shared knowledge 
to predict separate parameters for the foreground and background image regions.


The decoder, $G_D$,  applies the predicted parameters to the input image. $G_D$ consists of a sequence of fixed (non-trainable) differentiable functions applied separately to the foreground and background image regions, demarcated by the mask $\m$.  
Specifically, we sequentially generate a pair of intermediate images $I'_f(i)$ and $I'_b(i)$ from $\p_f$ and $\p_b$, respectively, based on a fixed order of operations. At iteration $i$: 
\begin{align}
    I'_f(i) = G_D(I'_f(i-1), \p_f(i)) \circ \m + G_D(I'_b(i-1), \p_b(i)) \circ (1-\m), \\ 
    \label{eq:iiter_f}
    I'_b(i) = G_D(I'_b(i-1), \p_b(i)) \circ \m + G_D(I'_f(i-1), \p_f(i)) \circ (1-\m), 
\end{align}
where $\circ$ refers to element-wise multiplication, and $I'_f(0)= I'_b(0) = I$. This decoding process is illustrated on the right side of Fig.~\ref{fig:graph}. The final products of this sequential editing process process, $I'_f$ and $I'_b$, are blended to synthesize the final image $I'$, using $\m$:
\begin{align}
    I' = I'_f\circ \m + I'_b\circ(1-\m).
    \label{eq:iprime}
\end{align}




\textbf{Losses:} We constrain our image transformations to respect the following properties: (1) $I'$ should be a realistic image, without deviating significantly from the input $I$,
(2) viewer attention should be redirected towards the image region specified by $\mathbf{m}$. 
To ensure the first property, we use a critic $D$, which differentiates between real and generated images. We train $D$ adversarially with $G_E$ using a hinge GAN loss known for its stability~\cite{jolicoeur2018relativistic,miyato2018spectral,zhang2018self}:
\begin{align}
    \calL_{D}(\Theta_D) =& - \E_{I,\m} \big[\min\{0, -D(I')-1\}\big]- \E_{I}\big[\min\{0, D(I)-1\}\big], \nonumber \\
    \calL_{G}(\Theta_G) =& -\E_{I,\m}\big[D(I')\big],
    \label{eq:hingeLSGANgt}
\end{align}
where $\Theta_D$ and $\Theta_G$ are the learnable weights of $D$ and $G_E$, respectively.


To ensure the second property, 
we use a state-of-the-art deep saliency model~\cite{Fosco_2020_CVPR}, termed $\S$ throughout the paper, as a proxy of viewer attention. 
We use $\S$ to compute the saliency map of the output image, $S_{I'} = \S(I', \Theta_S)$, where $\Theta_S$ are the parameters of $\S$, and 
calculate the attention loss in the mask as:
\begin{align}
    \calL_{att}(S_{I'}, \m) = -\frac{1}{\sum\limits_{i,j} \m_{[ij]}} \sum_{(i,j) \in \m}{S_{I'}}_{[ij]}\m_{[ij]},
    \label{eq:sloss}
\end{align}
by iterating over all mask pixels $(i,j)$. 
Normalization by the area of the masked region gives equal importance to regions of any size. The saliency maps obtained from $\S$ are normalized via softmax such that they sum to one. This ensures that increasing the saliency of the foreground necessarily decreases the saliency of the background. 

Minimizing Eq.~\ref{eq:sloss} alone would lead to unrealistic results, but in combination with the adversarial loss in Eq.~\ref{eq:hingeLSGANgt}, the algorithm tries to redirect viewer attention as much as possible while maintaining the realism of the output $I'$. 
The overall loss becomes:
\begin{align}
    \calL(\Theta_G, \Theta_D) =& \calL_{G}(\Theta_G) + \calL_{D}(\Theta_D)  + \lambda_s \calL_{att}(S_{I'}, \m), 
    \label{eq:finalloss}
\end{align}
where $\lambda_s$ controls the weight of the attention loss, which we empirically set to $2.5\times10^4$. Attention loss values are very small due to softmax normalization of the saliency maps. 

\textbf{Training dataset:}
We selected the MS-COCO dataset~\cite{lin2014microsoft} because of the semantic diversity it contains, and the object segmentations that could be used as input masks to our algorithm. We curated the dataset to produce a set of image-mask pairs appropriate for our task.
Specifically, we used the saliency model $\S$ to select images where the masked region is not already too salient, so that we can train our algorithm to shift attention to these regions. 
In addition, we discarded images containing only one mask instance or where the mask was too small or too large. Shifting attention to very small regions would likely require unrealistic image edits. Instances that are too large are often already quite salient. For the same reason, images with only one object segment (e.g., a single plane in the sky) may have no other regions to shift attention towards/away from. If a training image contained multiple masks satisfying all these conditions, we randomly picked one and discarded the others to ensure a diverse training set without over-representing any images. 
Following this process, we ended up with 49,949 image-mask pairs for the training set and 6,519 pairs for the validation set. We call this curated dataset `CoCoClutter' to emphasize that the images contain multiple objects, without being dominated by any one object. We trained our model with a batch size of 4 for 37,500 iterations, sufficient for the model to converge. We used a learning rate of 1e-5, linearly decayed towards 0 starting halfway through training.

\section{Evaluation}
In this section, we compare GazeShiftNet to competing approaches on two datasets, to measure the ability of each method to shift visual attention to the target image region, and to produce image edits that are realistic and faithful to the original image. We first consider representative examples from all the methods and discuss common behaviors. Next, we evaluate the methods using computational metrics that measure saliency shift and fidelity, and finally, we present the results of three user studies to measure attention shift, image realism, and fidelity. We conclude with runtime comparisons of all the methods, and ablation experiments run on our model.

\begin{figure*}
\centering
\setlength{\imwidth}{0.16\linewidth}
\setbox1=\hbox{\includegraphics[width=\imwidth]{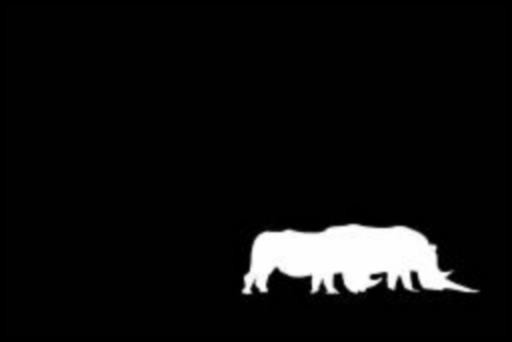}}
\setlength{\tabcolsep}{1pt}
\begin{tabular}{cccccc}
Input & Our & MEC~\cite{mechrez2019saliency} & HAG~\cite{hagiwara2011saliency} & HOR~\cite{mateescu2014attention} & GAT~\cite{gatys2017guiding} 
\\
\includegraphics[width=\imwidth]{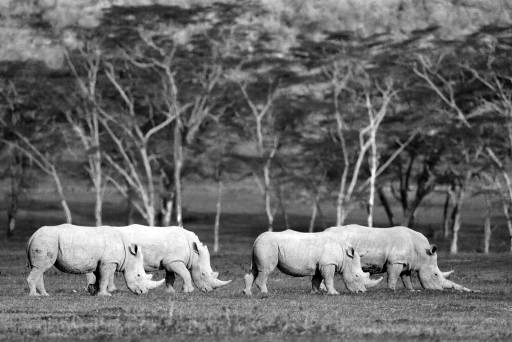}\llap{\makebox[\wd1][l]{\raisebox{0.4\imwidth}{\includegraphics[cfbox=red, width=0.4\imwidth]{Ims/compFig/mask0.jpg}}}}&
\includegraphics[width=\imwidth]{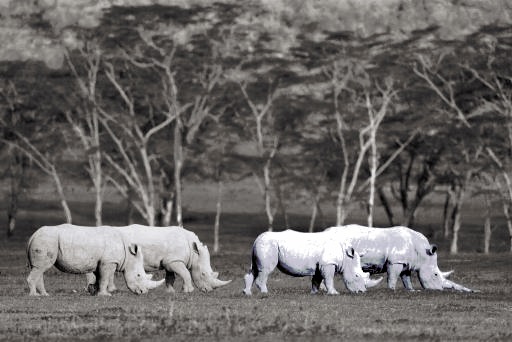} &
\includegraphics[width=\imwidth]{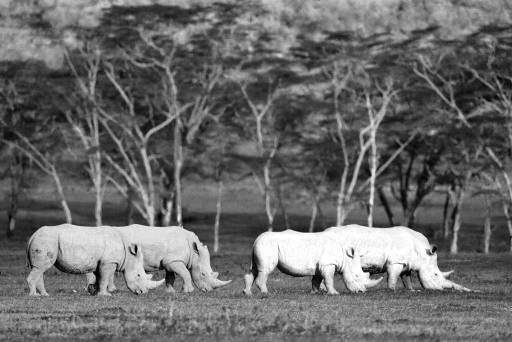} &
\includegraphics[width=\imwidth]{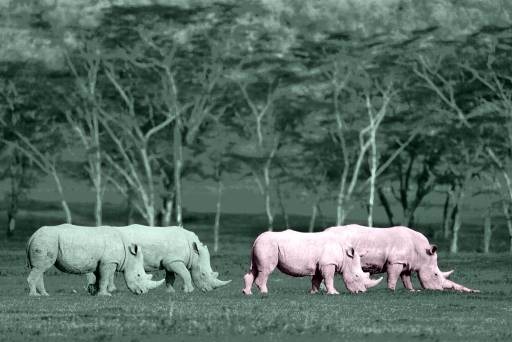} &
\includegraphics[width=\imwidth]{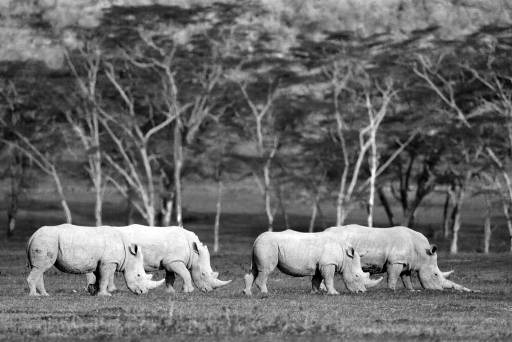} &
\includegraphics[width=\imwidth]{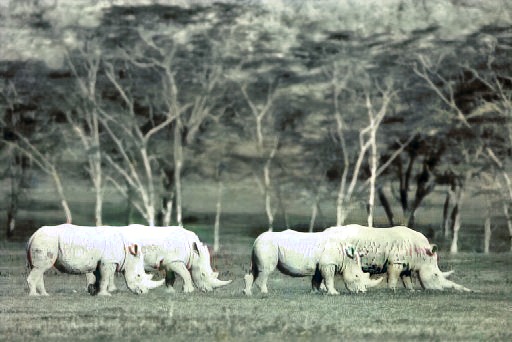}
\\
\includegraphics[width=\imwidth]{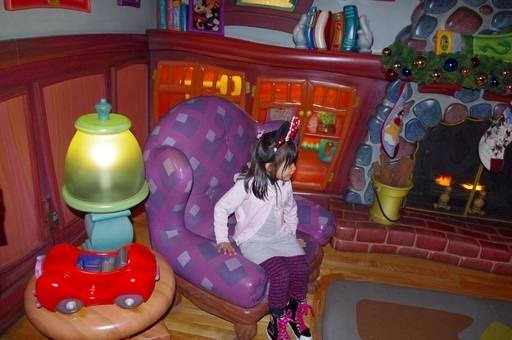}\llap{\makebox[\wd1][l]{\raisebox{0.4\imwidth}{\includegraphics[cfbox=red, width=0.4\imwidth]{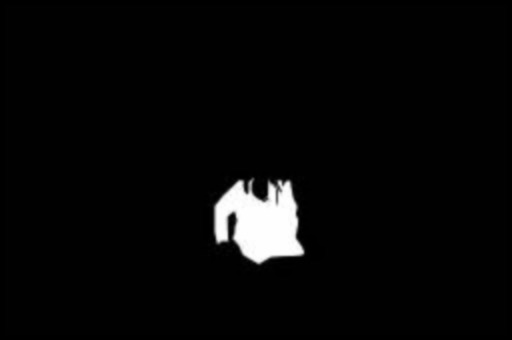}}}}&
\includegraphics[width=\imwidth]{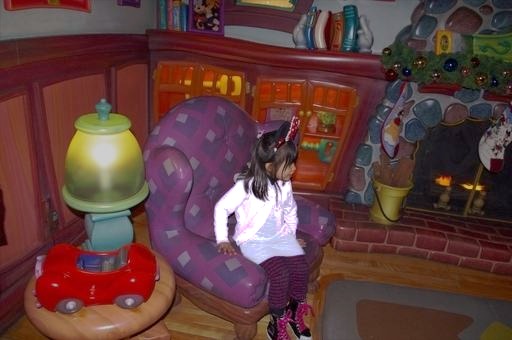} &
\includegraphics[width=\imwidth]{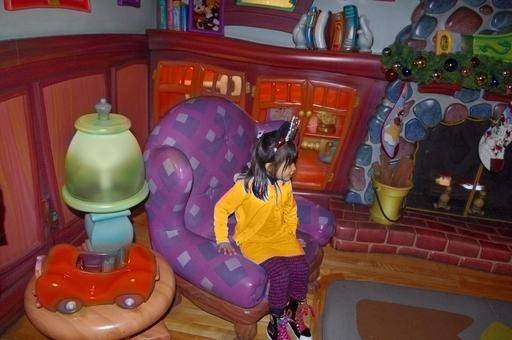} &
\includegraphics[width=\imwidth]{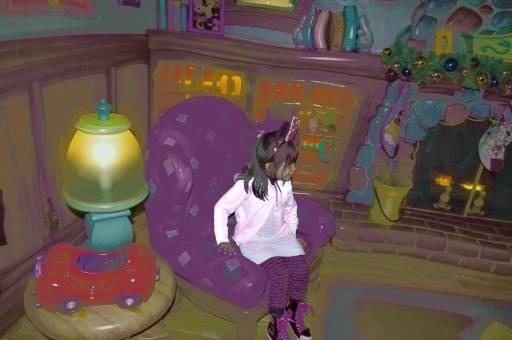} &
\includegraphics[width=\imwidth]{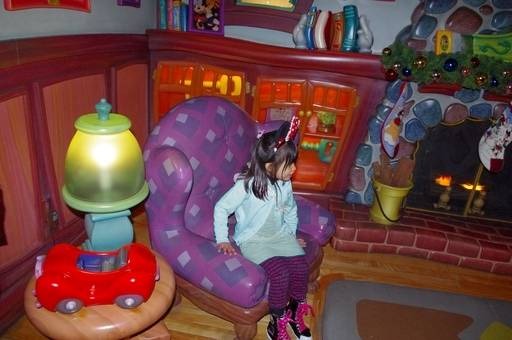} &
\includegraphics[width=\imwidth]{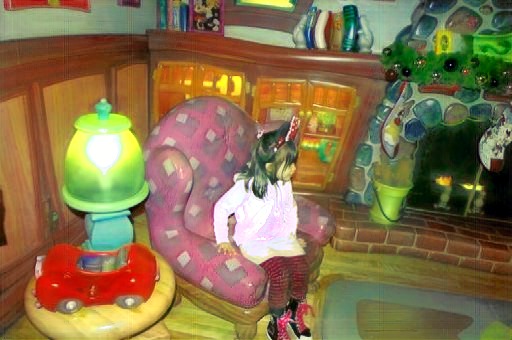}
\\
\includegraphics[width=\imwidth]{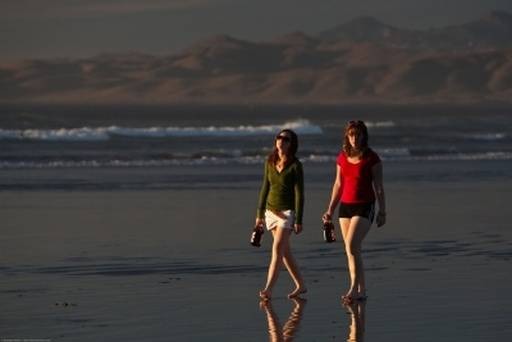}\llap{\makebox[\wd1][l]{\raisebox{0.4\imwidth}{\includegraphics[cfbox=red, width=0.4\imwidth]{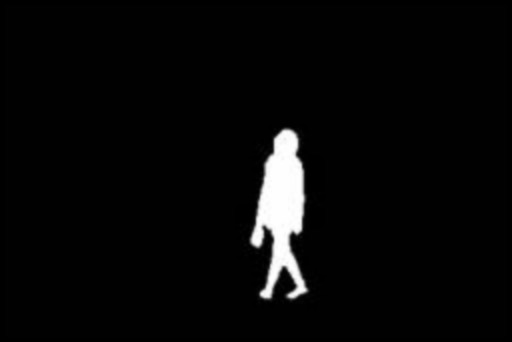}}}}&
\includegraphics[width=\imwidth]{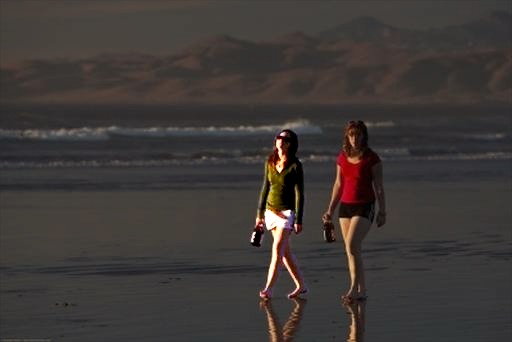} &
\includegraphics[width=\imwidth]{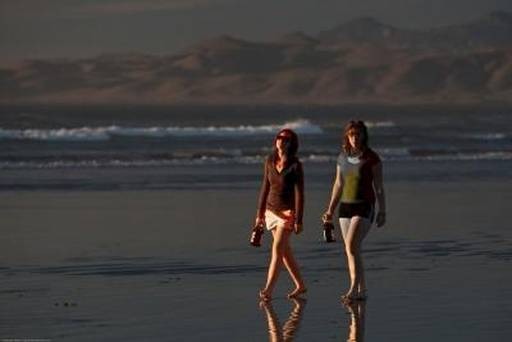} &
\includegraphics[width=\imwidth]{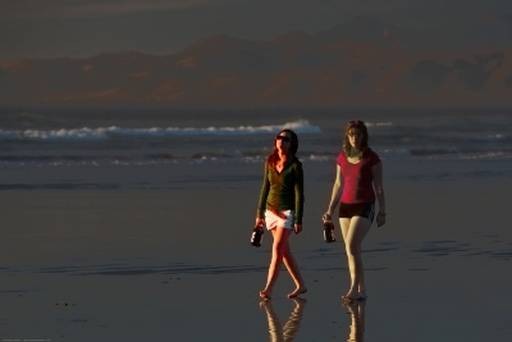} &
\includegraphics[width=\imwidth]{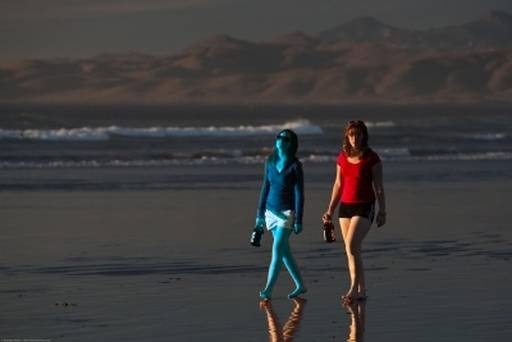} &
\includegraphics[width=\imwidth]{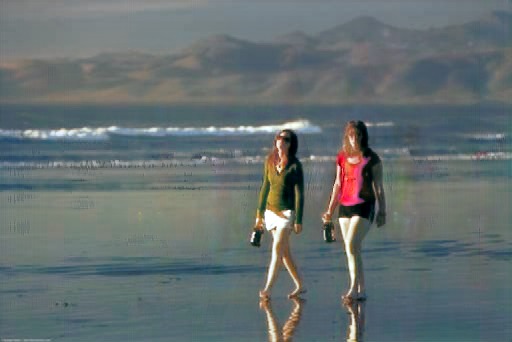}
\\
\includegraphics[width=\imwidth]{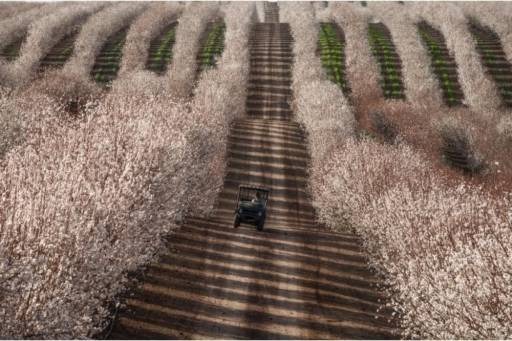}\llap{\makebox[\wd1][l]{\raisebox{0.4\imwidth}{\includegraphics[cfbox=red, width=0.4\imwidth]{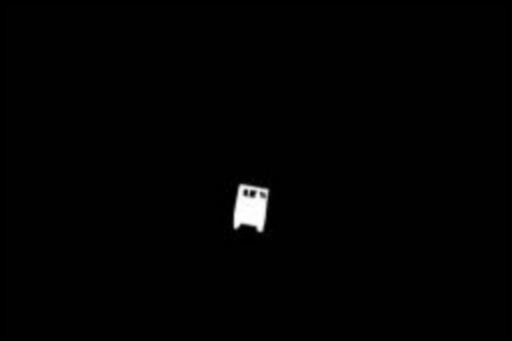}}}}&
\includegraphics[width=\imwidth]{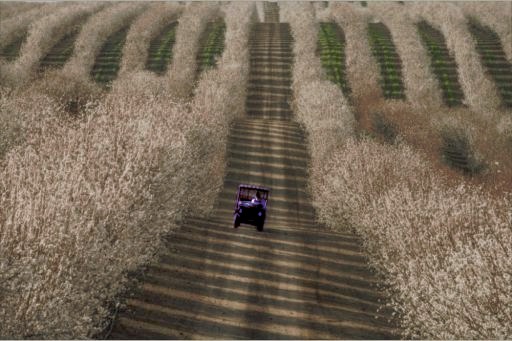} &
\includegraphics[width=\imwidth]{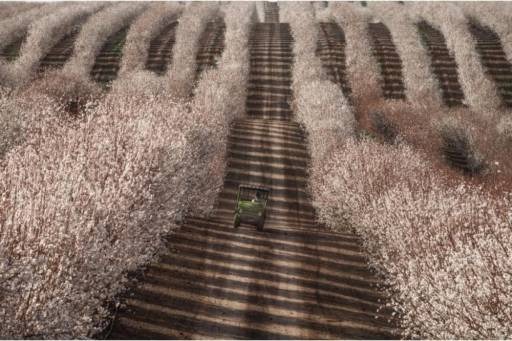} &
\includegraphics[width=\imwidth]{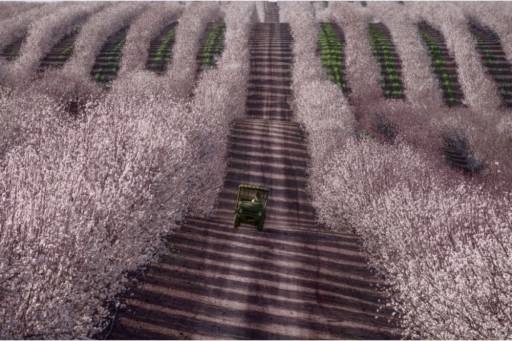} &
\includegraphics[width=\imwidth]{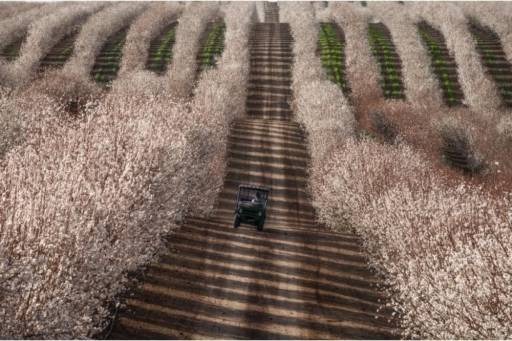} &
\includegraphics[width=\imwidth]{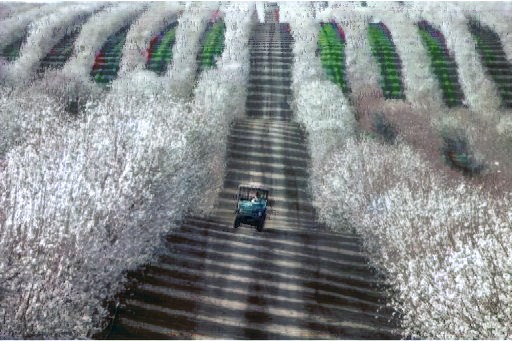}
\\
\includegraphics[width=\imwidth]{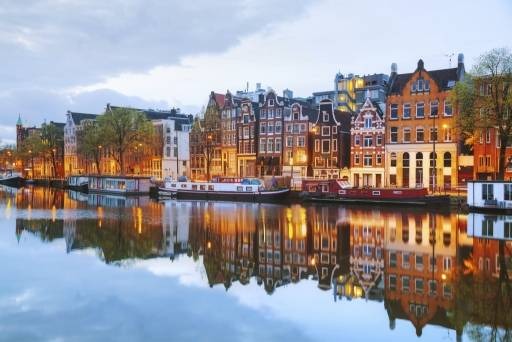}\llap{\makebox[\wd1][l]{\raisebox{0\imwidth}{\includegraphics[cfbox=red, width=0.4\imwidth]{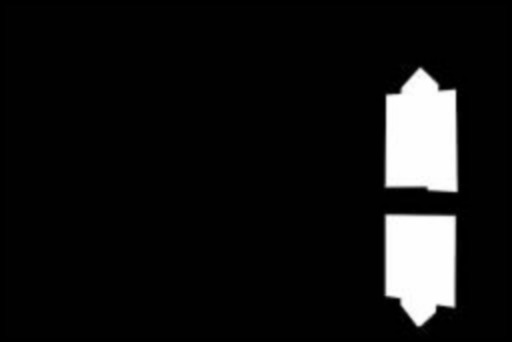}}}}&
\includegraphics[width=\imwidth]{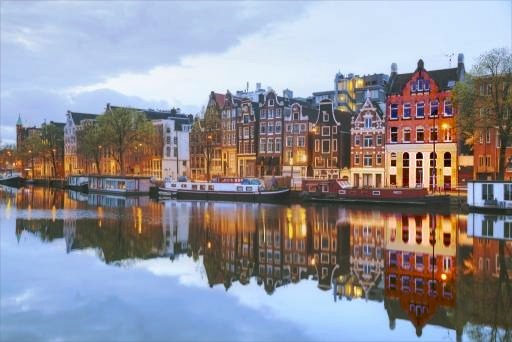} &
\includegraphics[width=\imwidth]{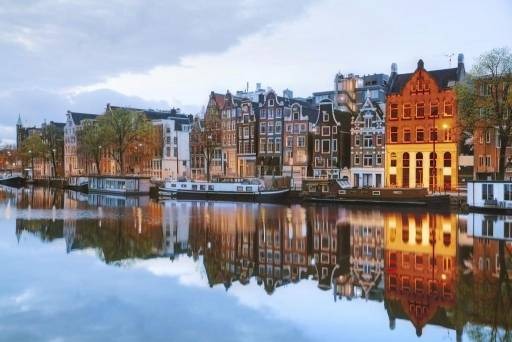} &
\includegraphics[width=\imwidth]{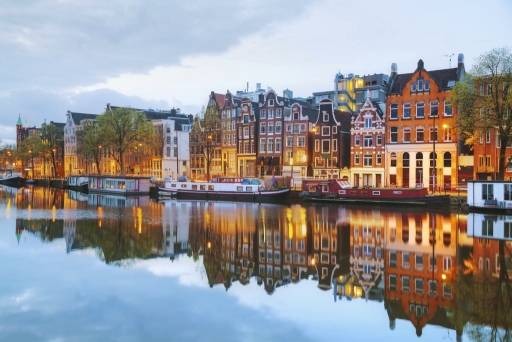} &
\includegraphics[width=\imwidth]{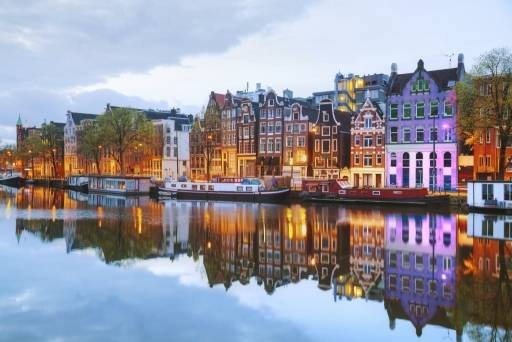} &
\includegraphics[width=\imwidth]{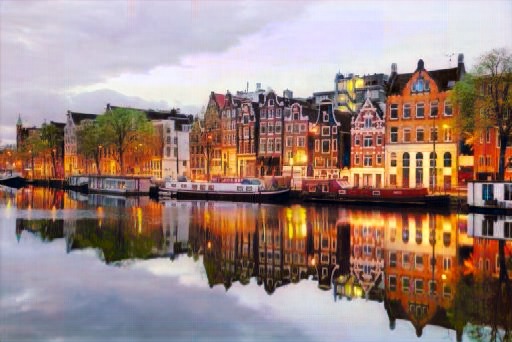}
\\

 \end{tabular}
\caption{Model comparisons on the Mechrez dataset. More in the Supplemental Material.}
\label{fig:comparisons}
\end{figure*}%

\subsection{Qualitative comparisons}
 
\begin{figure*}
\centering
\setlength{\imwidth}{0.16\linewidth}

\setbox1=\hbox{\includegraphics[width=\imwidth]{Ims/compFig/mask0.jpg}}
\setlength{\tabcolsep}{1pt}
\begin{tabular}{cccccc}

Input & Mask & Our & HAG~\cite{hagiwara2011saliency} & HOR~\cite{mateescu2014attention} & GAT~\cite{gatys2017guiding} 
\\
\includegraphics[width=\imwidth]{Ims/compFig_coco/im337.jpg}&
\includegraphics[width=\imwidth]{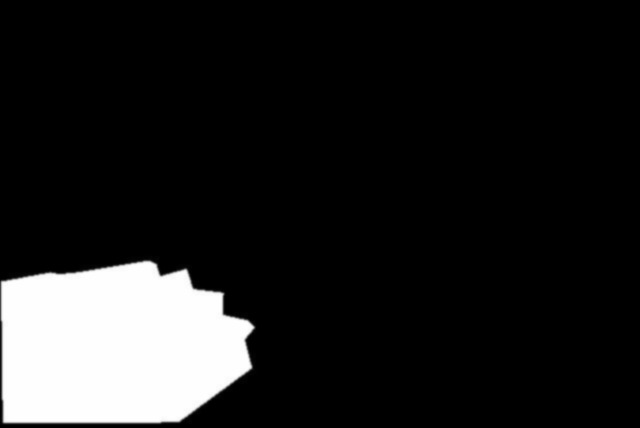}&
\includegraphics[width=\imwidth]{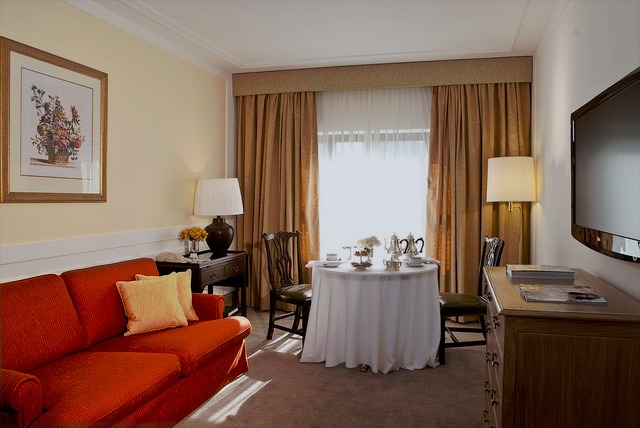}&
\includegraphics[width=\imwidth]{Ims/compFig_coco/im_hag_inc337_-0000.jpg}&
\includegraphics[width=\imwidth]{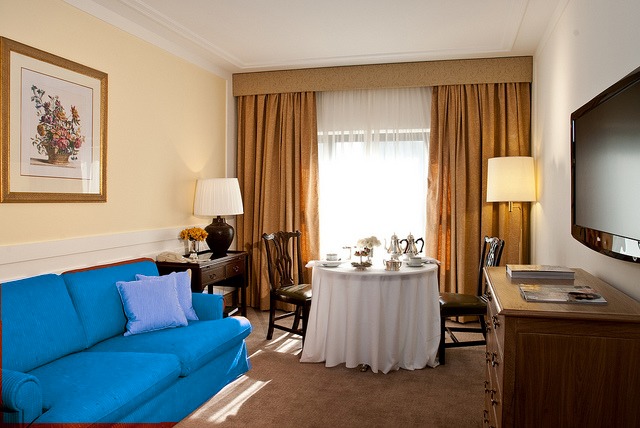}&
\includegraphics[width=\imwidth]{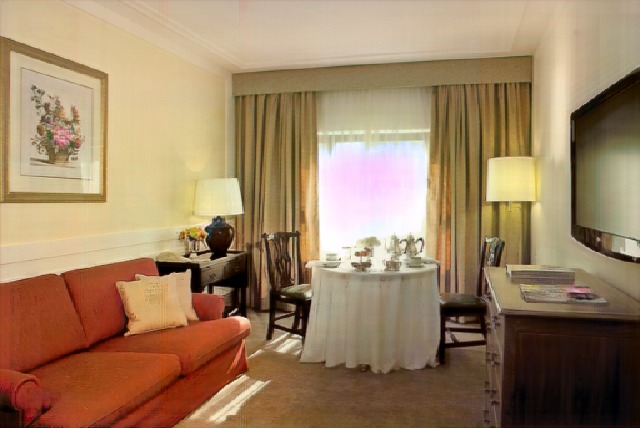}
\\
\includegraphics[width=\imwidth]{Ims/compFig_coco/im403.jpg}&
\includegraphics[width=\imwidth]{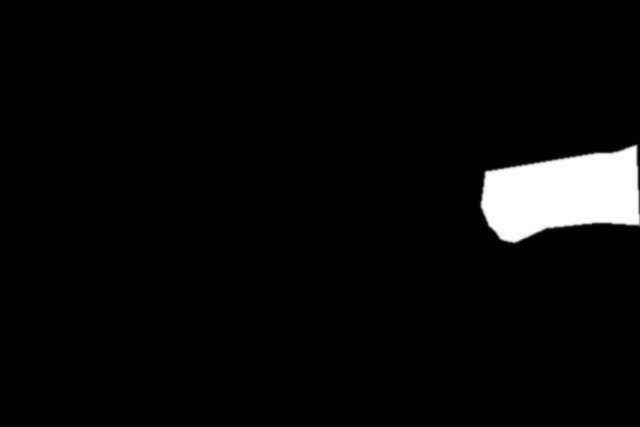}&
\includegraphics[width=\imwidth]{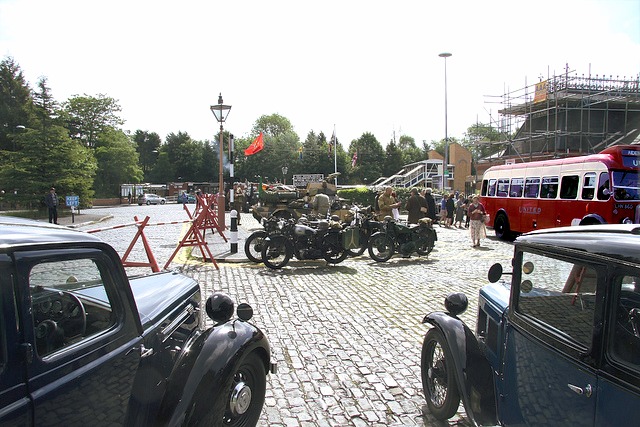}&
\includegraphics[width=\imwidth]{Ims/compFig_coco/im_hag_inc403_0081.jpg}&
\includegraphics[width=\imwidth]{Ims/compFig_coco/im_hor_inc403_-0039.jpg}&
\includegraphics[width=\imwidth]{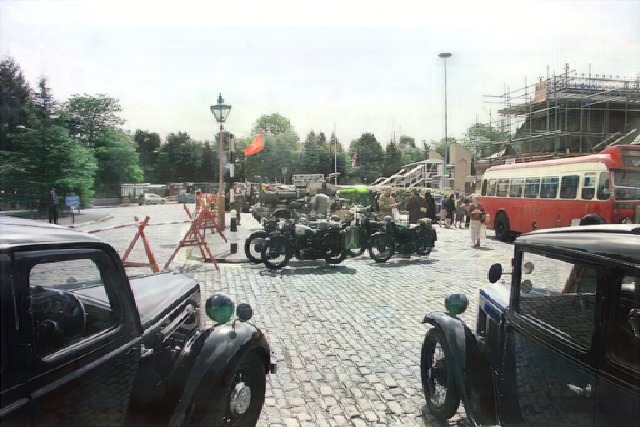}
\\
\includegraphics[width=\imwidth]{Ims/compFig_coco/im510.jpg}&
\includegraphics[width=\imwidth]{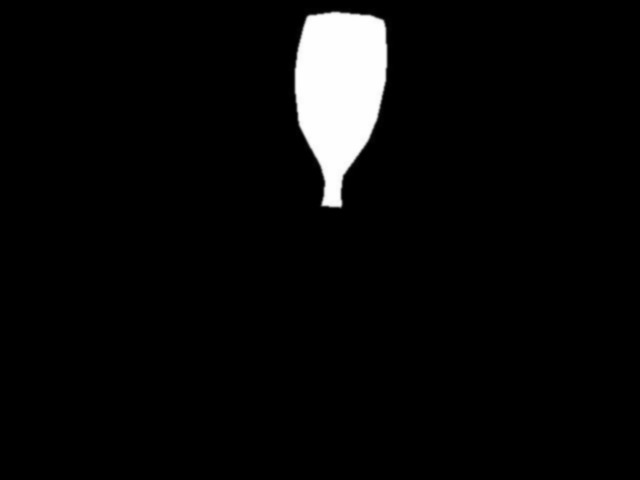}&
\includegraphics[width=\imwidth]{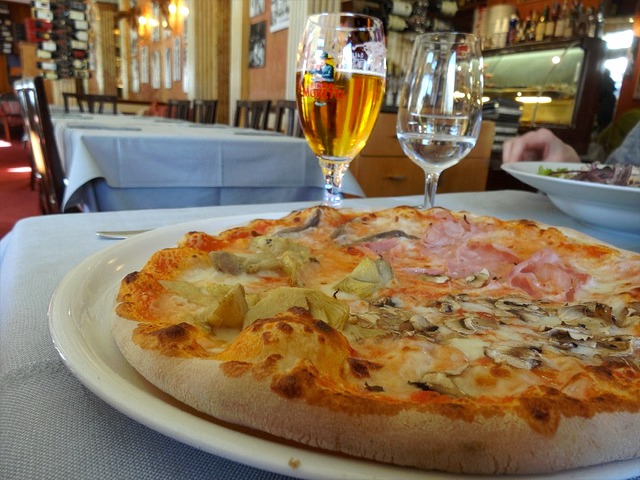}&
\includegraphics[width=\imwidth]{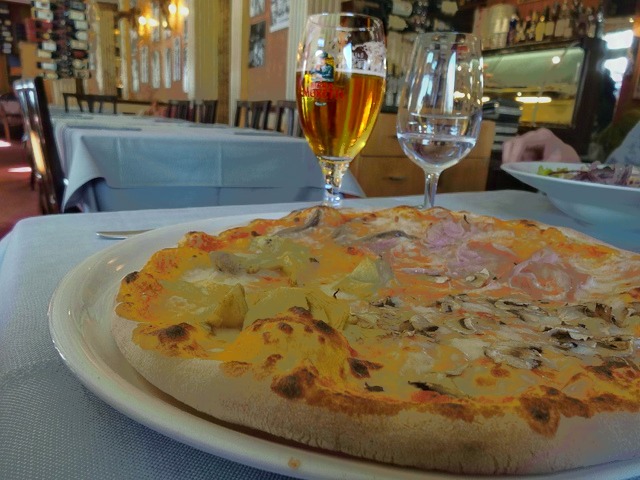}&
\includegraphics[width=\imwidth]{Ims/compFig_coco/im_hor_inc510_-0299.jpg}&
\includegraphics[width=\imwidth]{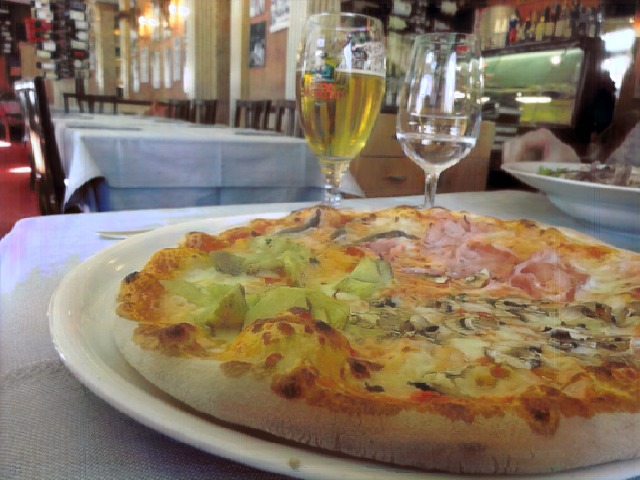}
\\
\includegraphics[width=\imwidth]{Ims/compFig_coco/im566.jpg}&
\includegraphics[width=\imwidth]{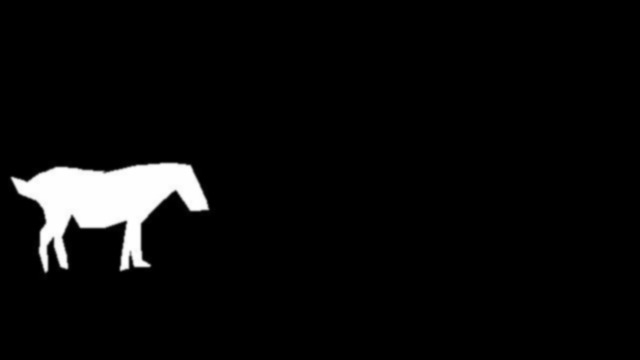}&
\includegraphics[width=\imwidth]{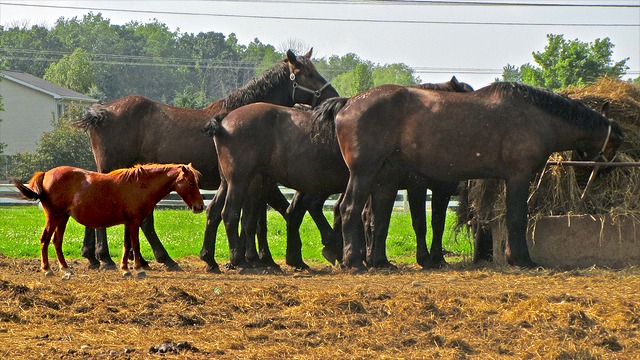}&
\includegraphics[width=\imwidth]{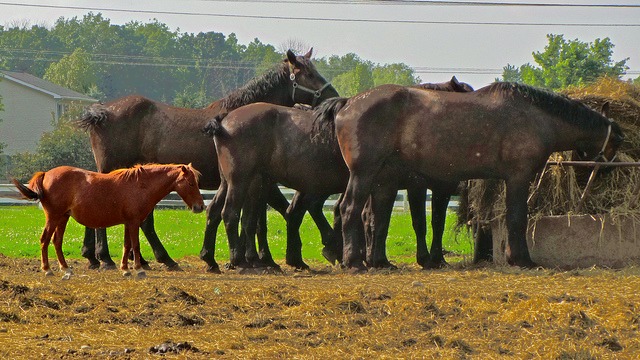}&
\includegraphics[width=\imwidth]{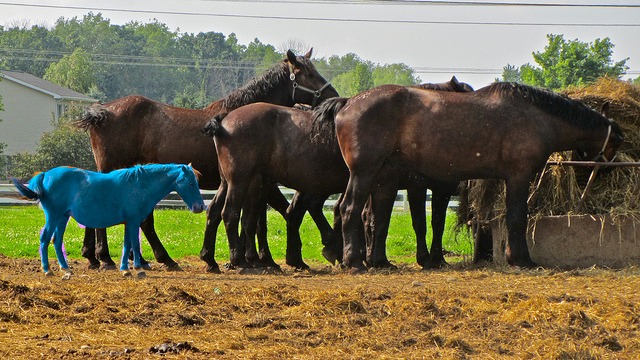}&
\includegraphics[width=\imwidth]{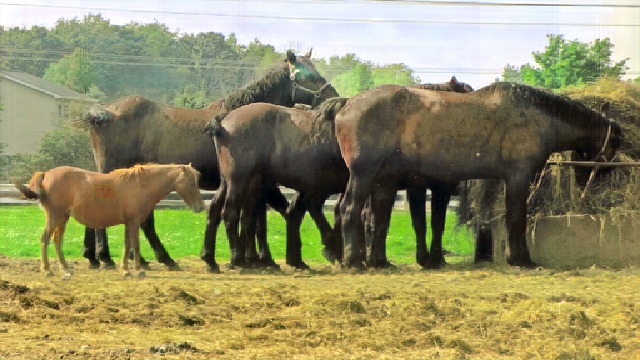}
\\
 \end{tabular}
\caption{Results on CoCoClutter dataset. More in the Supplemental Material.}
\label{fig:CoCoClutter_comp}
\end{figure*}

 We present visual results obtained from our approach and competing methods on the Mechrez and CoCoClutter datasets in Figs.~\ref{fig:comparisons} and~\ref{fig:CoCoClutter_comp}, respectively. Here we discuss the key properties of each method.

\textbf{MEC (Mechrez et al.~\cite{mechrez2019saliency}\footnote{Because the provided code could not reproduce the high quality results presented in their paper, for favorable comparison, we directly used images from their project page: \url{https://webee.technion.ac.il/labs/cgm/Computer-Graphics-Multimedia/Software/saliencyManipulation/}}):} Being patch-based, this approach is limited to reusing colors and textures from the same image. 
This approach does not always preserve the authenticity of the original photograph (note the change in shirt colors in Fig.~\ref{fig:comparisons}, rows 2 and 3). Changing image semantics may be unwanted behavior for some applications. MEC is not included in Fig.~\ref{fig:CoCoClutter_comp} because the code they provided did not reproduce their results, and led to significantly subpar quality images.

\textbf{HAG (Hagiwara et al.~\cite{hagiwara2011saliency}):} This approach alters the intensity and color values in an image, most often relying on changing the background, with only slight modifications to the foreground. This can result in big changes to the image content and low fidelity to the original (note the change of background image hue in Fig.~\ref{fig:comparisons}, rows 1-4, which has removed background texture in row 2).

\textbf{HOR (Mateescu et al.~\cite{mateescu2014attention}):} This approach modifies the masked region only to maximize the distribution separation between colors of the foreground and background, which often results in substantial color anomalies and loss of realism (striking examples can be found in Fig.~\ref{fig:comparisons}, rows 3 and 5, and Fig.~\ref{fig:CoCoClutter_comp}, all rows). 

\textbf{GAT (Gatys et al.~\cite{gatys2017guiding}):} This approach is trained using a probabilistic attention model, and requires 
a full target saliency map as input, which is not practical for a user to produce. 
To make this approach comparable with ours, we used their attention model~\cite{gatys2017guiding} to compute a map for our generated images, which we then used as the input target map for their method. 
From the last column of Figs.~\ref{fig:comparisons} and~\ref{fig:CoCoClutter_comp},
we see that GAT both struggles with making objects more salient and 
produces many artifacts.

In comparison to these approaches, our method does not change the hue of objects or semantics of the image, remaining faithful to the input image while succeeding to emphasize the desired image region. In the next section, we will quantify these observations. 

\subsection{Quantitative comparisons}\label{ssec:quantitative}

\newcommand\Mcomparison{1a}
\newcommand\Ccomparison{1b}
\newcommand\Tcomparison{1c}
\newcommand\Acomparison{1d}

We compare our approach to competing methods based on two criteria: (1)~whether a given method successfully shifts visual attention towards desired image regions, and (2)~whether the generated images remain realistic and maintain fidelity/faithfulness to the original photograph. We first performed a set of analyses using computational measures - i.e., with a computational model of saliency as a proxy for visual attention, and the LPIPS similarity metric as a measure of fidelity. Next, we ran a set of user studies on the top-performing models to (1) validate whether human attention indeed shifts towards the desired region, and (2) whether humans rate the generated images as both being realistic and having high fidelity to the original photographs. 

We performed all quantitative comparisons on two evaluation datasets: a subset of images from the Mechrez dataset~\cite{mechrez2019saliency} and images from our CoCoClutter validation set. We sampled 64 images from the Mechrez dataset, corresponding to 46 images from their \emph{object enhancement} collection and 18 images from their \emph{saliency shift} collection. We selected images containing multiple objects, in order to evaluate the ability of a given algorithm to shift attention to different image regions. For the computational measures, we used the entire CoCoClutter validation set, while for the user studies, we randomly sampled a set of 50 images with clean masks (well-segmented objects).

\subsubsection{Computational evaluation:}
To evaluate the ability of algorithms to successfully shift attention, we first use the initial and final saliency maps, corresponding to the original and edited images, to measure the mean increase of attention inside the mask area (Tables \Mcomparison,\Ccomparison, \emph{Saliency increase - Absolute}). We also measure the relative increase of attention by normalizing by the initial saliency map (\emph{Saliency increase - Relative}\footnote{Relative saliency increases can grow large when the corresponding instance has an average initial saliency value near zero.}). 
Second, we use measures previously used to evaluate attention shift~\cite{mechrez2019saliency}, the Pearson correlation and weighed F-beta~\cite{margolin2014evaluate} between the final saliency map and the binary mask (\emph{Similarity to mask}). 
Finally, as a measure of fidelity of the edited image to the original image, we use the LPIPS metric~\cite{zhang2018unreasonable}. To have more insight about the behaviour of each algorithm, we compute LPIPS on the entire image (\emph{Full}), on the background only (\emph{BG}) and on the foreground only (\emph{FG}). Note that Mechrez et al.~\cite{mechrez2019saliency} does not appear in Table \Ccomparison~as their code did not produce usable results on this dataset.

The three sets of computational measures used are complementary. \emph{Saliency increase} evaluates the increase in saliency values for the foreground, independently of changes to the background. \emph{Similarity to mask} considers the extreme case where the ground truth saliency map would be defined by the mask, hence taking into account changes in saliency both for the foreground and background. Finally, LPIPS measures how different the edited images are compared to the originals. 
An ideal model has high \emph{Saliency increase} and \emph{Similarity to mask}, and low LPIPS.    

Our approach performs the best on all computational measures of saliency across Mechrez (Table \Mcomparison) and CoCoClutter (Table \Ccomparison) datasets. 
MEC~\cite{mechrez2019saliency} comes in second due to its ability to replace foreground patches (from the mask) with salient patches from the same image. However, this often comes at the cost of sacrificing the colors and semantics of the original image. 
HAG~\cite{hagiwara2011saliency} is third overall, but suffers from high LPIPS scores (LPIPS \emph{Full} and \emph{BG}) as it often heavily modifies the original colors and details of the background. LPIPS scores for the foreground are generally very small, suggesting that HAG relies mostly on background modifications in its attention shifting pipeline. 
In contrast, HOR~\cite{mateescu2014attention} achieves very high LPIPS scores on the foreground due to the severe color artifacts it creates. At the same time, this method achieves the lowest LPIPS scores when considering the entire image (LPIPS \emph{Full}) because it leaves the background untouched, which usually covers the largest area of the image. 
Finally GAT~\cite{gatys2017guiding} performs the worst in terms of shifting attention, while simultaneously heavily modifying the image, resulting in the highest LPIPS scores and lowest \emph{Saliency increase} and \emph{Similarity to mask} scores. 
Overall, our algorithm is most successful at shifting computational attention, 
all while conserving original image properties, including, colors, textures, and semantics. It is in the top two models when considering LPIPS scores.

\begin{table}[!htb]
    \begin{subtable}[t]{.55\linewidth}
      \centering
    \resizebox{\textwidth}{!}{%
    (a)\begin{tabular}[t]{|c|ccc|cc|cc|}
    \hline
    \multirow{2}{*}{Model} & \multicolumn{3}{c|}{LPIPS $\downarrow$} & \multicolumn{2}{c|}{Saliency increase $\uparrow$} & \multicolumn{2}{c|}{Similarity to mask $\uparrow$} \\
     & Full & BG & FG & Absolute & Relative & WFB & CC  \\ \hline
Our    &   \cellcolor{green!35}5.96 &  \cellcolor{green!35}5.10 &  \cellcolor{green!35}0.81 &  \cellcolor{green!80}3.80 &  \cellcolor{green!80}35.77 &  \cellcolor{green!80}11.30 &  \cellcolor{green!80}30.43 \\
MEC~\cite{mechrez2019saliency} &   8.95 &  7.65 &  1.29 &  \cellcolor{green!35}3.42 &  \cellcolor{green!35}28.39 &  \cellcolor{green!35}10.42 &  \cellcolor{green!35}26.25 \\
HOR~\cite{mateescu2014attention}     &   \cellcolor{green!80}1.60 &  \cellcolor{green!80}0 &  1.41 &  1.74 &  16.29 &  9.85 &  24.43 \\
HAG~\cite{hagiwara2011saliency}      &   11.08 &  10.68 &  \cellcolor{green!80}0.37 &  2.24 &  21.75 &  10.33 &  26.12 \\
GAT~\cite{gatys2017guiding}     &   25.64 &  24.72 &  1.11 &  0.34 &  3.05 &  9.59 &  22.98 \\
\hline
    \end{tabular}}
    \label{tab:comp_mechrez}
    
    \resizebox{\textwidth}{!}{%
    (b)\begin{tabular}{|c|ccc|cc|cc|}
    \hline
    \multirow{2}{*}{Model} & \multicolumn{3}{c|}{LPIPS $\downarrow$} & \multicolumn{2}{c|}{Saliency increase $\uparrow$} & \multicolumn{2}{c|}{Similarity to mask $\uparrow$} \\
     & Full & BG & FG & Absolute & Relative & WFB & CC  \\ \hline
Our &   \cellcolor{green!35}4.87 &  \cellcolor{green!35}2.84 &  \cellcolor{green!35}1.95 & \cellcolor{green!80} 1.99 &  \cellcolor{green!80}25957.92 &  \cellcolor{green!80}11.81 &  \cellcolor{green!80}20.65 \\
HAG~\cite{hagiwara2011saliency}  &   7.37 &  6.58 &  \cellcolor{green!80}0.69 &  \cellcolor{green!35}1.30 &  \cellcolor{green!35}24419.15 &  11.22 &  18.89 \\
HOR~\cite{mateescu2014attention}  &   \cellcolor{green!80}4.00 &  \cellcolor{green!80}0 &  3.61 &  1.27 &  11065.47 &  \cellcolor{green!35}11.27 &  \cellcolor{green!35}18.98 \\
GAT~\cite{mechrez2019saliency}  &   30.07 &  27.08 &  3.61 & -0.05 &    2920.75 &  10.23 &  15.85 \\
    \hline
    \end{tabular}}
    \label{tab:comp_coco_clutter}

    \resizebox{0.4\textwidth}{!}{
    (c)\begin{tabular}{|c|c|}
    \hline
Model & Avg. run time \\
\hline
Our & \cellcolor{green!80}8.87s \\
HOR~\cite{mateescu2014attention} & 31.82s\\
HAG~\cite{hagiwara2011saliency} & 4743.54s\\
MEC~\cite{mechrez2019saliency} & $>$1 day \\
\hline
\end{tabular}}
\label{tab:time}
\resizebox{0.35\textwidth}{!}{
    \begin{tabular}{|c|c|}
    \hline
Model & Avg. run time \\
\hline
Our & \cellcolor{green!80}1.54s \\
GAT~\cite{gatys2017guiding} & 4.34s \\
\hline
\end{tabular}}
\label{tab:time}

    \end{subtable} 
    \begin{subtable}[t]{.45\linewidth}
      \centering
    \resizebox{.85\textwidth}{!}{%
    (d)\begin{tabular}[t]{|c|c|cc|}
    \hline
    \multirow{2}{*}{Model} & \multicolumn{1}{c|}{LPIPS $\downarrow$} & \multicolumn{2}{c|}{Saliency increase $\uparrow$}\\
     & Full  & Absolute & Relative \\ \hline
Our      &   \cellcolor{green!20}5.96  &  \cellcolor{green!35}3.80 &  \cellcolor{green!20}35.77  \\
\hline
\multicolumn{4}{|c|}{Parameter ablations}\\
\hline
tone + color &   \cellcolor{red!70}10.97  &  \cellcolor{green!80}5.33 &  \cellcolor{green!80}56.19  \\
sharp + exp + cont &   \cellcolor{green!65}1.76 &  \cellcolor{red!55}1.47 &  \cellcolor{red!55}13.00 \\
color &   \cellcolor{red!55}9.95  &  \cellcolor{red!10}3.50 &  \cellcolor{green!50}37.25  \\
our + saturation &   \cellcolor{red!40}9.70  &  \cellcolor{green!65}4.84 &  \cellcolor{green!65}48.76  \\
Fixed parameters &   \cellcolor{green!50}2.28 &  \cellcolor{red!40}1.95 &  \cellcolor{red!40}17.23  \\
\hline
\multicolumn{4}{|c|}{fg/bg ablations}\\
\hline
bg-only   &   \cellcolor{green!35}2.31  &  \cellcolor{red!70}0.54 &  \cellcolor{red!70}3.50  \\
fg-only   &   \cellcolor{green!80}1.19  &  \cellcolor{red!25}2.53 &  \cellcolor{red!25}29.40 \\
\hline
\multicolumn{4}{|c|}{order of operations ablations}\\
\hline
col,ton,con,exp,sha    &   6.74  &  \cellcolor{green!50}4.01 &  35.36  \\
ton,col,sha,exp,con    &   \cellcolor{red!25}8.04  & \cellcolor{green!20}3.73 &  32.47 \\
con,exp,sha,ton,col    &   6.55  &  3.65 &  \cellcolor{green!35}36.98 \\
ton,col,sha,con,exp    &   \cellcolor{red!10}7.87  &  3.72 &  \cellcolor{red!10}30.49  \\
\hline
    \end{tabular}}
    \label{tab:ablation}
    \end{subtable}%
\caption{(a) Computational evaluation on Mechrez dataset. (b) Computational evaluation on CoCoClutter dataset. The top two performing models according to each metric are highlighted in green (darker is better). (c) Left: Run time averaged over 30 high resolution images; Right: As GAT is unable to run directly on high resolution images, we use low resolution versions of the same images. (d) Ablation studies on Mechrez dataset. Our chosen method involves applying sharpening, exposure, contrast, tone adjustment, and color adjustment (in that order) to both foreground and background. Ablations consider a subset of parameters and different orderings, as well as application to either one of foreground/background. Darker green colors indicate better scores, darker red colors indicate worse scores. All metrics except run time are multiplied by 100 for legibility.}
\end{table}






\subsubsection{User studies:}
We reinforce the findings from the computational measures by collecting human data using Amazon's Mechanical Turk for three tasks: visual attention, image fidelity, and image realism. 
Our first set of studies measured shifts in human attention, when images are modified by the various methods compared. We used the same crowdsourced gaze tracking method, \emph{CodeCharts}~\cite{newman2020turkeyes}, that was used to collect training data for the saliency model~\cite{Fosco_2020_CVPR} we adapted in GazeShiftNet.
In CodeCharts, participants are asked to look at an image for a few seconds, then a jittered grid of alphanumeric triplets (``codes") is flashed for a brief interval, and the participant is subsequently prompted to enter the code seen last. This task design captures the area where a participant was looking at the moment when the image disappeared. CodeCharts has been shown to approximate human eye movements collected using an eye tracker~\cite{newman2020turkeyes}, which we use to evaluate the ability of algorithms to shift human attention by modifying images. 

We used CodeCharts to collect human attention data on 64 images from the Mechrez dataset and 50 from the CoCoClutter dataset. We collected attention data on the original images and on the edited images produced by each method. 
We obtained gaze points from an average of 50 participants per image, that we converted into an attention heatmap for the image. We then compared the attention heatmaps of the original images to those of the edited images, to evaluate the relative attention increase achievable by each method. These values are plotted on the x-axes of Figs.~\ref{fig:userstudies}a and b (left) for the CoCoClutter and Mechrez datasets, respectively. The table of values is also available in the Supplemental Material. Based on these study results, GazeShiftNet achieves a greater attention shift than HAG, a smaller one than HOR, and a comparable one to MEC. These results are unsurprising given the algorithm behaviors: HAG mostly modifies the background which leads to less noticeable change to the masked objects, whereas HOR often drastically changes the color of the foreground object making it significantly stand out from the rest of the image. From Fig.~\ref{fig:userstudies}, we can see that our method achieves a balance between shifting attention and maintaining image fidelity, as described below.

The next set of user studies measured image fidelity. We asked an average of 25 participants to evaluate each edited image compared to its original using the following options: not edited (3), slightly edited (2), moderately edited (1), definitely edited (0). In parentheses are the fidelity scores we assigned to each answer (higher is better). Fidelity score distributions across images, averaged over study participants, are visualized as box plots in Fig.~\ref{fig:userstudies}a and b (middle) for the CoCoClutter and Mechrez datasets, respectively. Both GazeShiftNet and HAG achieve fidelity scores that are statistically significantly higher than HOR on the CoCoClutter dataset, and higher than MEC on the Mechrez dataset ($p<0.001$). No other pairwise comparisons were significant. 
The human fidelity judgements are intended to evaluate the same aspect of models as the LPIPS computational metric. Because the fidelity scores of the models correspond most to their LPIPS FG scores from Tables \Mcomparison~and \Ccomparison, we suspect that study participants focused more on the foreground regions of the edited images when judging fidelity.    

While fidelity measured how similar an edited image is compared to its original, we also ran a user study to evaluate the realism of the edited images in isolation. We asked an average of 25 participants to evaluate whether each image is: 
definitely not edited~(3), probably not edited (2), probably edited (1), definitely edited (0). In parentheses are the realism scores we assigned to each answer (higher is better). Realism score distributions across images, averaged over study participants, can be found in Fig.~\ref{fig:userstudies}a and b (right) for the CoCoClutter and Mechrez image datasets, respectively. On Mechrez, all methods perform comparably in terms of realism. This supports the findings from Mechrez et al.~\cite{mechrez2019saliency}, who similarly found that when evaluated using a realism user study, the algorithms performed comparably.
However, on the CoCoClutter dataset, differences across methods are more pronounced. GazeShiftNet and HAG obtained statistically significantly higher realism scores than HOR ($p<0.001$). The reason for these differences is that the Mechrez dataset masks often cover only part of an object, which conserves realism when edits are applied (e.g., an edited blue shirt is no less realistic than the original gray one; though fidelity suffers in this case), but the CoCoClutter dataset masks include full objects, which exposes significant failure methods of other approaches (such as when the HOR method turns entire people blue).

Across fidelity and realism, our approach achieves the smallest standard deviations in scores, and the lower range of scores starts higher (Fig.~\ref{fig:userstudies}). This shows that GazeShiftNet behaves more consistently across a variety of image types, with fewer catastrophic failures than some of the other methods, making it a more practical method overall.

As additional validation of our approach, we compared its performance to that of professional users on the 30 high-resolution images from our exploratory studies (Sec.~\ref{sec:method}). The results show that while users produce edits that are judged to have more realism and fidelity (and correspondingly lower LPIPS \emph{FG}) scores, our model is able to achieve both higher saliency and attention shift scores than users on average (although some users produce quite effective attention shifts in photos). We note that our model achieves the highest increase in saliency/attention than HAG and HOR on this image dataset. Detailed study results can be found in the Supplemental Material.


\begin{figure}[!ht]
\centering
\includegraphics[width=0.48\linewidth]{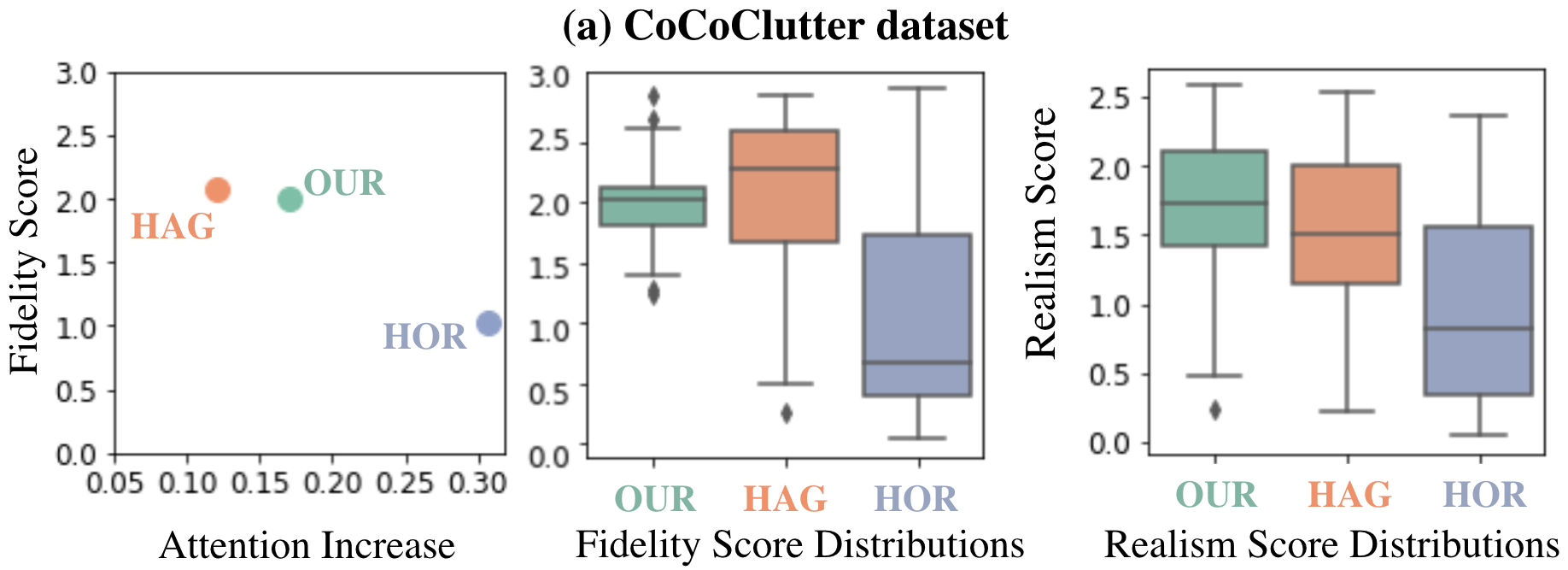} 
\includegraphics[width=0.48\linewidth]{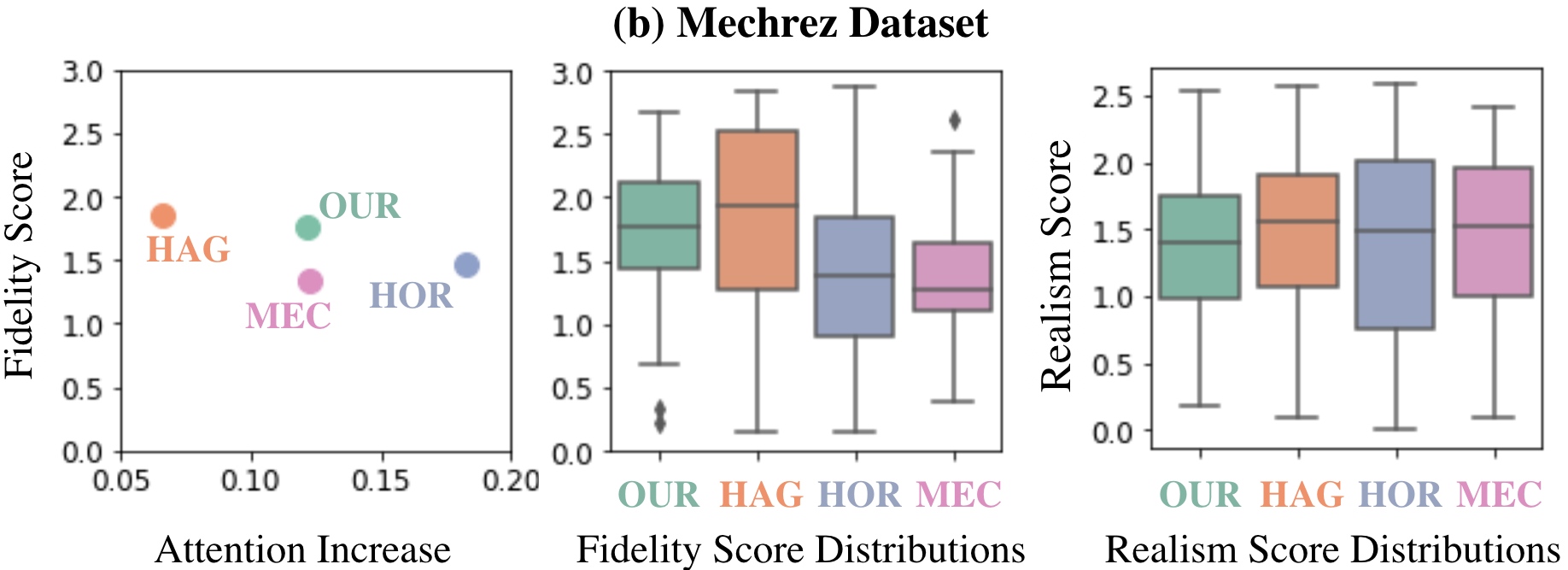} 
\caption{Results of user studies on three separate crowdsourcing tasks - measuring image fidelity, realism, and human attention - on two datasets: (a)~CoCoClutter and (b)~Mechrez. Different methods trade-off average image fidelity for average attention increase, and vice versa, whereas our approach achieves a good balance of both (left). 
 Box plots of the fidelity score distributions (middle) and realism score distributions (right) demonstrate the variability in the performance of each method. Ours performs most consistently, with the smallest range of scores, i.e., fewer failure modes. 
 All methods perform similarly in realism on the Mechrez dataset because of the nature of the objects selected for emphasis (see text for details).}
\label{fig:userstudies}
\end{figure}

\subsection{Run time comparisons}
We timed each algorithm on 30 high-resolution images ($[648, 1332]\times1500$ pixels)
obtained from Adobe Stock and manually annotated to include a masked object per image. Average run times over the 30 images are listed in Table \Tcomparison. Our algorithm is significantly faster than HAG and MEC, which are iterative optimization-based algorithms. 
In fact, MEC took more than one day to process the 30 images. HOR is fairly quick but still takes more than a second per image. GAT cannot run on high resolution images, so for this comparison we used the same 30 images in resolutions corresponding to the required neural network input sizes. Because their network is deeper and their method additionally requires computing a target saliency map as an input, their computation time is nearly 3 times ours.

\subsection{Ablation studies}\label{ssec:ablations}
Table \Acomparison~includes a summary of the ablation tests evaluating our design choices: (1) the set of parametric transformations, (2) whether to apply parametric transformations to foreground, background, or both, (3) the ordering of transformations applied to images. As it is not immediately obvious how to balance the trade-off between realism and attention shift across our different ablations, we identify the five best models according to three metrics (LPIPS, absolute and relative saliency increase), and choose a high-performing model across all criteria. In Table \Acomparison~the best performing models are highlighted in green (darker is better), and the worst performing models are highlighted in red (darker is worst). Our final model selection achieves good performance across the metrics. A detailed discussion of the ablation experiments can be found in the Supplemental Material. 

\section{Extensions}
Our method could be extended further, making it even more flexible. We discuss additional use cases and extensions below.

\begin{figure*}
 \centering
\includegraphics[width=0.95\linewidth]{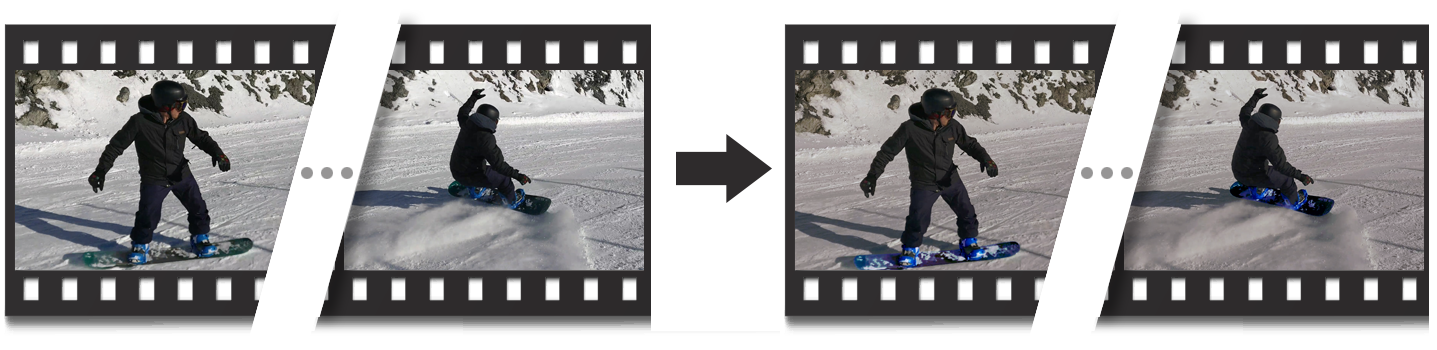} 
\caption{Our method can be seamlessly applied to video frames. We predict parameters for the first frame and transfer them to all subsequent frames containing the same segmented object (in this case, the snowboard).}
\label{fig:video}
\end{figure*}

\textbf{Application to videos:}
GazeShiftNet can be seamlessly generalized to videos by predicting parameters on the first frame, and applying them to subsequent video frames (provided we have a common, segmented object across frames). We find this produces good results on the rest of the frames while avoiding flickering. 
In contrast, most competing attention shifting algorithms~\cite{gatys2017guiding,hagiwara2011saliency,mechrez2019saliency} would need to be run on each frame separately, as their transformations cannot be transferred across images. 
We show in Fig.~\ref{fig:video} snapshots from a video from the DAVIS dataset \cite{Perazzi_CVPR_2016}.

\textbf{Interactive image editing:}
GazeShiftNet, being parametric, makes it easy to hand control of the edited image back to the user. By interpolating between the predicted parameters and the set of parameters where $I'=I$, we can let a user dial the edits up or down. We built a prototype of an interface where a user can interact with a single saliency slider that interpolates the parameters at interactive rates (see Fig.~\ref{fig:teaser} and the Supplemental Material). This form of interpolation results in smooth, artifact-free image transformations. The user can also adjust each of the parameter sliders separately. Importantly, while parameters can be predicted on smaller image sizes, final transformations can be applied to professional high-resolution photography at interactive rates. These properties do not hold for non-parameteric methods~\cite{gatys2017guiding,mechrez2019saliency}. Optimization-based methods~\cite{hagiwara2011saliency,mateescu2014attention} do not allow for interactive and artifact-free post-processing edits.

\begin{figure*}
\centering
\setlength{\imwidth}{0.16\linewidth}
\setbox1=\hbox{\includegraphics[width=\imwidth]{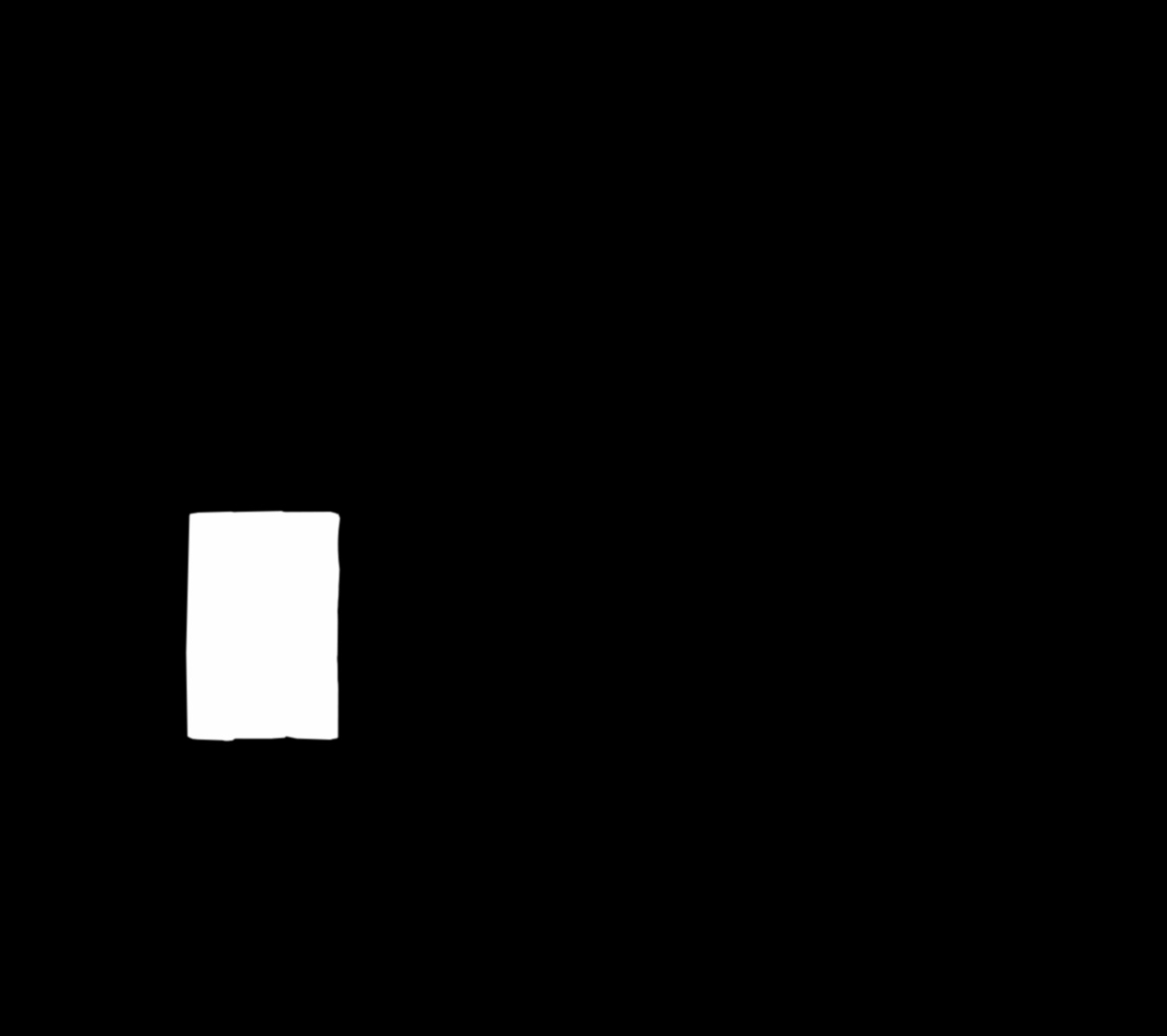}}
\setlength{\tabcolsep}{1pt}
\begin{tabular}{cccccc}

Input & $z_1$  & $z_2$ & $z_3$ & user 1 & user 2
\\



\includegraphics[width=\imwidth]{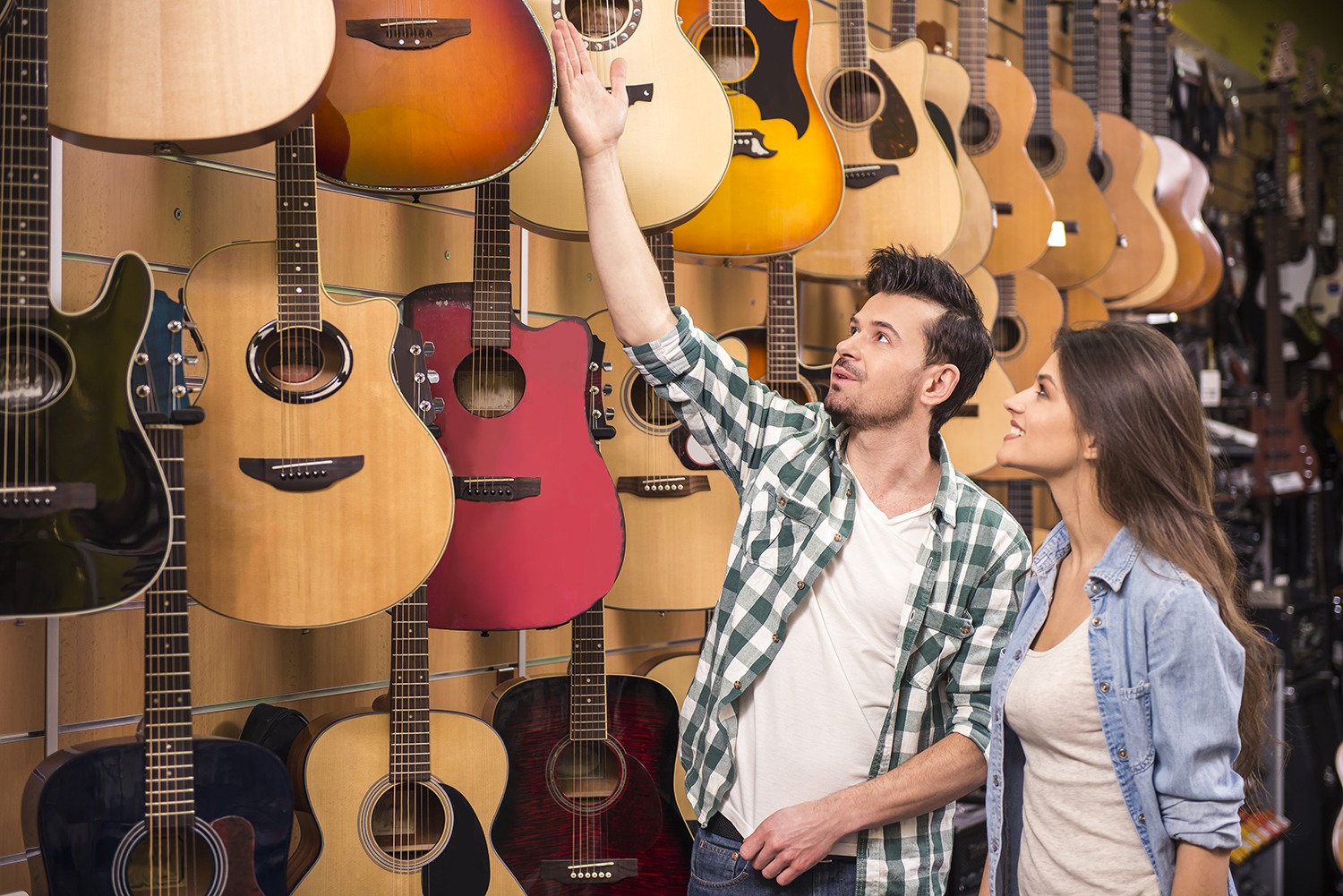}\llap{{\raisebox{0.465\imwidth}{\includegraphics[cfbox=red, width=0.3\imwidth]{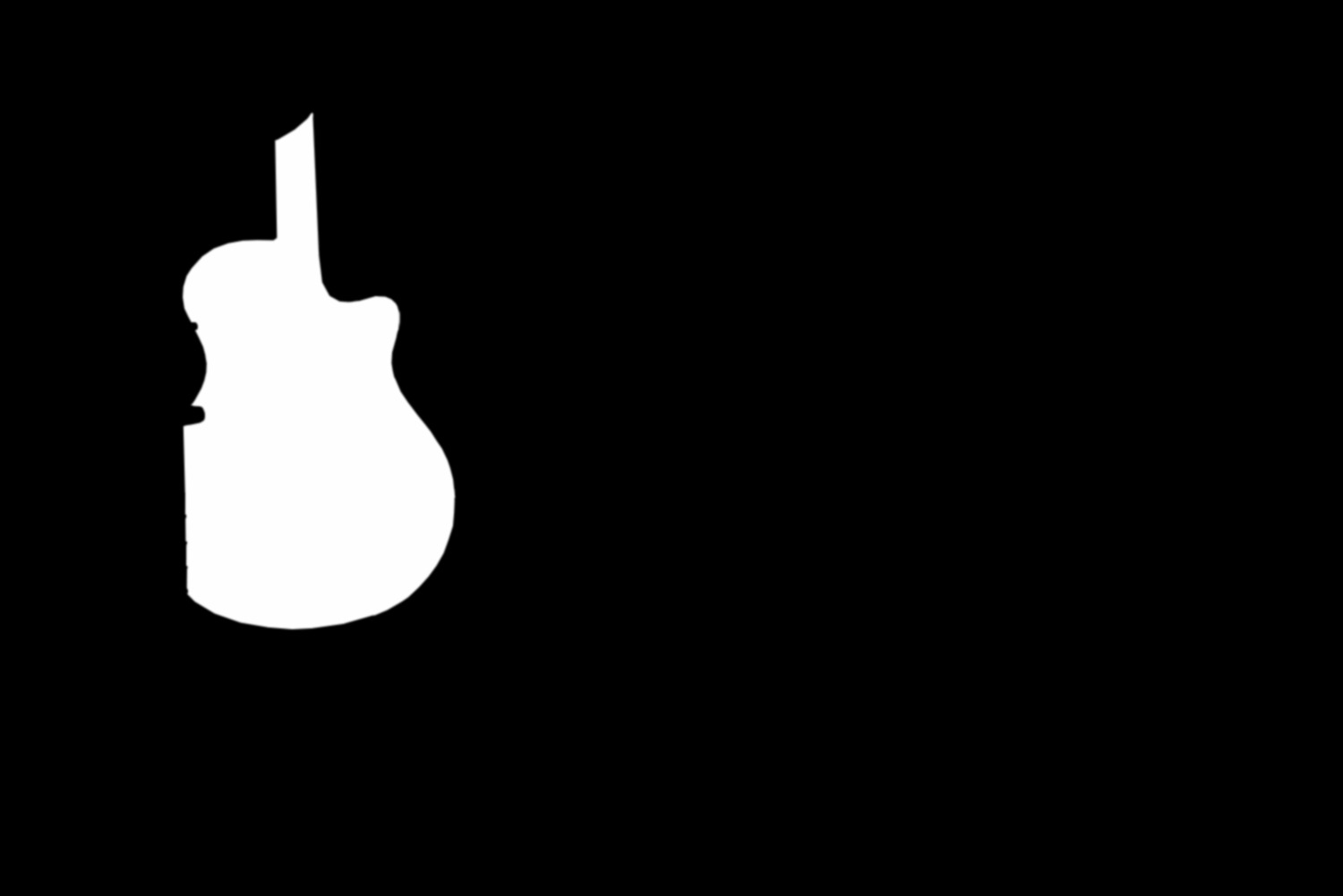}}}} &
\includegraphics[width=\imwidth]{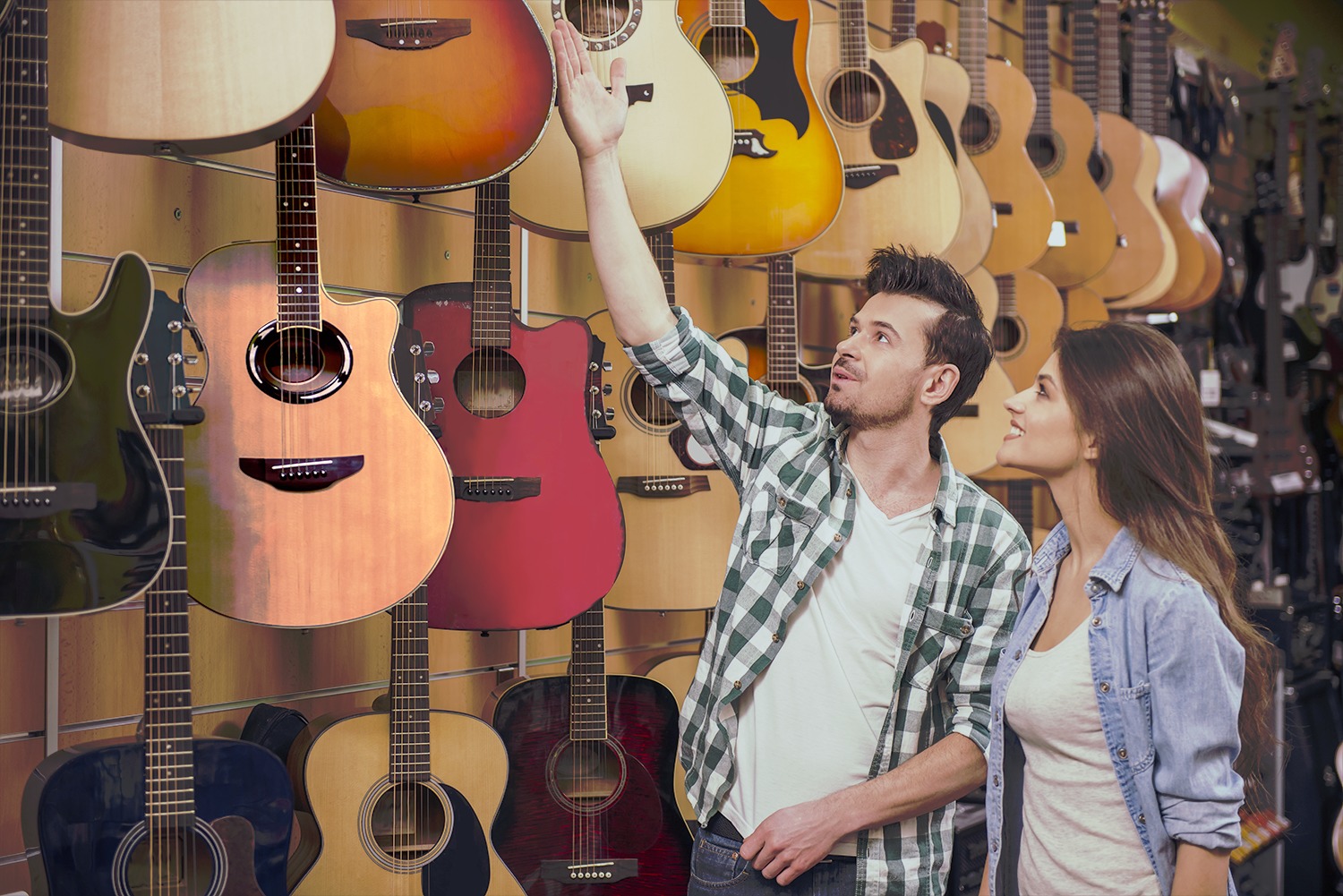}&
\includegraphics[width=\imwidth]{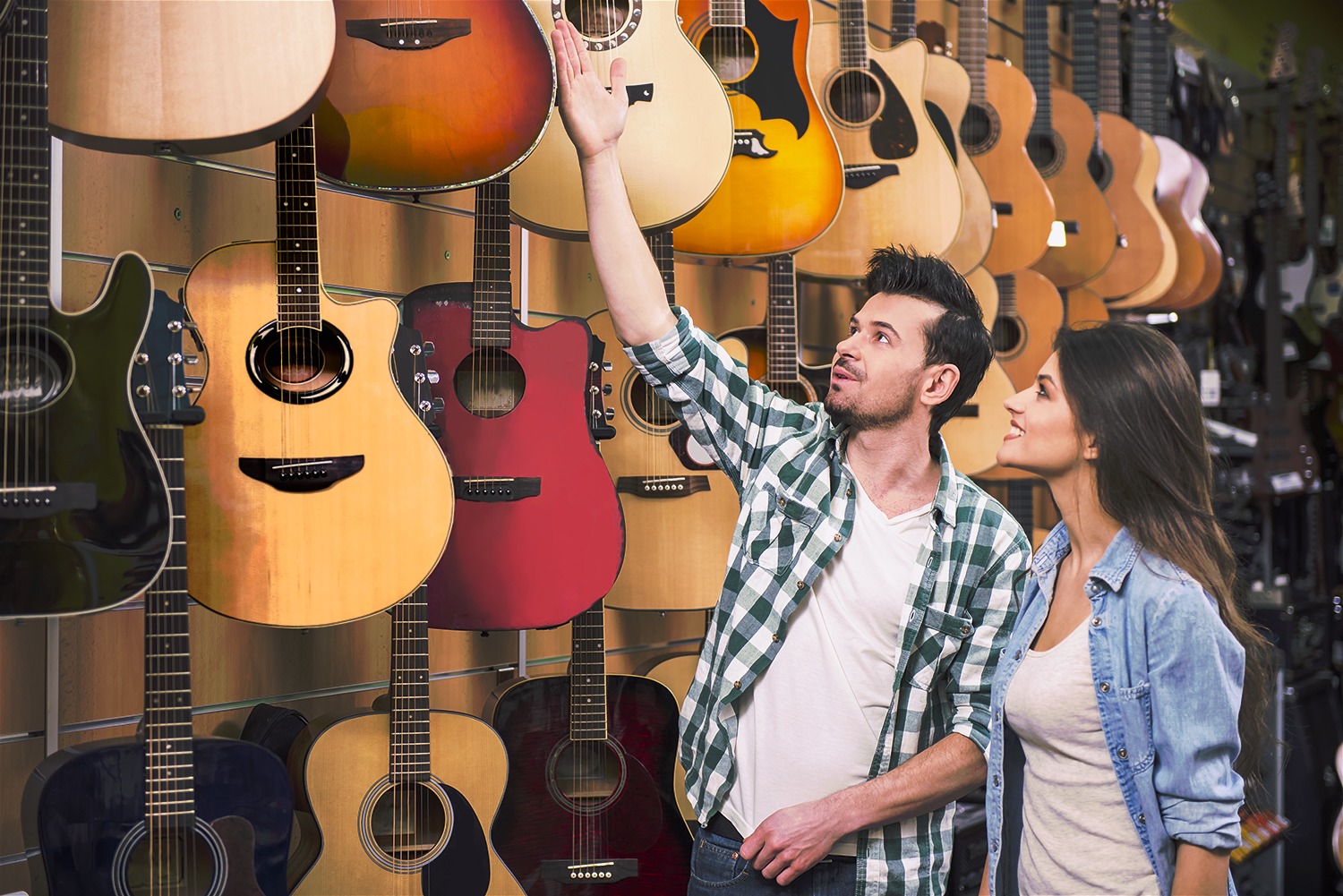}&
\includegraphics[width=\imwidth]{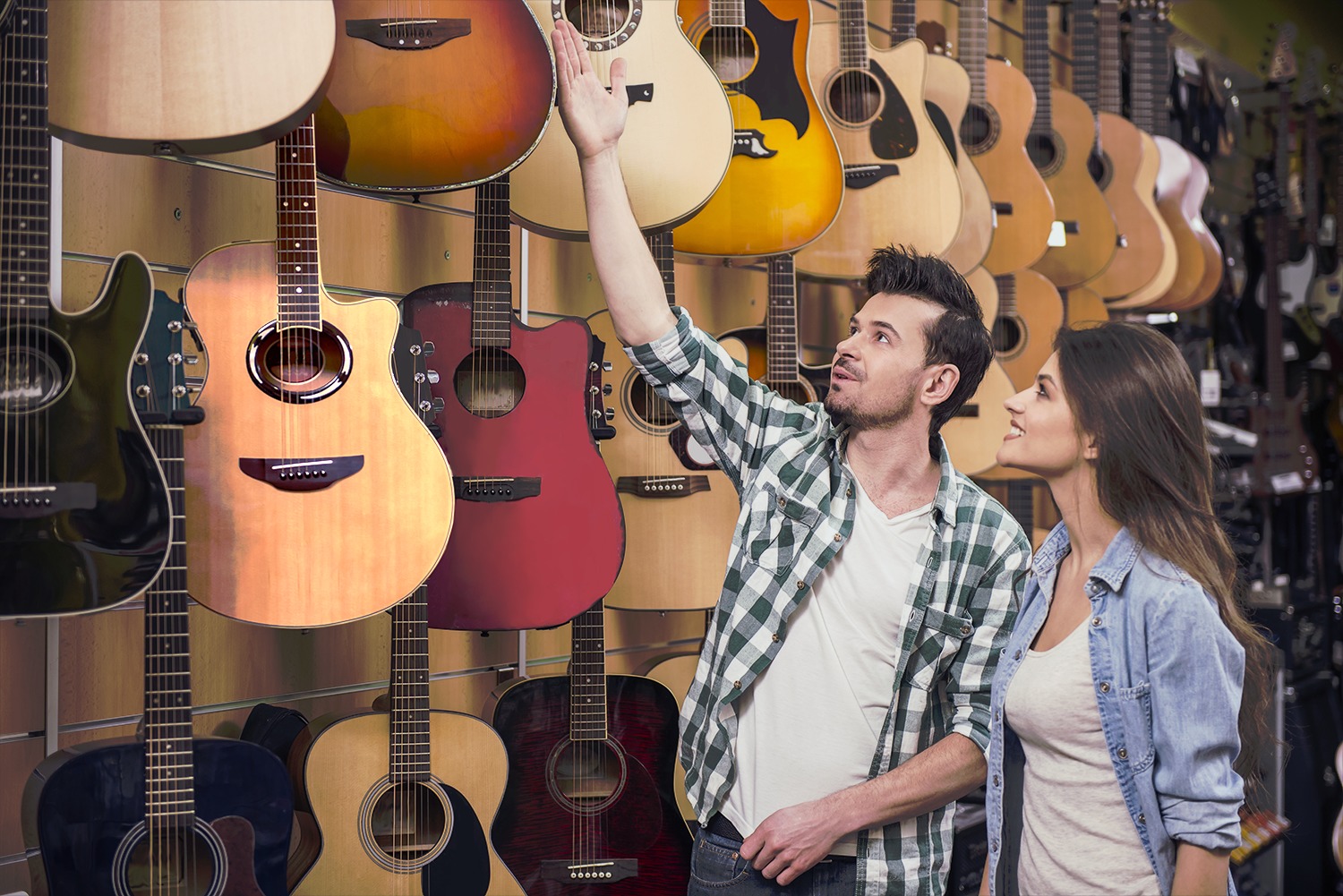}&
\includegraphics[width=\imwidth]{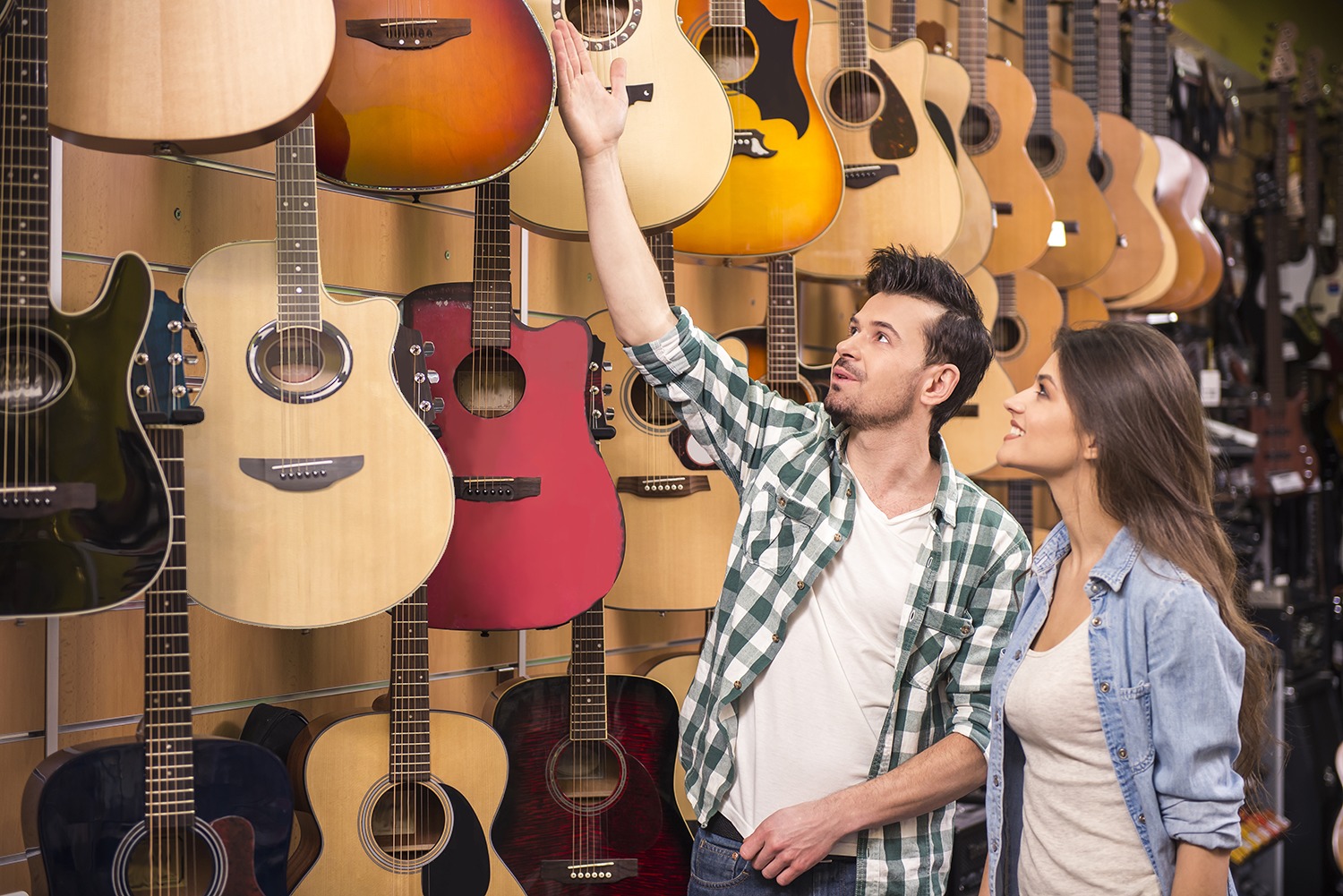}&
\includegraphics[width=\imwidth]{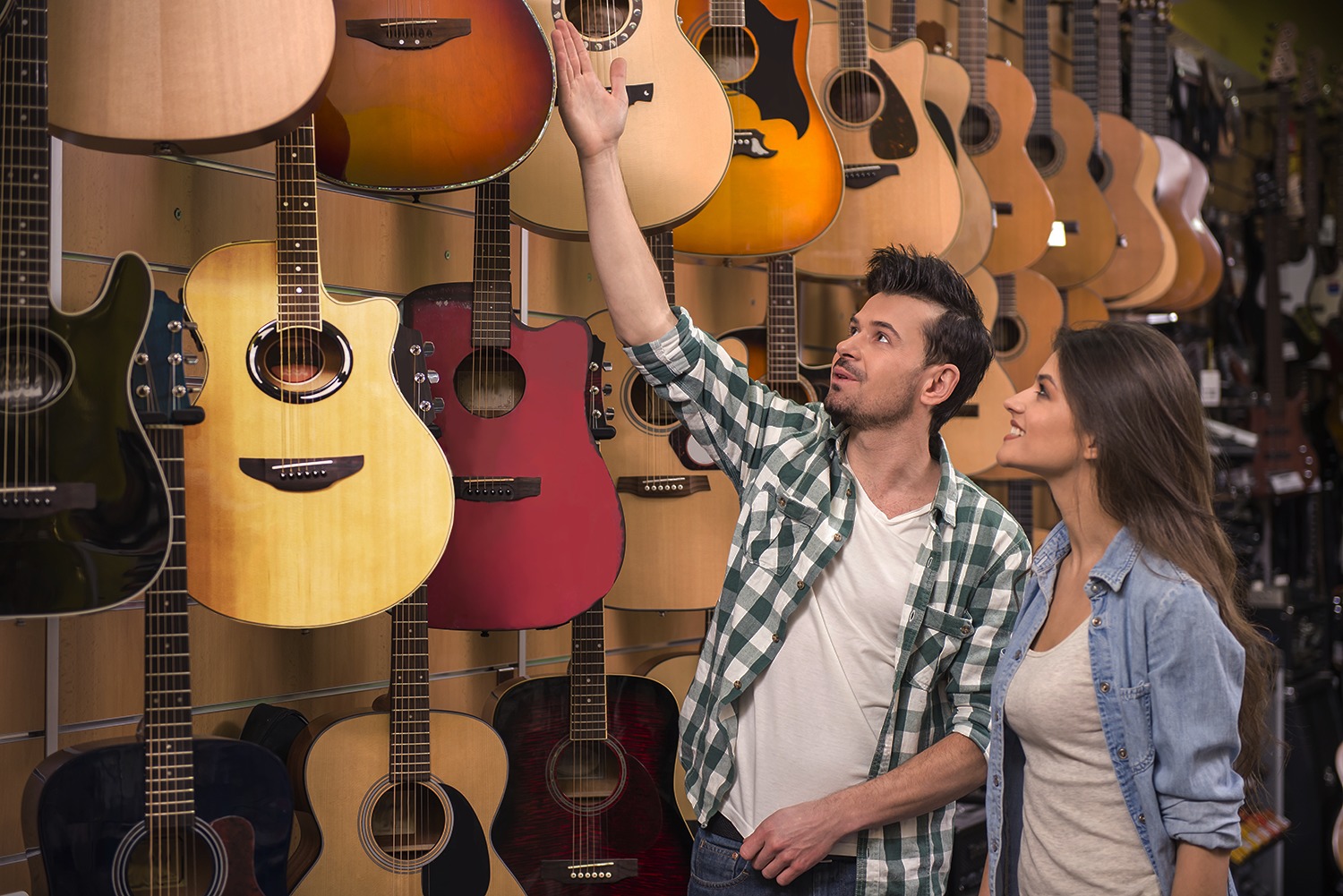}\\

\includegraphics[width=\imwidth]{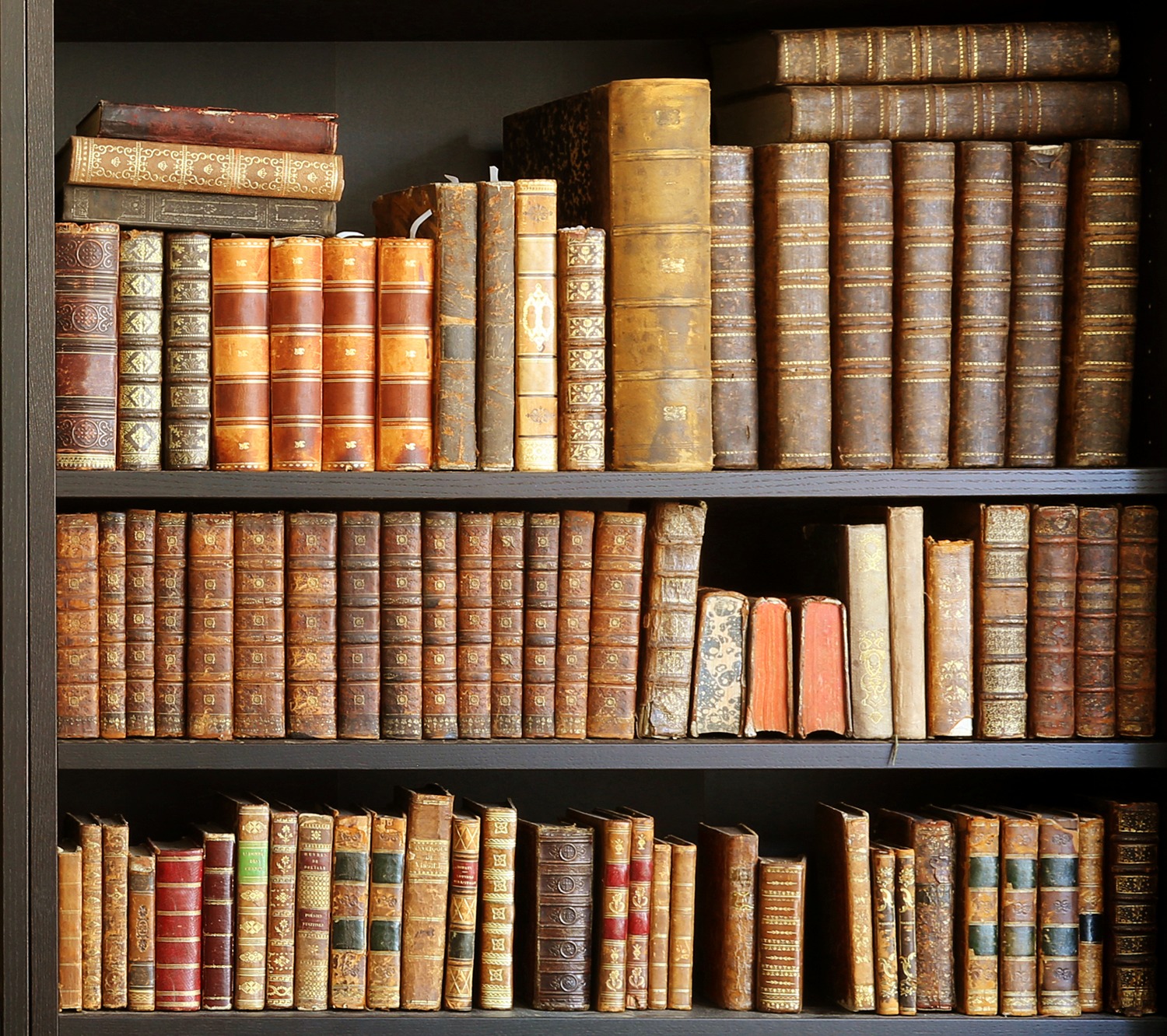}\llap{{\raisebox{0.53\imwidth}{\includegraphics[cfbox=red, width=0.4\imwidth]{Ims/MultiStyle_celso/mask8.jpg}}}} &
\includegraphics[width=\imwidth]{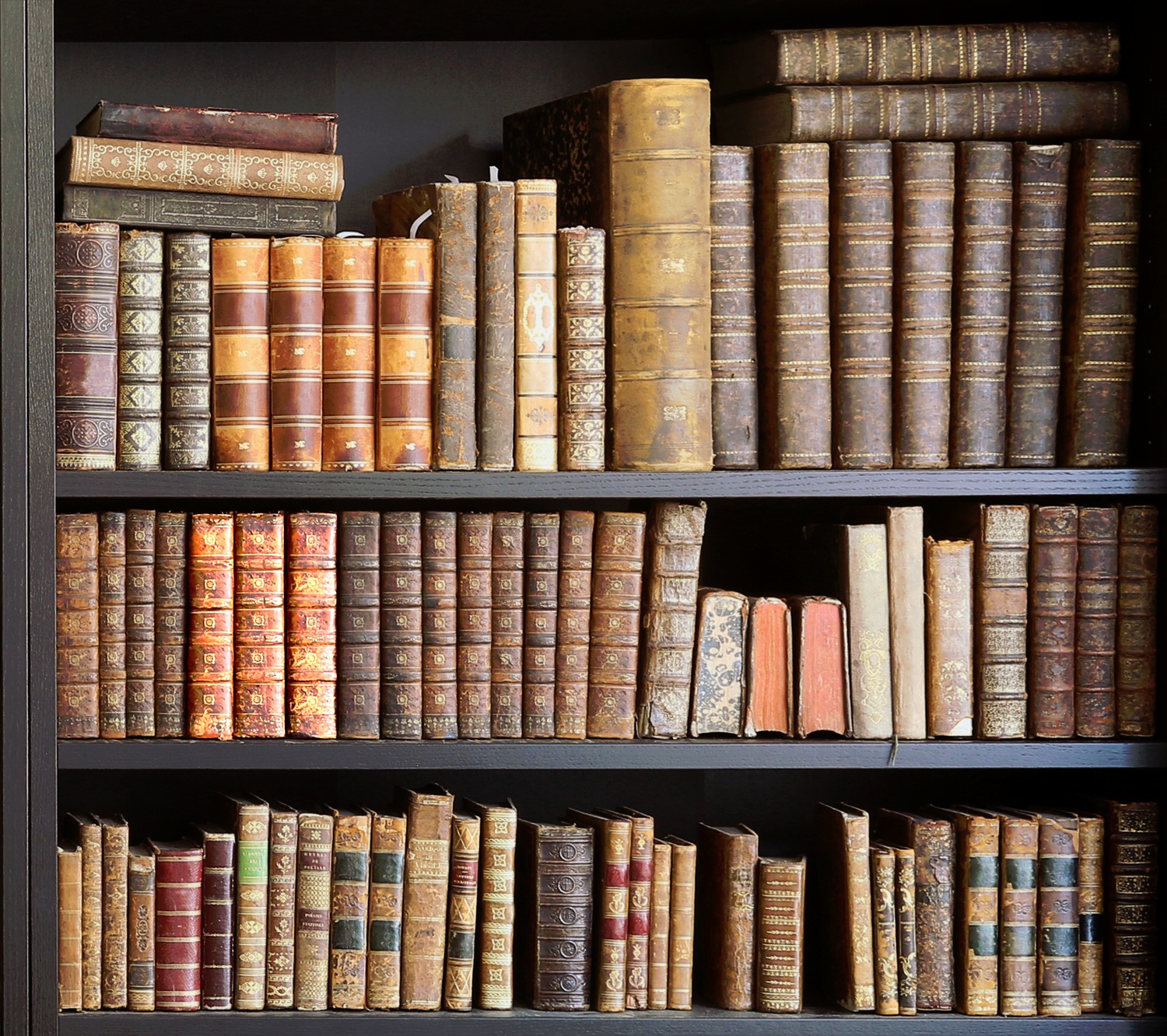}&
\includegraphics[width=\imwidth]{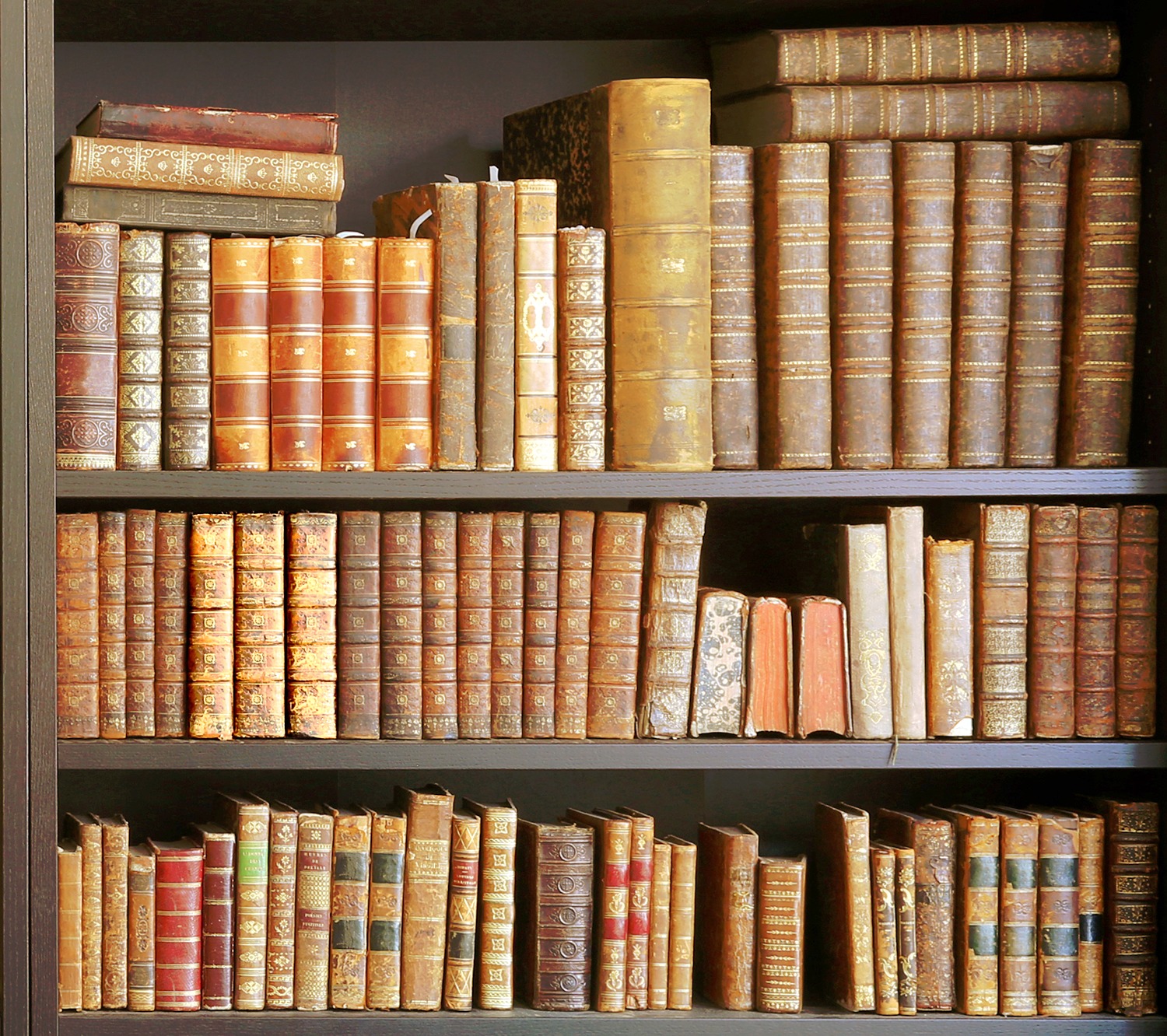}&
\includegraphics[width=\imwidth]{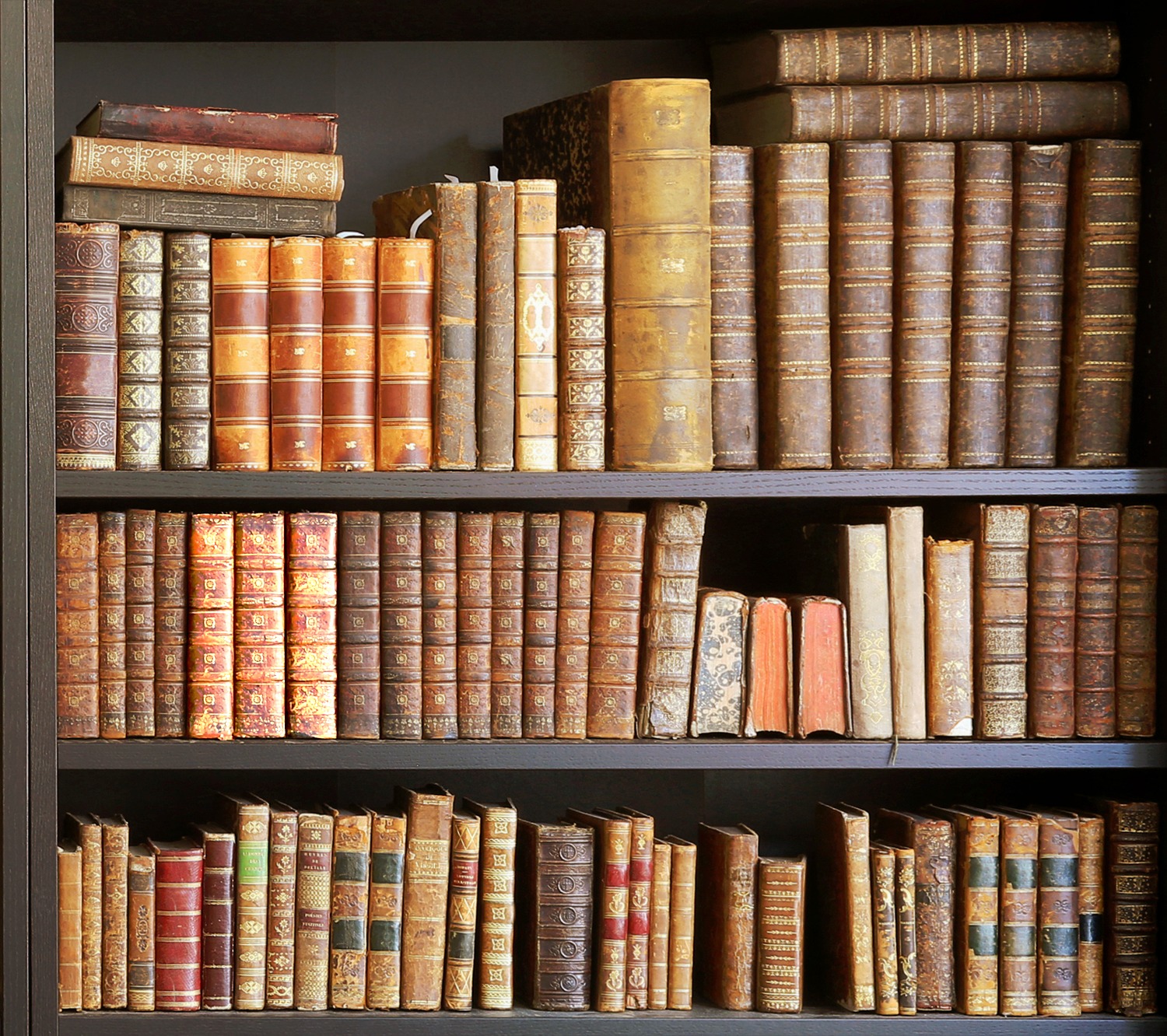}&
\includegraphics[width=\imwidth]{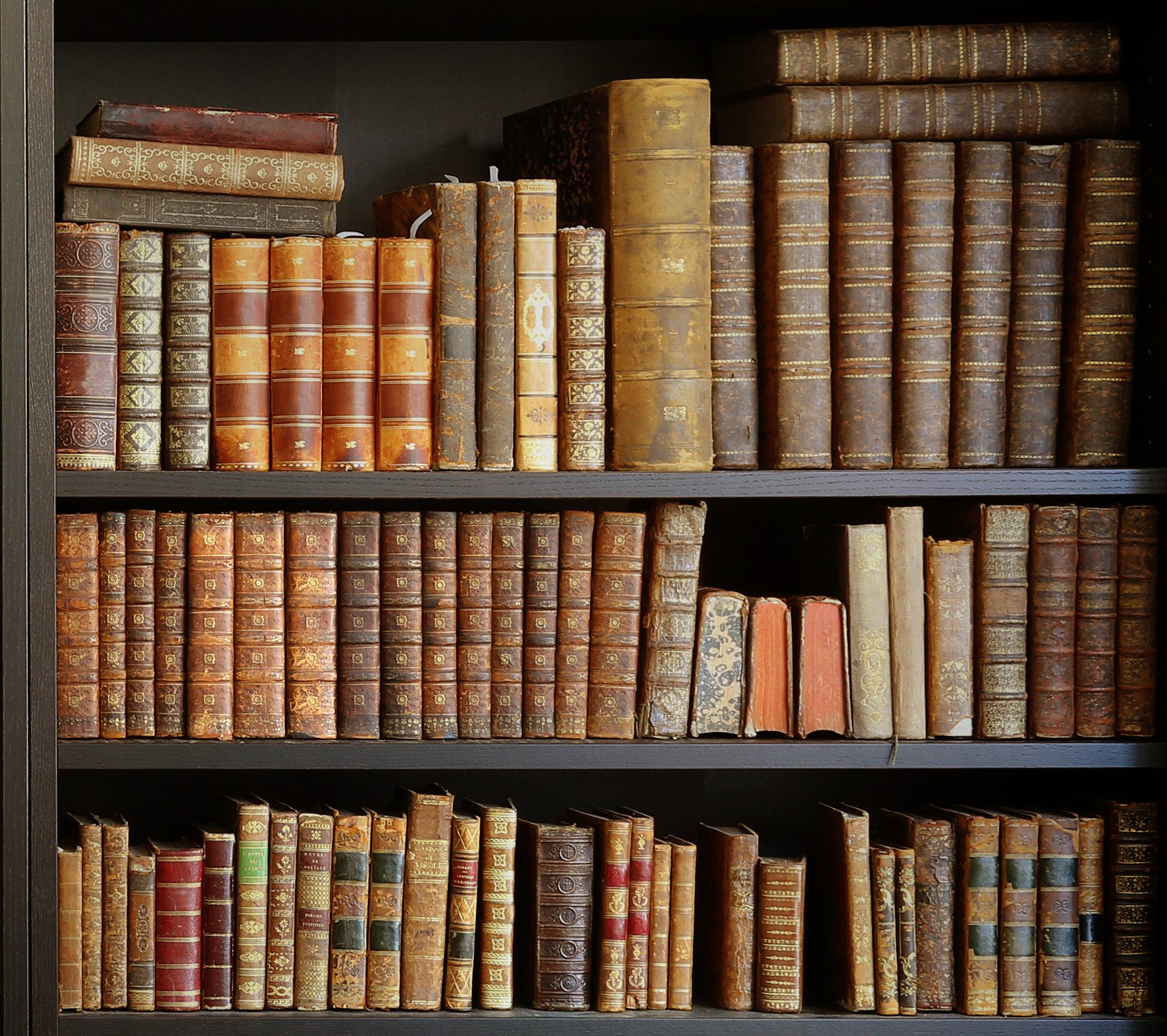}&
\includegraphics[width=\imwidth]{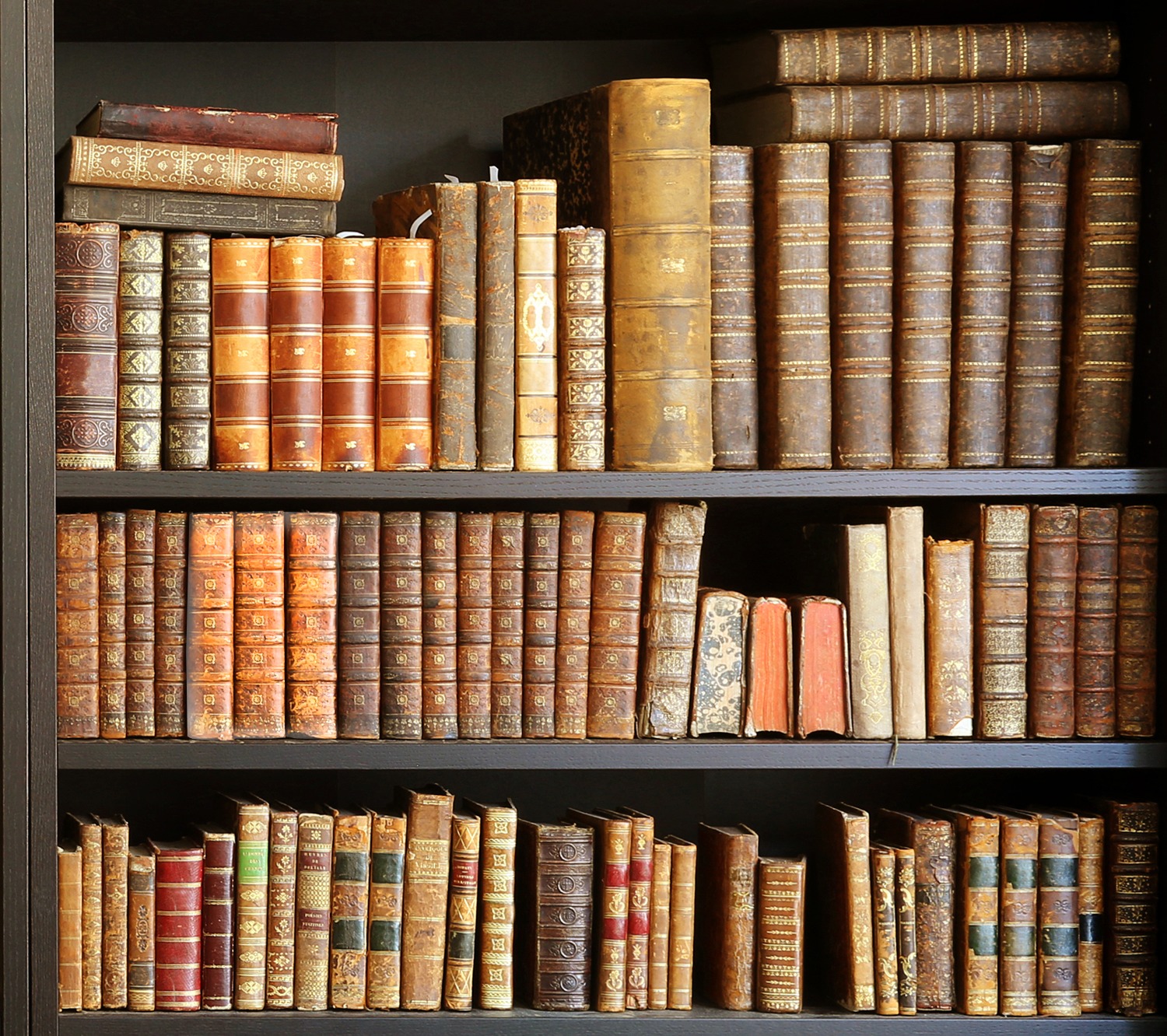}\\

 \end{tabular}
\caption{Sampling different latent $z_i$ in our model results in stochastic variations, all of which achieve the same saliency objective, but with different `styles' of edits. We include two sample edits done by professional users from our exploratory studies.}
\label{fig:MultiStyle}
\end{figure*}

\textbf{Stochastic parameter generation:} There isn't a single way to enhance an object in an image, and different users may choose to apply different sets of transformations when editing an image (two far right columns in Fig.~\ref{fig:MultiStyle}). Motivated by such natural variability, GazeShiftNet can be extended to produce results in `multiple styles' by predicting different parameter values for the transformations.
We introduce stochasticity using a latent vector $z \in \mathbb{R}^{10}$, randomly sampled from a Gaussian distribution. This vector $z$ is first tiled to match the mask dimensions, and then concatenated as an input, similarly to how the mask is handled. In fact, the architecture is the same as in Fig.~\ref{fig:graph}, but replacing $\m$ with $\m \mathbf{z}$ (where $\mathbf{mz}$ is the concatenation of $\m$ and tiled $z$: $\z \in \mathbf{R}^{h \times w \times 10}$). To force the network to actually use $\z$ in producing the output parameters $\p_f$ and $\p_b$, 
we add an additional loss encouraging $z$ to be reconstructed from $\p$ (the concatenation of $\p_f$ and $\p_b$): $\calL_r(z, \p_f, \p_b) = \frac{1}{10}\|z- \text{ENC}(\p) \|_1$, where $\text{ENC}$ is an encoder formed by a series of fully connected layers. We can then randomly sample the latent vector $z$ to produce diverse edits (Fig.~\ref{fig:MultiStyle}).

\textbf{Decreasing human attention:} While GazeShiftNet has been trained to shift viewer attention \emph{towards} a specific region in the image, it can also be trained to achieve the opposite: shifting attention \emph{away} from an image region. This can be useful for distractor attenuation~\cite{fried2015distractors,kolkin2017spatially}, e.g., reducing the prominence of passerbys in personal photo collections. Towards this goal, we extend our network to have two additional heads to output the parameters $\p_f^{dec}$ and $\p_b^{dec}$. We then generate two images, $I'_{dec}$ using  $\p_f^{dec}$ and $\p_b^{dec}$ and $I'_{inc}$ using $\p_f^{inc}$ and $\p_b^{inc}$. The new attention loss (Eq.~\ref{eq:sloss}) becomes: $\calL'_{att}(S_{I'_{inc}}, S_{I'_{dec}}, \m) = \calL_{att}(S_{I'_{inc}}, \m) - \calL_{att}(S_{I'_{dec}}, \m)$. 
To successfully train this model, we had to discard from the training set all masks where the average computational saliency was too small. Such masks were not suitable for the objective of decreasing saliency. In addition, we trained this model for double the amount of time, with hyper-parameter $\lambda_s=2\times10^4$. Results from this network are visualized in Fig.~\ref{fig:incdec}.

\newcommand\wc{0.8}

\begin{figure*}
\centering
\setlength{\imwidth}{0.24\linewidth}
\setbox1=\hbox{\includegraphics[width=\wc\imwidth]{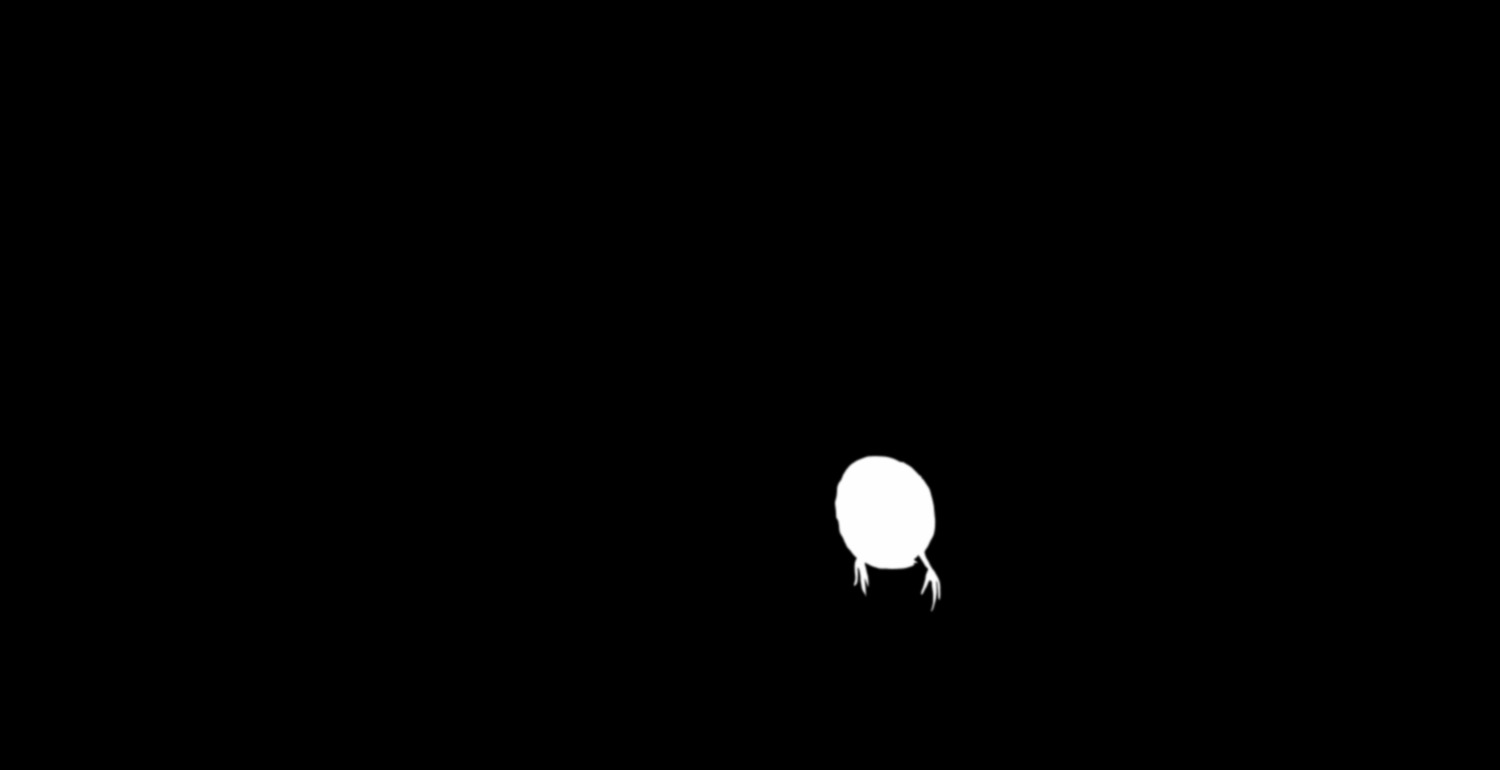}}
\setlength{\tabcolsep}{1pt}
\begin{tabular}{ccccc}

Input & $\uparrow$ Attention & Saliency Map & $\downarrow$ Attention & Saliency Map
\\
\includegraphics[width=\wc\imwidth]{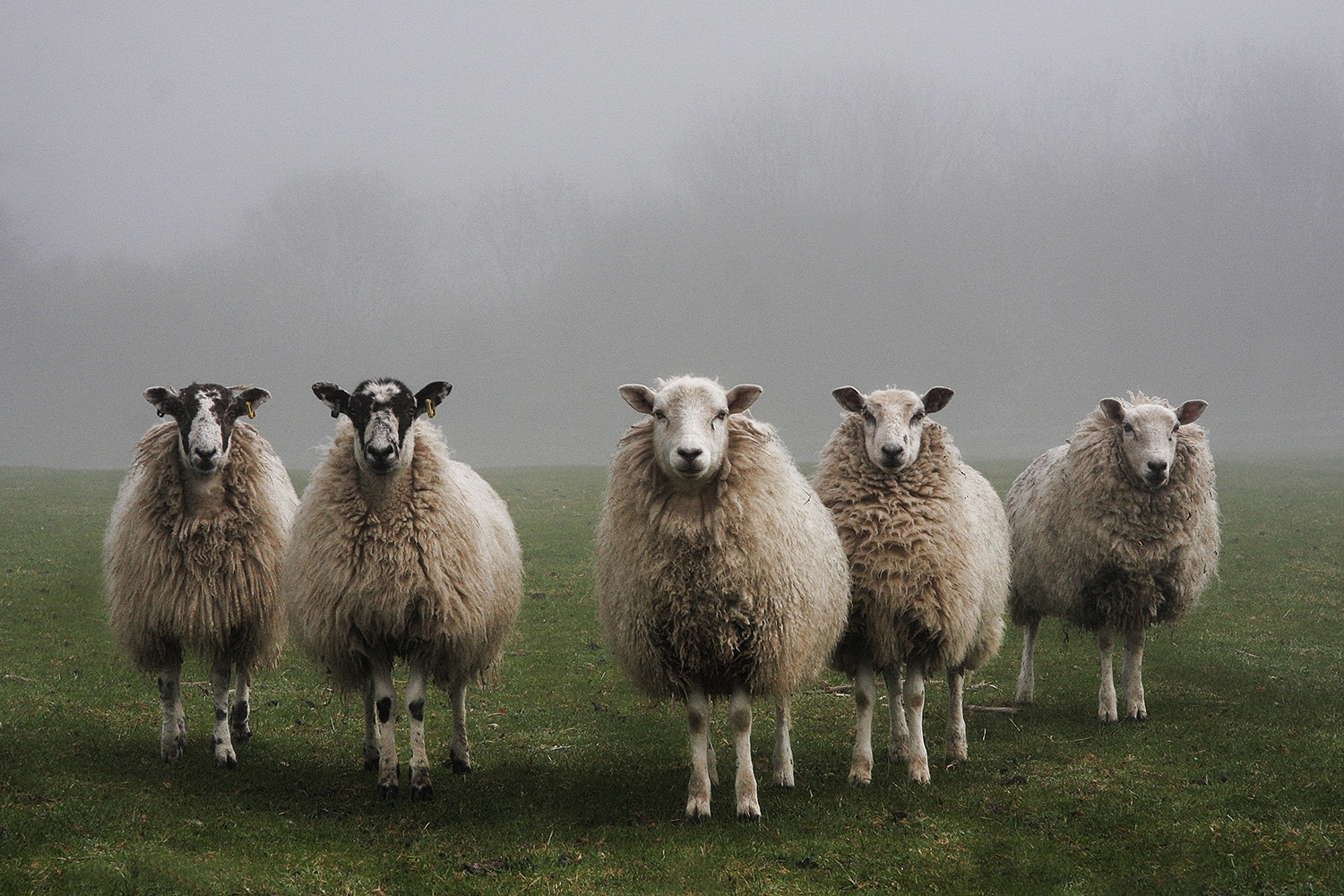}\llap{{\raisebox{0.33\imwidth}{\includegraphics[cfbox=red, width=0.3\imwidth]{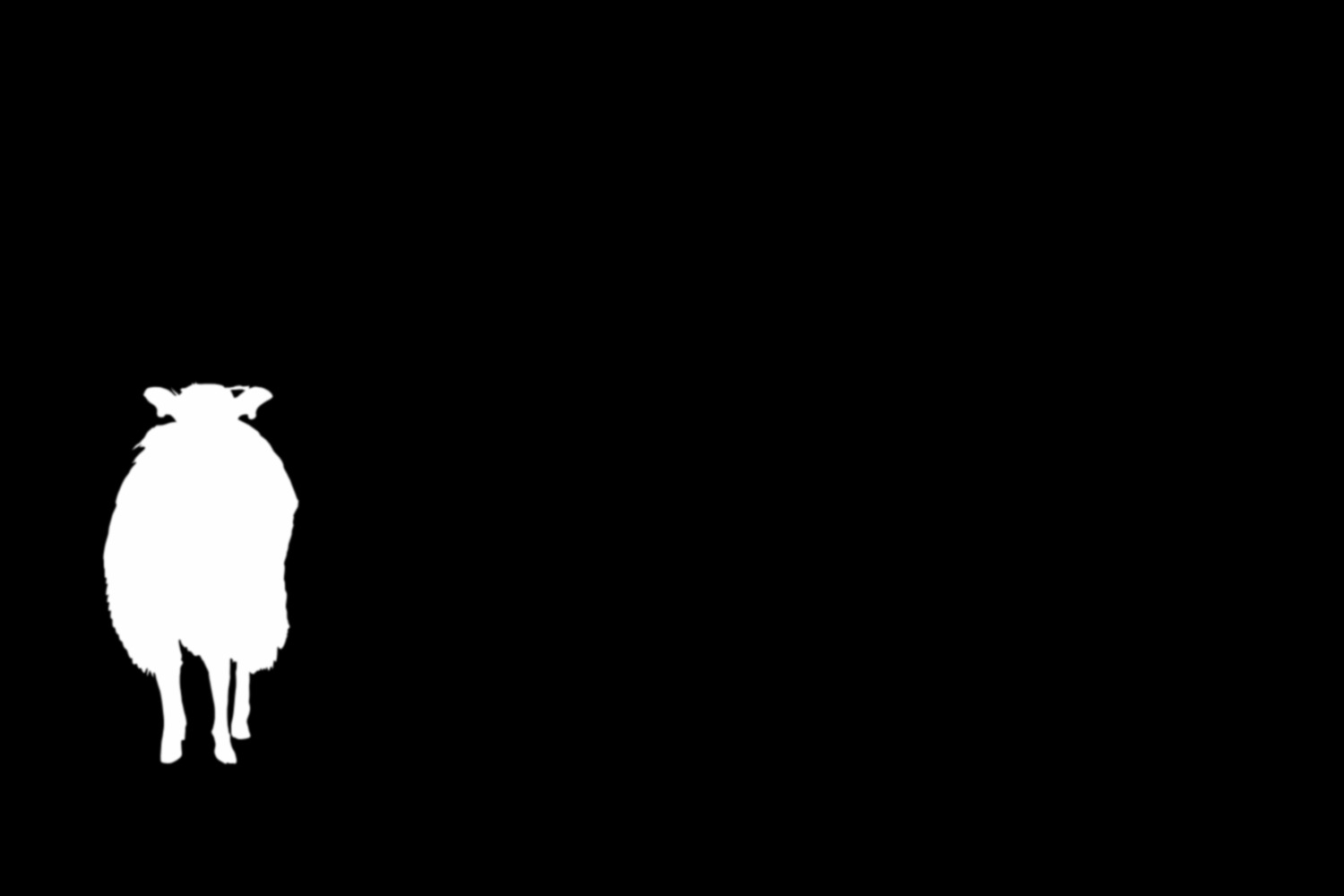}}}}&
\includegraphics[width=\wc\imwidth]{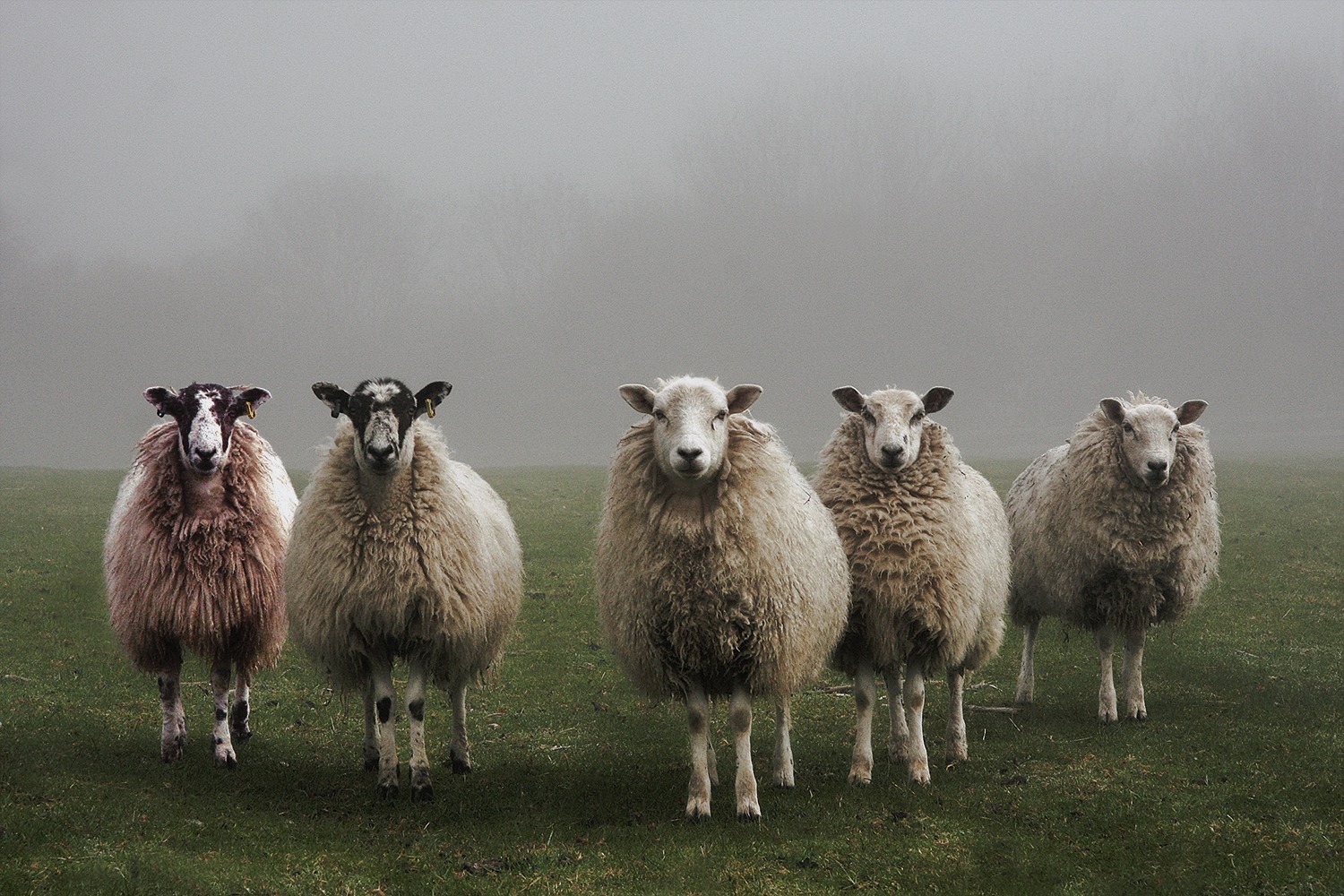} &
\includegraphics[width=\wc\imwidth, height = 1.57cm]{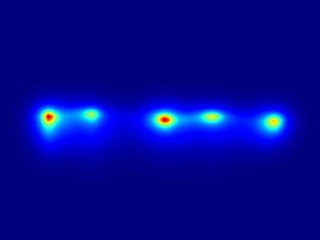} &
\includegraphics[width=\wc\imwidth]{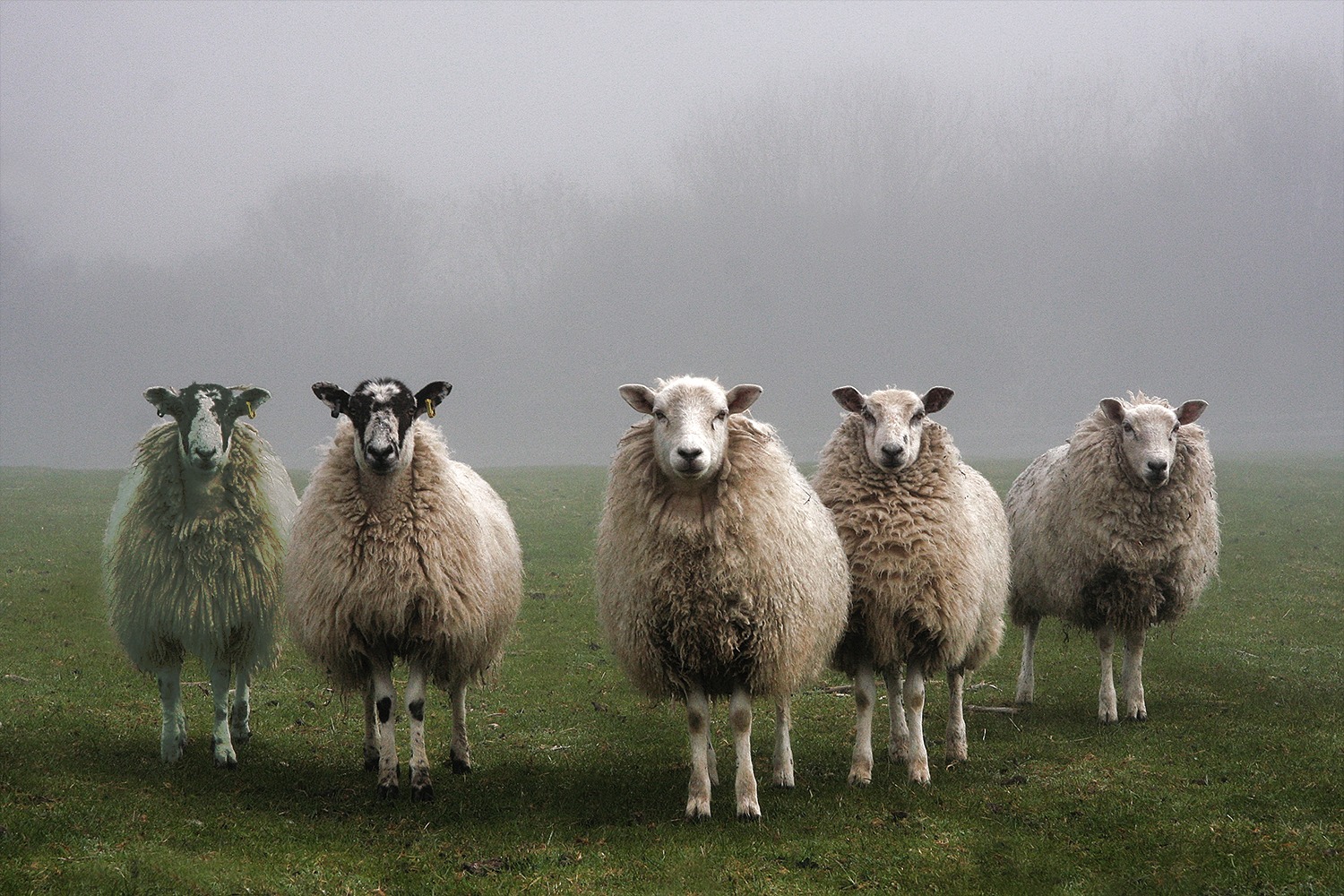}&
\includegraphics[width=\wc\imwidth, height = 1.57cm]{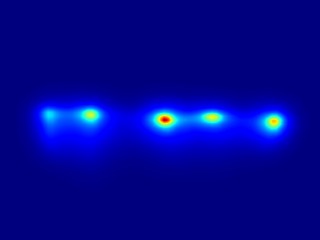}
\\
\includegraphics[width=\wc\imwidth]{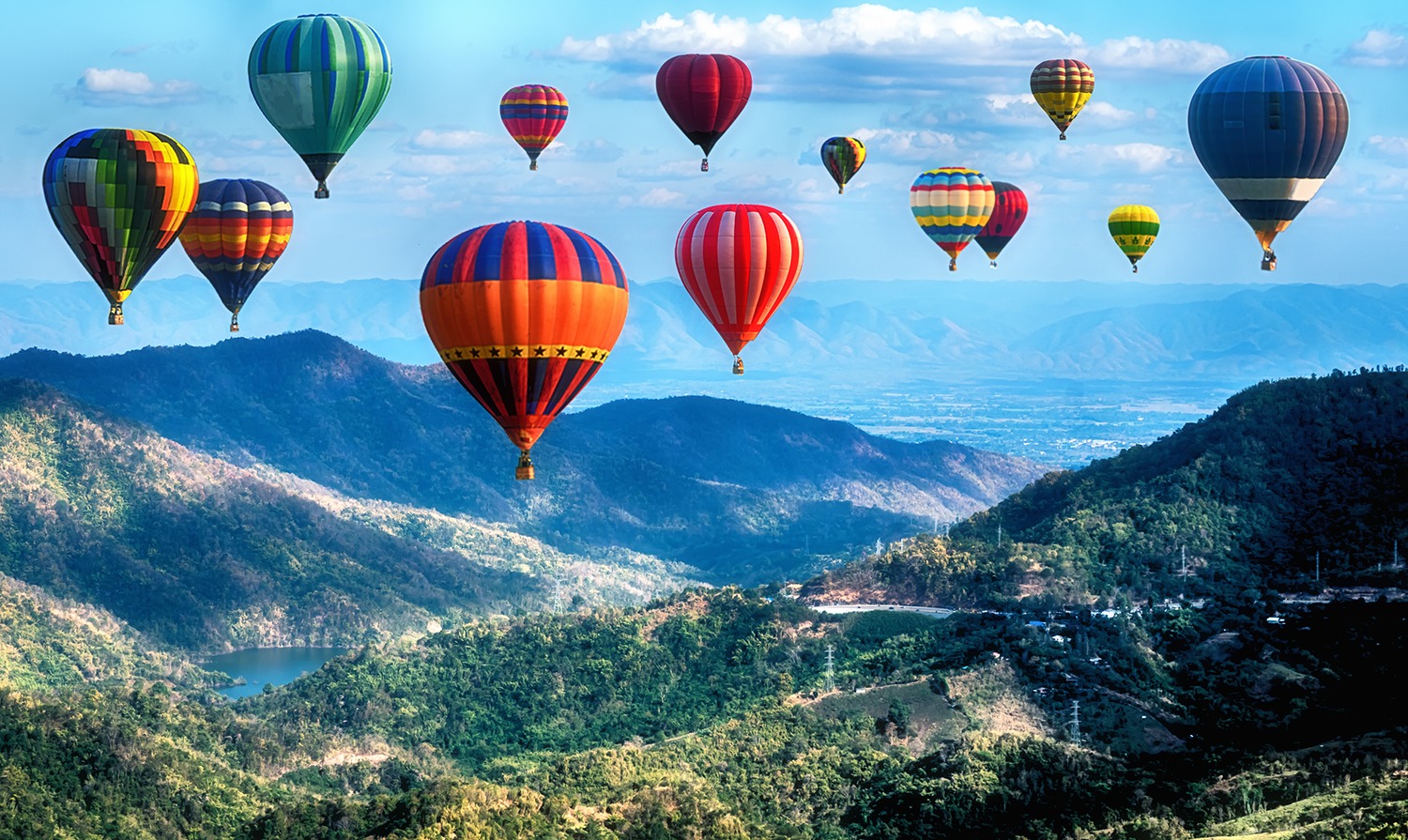}\llap{{\raisebox{0\imwidth}{\includegraphics[cfbox=red, width=0.3\imwidth]{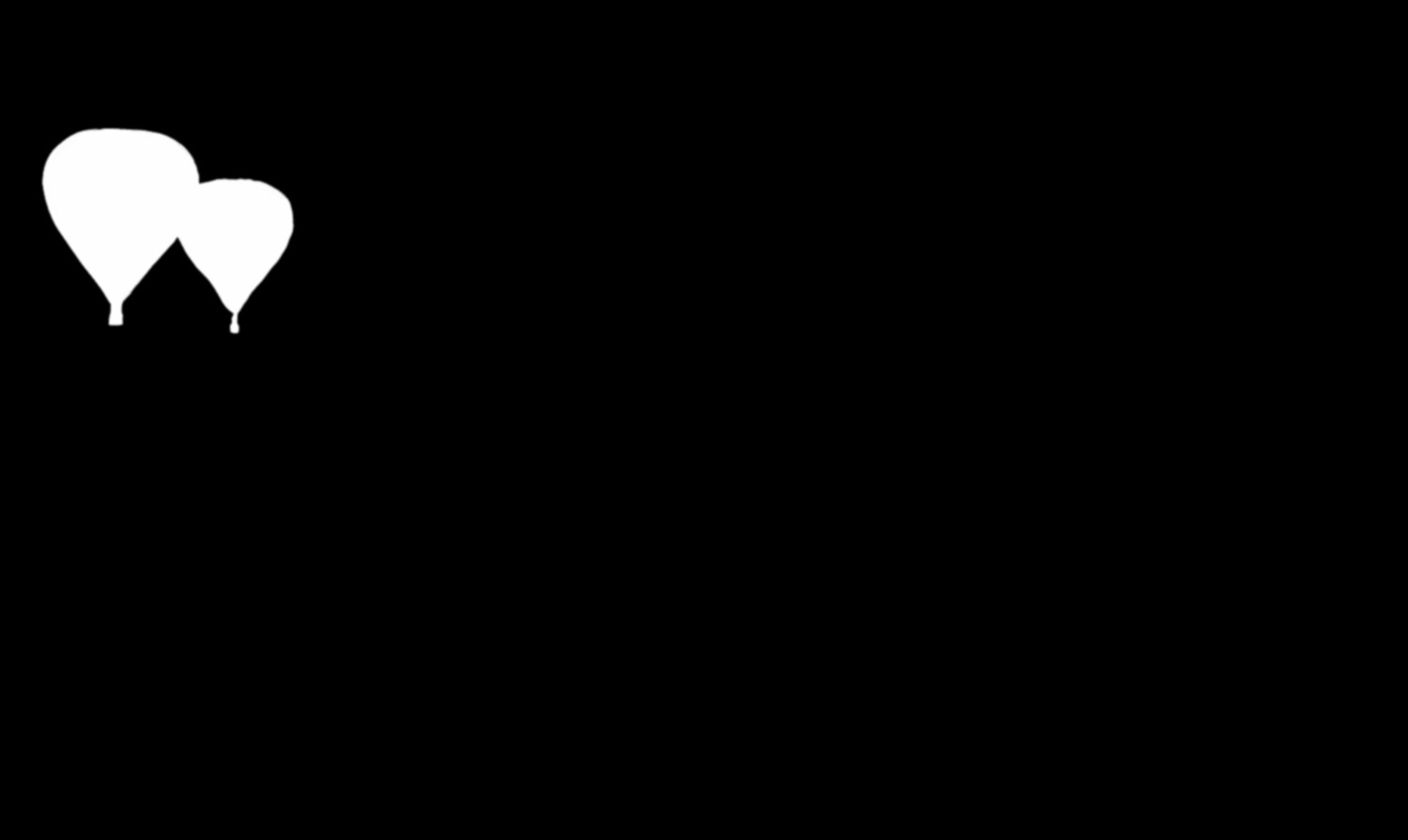}}}}&
\includegraphics[width=\wc\imwidth]{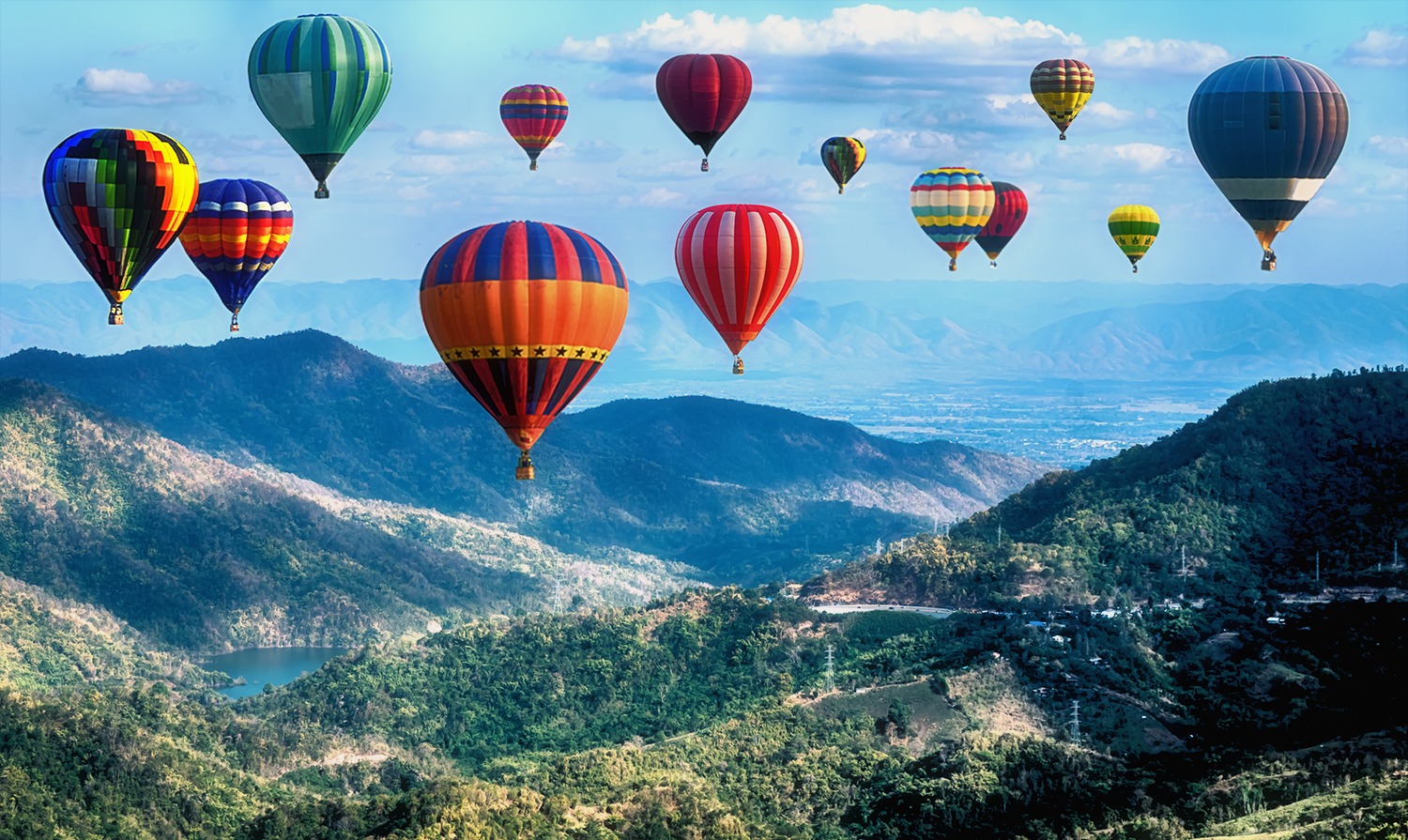} &
\includegraphics[width=\wc\imwidth, height = 1.395cm]{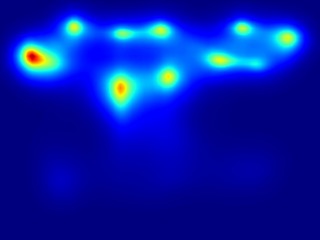} &
\includegraphics[width=\wc\imwidth]{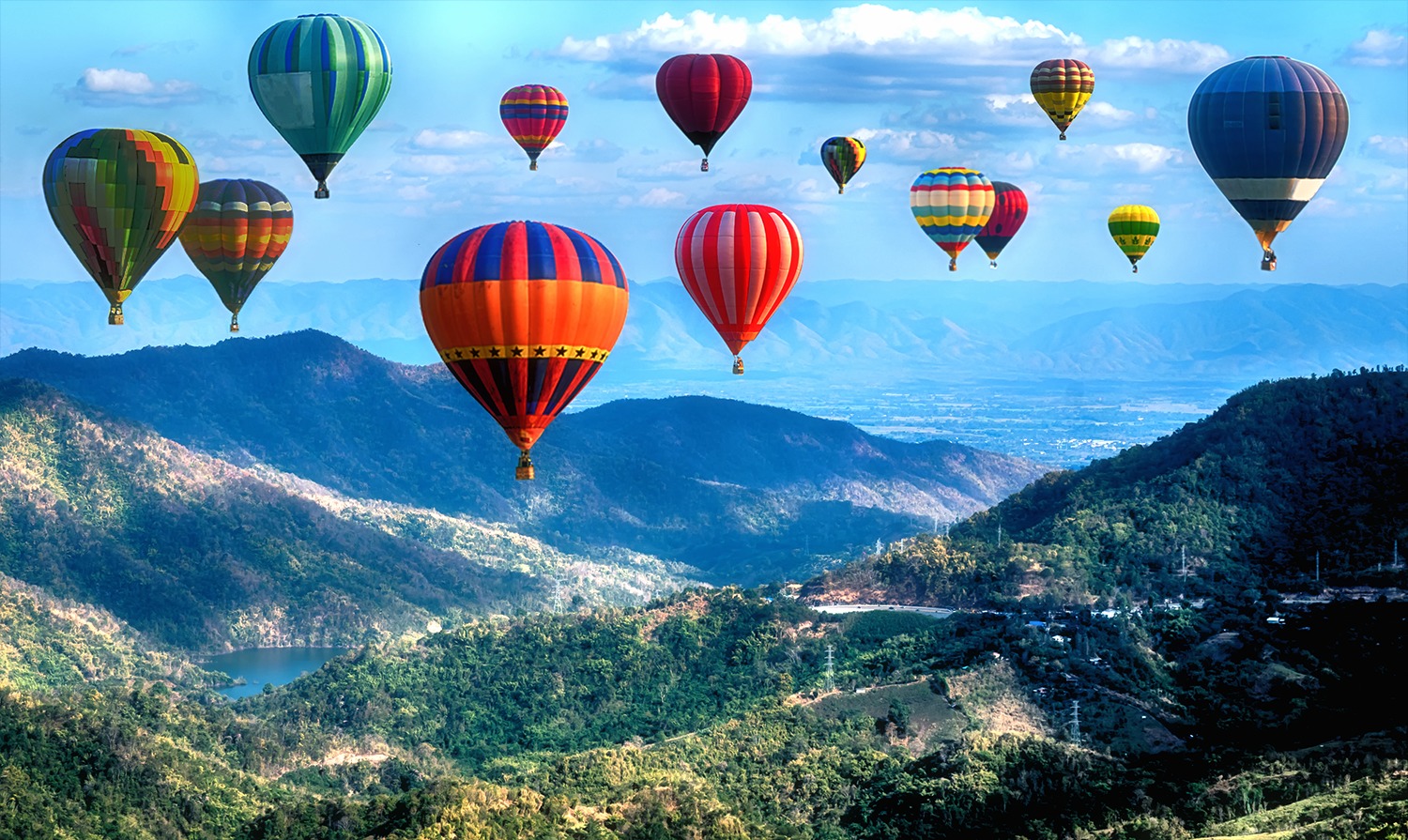}&
\includegraphics[width=\wc\imwidth, height = 1.395cm]{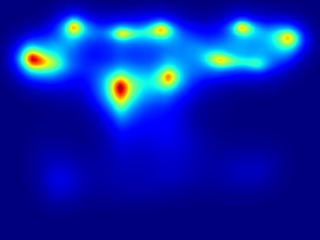}
\\

 \end{tabular}
\caption{Using the input image and corresponding mask, we generate two images, one to shift visual attention towards the mask (col. 2) as seen from the saliency map in col. 3, and one to shift attention away from the mask (col. 4, saliency map in col. 5).}
\label{fig:incdec}
\end{figure*}%

\section{Conclusion}

We presented a practical method for automatically editing images (with extensions to videos) in a subtle way, while effectively redirecting viewer attention. We demonstrated that our method achieves a good balance between shifting the attention - of saliency models and human participants alike - while maintaining the realism and fidelity of the original image (i.e., the photographer's intent, semantics/colors of the original image). 
Having a practical method depends on balancing these objectives, and achieving consistent and robust results across a variety of images. We also showed significant improvements in computation time over past approaches. Most importantly, our global parametric approach allows running our method on high-resolution images and videos at interactive rates, as well as interpolating in parameter space to hand control of the edits back to a user within an editing interface.
Such an effective and practical approach to image editing can benefit many applications, including website design, effective marketing campaigns, and image enhancement.
These tasks are typically completed by professionals using advanced image editing software. Our automated approach can simplify, and has the potential to replace, some professional image editing workflows.

\textbf{Acknowledgements:} Y.~A.~Mejjati thanks the Marie Sklodowska-Curie grant No 665992, and the Centre for Doctoral Training in Digital Entertainment (CDE), EP/L016540/1. K.~I.~Kim thanks Institute of Information \& communications Technology Planning  Evaluation (IITP) grant (No.20200013360011001, Artificial Intelligence Graduate School support (UNIST)) funded by the Korea government (MSIT).
\clearpage
%
%
\bibliographystyle{splncs04}
\bibliography{egbib}

\begin{thebibliography}{10}
\providecommand{\url}[1]{\texttt{#1}}
\providecommand{\urlprefix}{URL }
\providecommand{\doi}[1]{https://doi.org/#1}

\bibitem{achanta2009saliency}
Achanta, R., S{\"u}sstrunk, S.: Saliency detection for content-aware image
  resizing. In: ICIP (2009)

\bibitem{barnes2009patchmatch}
Barnes, C., Shechtman, E., Finkelstein, A., Goldman, D.B.: Patchmatch: A
  randomized correspondence algorithm for structural image editing. In: TOG
  (2009)

\bibitem{bianco2019learning}
Bianco, S., Cusano, C., Piccoli, F., Schettini, R.: Learning parametric
  functions for color image enhancement. In: International Workshop on
  Computational Color Imaging (2019)

\bibitem{borji2019saliency}
Borji, A.: Saliency prediction in the deep learning era: Successes and
  limitations. TPAMI  (2019)

\bibitem{bylinskii2017learning}
Bylinskii, Z., Kim, N.W., O'Donovan, P., Alsheikh, S., Madan, S., Pfister, H.,
  Durand, F., Russell, B., Hertzmann, A.: Learning visual importance for
  graphic designs and data visualizations. In: UIST (2017)

\bibitem{bylinskii2016should}
Bylinskii, Z., Recasens, A., Borji, A., Oliva, A., Torralba, A., Durand, F.:
  Where should saliency models look next? In: ECCV (2016)

\bibitem{chandakkar2016structured}
Chandakkar, P.S., Li, B.: A structured approach to predicting image enhancement
  parameters. In: WACV (2016)

\bibitem{chen1993view}
Chen, S.E., Williams, L.: View interpolation for image synthesis. In: SIGGRAPH
  (1993)

\bibitem{chenguide}
Chen, Y.C., Chang, K.J., Tsai, Y.H., Wang, Y.C.F., Chiu, W.C.: Guide your eyes:
  Learning image manipulation under saliency guidance. In: BMVC (2019)

\bibitem{cornia2019show}
Cornia, M., Baraldi, L., Cucchiara, R.: Show, control and tell: a framework for
  generating controllable and grounded captions. In: CVPR (2019)

\bibitem{cornia2018sam}
Cornia, M., Baraldi, L., Serra, G., Cucchiara, R.: {SAM: Pushing the Limits of
  Saliency Prediction Models}. In: CVPR Workshops (2018)

\bibitem{Fosco_2020_CVPR}
Fosco, C., Newman, A., Sukhum, P., Zhang, Y.B., Zhao, N., Oliva, A., Bylinskii,
  Z.: How much time do you have? modeling multi-duration saliency. In: CVPR
  (2020)

\bibitem{fried2015distractors}
Fried, O., Shechtman, E., Goldman, D.B., Finkelstein, A.: Finding distractors
  in images. In: CVPR (2015)

\bibitem{gatys2016image}
Gatys, L.A., Ecker, A.S., Bethge, M.: Image style transfer using convolutional
  neural networks. In: CVPR (2016)

\bibitem{gatys2017guiding}
Gatys, L.A., K{\"u}mmerer, M., Wallis, T.S., Bethge, M.: Guiding human gaze
  with convolutional neural networks. arXiv preprint arXiv:1712.06492  (2017)

\bibitem{goodfellow2014generative}
Goodfellow, I., Pouget-Abadie, J., Mirza, M., Xu, B., Warde-Farley, D., Ozair,
  S., Courville, A., Bengio, Y.: Generative adversarial nets. In: NeurIPS
  (2014)

\bibitem{hagiwara2011saliency}
Hagiwara, A., Sugimoto, A., Kawamoto, K.: Saliency-based image editing for
  guiding visual attention. In: Proceedings of the 1st international workshop
  on pervasive eye tracking \& mobile eye-based interaction (2011)

\bibitem{hertzmann2001image}
Hertzmann, A., Jacobs, C.E., Oliver, N., Curless, B., Salesin, D.H.: Image
  analogies. In: SIGGRAPH (2001)

\bibitem{hu2018exposure}
Hu, Y., He, H., Xu, C., Wang, B., Lin, S.: Exposure: A white-box photo
  post-processing framework. SIGGRAPH  (2018)

\bibitem{huang2017arbitrary}
Huang, X., Belongie, S.: Arbitrary style transfer in real-time with adaptive
  instance normalization. In: ICCV (2017)

\bibitem{jolicoeur2018relativistic}
Jolicoeur-Martineau, A.: The relativistic discriminator: a key element missing
  from standard gan. In: ICLR (2019)

\bibitem{karras2019style}
Karras, T., Laine, S., Aila, T.: A style-based generator architecture for
  generative adversarial networks. In: CVPR (2019)

\bibitem{kaufman2012content}
Kaufman, L., Lischinski, D., Werman, M.: Content-aware automatic photo
  enhancement. In: Computer Graphics Forum (2012)

\bibitem{kolkin2017spatially}
Kolkin, N.I., Shakhnarovich, G., Shechtman, E.: Training deep networks to be
  spatially sensitive. In: ICCV (2017)

\bibitem{koutras2019susinet}
Koutras, P., Maragos, P.: Susinet: See, understand and summarize it. In: CVPR
  Workshops (2019)

\bibitem{kummerer2016deepgaze}
K{\"u}mmerer, M., Wallis, T.S., Bethge, M.: Deepgaze ii: Reading fixations from
  deep features trained on object recognition. arXiv preprint arXiv:1610.01563
  (2016)

\bibitem{li2016objects}
Li, N., Zhao, X., Yang, Y., Zou, X.: Objects classification by learning-based
  visual saliency model and convolutional neural network. Computational
  intelligence and neuroscience  (2016)

\bibitem{lin2014microsoft}
Lin, T.Y., Maire, M., Belongie, S., Hays, J., Perona, P., Ramanan, D.,
  Doll{\'a}r, P., Zitnick, C.L.: Microsoft coco: Common objects in context. In:
  ECCV (2014)

\bibitem{margolin2014evaluate}
Margolin, R., Zelnik-Manor, L., Tal, A.: How to evaluate foreground maps? In:
  CVPR (2014)

\bibitem{mateescu2014attention}
Mateescu, V.A., Baji{\'c}, I.V.: Attention retargeting by color manipulation in
  images. In: Proceedings of the 1st International Workshop on Perception
  Inspired Video Processing (2014)

\bibitem{mathe2012dynamic}
Mathe, S., Sminchisescu, C.: Dynamic eye movement datasets and learnt saliency
  models for visual action recognition. In: ECCV (2012)

\bibitem{mechrez2019saliency}
Mechrez, R., Shechtman, E., Zelnik-Manor, L.: Saliency driven image
  manipulation. WACV  (2019)

\bibitem{mejjati2018unsupervised}
Mejjati, Y.A., Richardt, C., Tompkin, J., Cosker, D., Kim, K.I.: Unsupervised
  attention-guided image-to-image translation. In: NeurIPS (2018)

\bibitem{miyato2018spectral}
Miyato, T., Kataoka, T., Koyama, M., Yoshida, Y.: Spectral normalization for
  generative adversarial networks. In: ICLR (2018)

\bibitem{moosmann2006learning}
Moosmann, F., Larlus, D., Jurie, F.: Learning saliency maps for object
  categorization. In: ECCV (2006)

\bibitem{newman2020turkeyes}
Newman, A., McNamara, B., Fosco, C., Zhang, Y.B., Sukhum, P., Tancik, M., Kim,
  N.W., Bylinskii, Z.: Turkeyes: A web-based toolbox for crowdsourcing
  attention data. In: ACM CHI Conference on Human Factors in Computing Systems
  (2020)

\bibitem{park2019gaugan}
Park, T., Liu, M.Y., Wang, T.C., Zhu, J.Y.: Gaugan: semantic image synthesis
  with spatially adaptive normalization. In: SIGGRAPH (2019)

\bibitem{Perazzi_CVPR_2016}
Perazzi, F., Pont-Tuset, J., McWilliams, B., {Van Gool}, L., Gross, M.,
  Sorkine-Hornung, A.: A benchmark dataset and evaluation methodology for video
  object segmentation. In: CVPR (2016)

\bibitem{perez2003poisson}
P{\'e}rez, P., Gangnet, M., Blake, A.: Poisson image editing. TOG  (2003)

\bibitem{shen2014webpage}
Shen, C., Zhao, Q.: Webpage saliency. In: ECCV (2014)

\bibitem{su2005emphasis}
Su, S.L., Durand, F., Agrawala, M.: De-emphasis of distracting image regions
  using texture power maps. In: ICCV workshops (2005)

\bibitem{tsai2017deep}
Tsai, Y.H., Shen, X., Lin, Z., Sunkavalli, K., Lu, X., Yang, M.H.: Deep image
  harmonization. In: CVPR (2017)

\bibitem{wang2018high}
Wang, T.C., Liu, M.Y., Zhu, J.Y., Tao, A., Kautz, J., Catanzaro, B.:
  High-resolution image synthesis and semantic manipulation with conditional
  gans. In: CVPR (2018)

\bibitem{Wang_2019_CVPR}
Wang, W., Shen, J., Cheng, M.M., Shao, L.: An iterative and cooperative
  top-down and bottom-up inference network for salient object detection. In:
  CVPR (2019)

\bibitem{Wang_2019_cv}
Wang, W., Zhao, S., Shen, J., Hoi, S.C.H., Borji, A.: Salient object detection
  with pyramid attention and salient edges. In: CVPR (2019)

\bibitem{wong2011saliency}
Wong, L.K., Low, K.L.: Saliency retargeting: An approach to enhance image
  aesthetics. In: WACV-workshop (2011)

\bibitem{yan2016automatic}
Yan, Z., Zhang, H., Wang, B., Paris, S., Yu, Y.: Automatic photo adjustment
  using deep neural networks. TOG  (2016)

\bibitem{zhang2018self}
Zhang, H., Goodfellow, I., Metaxas, D., Odena, A.: Self-attention generative
  adversarial networks. ICML  (2019)

\bibitem{zhang2018unreasonable}
Zhang, R., Isola, P., Efros, A.A., Shechtman, E., Wang, O.: The unreasonable
  effectiveness of deep features as a perceptual metric. In: CVPR (2018)

\bibitem{zheng2018task}
Zheng, Q., Jiao, J., Cao, Y., Lau, R.W.: Task-driven webpage saliency. In: ECCV
  (2018)

\bibitem{zhou2020}
{Zhou}, L., {Zhang}, Y., {Jiang}, Y., {Zhang}, T., {Fan}, W.: Re-caption:
  Saliency-enhanced image captioning through two-phase learning. IEEE
  Transactions on Image Processing  (2020)

\bibitem{zhu2017unpaired}
Zhu, J.Y., Park, T., Isola, P., Efros, A.A.: Unpaired image-to-image
  translation using cycle-consistent adversarial networks. In: ICCV (2017)

\bibitem{zund2013content}
Z{\"u}nd, F., Pritch, Y., Sorkine-Hornung, A., Mangold, S., Gross, T.:
  Content-aware compression using saliency-driven image retargeting. In: ICIP
  (2013)

\end{thebibliography}

\clearpage
\appendix
\section*{Appendix}

\section{Parameter definition}
GazeShiftNet implements the following parametric transformations: sharpening, exposure, contrast, tone and color curve adjustment, as defined below. \\

\textbf{Sharpness:} Given an input image $I$ and the predicted sharpness parameter $p_1 \in [-2, 2]$, the output image is obtained by first computing image edges using the Sobel filters $f_1$ and $f_2$ as follows:
$I_{edge} = \sqrt{(I*f_1)^2 + (I*f_2)^2}$, where $*$ is the convolution operation, and $f_1, f_2$ are the filters:
$
 f_1 = \frac{1}{8}\begin{bmatrix} -1& 0 & 1\\ -2& 0 & 2\\ -1& 0 & 1\\ \end{bmatrix}, 
 f_2 = \frac{1}{8}\begin{bmatrix} -1& -2 & -1\\ 0& 0 & 0\\ 1& 2 & 1\\ \end{bmatrix}.
$

Finally, the output image $I'$ is calculated as $I' = I + p_1 I_{edge} I $.\\

\textbf{Exposure:} Given an input image $I$ and the predicted exposure parameter $p_2 \in [-3, 3]$, the output image is computed as $I'=I\exp(p_2*\log(2))$, as in~\cite{hu2018exposure}.\\

\textbf{Contrast:}  Given an input image $I$ and the predicted contrast parameter $p_3 \in [-1, 1]$, the output image is obtained by the linear interpolation: $I'=(1-p_3)I + p_3 I'' $, where $I'' = I \frac{\frac{1}{2} (1-\cos(\pi I_{lum}))}{I_{lum}}$, and $I_{lum} = = 0.27R + 0.67G + 0.06B$, as in~\cite{hu2018exposure}.\\

\textbf{Tone and Color adjustment:}
We define color and tone adjustment using monotonic and piece-wise curve representations, in the same way as \cite{hu2018exposure}. 
The curve is represented using $L$ different parameters, e.g. for tone adjustment $\mathbf{p}^t = \{p^t_0, p^t_1, \dots, p^t_L\}$. In this case, $I' = \frac{1}{\sum_{i=0}^L p^t_i}\sum_{i=0}^{L-1}\text{clip}(L.I - i, 0, 1) p^t_i$.

For tone adjustment, we define the same set of $L$ parameters ($\in [0, 3]$) for R, G, B. While, for color adjustment three distinct sets of $L$ parameters ($\in [0, 3]$) are defined for R and G and B. We set $L=8$ for all our experiments. \\

In addition to the parameters defined above we experimented with additional parametric transformations, including white balance, blur, saturation, and gamma correction. For white balance and gamma correction, we found that similar effects could be reproduced by a combination of our other simpler parameters. We decided to focus on a smaller but sufficient subset, to avoid unnecessarily bulking up our model. Regarding saturation, we found that it can have destructive artifacts, as discussed in our ablation study below. We also tried using Gaussian blur, but this parametric transformation was not used by our model. We believe this is due to the lack of examples in our training set containing blur, and thus blur being penalized by the discriminator. One solution would be to collect a large dataset with large defocus blur or add artificial blur to the background of our training set in a realistic manner. Both these solutions can be explored as an interesting future work direction.

\section{User studies}

\subsection{Exploratory study: professional edits}

\begin{figure*}
\centering
\setlength{\imwidth}{0.16\linewidth}
\setlength{\tabcolsep}{1pt}
\begin{tabular}{cccccc}
Input & Mask & user 1 & user 2 & user 3 & user 4 
\\
 \includegraphics[width=\imwidth]{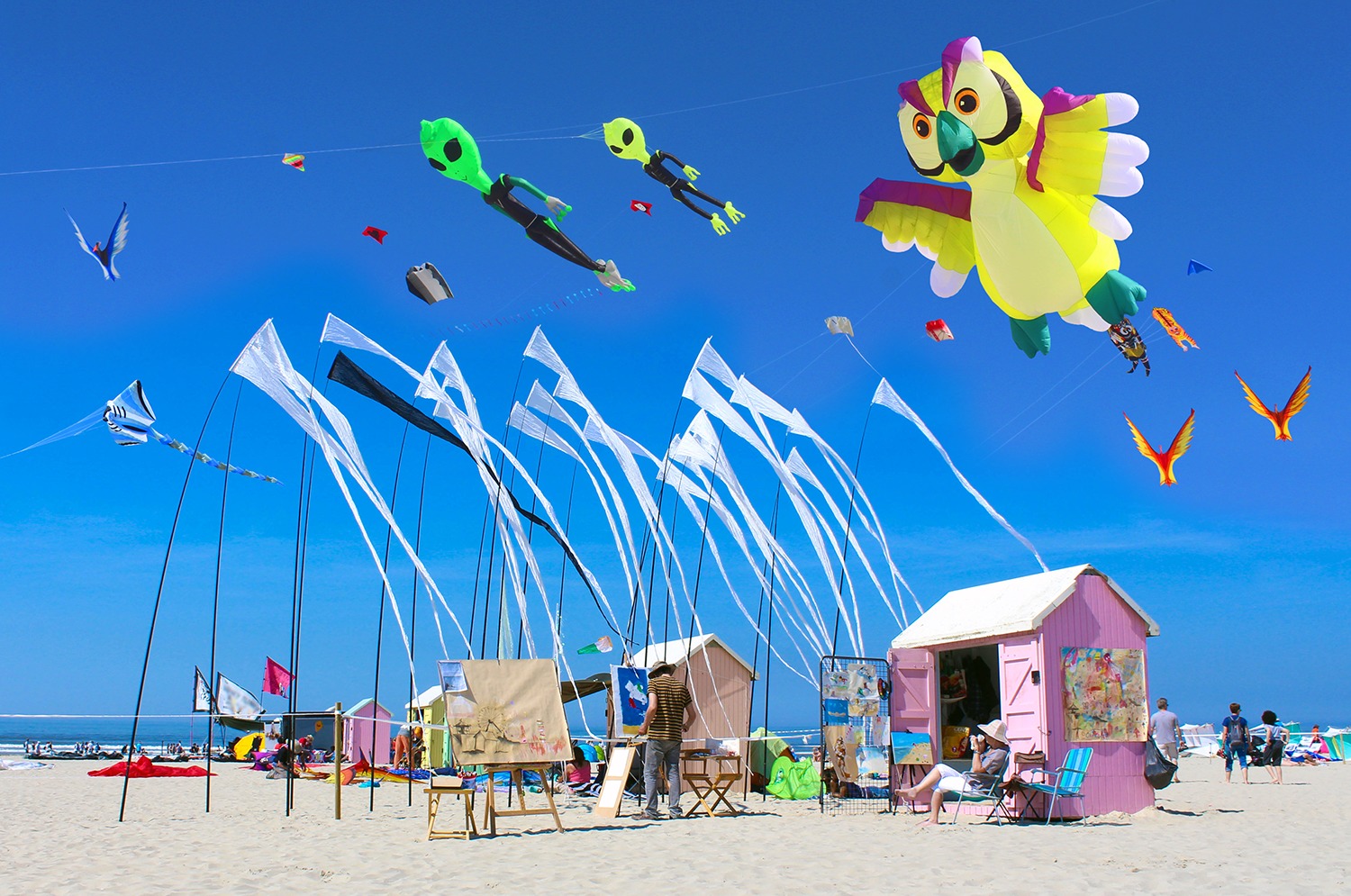}&
  \includegraphics[width=\imwidth]{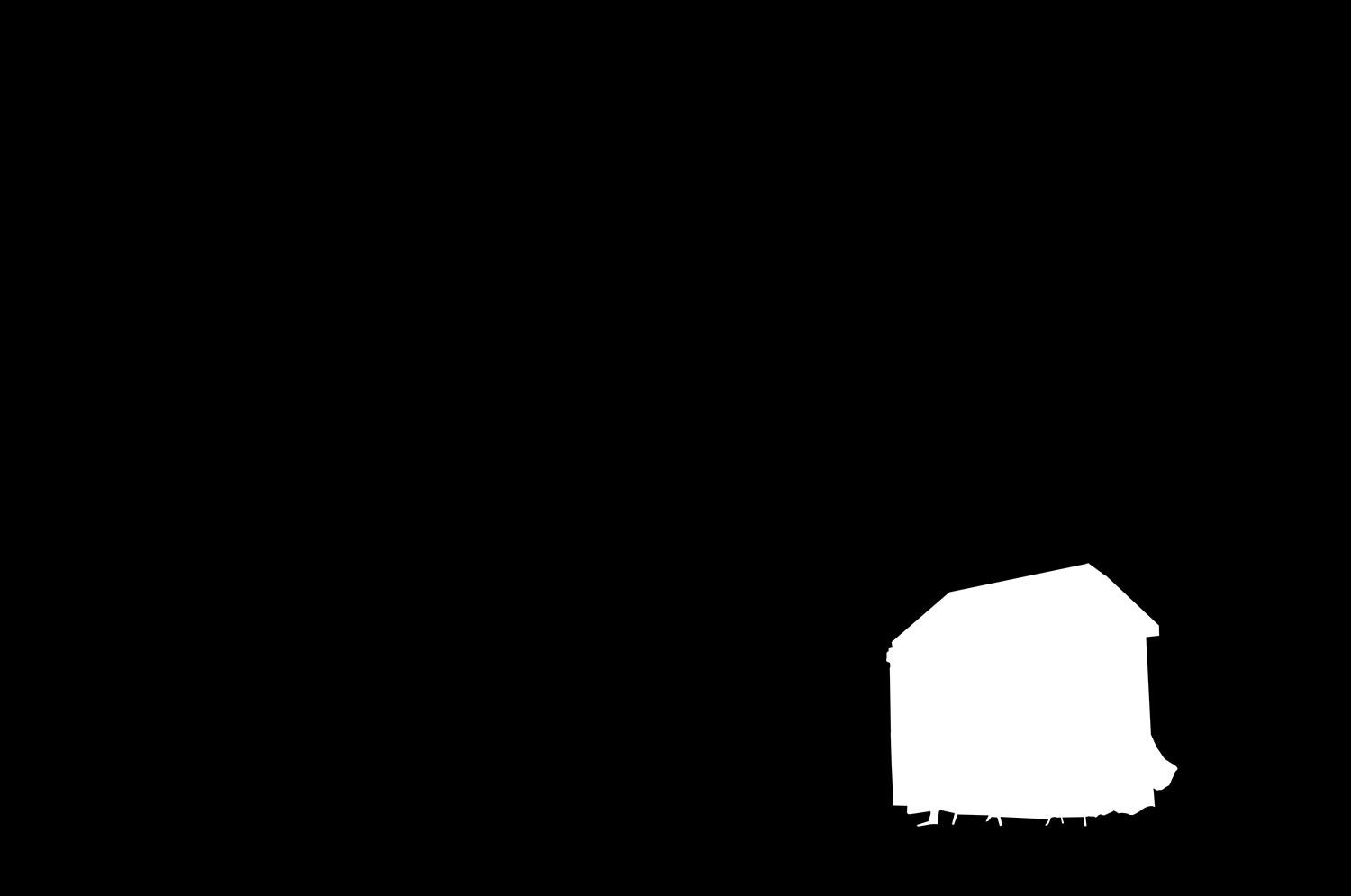}&
   \includegraphics[width=\imwidth]{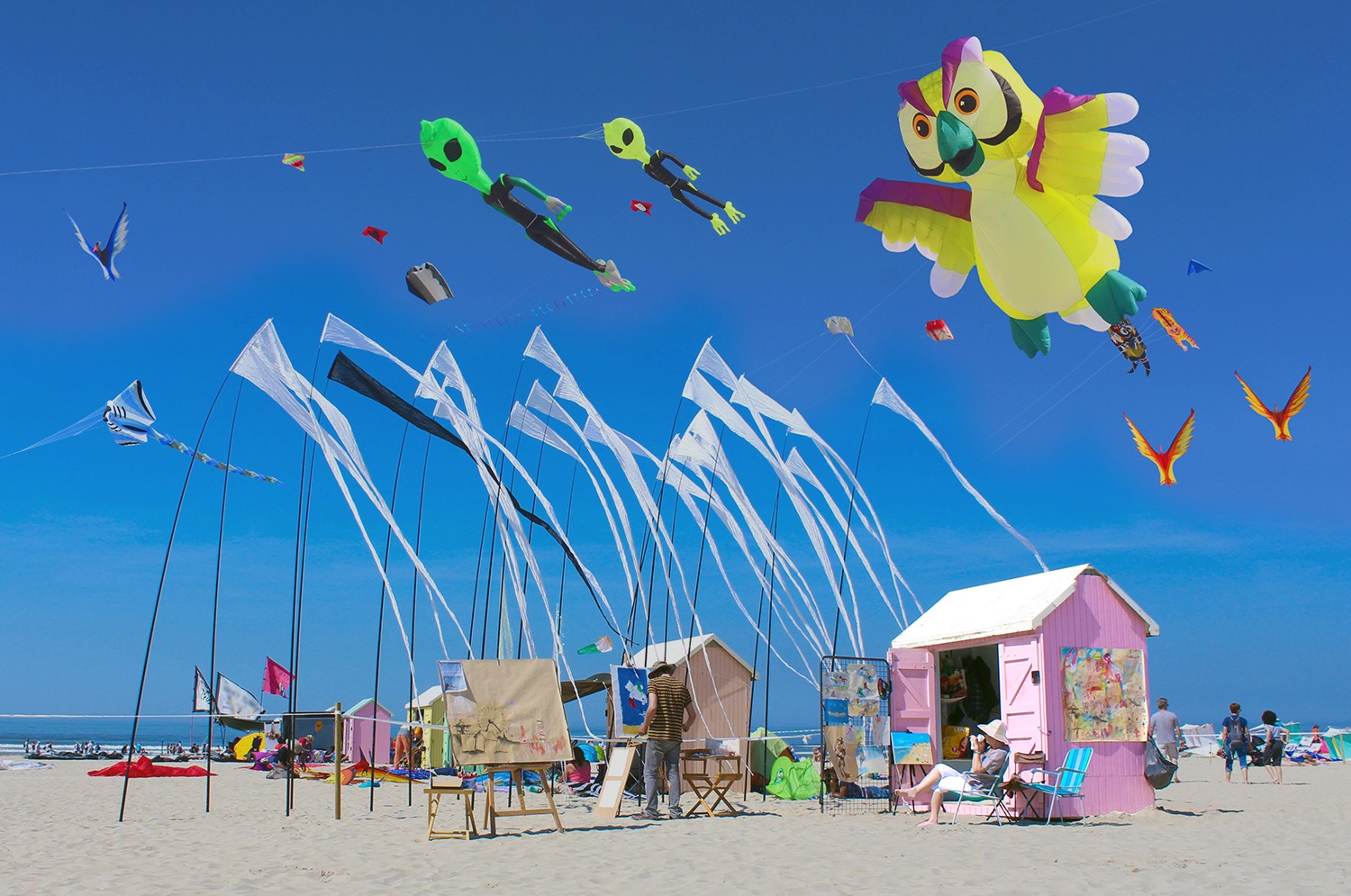}&
    \includegraphics[width=\imwidth]{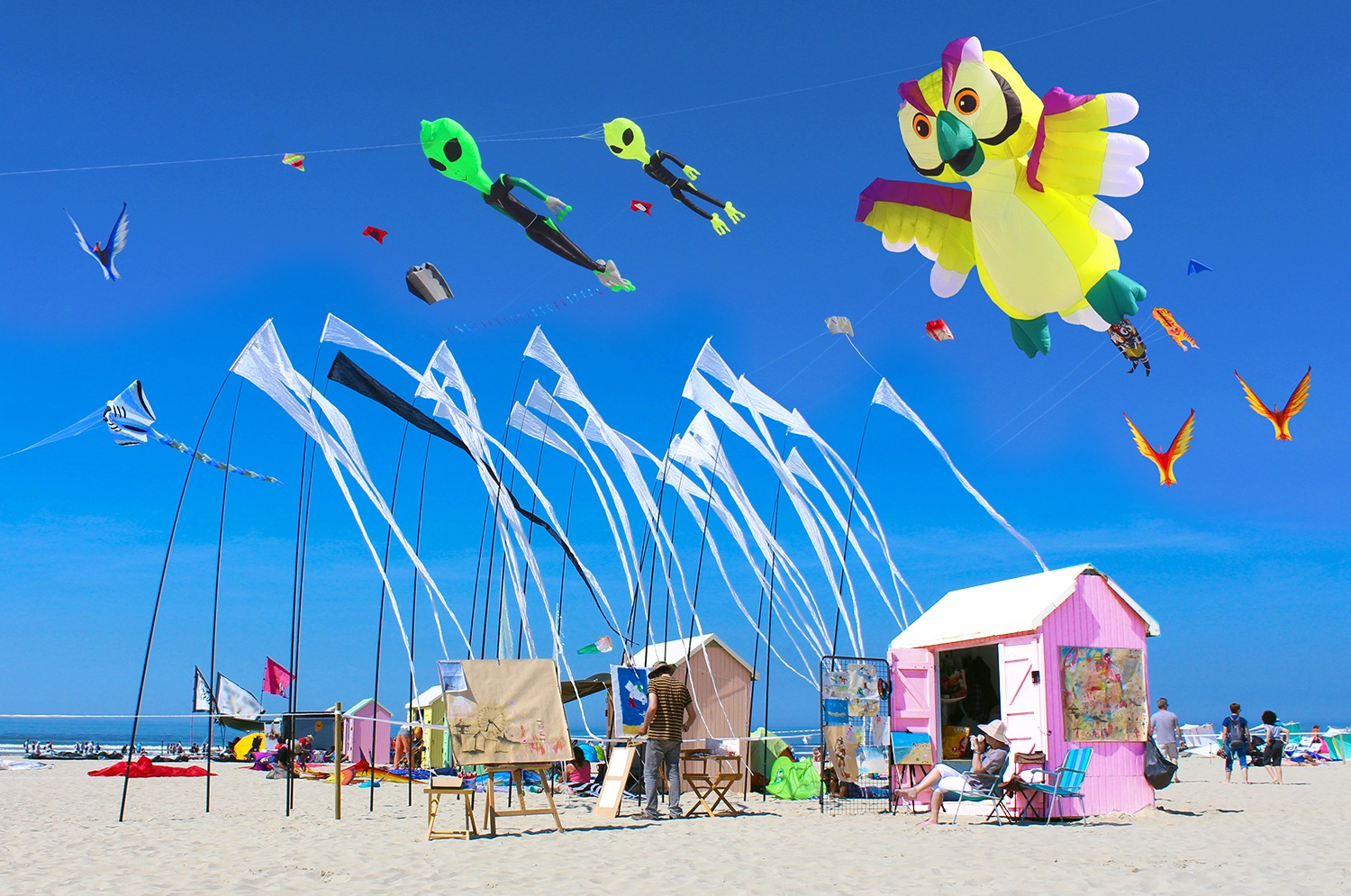}&
     \includegraphics[width=\imwidth]{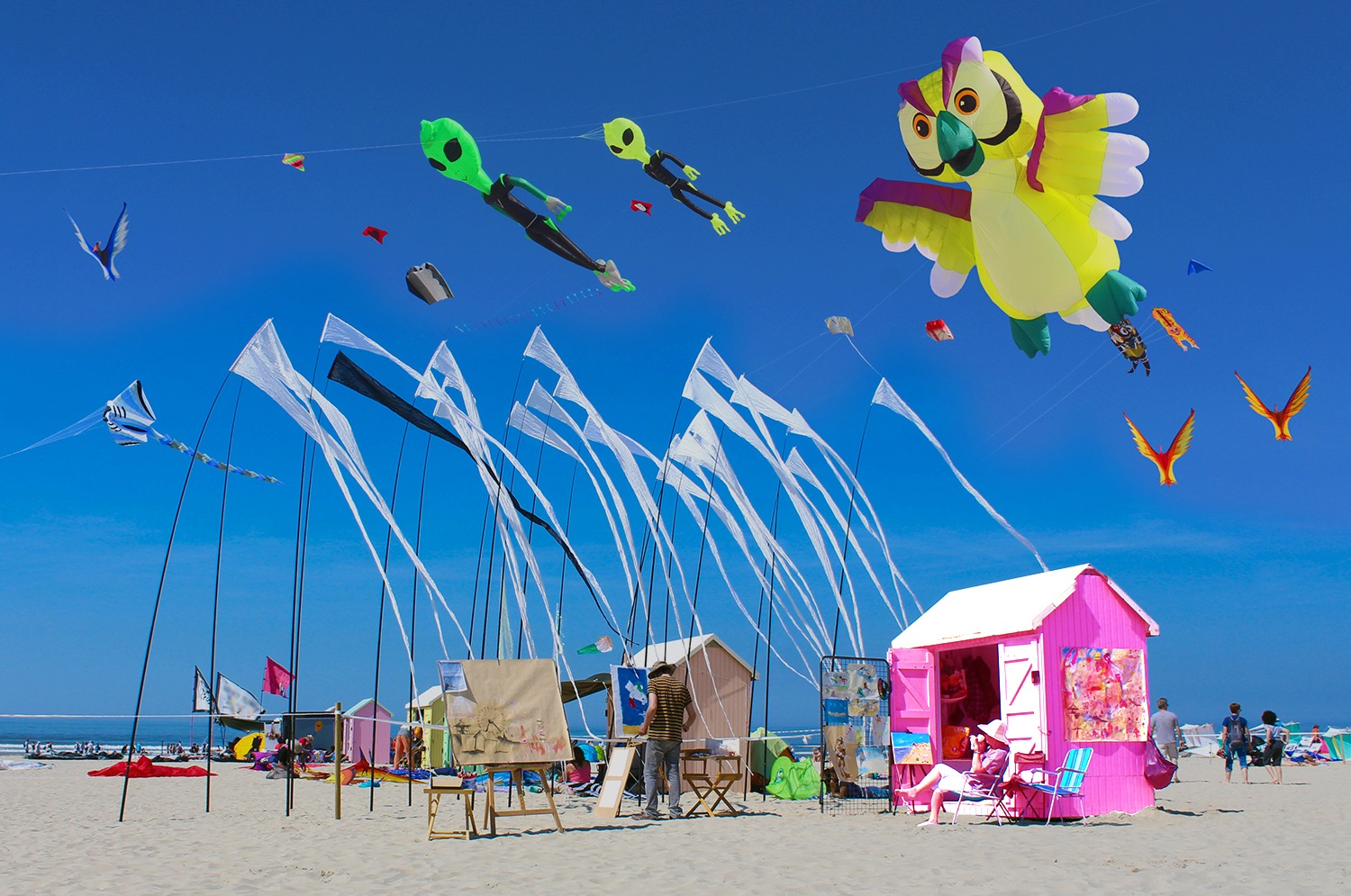}&
     \includegraphics[width=\imwidth]{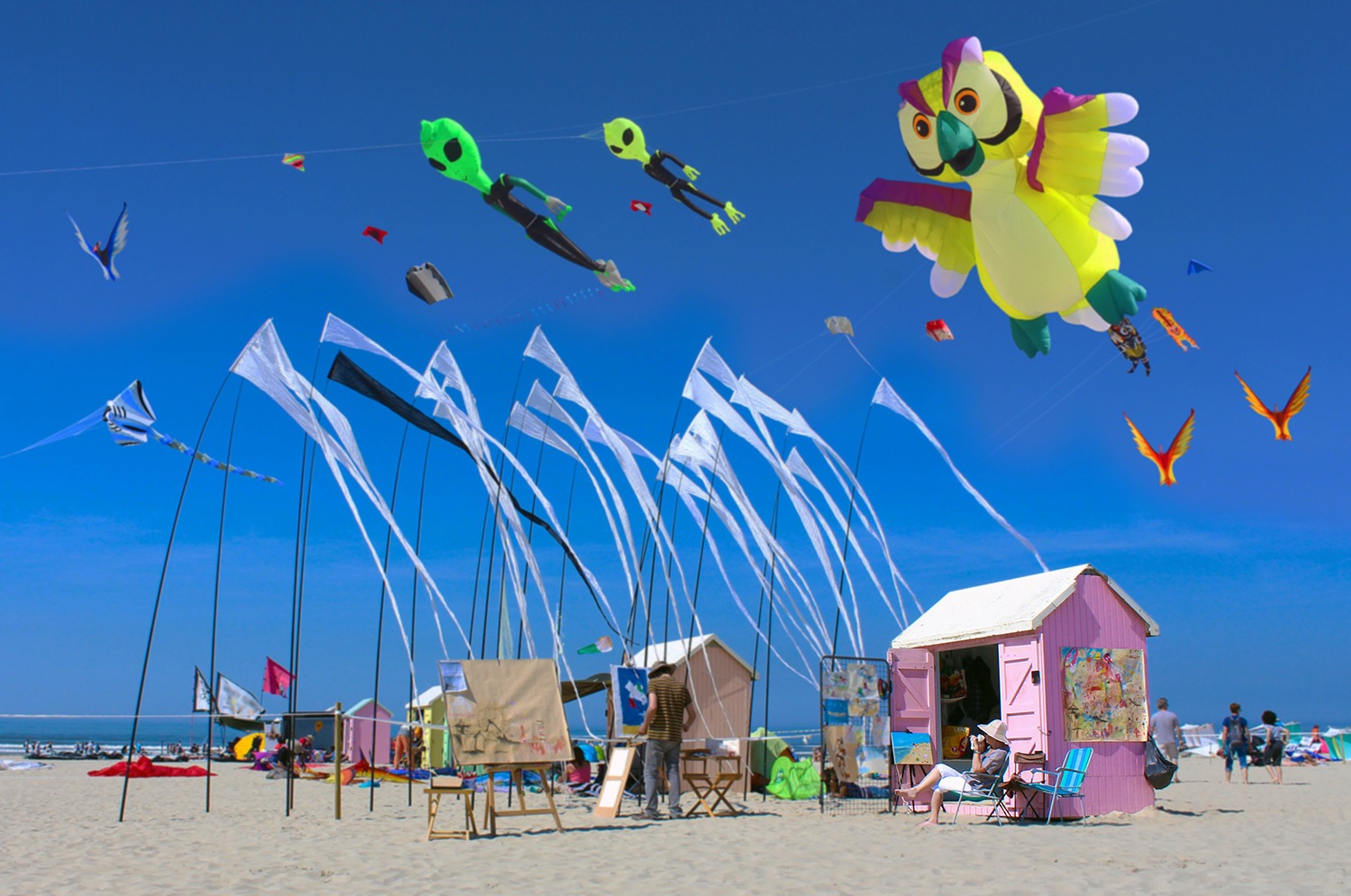}\\
     
 \includegraphics[width=\imwidth]{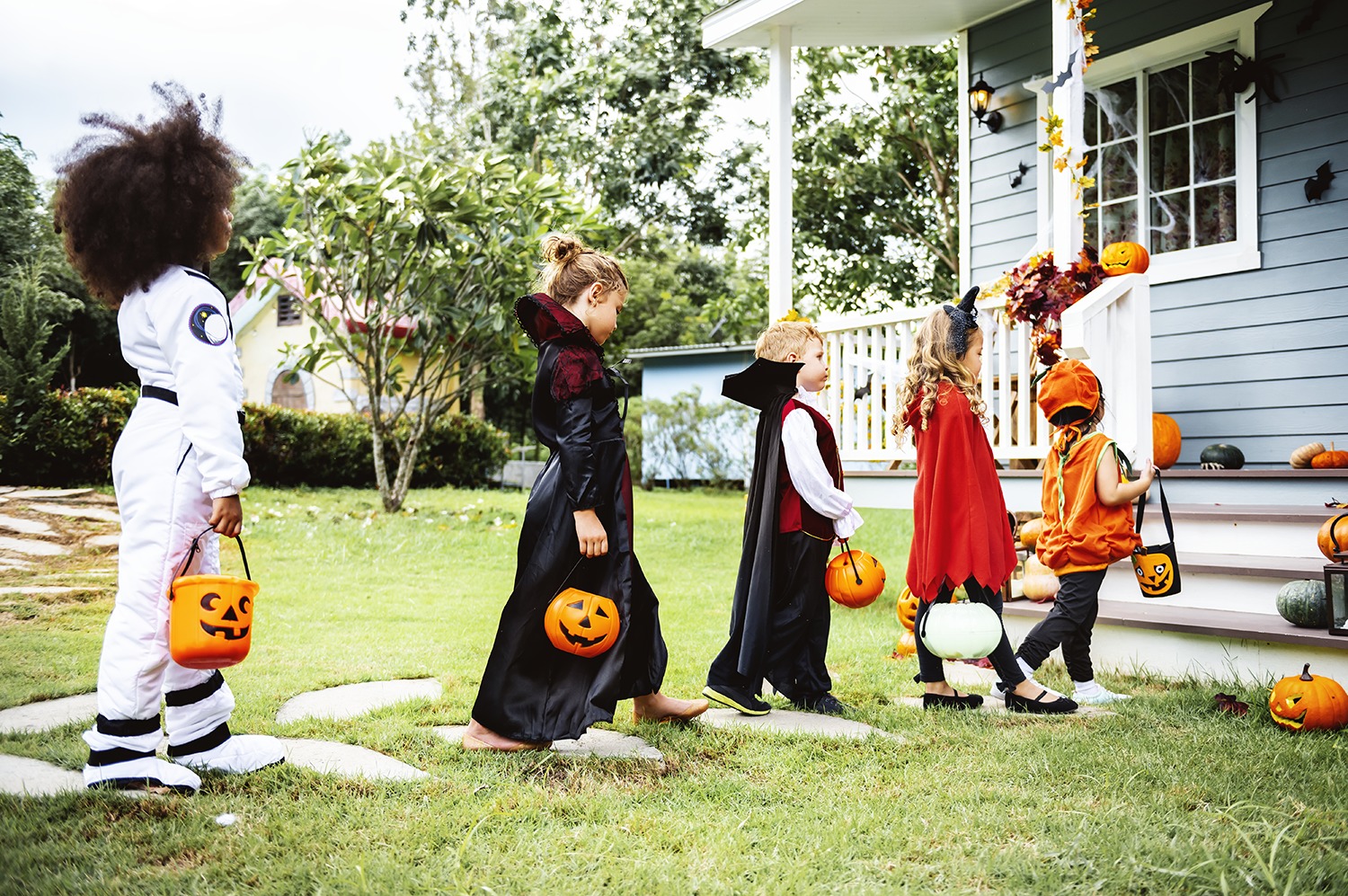}&
  \includegraphics[width=\imwidth]{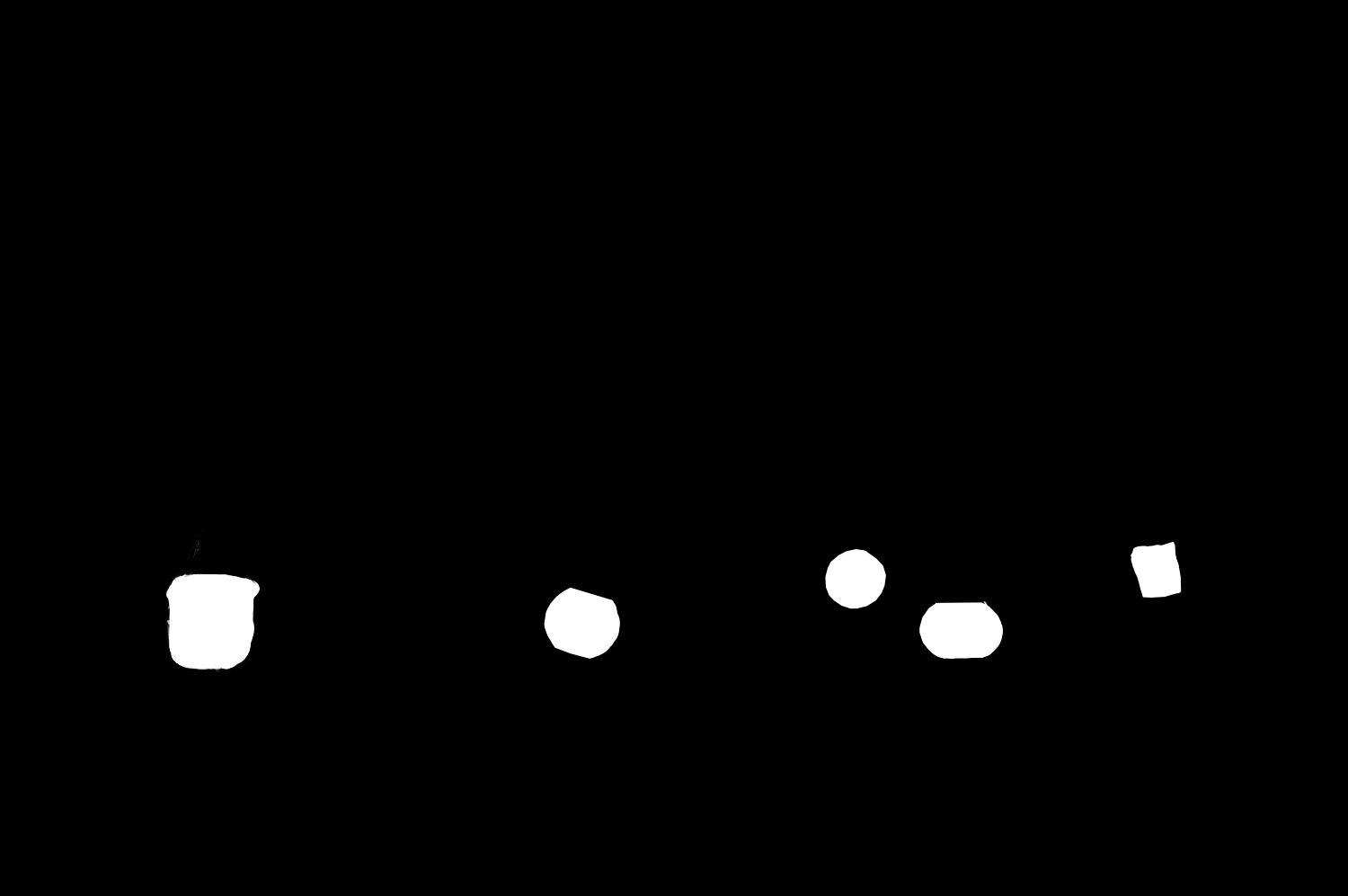}&
   \includegraphics[width=\imwidth]{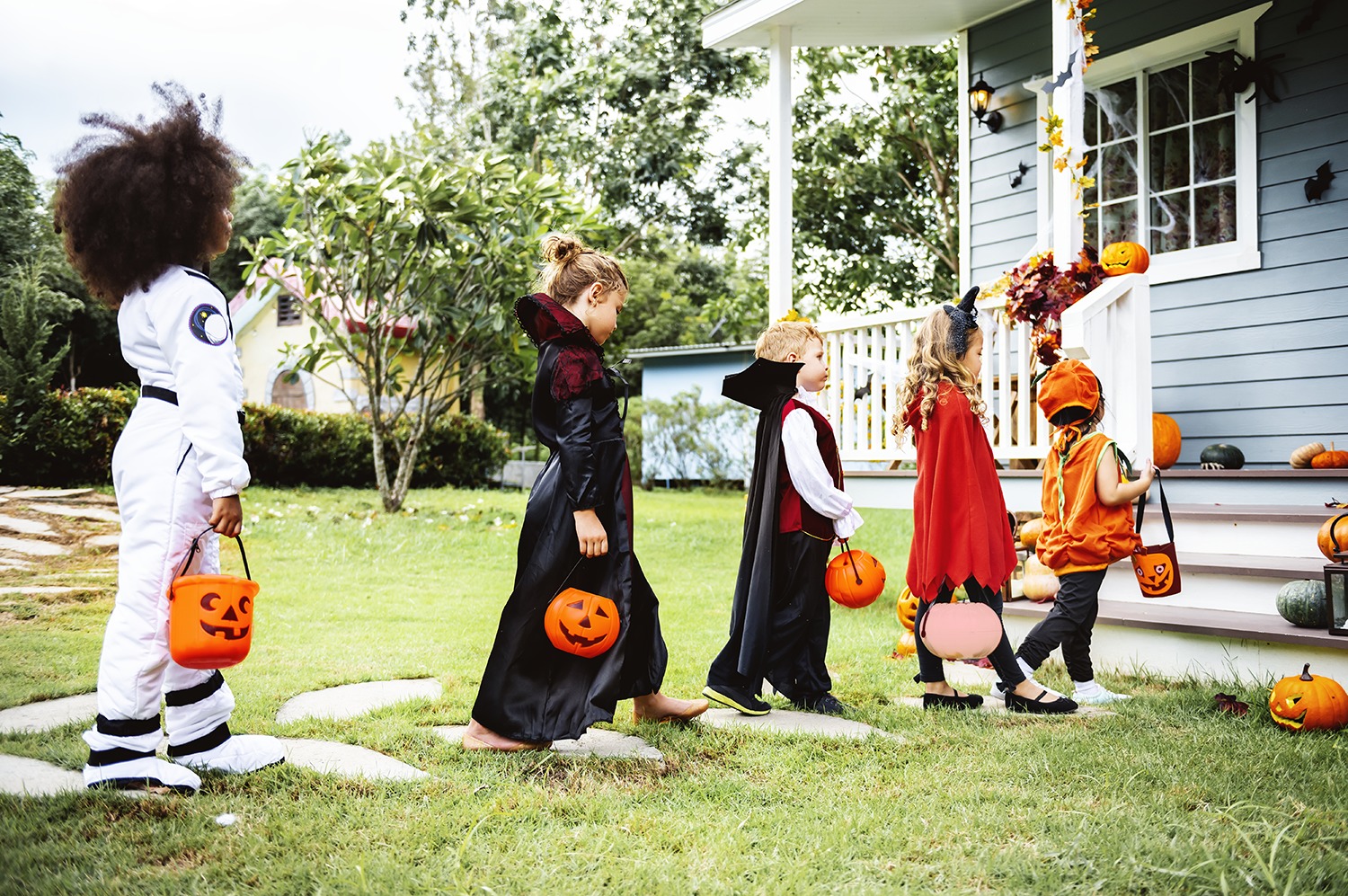}&
    \includegraphics[width=\imwidth]{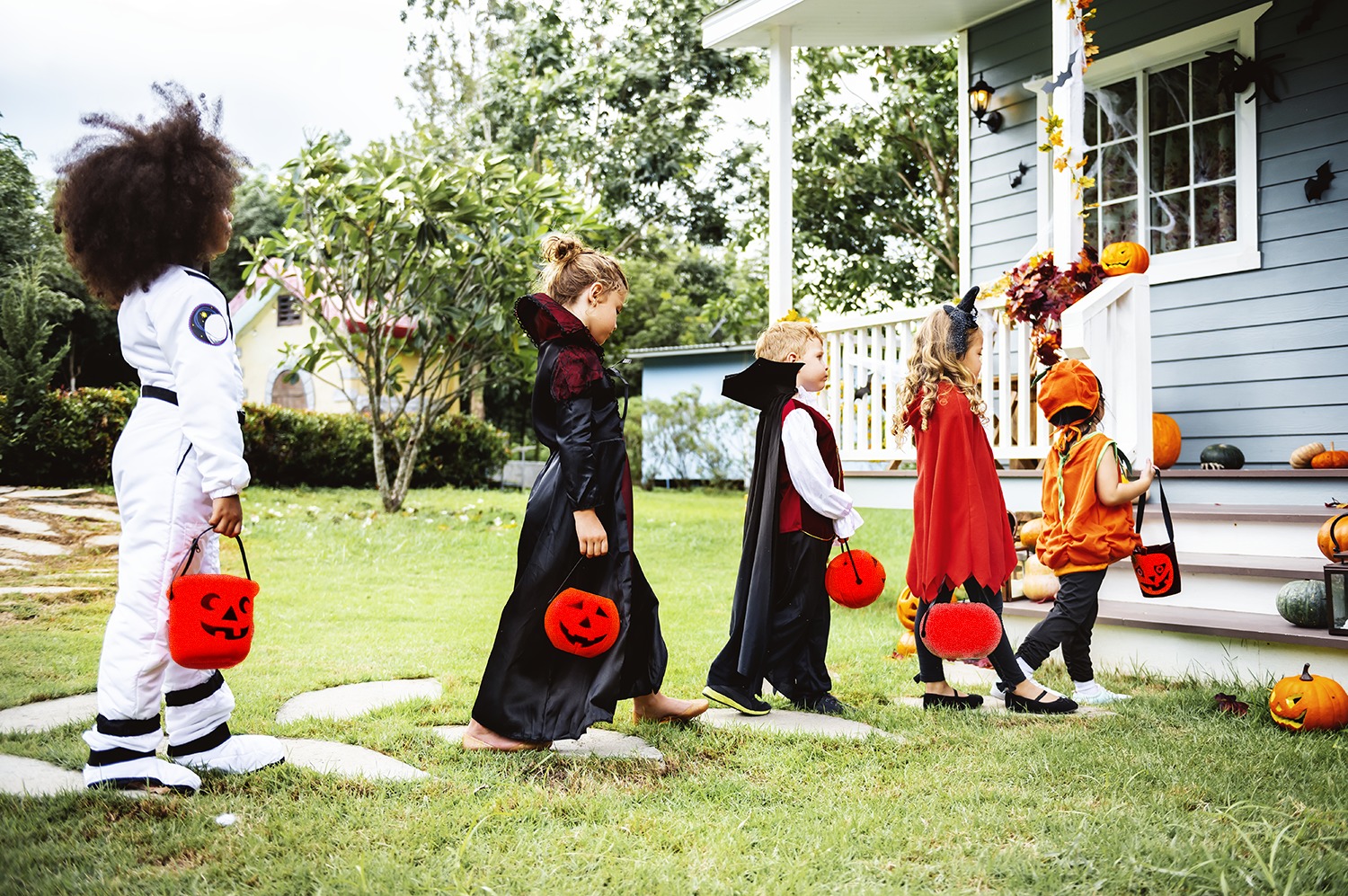}&
     \includegraphics[width=\imwidth]{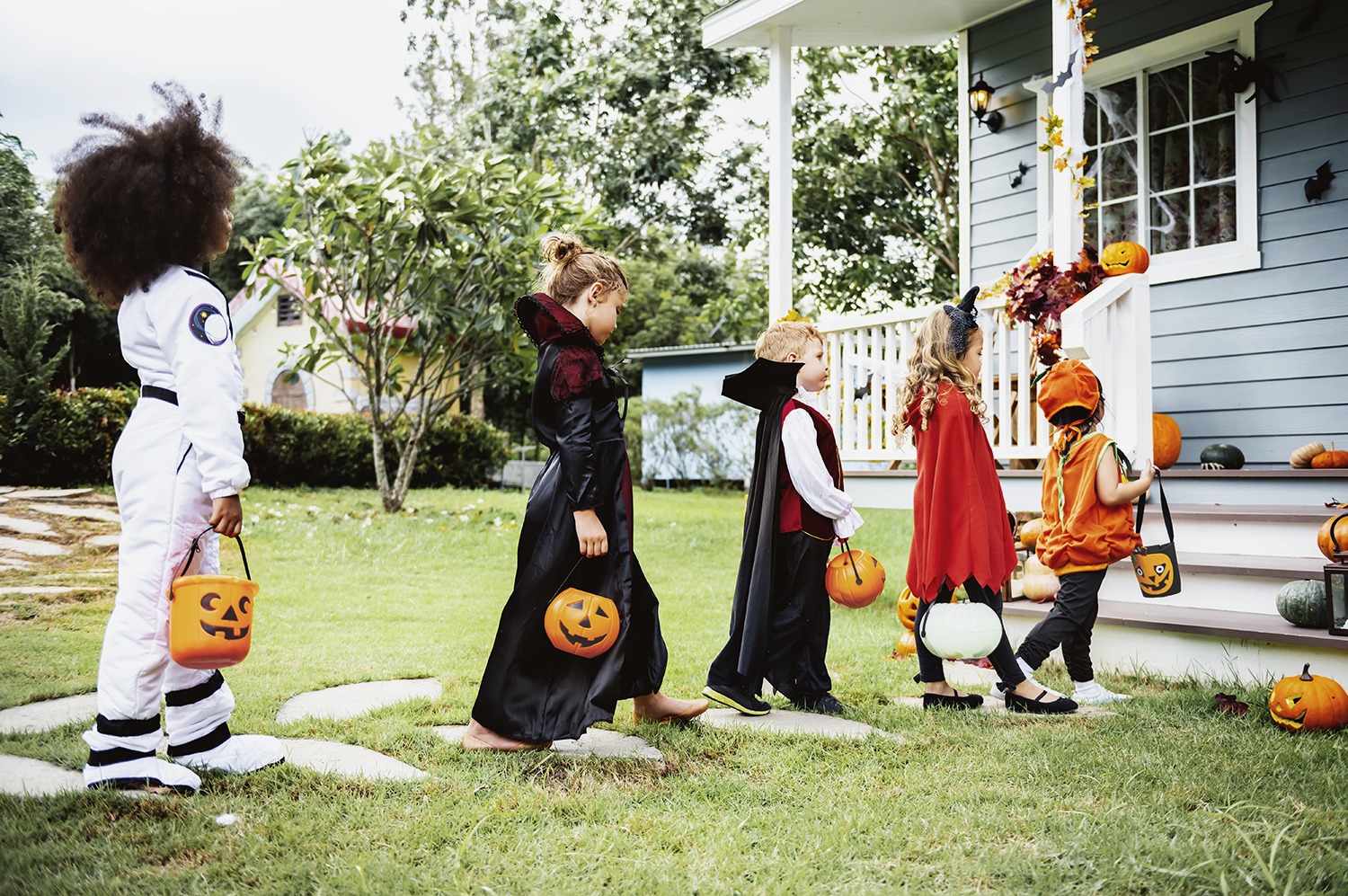}&
     \includegraphics[width=\imwidth]{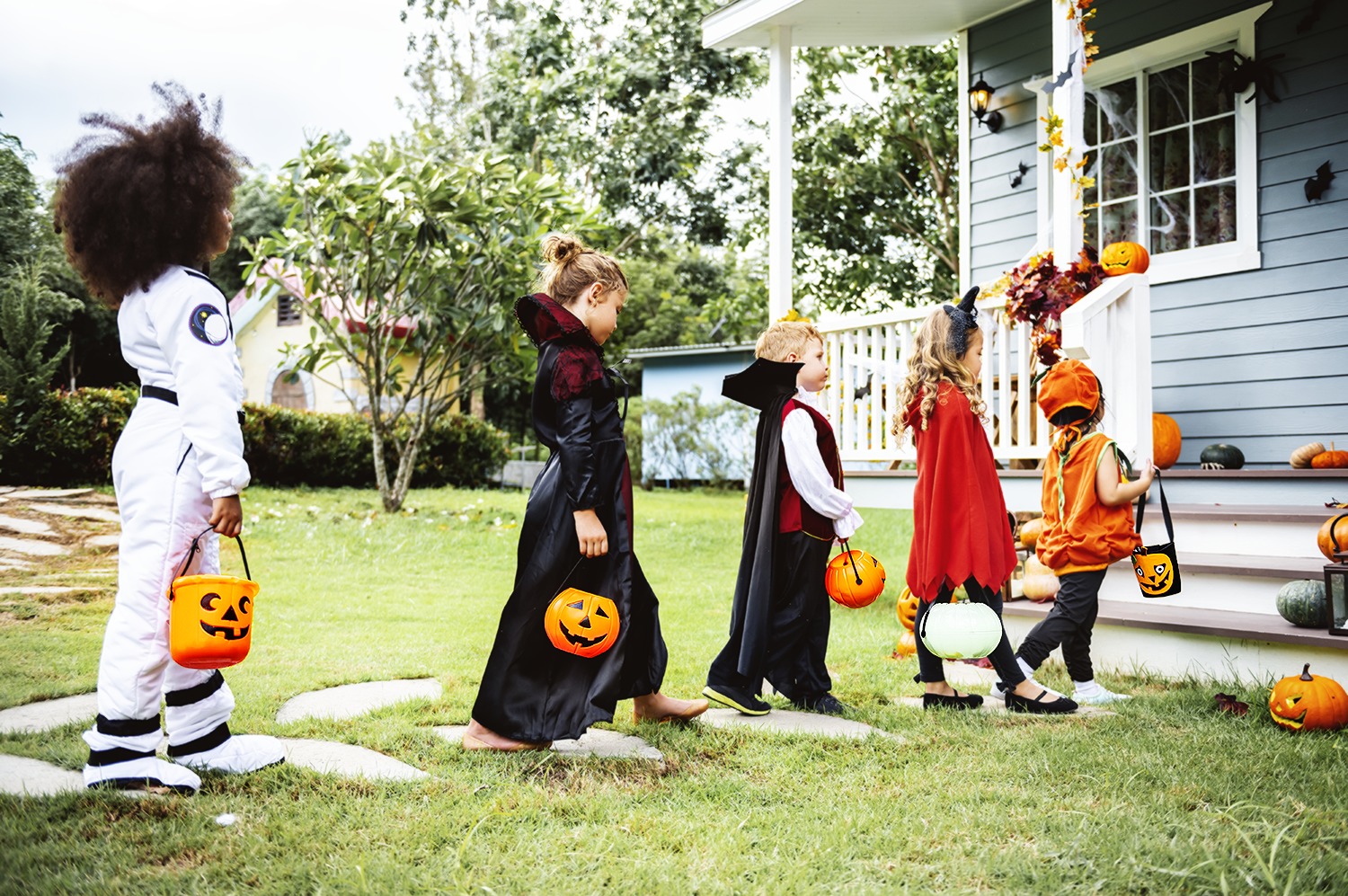}\\
     
 \includegraphics[width=\imwidth]{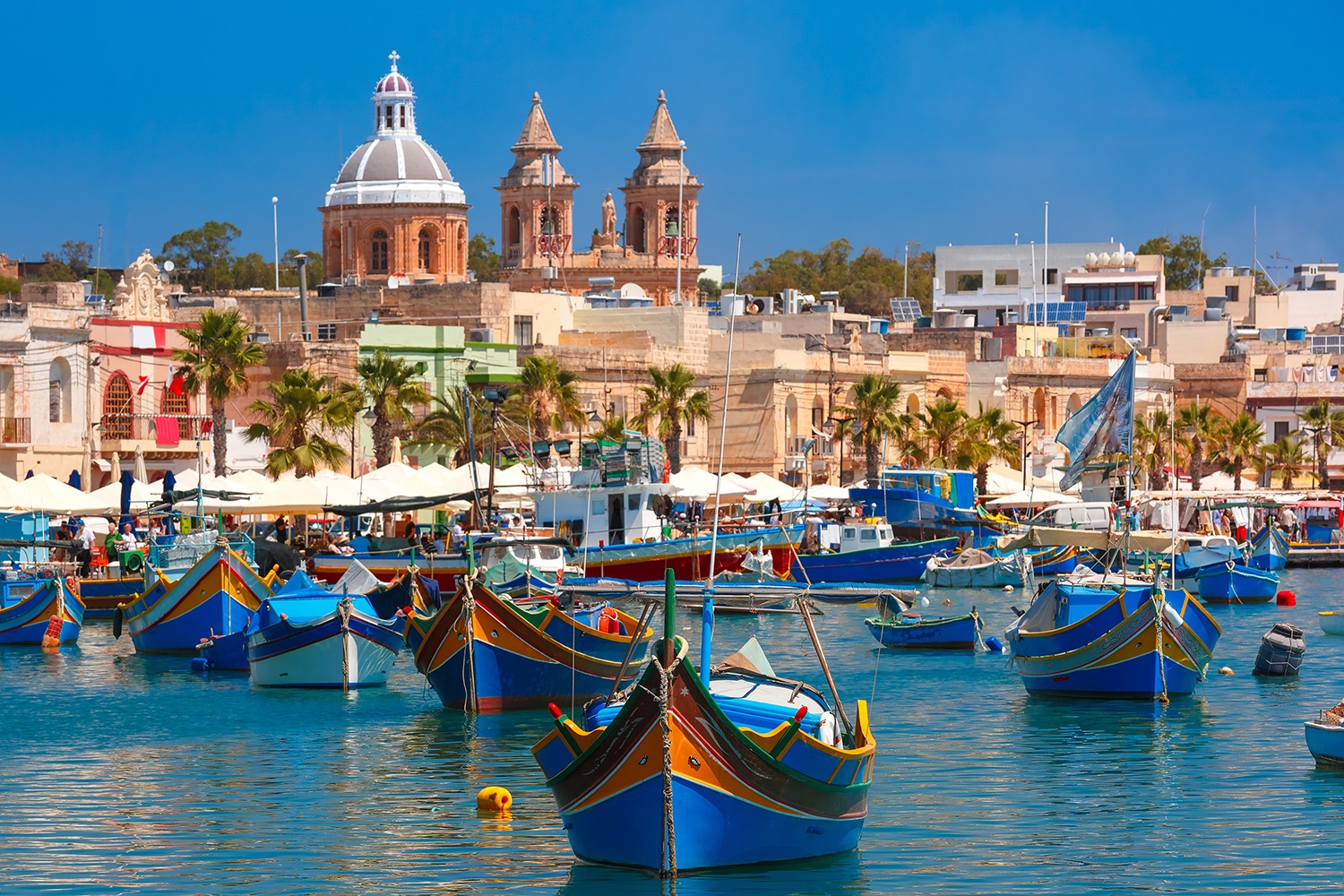}&
  \includegraphics[width=\imwidth]{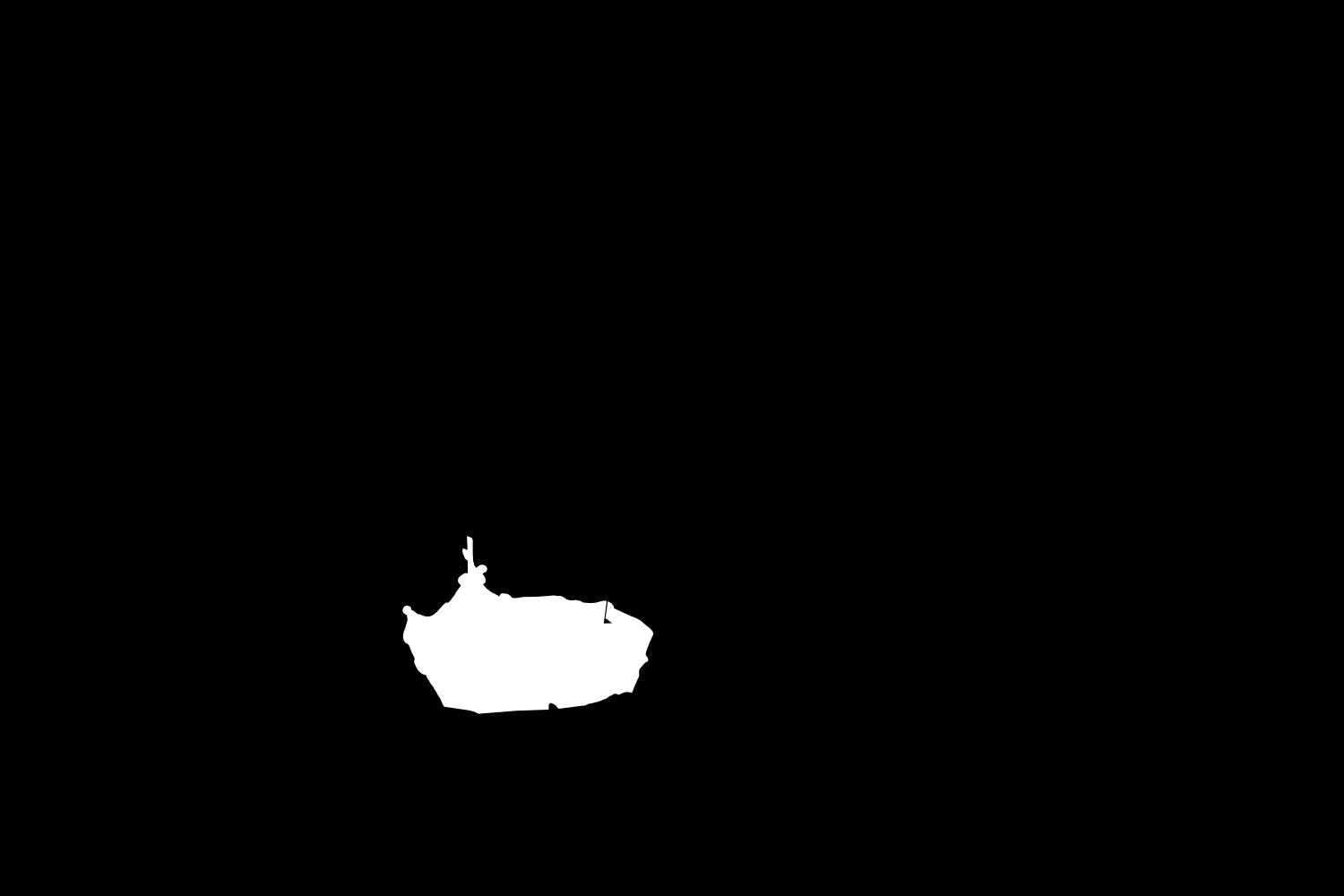}&
   \includegraphics[width=\imwidth]{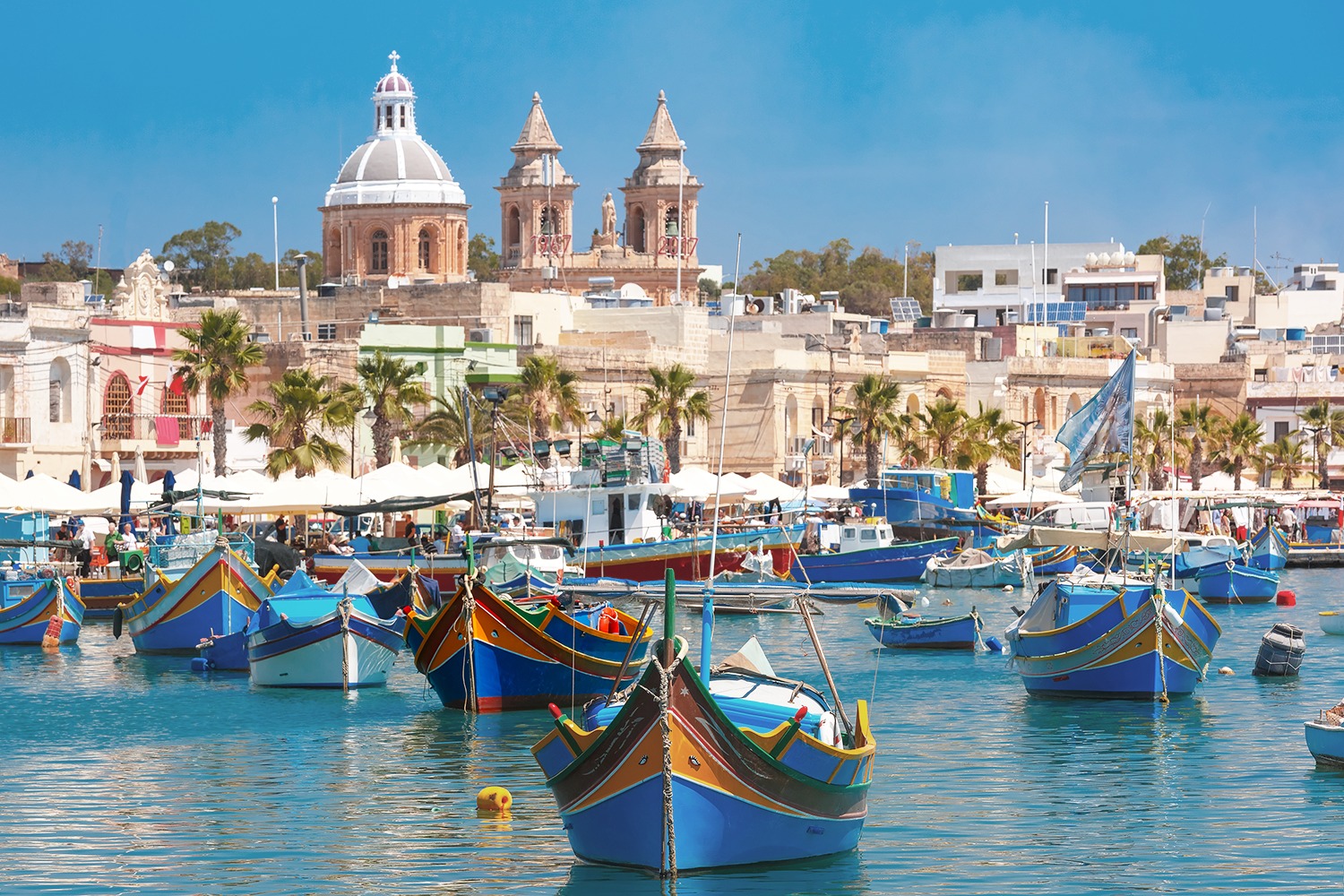}&
    \includegraphics[width=\imwidth]{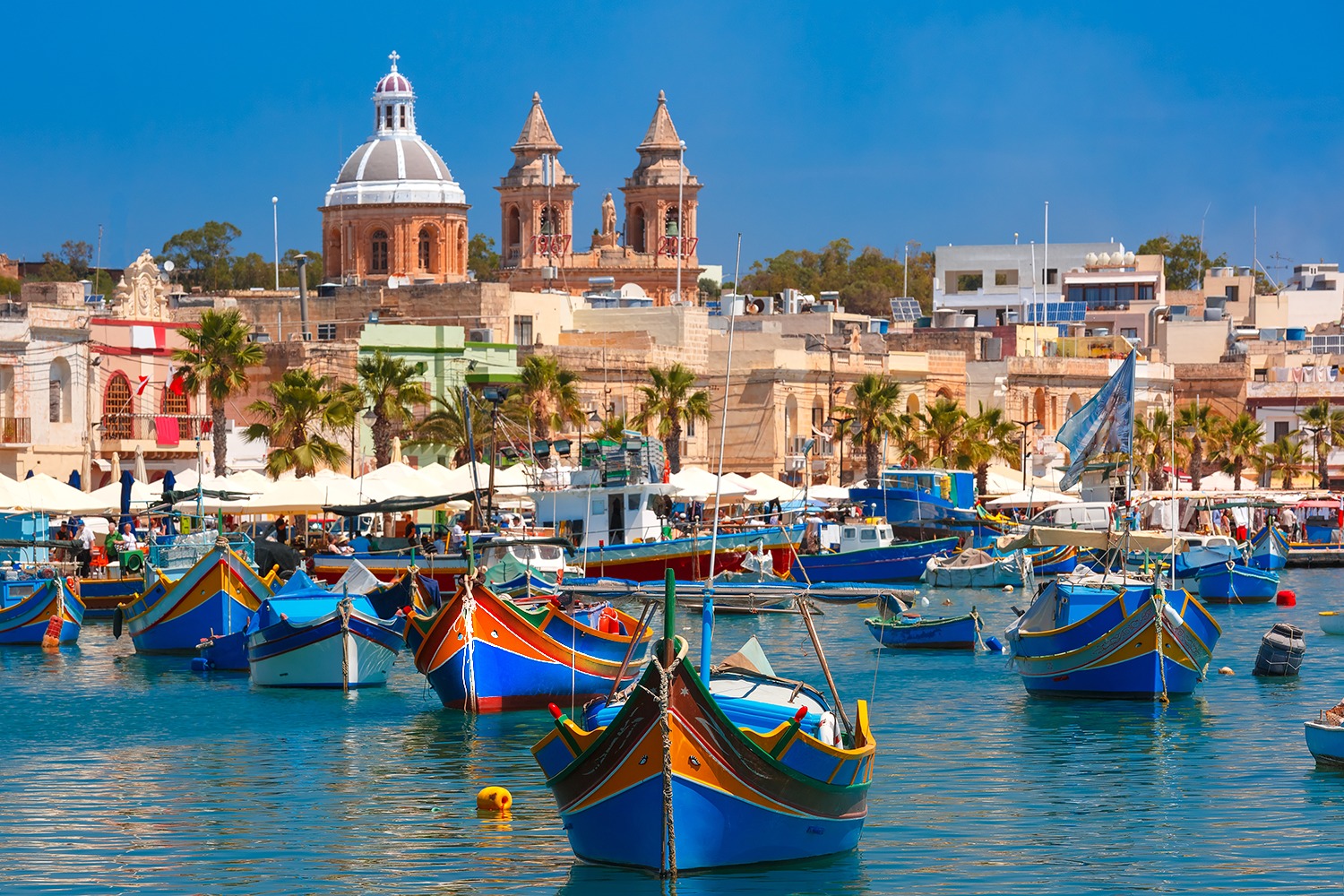}&
     \includegraphics[width=\imwidth]{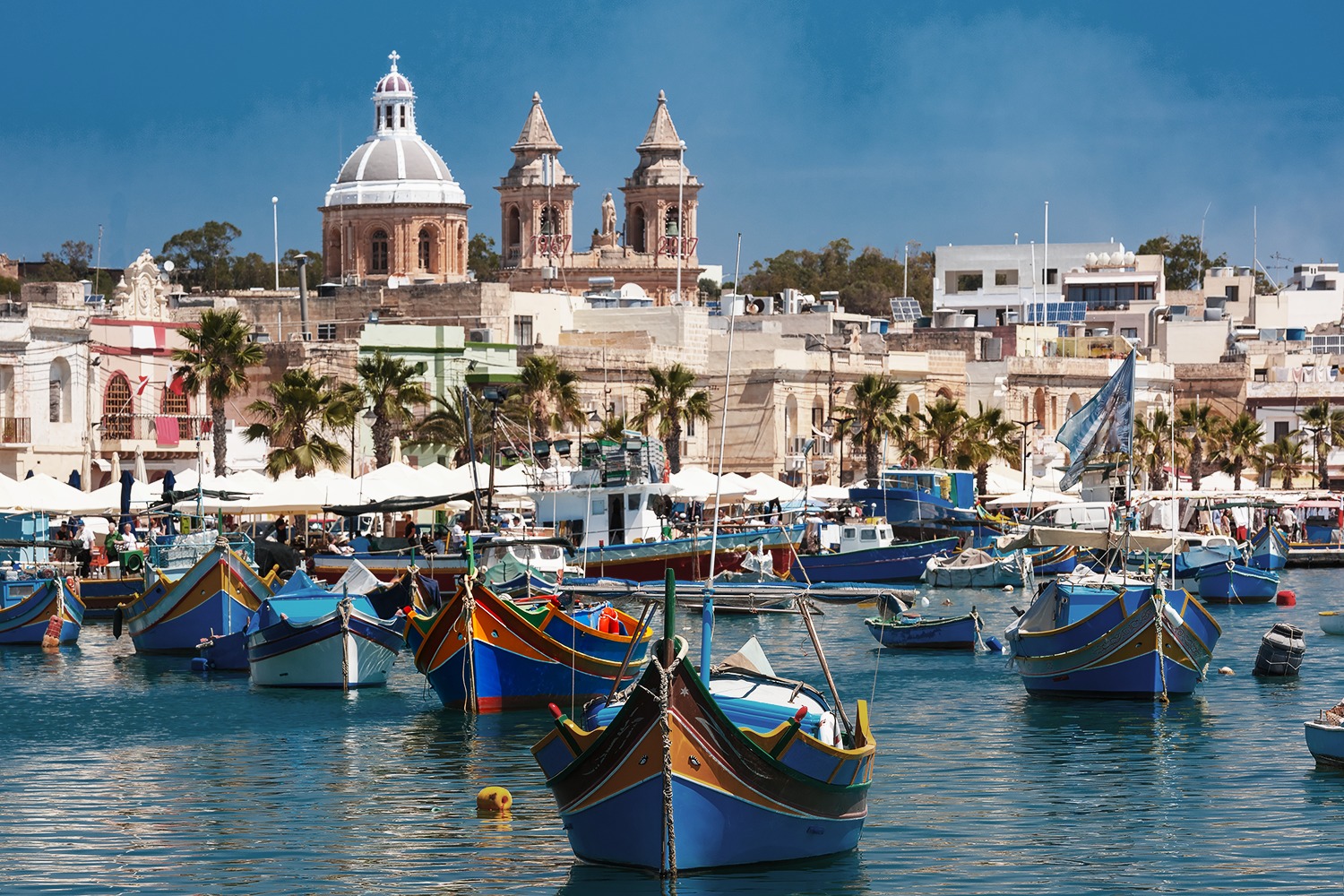}&
     \includegraphics[width=\imwidth]{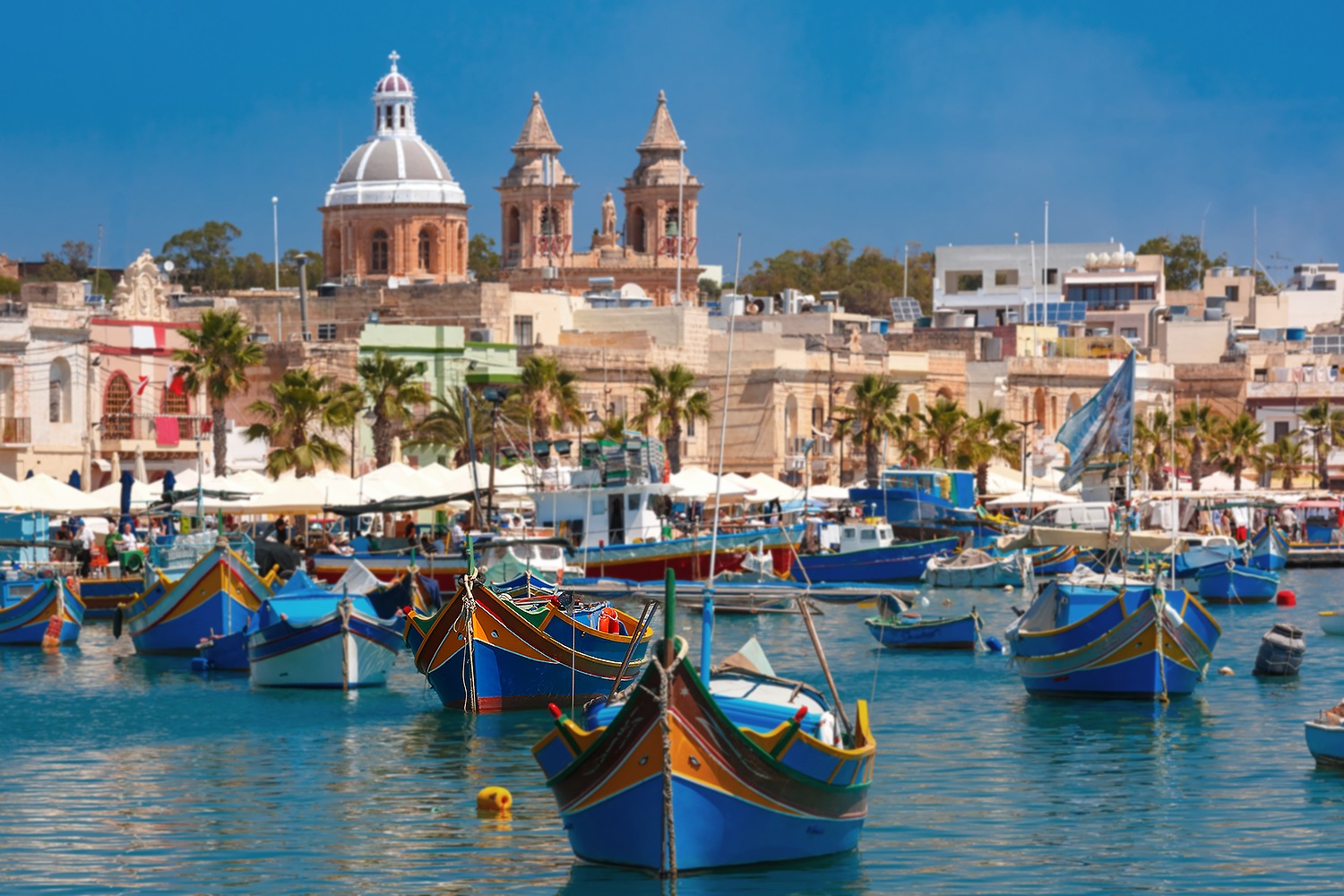}\\
     
 \includegraphics[width=\imwidth]{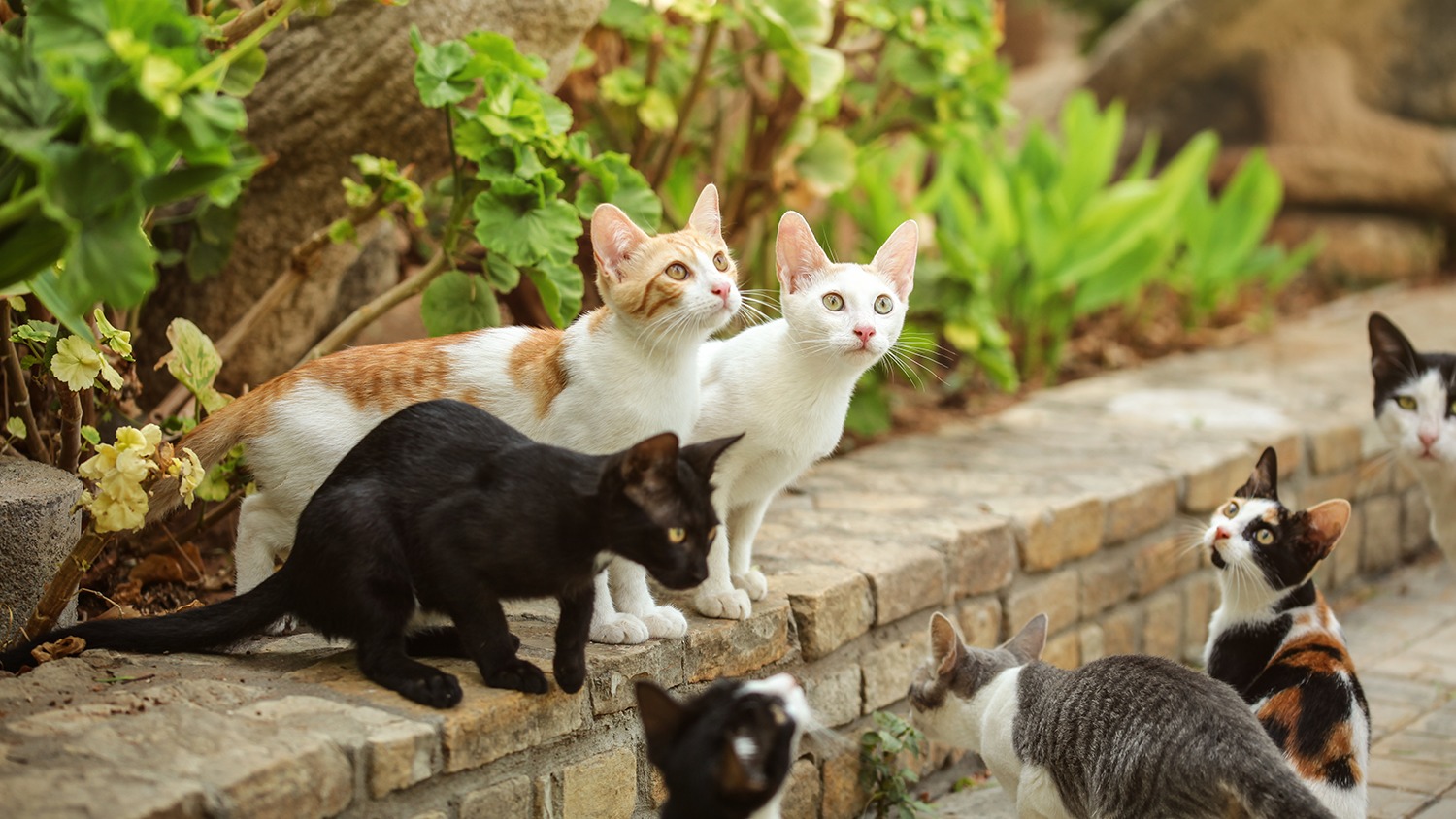}&
  \includegraphics[width=\imwidth]{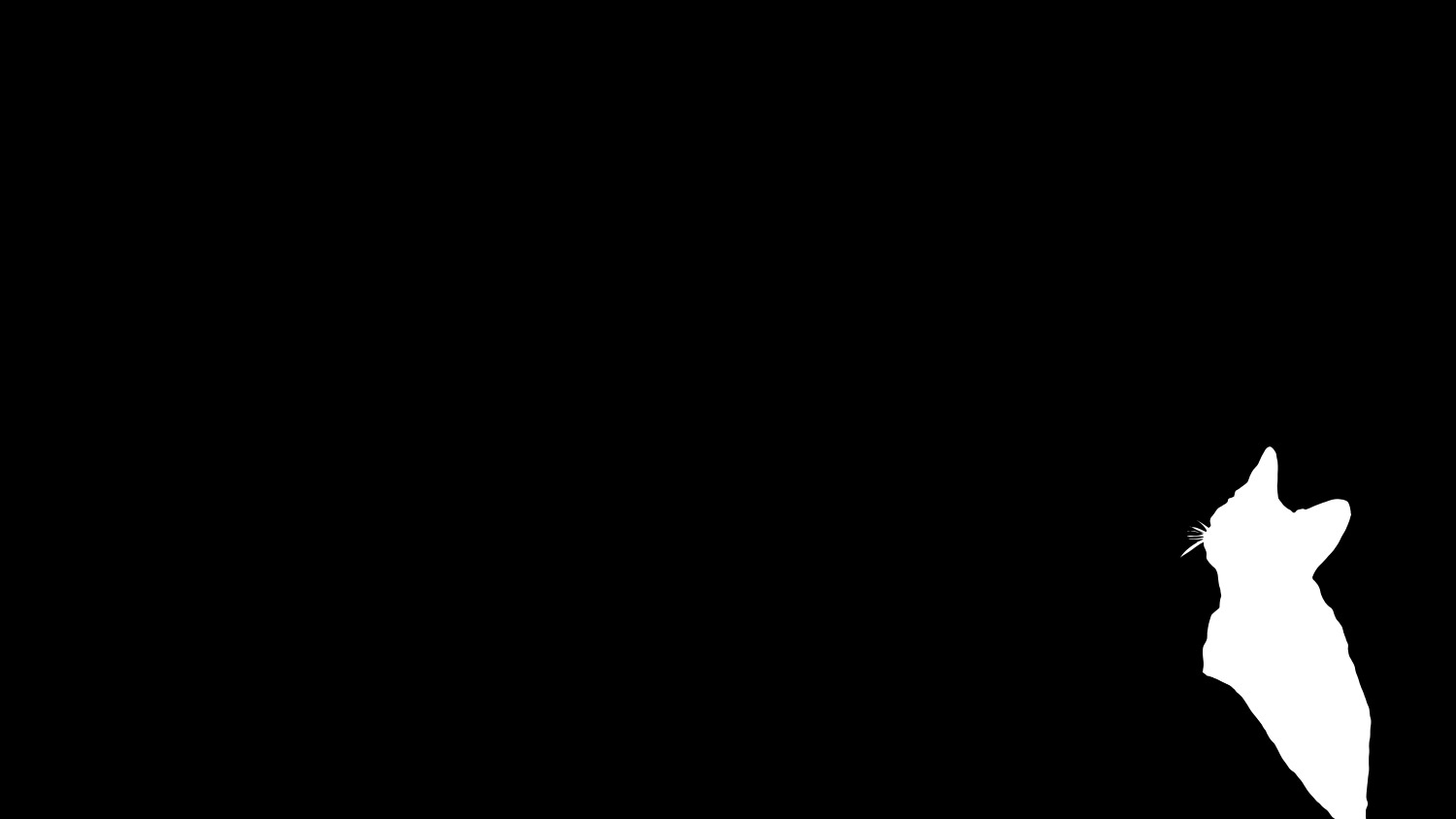}&
   \includegraphics[width=\imwidth]{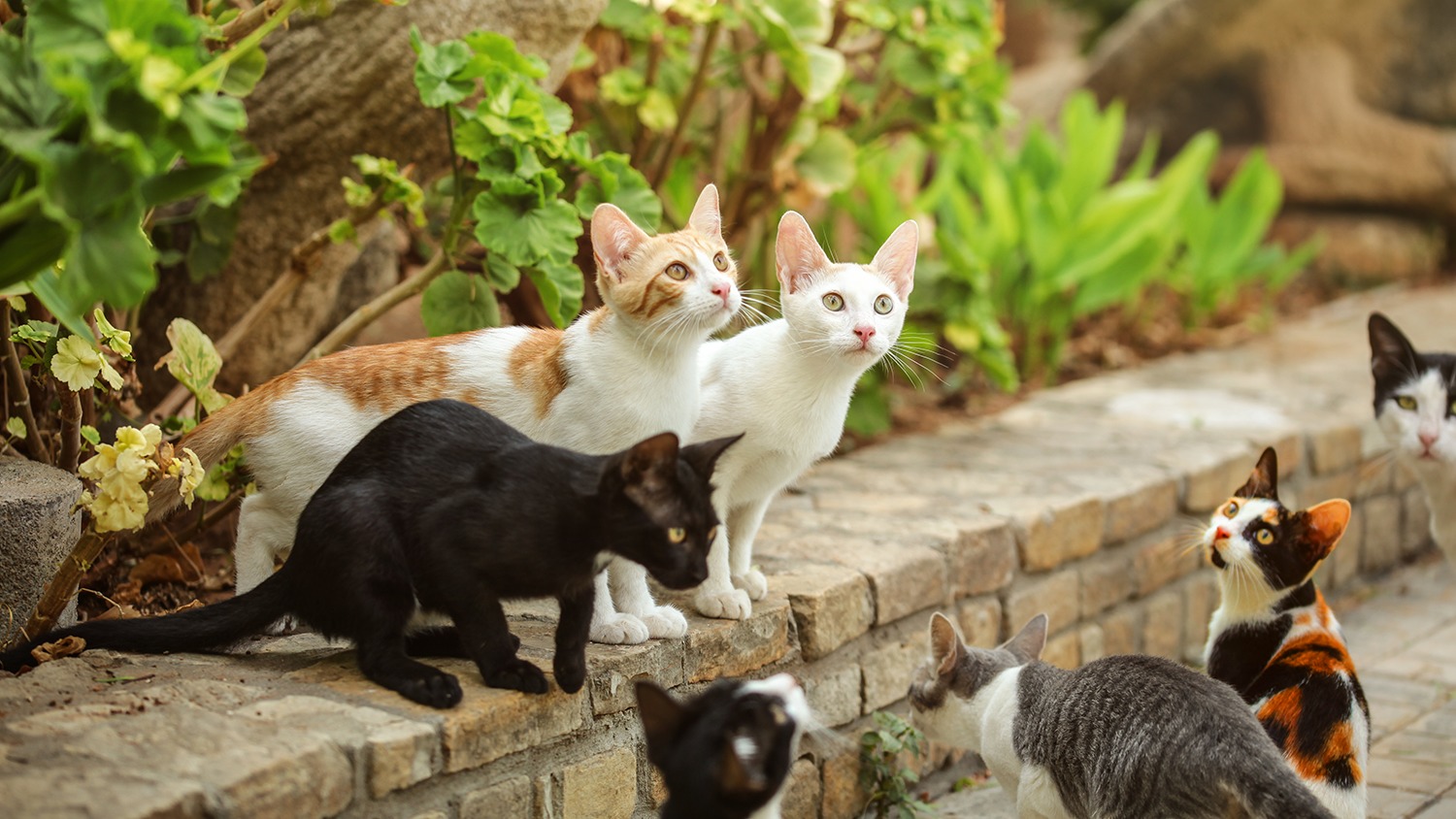}&
    \includegraphics[width=\imwidth]{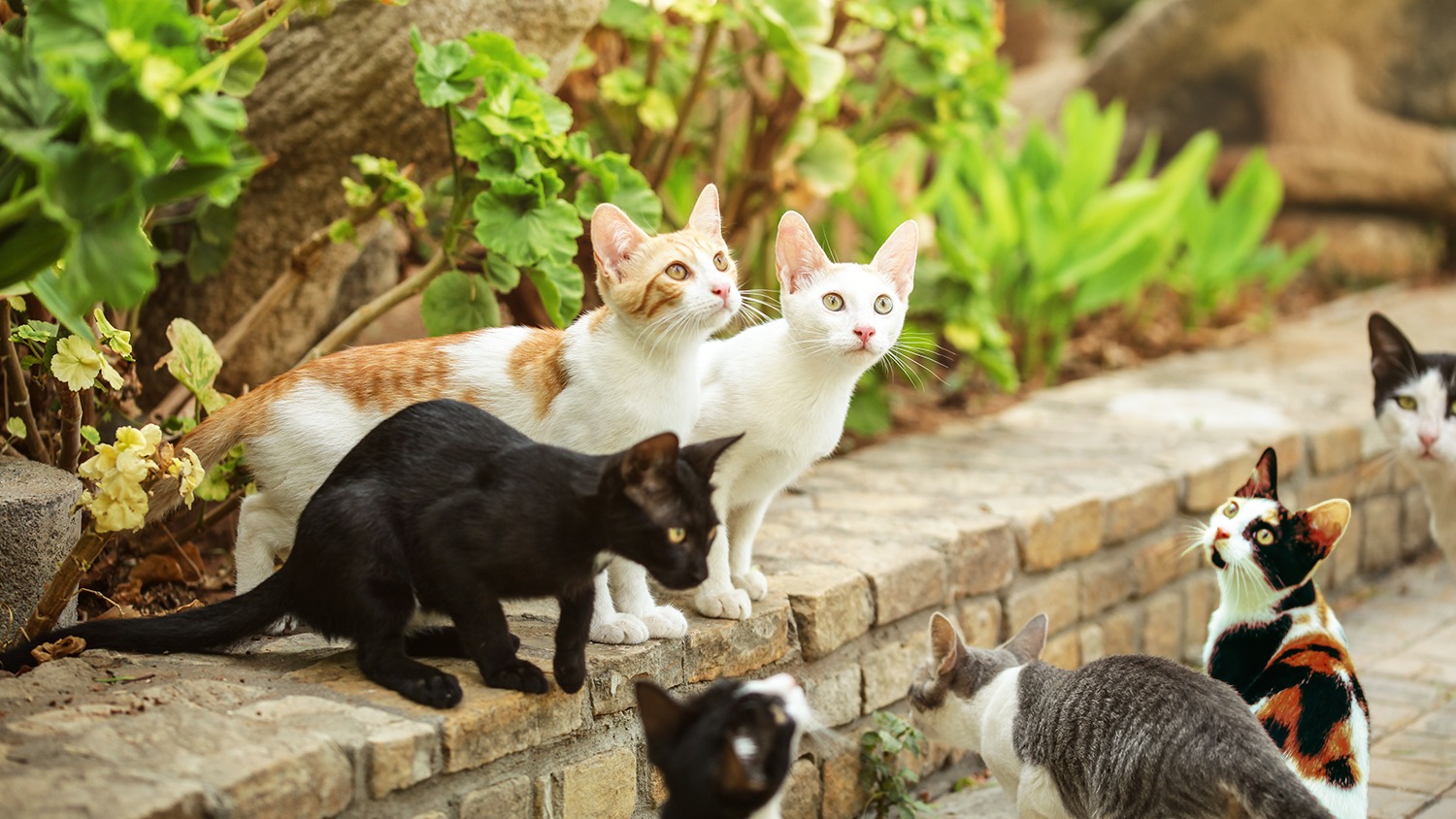}&
     \includegraphics[width=\imwidth]{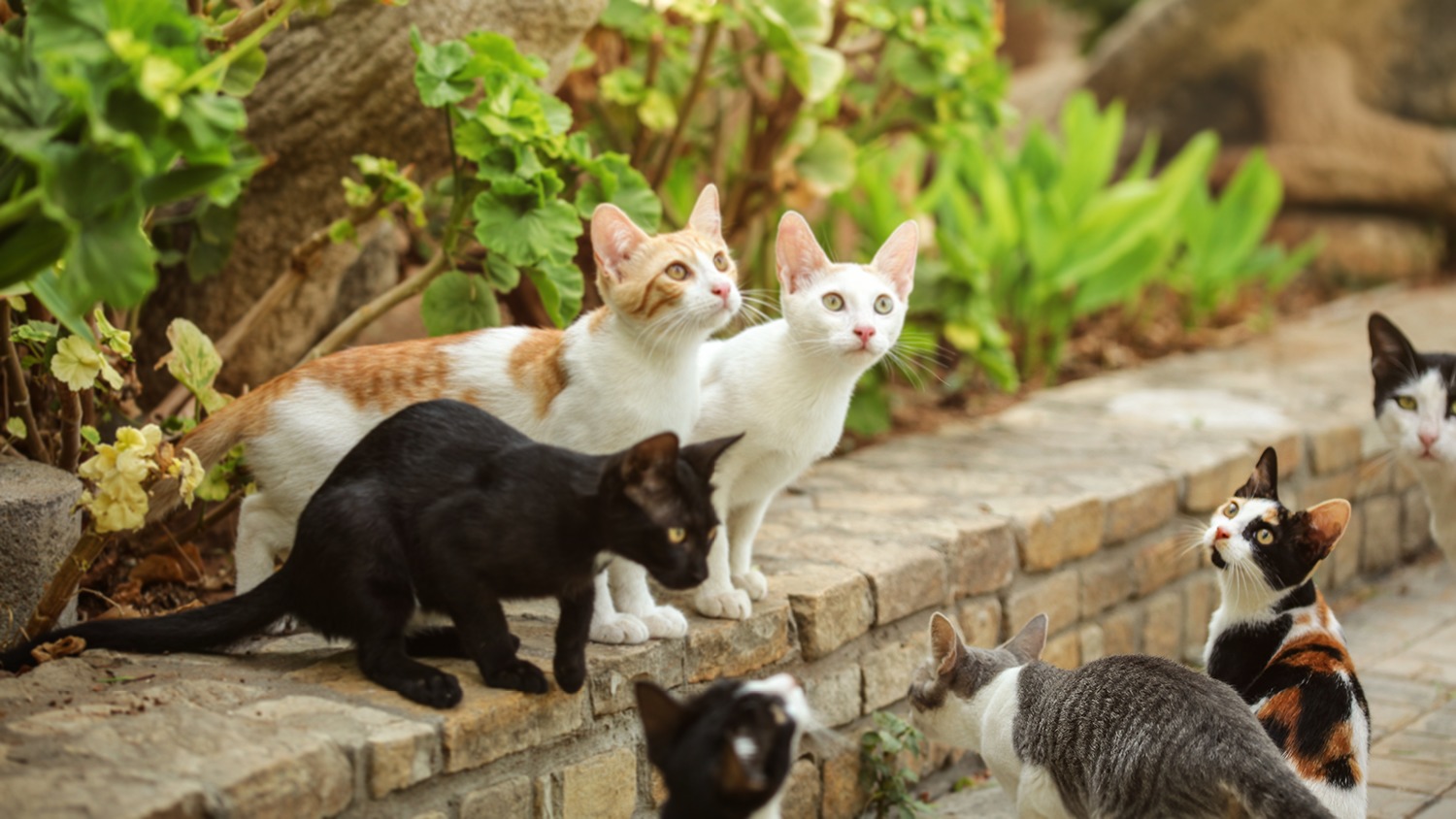}&
     \includegraphics[width=\imwidth]{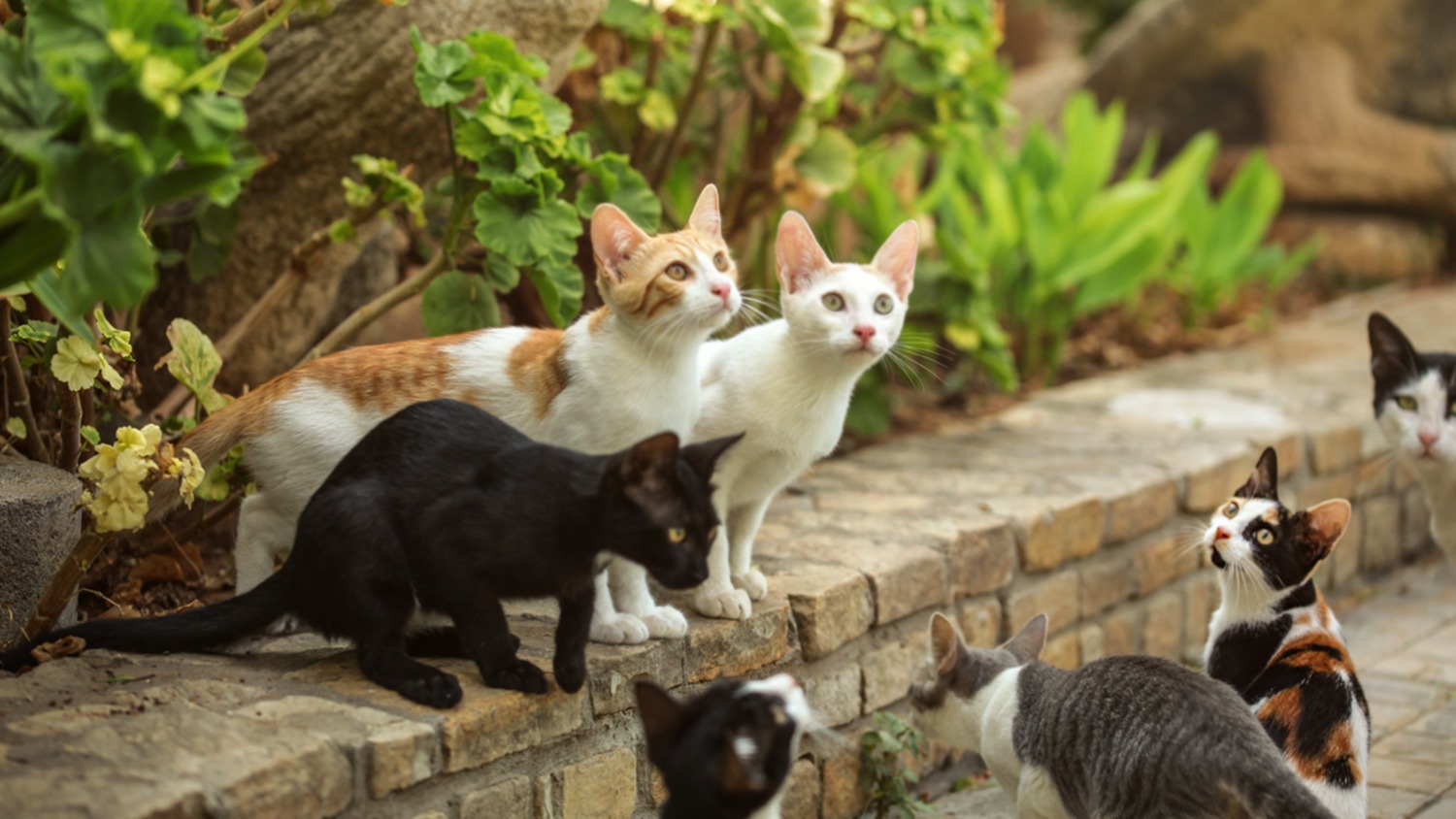}\\
     
 \includegraphics[width=\imwidth]{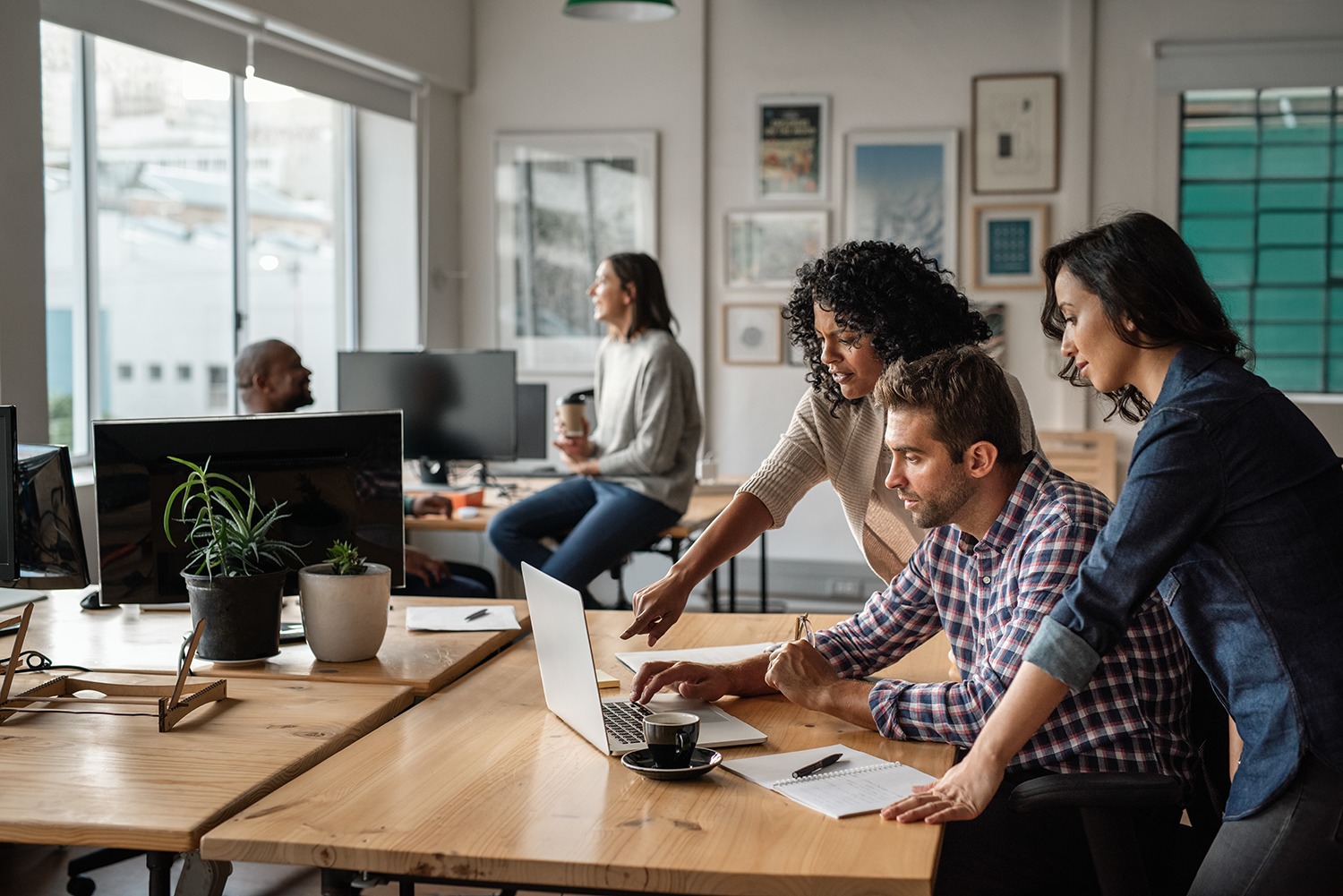}&
  \includegraphics[width=\imwidth]{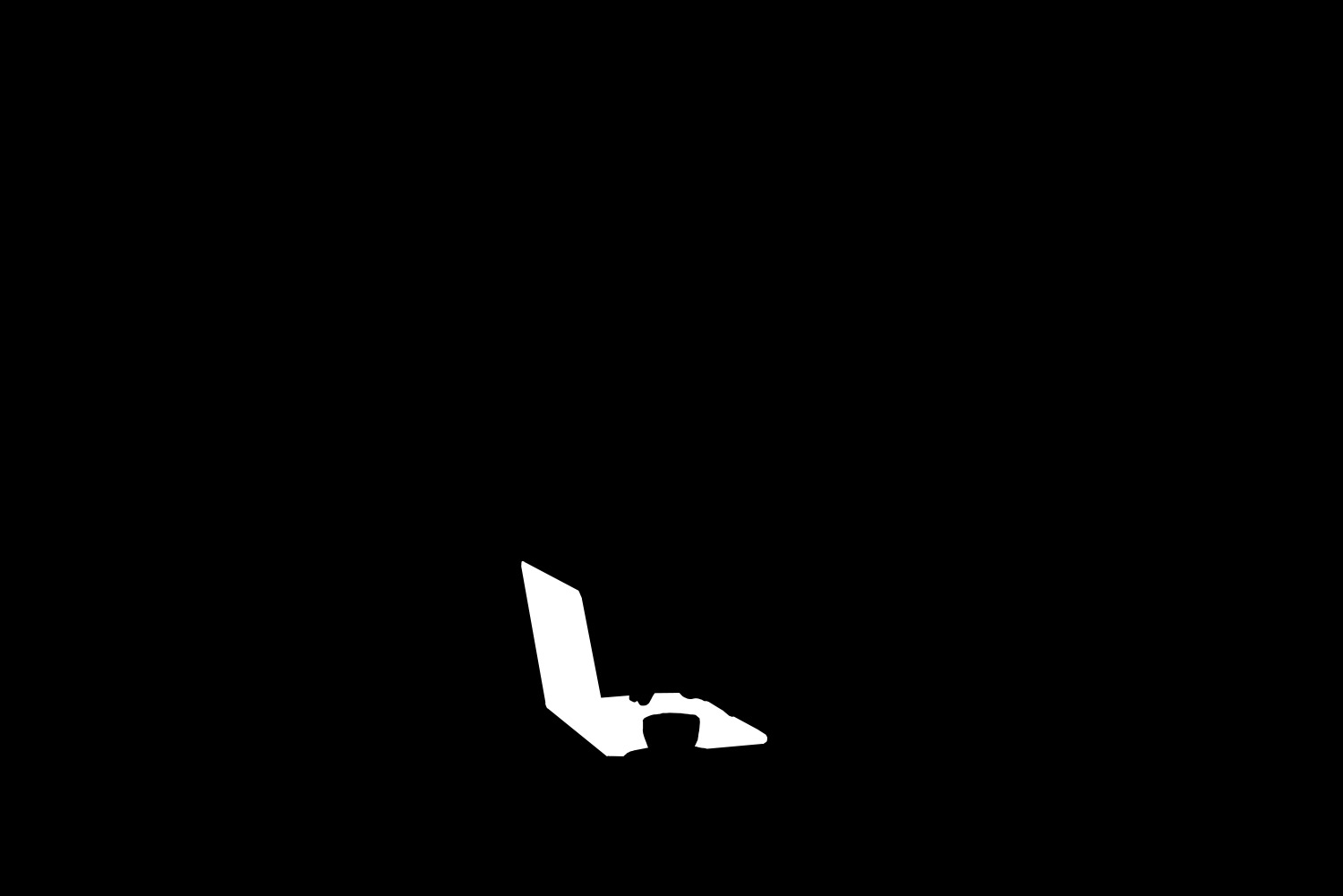}&
   \includegraphics[width=\imwidth]{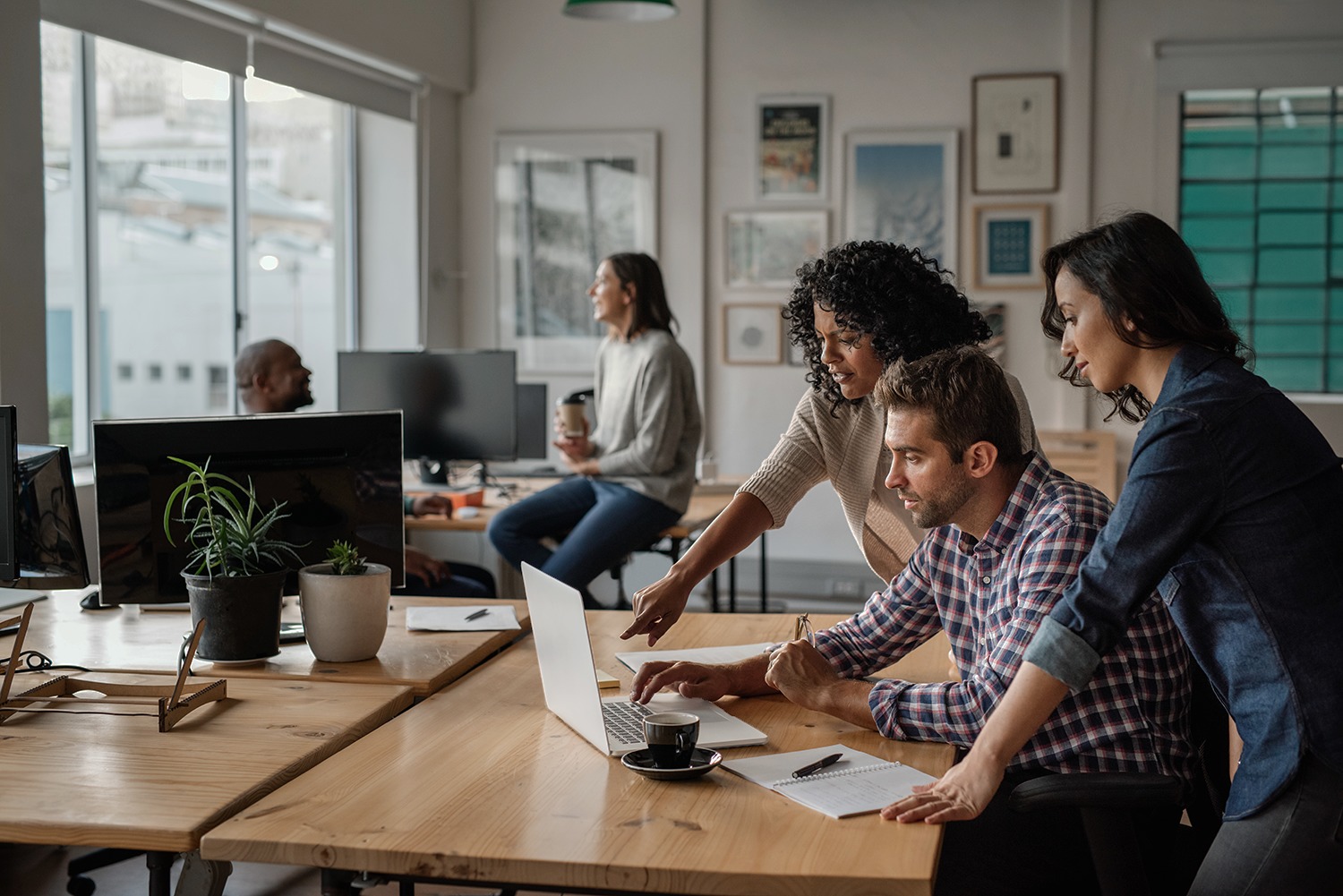}&
    \includegraphics[width=\imwidth]{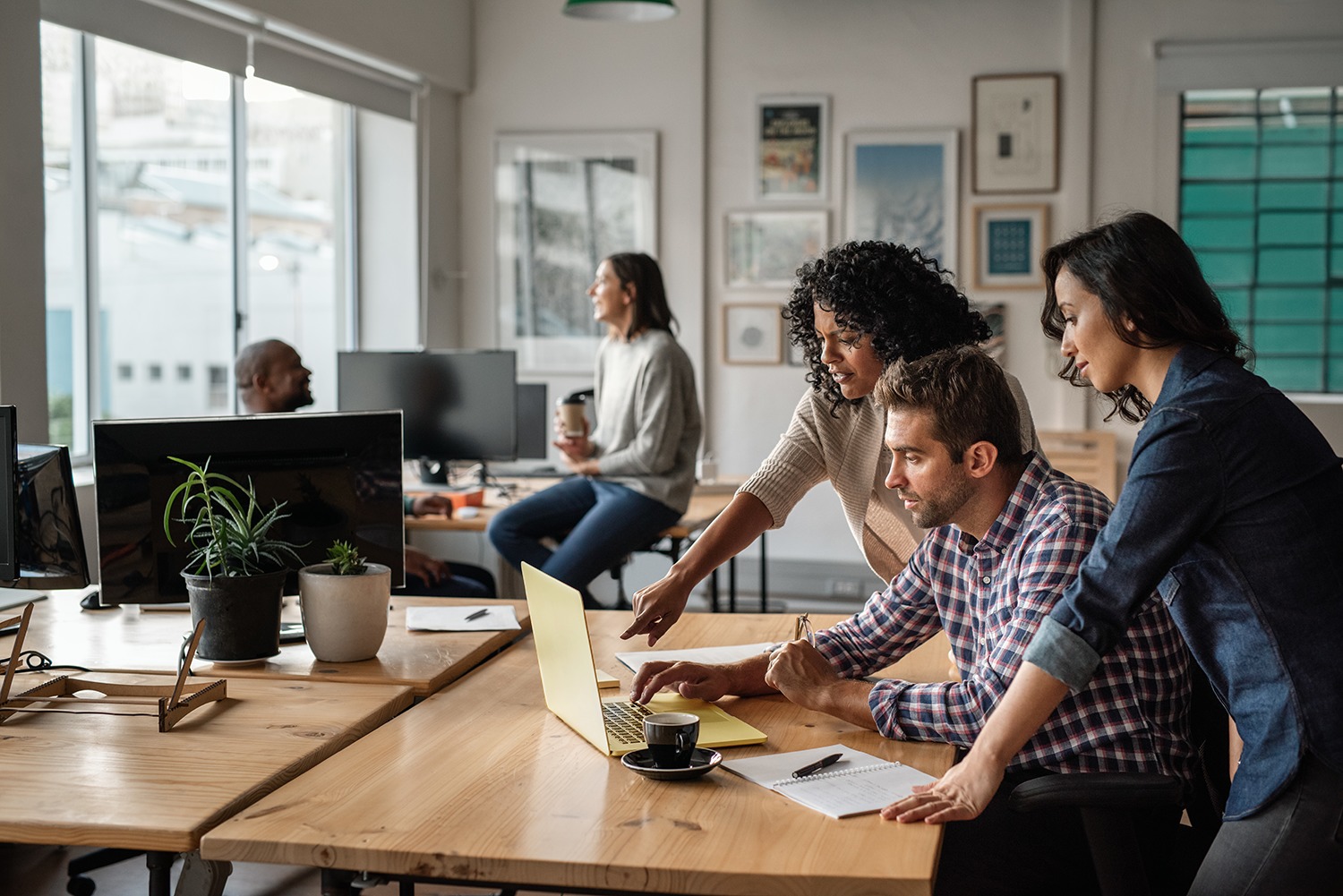}&
     \includegraphics[width=\imwidth]{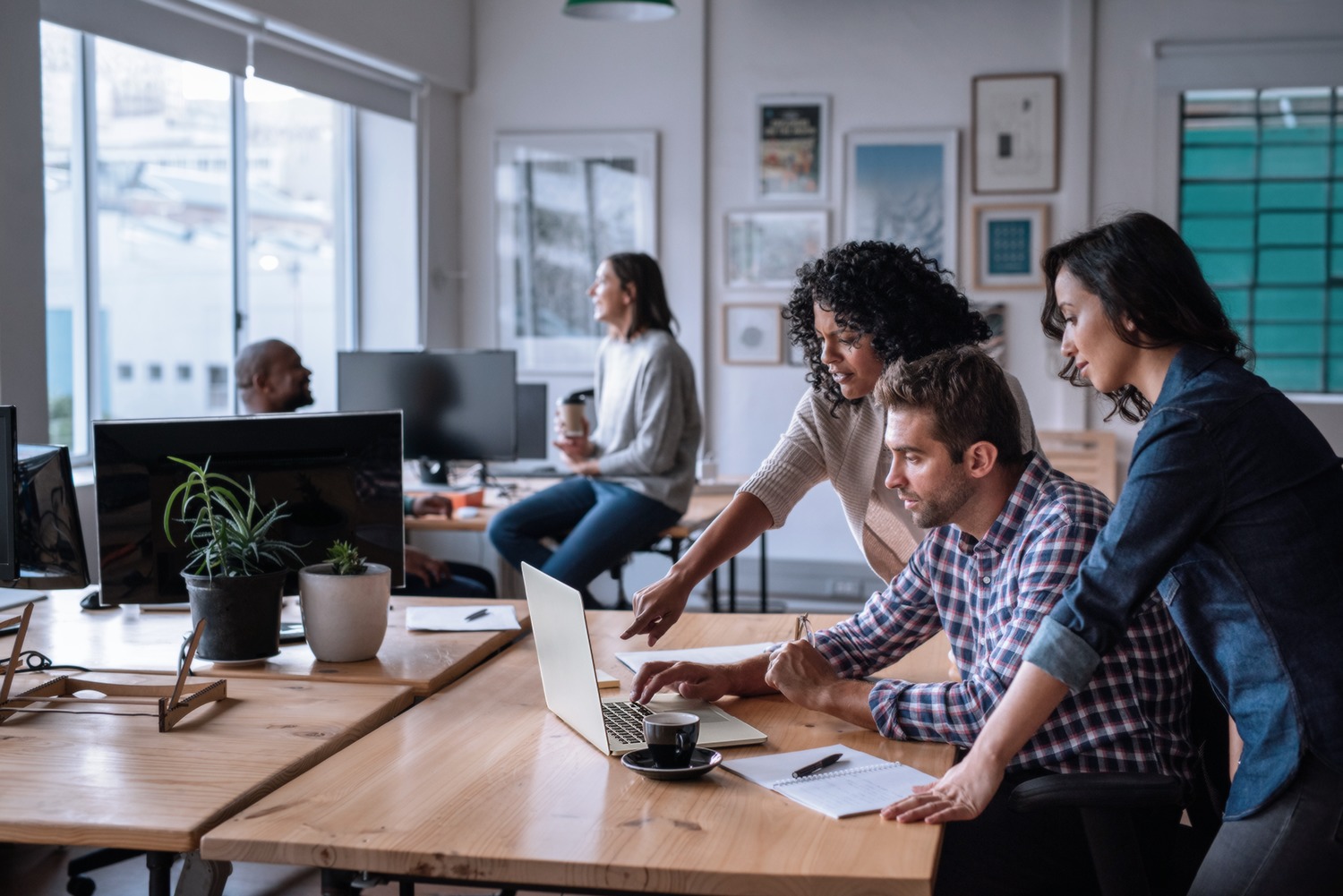}&
     \includegraphics[width=\imwidth]{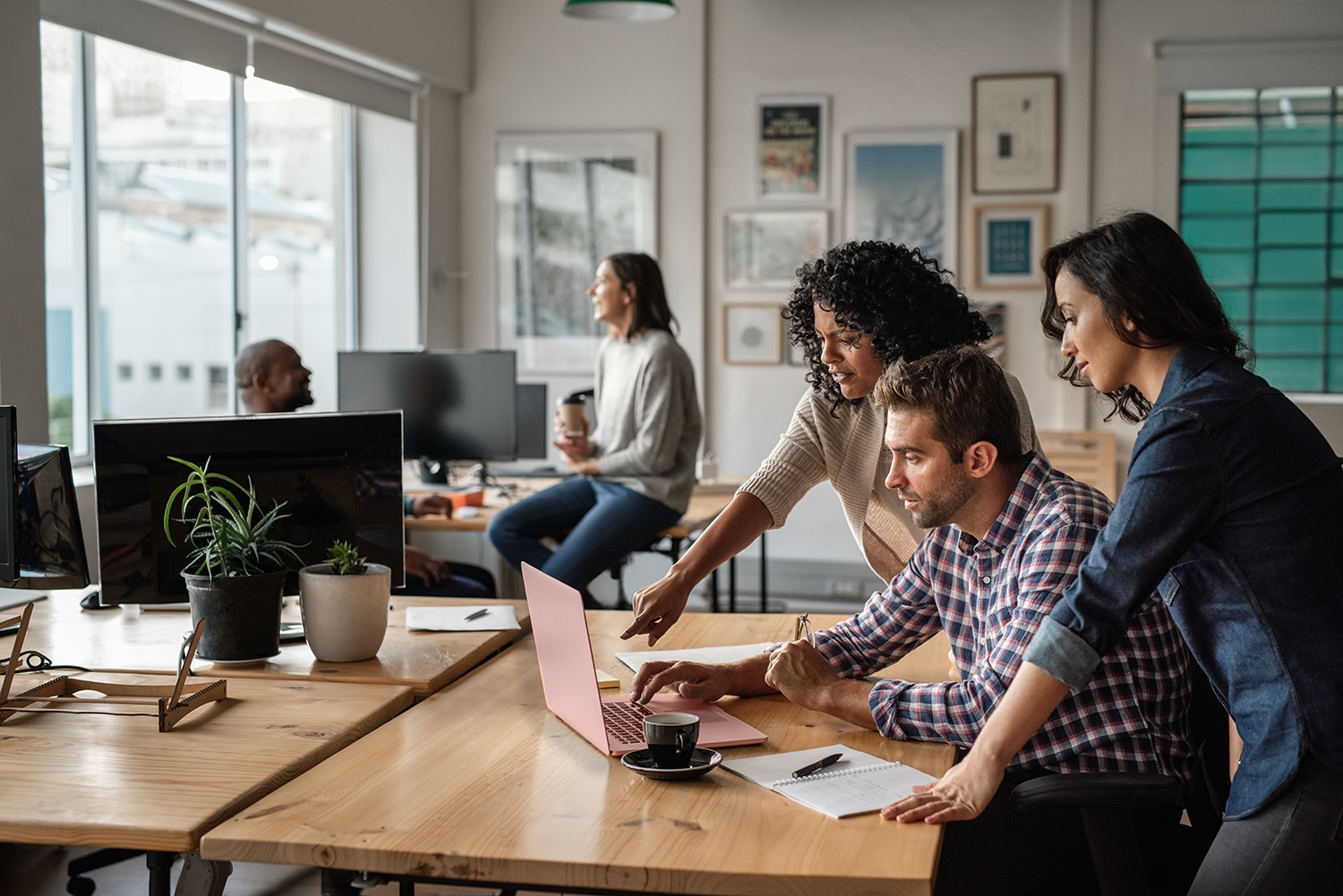}\\

 \end{tabular}
\caption{Professional edits sourced from usertesting.com for the HighResClutter dataset.}
\label{fig:users}
\end{figure*}

Our computational approach to attention-aware photo editing is motivated by real, professional workflows. For this study, we selected 30 high-resolution Adobe Stock photos covering a wide range of themes/genres and containing objects of varying sizes. Object masks were then manually created by segmenting an image region that was not already the main subject of the photograph, and was off-center to the image, so that we could later detect shifts in attention. We ran initial studies on the crowdsourcing platform \url{www.usertesting.com}, asking participants familiar with professional photo editing software to load the provided images and masks into Adobe's Photoshop, and edit the images to make the selected objects ``stand out (become more noticeable)" using any technique but encouraging ``more subtle edits". Participants were presented with 3 image-mask pairs, and submitted their 3 final edits. Results from these studies were manually filtered out if they contained very strong and obvious effects, if they showed no signs of edits, or if they reduced the prominence of the masked object, thus failing to follow instructions. We ran studies until we obtained a total of 5 different valid edits per photo. We collected a total of 150 professional edits (5 participants $\times$ 30 photos, see some examples in Fig.~\ref{fig:users}). We call this the HighResClutter dataset.

\subsection{Fidelity study}

\begin{figure}
\centering
\includegraphics[width=0.9\linewidth]{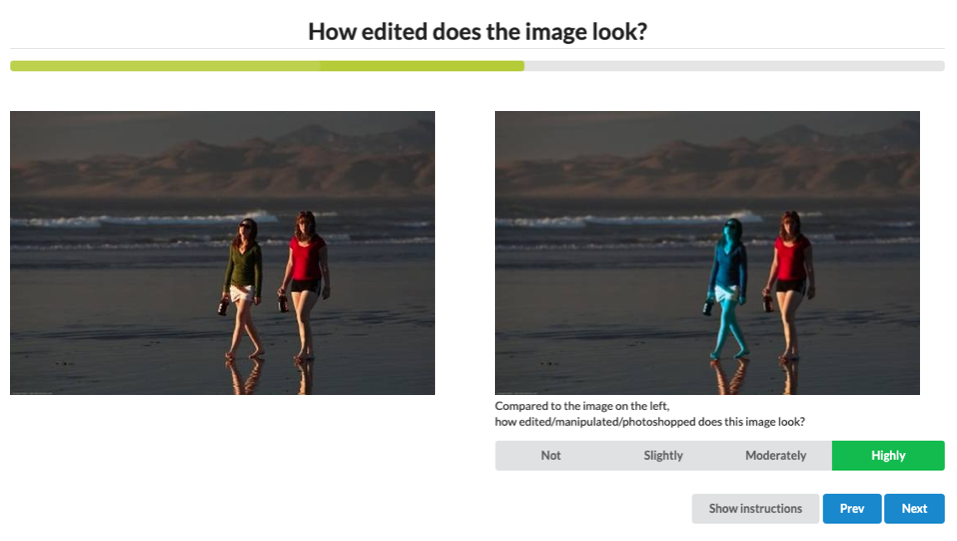} 
\caption{Given a pair of images, an original and another variant, participants rated how edited the right image looked compared to the original.}
\label{fig:fidelitystudy}
\end{figure}

In this study, participants were presented with pairs of images, an original photograph and an edited photograph (using one of multiple automated methods), and were asked to rate ``Compared to the image on the left, how edited/manipulated/photoshopped does this image look?": ``Not", ``Slightly", ``Moderately", or ``Highly" (Fig.~\ref{fig:fidelitystudy}).
Amazon's Mechanical Turk (MTurk) participants were recruited for the study, and assigned either 32 randomly selected images out of the 64 Mechrez dataset images, 25 randomly selected images out of the 50 CoCoClutter dataset images, or all 30 of the HighResClutter images. In all cases, participants were presented with an additional 5 sentinel images (the same for all datasets) which were randomly interspersed throughout the experiment. These sentinel images were chosen to be significantly different from the original images (as in the example in Fig.~\ref{fig:fidelitystudy}), and were used as a quality filter for participant data. The study took approximately 5 minutes to complete, and participants were paid \$1 for their time. We kept track of screen size used, time spent per image pair, and selections made. Participants were filtered out if screen size was less than $1000\times1000$, if screen size was adjusted at any point throughout the experiment, if less than 2 seconds on average were spent per image pair, and if any of the sentinels were rated as not edited. These automatic criteria filtered out about 15\% of the participant data, leaving an average of 25 participant responses per image pair, that we averaged together to obtain a fidelity score per image and automatic method. For the Mechrez dataset, we ran this study for each of OUR, MEC, HAG, and HOR models. For the CoCoClutter dataset, we ran this study for each of OUR, HAG, and HOR models.
For the HighResClutter images, we ran the study on OUR, and 5 individual professional edits. 

\subsection{Realism study}

\begin{figure}
\centering
\includegraphics[width=0.9\linewidth]{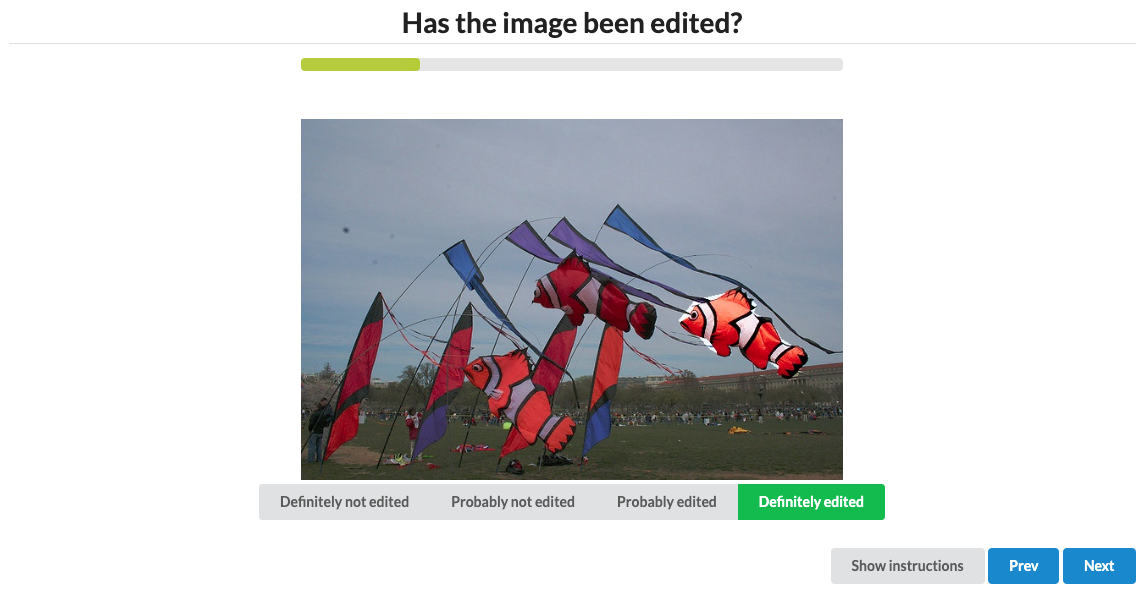} 
\caption{Given a sequence of images, participants rated how edited an image looked. Half of the images shown to participants were originals, and the other half were edited images.}
\label{fig:realismstudy}
\end{figure}

Participants were presented with a sequence of images and were asked to judge ``Has the image been edited?", with possible answers: ``Definitely not edited", "Probably not edited", ``Probably edited", ``Definitely edited" (Fig.~\ref{fig:realismstudy}). Half of the images in the sequence were edited, and the other half of the images were originals. Images were randomly sampled from either the Mechrez, CoCoClutter, or HighResClutter datasets (but all images in a single experiment came from a single dataset). All the images were randomly shuffled, and 5 sentinel images were randomly interspersed throughout the experiment. These sentinel images were chosen to be obviously edited. We collected an average of 30 MTurk participant responses for each image, which after automatic filtering, yielded an average of 20-25 raters per image. We filtered out participants who had screen sizes less than $1000\times600$, those who adjusted the screen size at any point throughout the experiment, spent less than 2 seconds per image on average, and rated any of the 5 sentinels as ``Definitely not edited". Participants spent roughly 2-3 minutes on the task, and were compensated \$0.45-\$0.65 depending on the number of images shown. We tested the same images and models as in the fidelity study. 

\subsection{Attention study}

\begin{figure}
\centering
\includegraphics[width=0.9\linewidth]{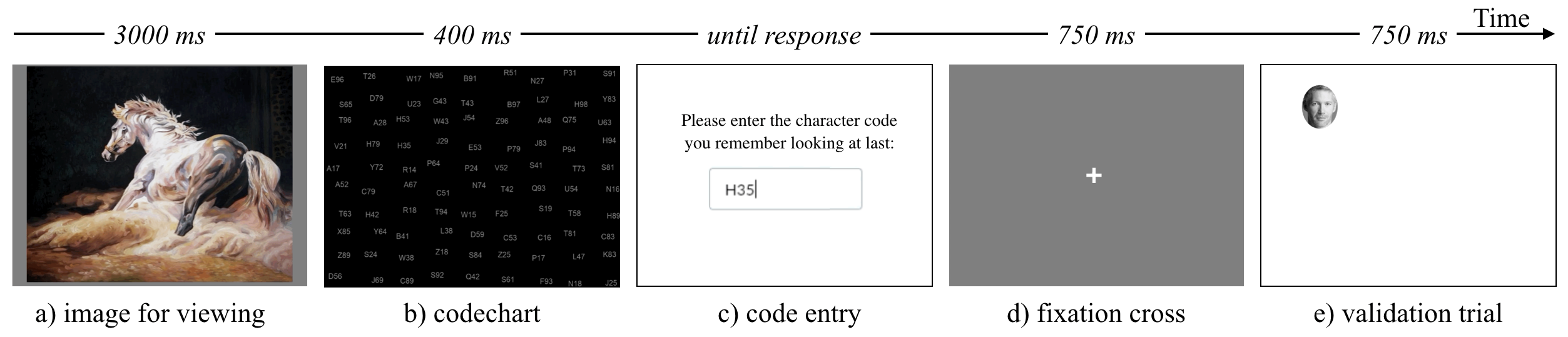}
\caption{CodeCharts user study experiment flow, adapted from \cite{Fosco_2020_CVPR}.}
\label{fig:attentionstudy}
\end{figure}

We used the CodeCharts methodology \cite{Fosco_2020_CVPR}\footnote{Using the code provided at \url{https://github.com/turkeyes/codecharts}} to collect ground truth attention data on all the original and edited images so that we could evaluate whether they successfully shifted viewer attention towards the desired image regions. Participants would be shown an image for 3 seconds, followed by a quickly-flashed chart of alphanumeric triplets (codechart), and they would be asked to report the last code they gazed at. This sequence repeated for many images in a row (30-60 in our experiments), and images were separated by fixation crosses to re-center participant gaze (Fig.~\ref{fig:attentionstudy}). A total of 5 sentinels (validation trials), in the form of cropped faces against a white background, were spaced throughout the image sequence. If the code reported for a sentinel did not overlap with the location of the face, that was considered a failed trial. Any image was also classified as a failed trial if participants entered a code that was nowhere to be found on the corresponding codechart. Participant data was filtered out if any of the sentinels were a failed trial, and if invalid codes were entered on more than 25\% of the other trials. A 6-image tutorial, including 3 natural images and 3 sentinel images preceded the rest of the experiment. Any participant that failed the tutorial, entering an invalid code for 2/6 images, was also filtered out. We filtered out a total of 15-20\% of participants, depending on the experiment, based on these criteria. After filtering, an average of 45-50 gaze points remained per image, for further analysis. We collected attention data for the same images from the Mechrez, CoCoClutter, and HighResClutter datasets for which we collected realism and fidelity scores.

\subsection{Results}

In Tables~\ref{tab:human_mechrez},~\ref{tab:my_label}, and ~\ref{tab:highres-human} are results from the realism, fidelity, and attention user studies on the Mechrez, CoCoClutter, and HighResClutter datasets. Plots corresponding to these numbers for the Mechrez and CocoClutter studies can be found in the main paper, and the ones for the HighResClutter dataset are included in Fig.~\ref{fig:userstudies} here. We note that the differences across the methods were not found to be statistically significant across any of the measures (realism, fidelity, attention) due to the small number of images tested. However, the pattern of results shows that our approach consistently has small standard deviations of realism and fidelity scores, indicating that the produced results are more robust and consistent across different image types and datasets. Moreover, we achieve a balance between realism/fidelity and attention shift, whereas other methods trade off one for the other (this is easier to see from the scatter plots in the main paper than from the tables here). Surprisingly, compared to professionals we are able to shift attention more effectively, however this comes at the cost of reducing realism and fidelity. This suggests that the task we aim to solve is a challenging one even for humans, and that there is indeed a trade off between attention increase and realism. 
 
\begin{table}[h]
     \centering
     \begin{tabular}{|p{1.3cm}|p{1.5cm}|p{1.5cm}|p{1.4cm}p{1.4cm}|p{1.4cm}p{1.4cm}|}
     \hline
     \multirow{2}{*}{Model}& \multirow{2}{*}{Realism} & \multirow{2}{*}{Fidelity} & \multicolumn{2}{c|}{Attention increase} & \multicolumn{2}{c|}{Similarity to mask} \\
       & & & Absolute & Relative & WFB & CC \\ \hline
     OUR & 1.38 (0.58) & 1.75 (0.52) & 0.55 & 12.16 & 8.02 & 21.72 \\
     MEC~\cite{mechrez2019saliency} & 1.47 (0.58) & 1.34 (0.47) & 0.42 & 12.22 & 7.40 & 19.30 \\
     HOR~\cite{mateescu2014attention} & 1.40 (0.70) & 1.46 (0.74) & 1.20 & 18.22 & 7.62 & 21.63 \\
     HAG~\cite{hagiwara2011saliency} & 1.46 (0.63) & 1.86 (0.75) & -0.28 & 6.64 & 6.62 & 16.34 \\
       \hline
     \end{tabular}
     \caption{Results from user studies on Mechrez dataset images. We include the average realism and fidelity scores (with standard deviations, in parentheses) across 25 human raters per image. All values are scaled by 100 apart from realism and fidelity.}
    \label{tab:human_mechrez}
 \end{table}

 \begin{table}[h]
     \centering
     \begin{tabular}{|p{1.3cm}|p{1.5cm}|p{1.5cm}|p{1.4cm}p{1.4cm}|p{1.4cm}p{1.4cm}|}
     \hline
     \multirow{2}{*}{Model} & \multirow{2}{*}{Realism} & \multirow{2}{*}{Fidelity} & \multicolumn{2}{c|}{Attention increase} & \multicolumn{2}{c|}{Similarity to mask} \\
       & & & Absolute & Relative & WFB & CC \\ \hline
     OUR & 1.67 (0.54) & 2.01 (0.34) & 0.44 & 17.08 & 9.57 & 13.43  \\
     HOR~\cite{mateescu2014attention} & 0.98 (0.73) & 1.03 (0.79) & 1.16 & 30.66 & 12.05 & 21.07 \\
     HAG~\cite{hagiwara2011saliency} & 1.50 (0.61) & 2.08 (0.66) & 0.43 & 12.12 & 9.04 & 13.46 \\
       \hline
     \end{tabular}
     \caption{Results from user studies on the CoCoClutter dataset. All values are scaled by 100 apart from realism and fidelity.}
     \label{tab:my_label}
 \end{table}
 
 \begin{table}[h]
     \centering
     \begin{tabular}{|p{1.3cm}|p{1.5cm}|p{1.5cm}|p{1.4cm}p{1.4cm}|p{1.4cm}p{1.4cm}|}
     \hline
     \multirow{2}{*}{Model} & \multirow{2}{*}{Realism} & \multirow{2}{*}{Fidelity} & \multicolumn{2}{c|}{Attention increase} & \multicolumn{2}{c|}{Similarity to mask} \\
       & & & Absolute & Relative & WFB & CC \\ \hline
     OUR & 1.07 (0.50) & 1.67 (0.42) & 11.79 & 149.86 & 7.12 & 17.93 \\
     PROF & 1.42 (0.39) & 1.94 (0.47) & 4.27 & 134.47 & 5.27 & 11.39  \\
       \hline
     \end{tabular}
     \caption{Results from user studies on HighResClutter, where PROF refers to professionals recruited via the www.usertesting.com crowdsourcing platform. All values are scaled by 100 apart from realism and fidelity.}
     \label{tab:highres-human}
 \end{table}
 
 \begin{table}[h]
    \centering
    \resizebox{\textwidth}{!}{%
    \begin{tabular}{|c|ccc|cc|cc|}
    \hline
    \multirow{2}{*}{Model} & \multicolumn{3}{c|}{LPIPS $\downarrow$} & \multicolumn{2}{c|}{Saliency increase $\uparrow$} & \multicolumn{2}{c|}{Similarity to mask $\uparrow$} \\
     & Full & BG & FG & Absolute & Relative & WFB & CC  \\ \hline
OUR &       6.16 &      5.35 &      0.82 &     4.19 &     38.82 &  9.13 &  25.64 \\
PROF &     6.93 &      6.58 &      0.36 &     1.74 &     15.77 &  8.0 &  21.42 \\
PROF-range & [1.15, 16.89]& [0.73, 16.74]& [0.04, 0.83] & [-0.46, 4.31] & [-4.33, 35.39] & [7.03, 9.12] & [17.72, 25.53]\\ 
\hline
    \end{tabular}}
    \caption{Scores (scaled by 100) from computational measures on HighResClutter.}
    \label{tab:comp_HR}
\end{table}
 
 \begin{figure}[!ht]
\centering
\includegraphics[width=0.9\linewidth]{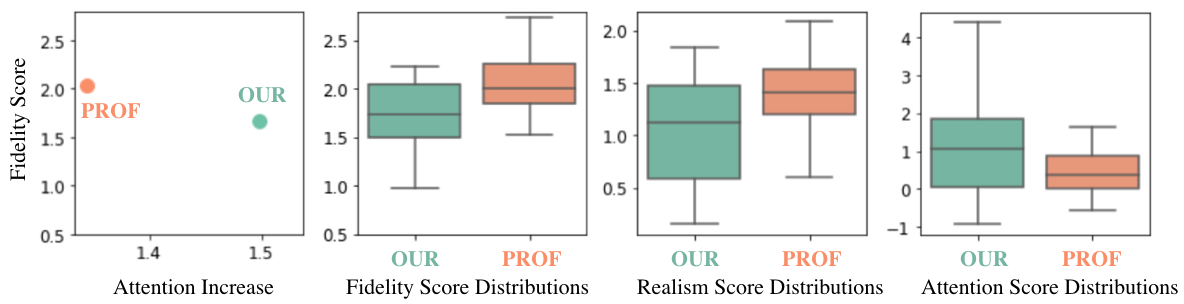}
\caption{Results of user studies run on three separate crowdsourcing tasks - measuring image fidelity, realism, and human attention - on 30 high-resolution images. The fidelity, realism, and attention score distributions are visualized as box plots. We compare to the results of professionals (PROF) recruited via the www.usertesting.com crowdsourcing platform. Compared to professionals, we are able to increase attention more effectively, but at the cost of reducing realism and fidelity.} 
\label{fig:userstudies}
\end{figure}


\section{Evaluations}
\subsection{Saliency model} 
We used a state-of-the-art saliency model~\cite{Fosco_2020_CVPR} to evaluate the ability of each model to shift computational attention. At the time of submission, this model was second overall on the LSUN 2017 challenge leaderboard\footnote{\url{https://competitions.codalab.org/competitions/17136\#results}} as of 09/24/2019, and leading in terms of NSS score. Our model choice criteria included having a top-performing model with a small footprint, so that our final model, in which saliency would be one of multiple sub-components, would be fit for practical use and not bulky. In contrast, other top-performing models are quite bulky: SAM~\cite{cornia2018sam} at 70M parameters, DeepGaze II~\cite{kummerer2016deepgaze} at 40M parameters. MD-SEM~\cite{Fosco_2020_CVPR} has 30M parameters while outperforming these models on the LSUN challenge, making it more attractive as a building block within larger systems. 

For completeness, we also provide the absolute and relative saliency increases of all methods using the DeepGaze model~\cite{gatys2017guiding} in Table~\ref{tab:comp_deepgaze}. Under this alternative saliency model, our approach still outperforms the alternatives on both datasets evaluated. 

\begin{table}[h]
    \centering
    \begin{tabular}{|c|cc|}
    \hline
    \multirow{2}{*}{Model} & \multicolumn{2}{c|}{Saliency increase} \\
     &Absolute & Relative \\ \hline
OUR &       31.96 &	2.96   \\
MEC	& 16.60 &	1.52 \\
HAG	& 17.68 &	16.06 \\
HOR	& 3.12 &	0.27 \\
GAT	& 10.41 &	0.95 \\
\hline
\end{tabular}
\quad
\begin{tabular}{|c|cc|}
\hline
\multirow{2}{*}{Model} & \multicolumn{2}{c|}{Saliency increase} \\
&Absolute & Relative \\ \hline
OUR	& 17.20	& 15.11 \\
HAG	& 15.28	& 1.31 \\
HOR	& 1.30	& 0.09 \\
GAT	& 8.13	& 0.68 \\
\hline
    \end{tabular}
    \caption{Absolute and relative saliency increase (scaled by 100) using DeepGaze as a measure of computational saliency on the Mechrez (left) and CoCoClutter (right) datasets.}
    \label{tab:comp_deepgaze}
    \vspace{-2em}
\end{table}

\subsection{Ablation studies}\label{ssec:ablations}
Here we report on the influence of different components of our model to its final performance. First, we consider which set of parametric transformations to use. Second, we consider what happens if we only predict parametric transformations for the foreground or background of the image, instead of both. Third, we discuss how the order of parameter application influences the results. 
As it is not immediately obvious how to balance the trade-off between realism and attention shift across our different ablations, we identify the five best models according to three criteria (LPIPS, absolute and relative saliency increase), and choose a high-performing model across all criteria. Results for all the following ablations refer to Table \Acomparison ~in the main paper, where the best performing models are highlighted in green (darker is better), and the least performing models are highlighted in red (darker is worst).

\begin{figure*}[!ht]
\centering
\setlength{\imwidth}{0.3\linewidth}
\setlength{\tabcolsep}{2pt}
\begin{tabular}{ccc}
Input & Mask  & tone+color
\\
 \includegraphics[width=\imwidth]{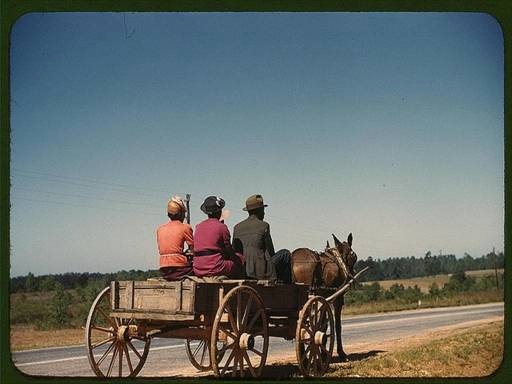}&
 \includegraphics[width=\imwidth]{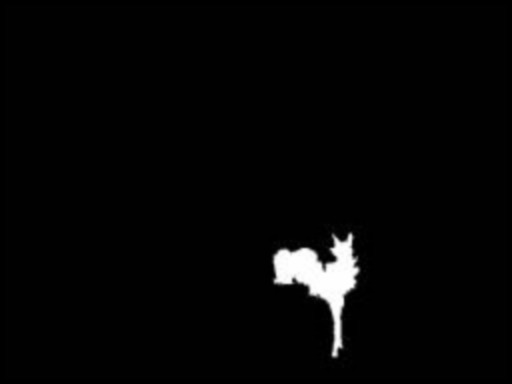}&
 \includegraphics[width=\imwidth]{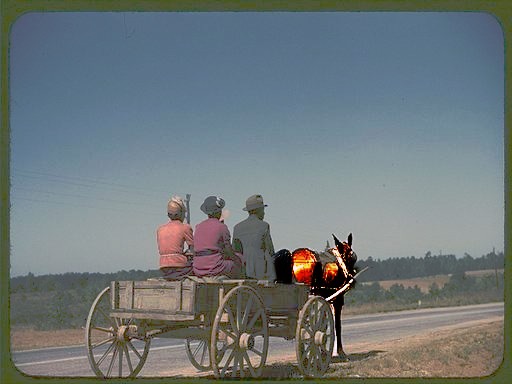}\\
  \includegraphics[width=\imwidth]{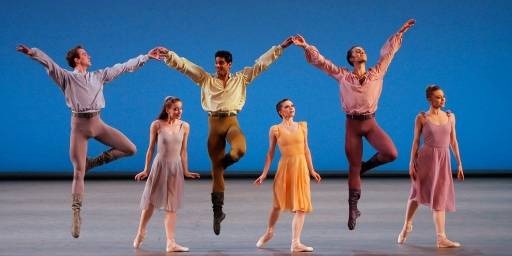}&
 \includegraphics[width=\imwidth]{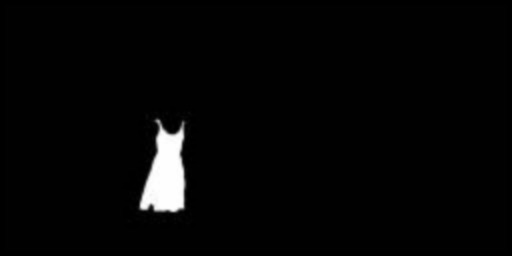}&
 \includegraphics[width=\imwidth]{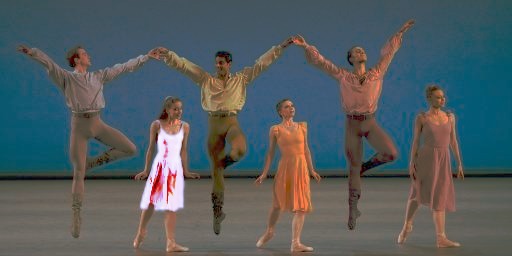}
 \\
Input & Mask  & color 
\\
 \includegraphics[width=\imwidth]{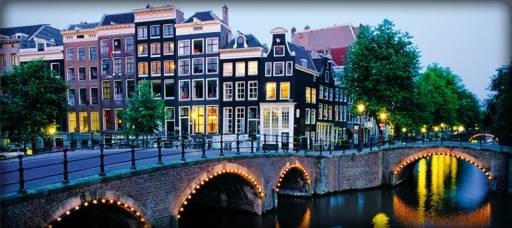}&
 \includegraphics[width=\imwidth]{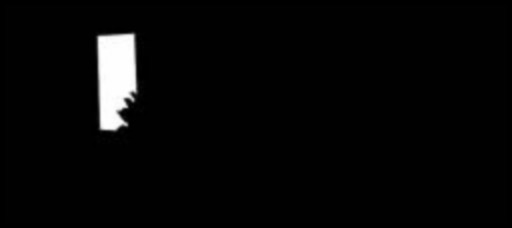}&
 \includegraphics[width=\imwidth]{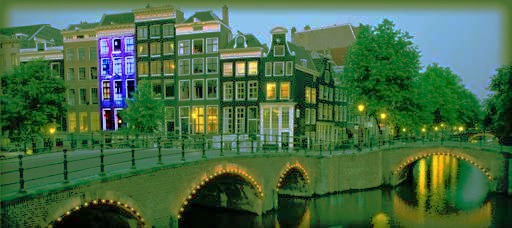}\\
  \includegraphics[width=\imwidth]{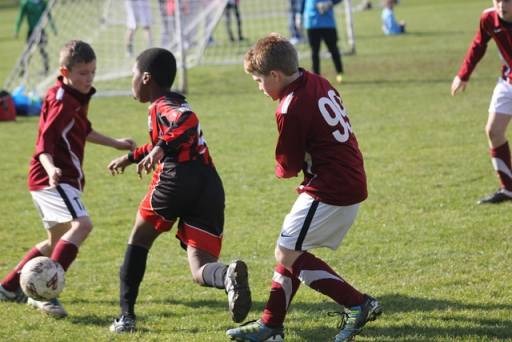}&
 \includegraphics[width=\imwidth]{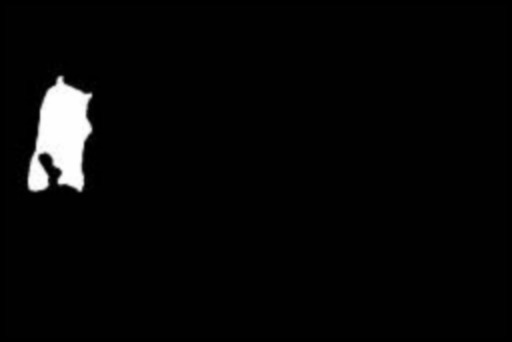}&
 \includegraphics[width=\imwidth]{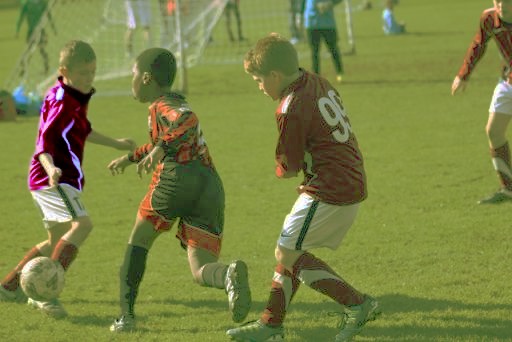}
 \\
 Input & Mask  & our + saturation 
 \\
  \includegraphics[width=\imwidth]{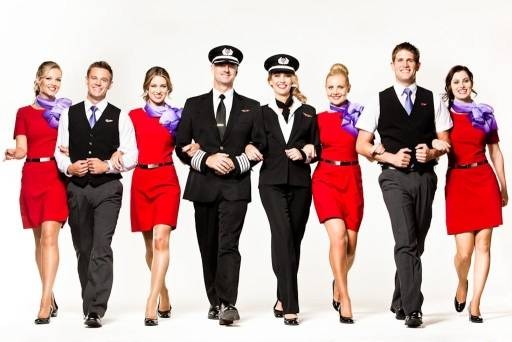}&
    \includegraphics[width=\imwidth]{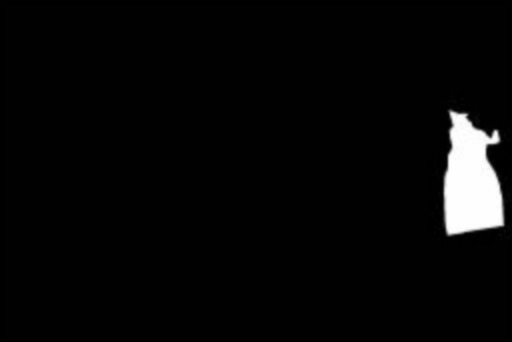}&
      \includegraphics[width=\imwidth]{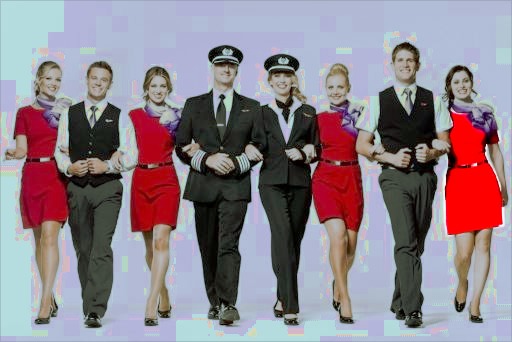}\\
        \includegraphics[width=\imwidth]{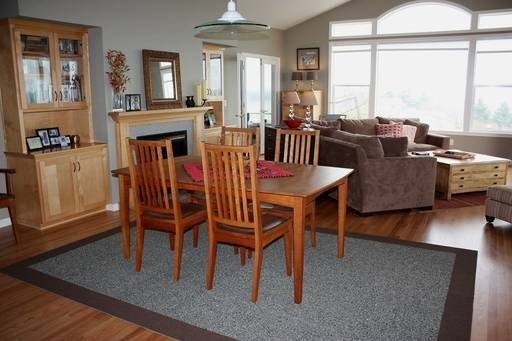}&
    \includegraphics[width=\imwidth]{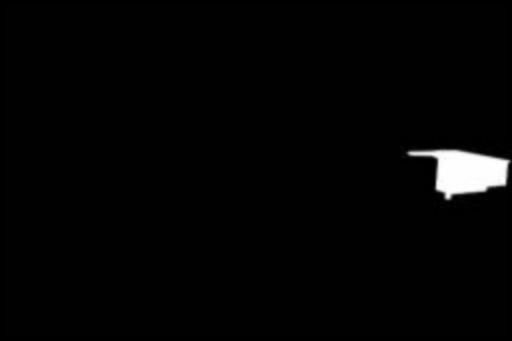}&
      \includegraphics[width=\imwidth]{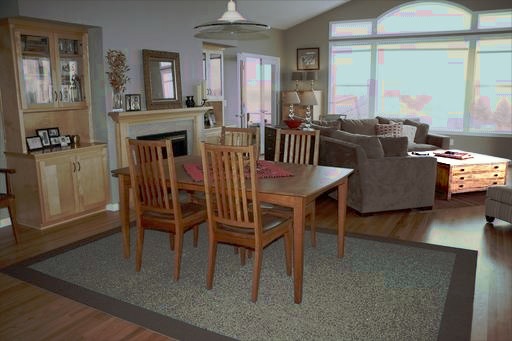}\\
 \end{tabular}
\caption{Artifacts created by some of the parameter ablations experiments.}
\label{fig:ablations}
\end{figure*}

\textbf{Parameter ablations:} Tone and color curve adjustments are two of our most powerful transformations as each is defined by multiple parameters describing a piece-wise linear function. 
If excluded (`sharp+exp+cont' in Table \Acomparison ~in the main paper), the model achieves only a small increase in saliency and small LPIPS values, indicating that the generated image is very similar to the original. On the other hand, a model that only uses tone and color adjustment (`tone+color') produces a significant increase in both saliency and LPIPS values. 
Generated images from this approach often look unrealistic as the network overuses these parameters to minimize the attention loss. A similar but smaller effect is noticeable when using only color adjustment (`color'). 
This suggests that combining tone and color with subtler transformations such as contrast, exposure and sharpening (`our') gives more freedom to the network for achieving high saliency shifts while maintaining image realism. 
We also consider adding saturation to our model (`our+saturation'), and find an increase in saliency corresponding to increased flexibility for our network. However, saturation often introduces destructive artifacts~\cite{mateescu2014attention}, 
resulting in very high LPIPS scores. Figure~\ref{fig:ablations} displays examples of such artifacts. 

Finally, we compute the mean of each predicted parameter value over the entire CoCoClutter validation set, and apply the resulting set of parameters to the input images (`Fixed parameters'). The motivation is to evaluate if our network learns content-aware transformations, or if similar performance is achievable by applying the same transformation independently of the image content. This approach indeed achieves a shift in saliency, showing such a set of transformations can generalize (e.g., brightening the foreground and darkening the background). However, this saliency shift is significantly weaker than `our', suggesting that our network is able to adapt a set of transformations to each image. 

\textbf{FG/BG ablations:} Our network predicts two sets of parameters, one for the foreground, and one for the background. Is similar performance achievable using only one set of parameters? Table \Acomparison ~(fg/bg ablations in the main paper) shows that only applying parameters to the background (`bg-only') leads to marginal saliency increase. In contrast, modifying the foreground while leaving the background untouched, performs better, but still fails to outperform the setting with both sets of parameters. Note that a lower LPIPS value is obtained by `fg-only' because when the background, which covers a larger area in the image, is left untouched, the final result is more similar to the input image.

\textbf{Order ablations:} We test 4 configurations in terms of order of parameter application, presented in Table \Acomparison ~in the main paper as `order of application ablations', where `sha' is sharpening, `exp' is exposure, `con' is contrast, `ton' is tone adjustment and `col' is color adjustment. Changing the order that parameters are applied in affects the final images generated. Some orders achieve higher saliency increases compared to others, but compromise in terms of image realism (higher LPIPS values). We chose the ordering that achieves the best trade off between saliency shift and realism. 
While we did not extensively test all possible order combinations, we leave the task of automatically finding such an optimal ordering for future work. 

\subsection{Model architectures}
Tables~\ref{tab:tab6},\ref{tab:tab7} summarize our generator and discriminator architectures.

\begin{table}[!htb]
    \begin{subtable}{.5\linewidth}
      \centering
    \begin{tabular}{|l|r|r|r|c|l|}
        \hline
         Layer & \#Filters & Size & Stride & InstNorm & Act.   \\
         \hline
         Conv.   & 64  & $7\times7$ & 2 & \checkmark & LReLU \\
         Conv.   & 128 & $3\times3$ & 2 & \checkmark & LReLU \\
         Conv.   & 256 & $3\times3$ & 2 & \checkmark & LReLU \\
         Conv.   & 512 & $3\times3$ & 2 & \checkmark & LReLU \\
         Conv.   & 1024 & $3\times3$ & 2 & - & LReLU \\
         AvgP.   & - & - & - & - & - \\
         \hline
    \end{tabular}
    \end{subtable}%
    \begin{subtable}{.5\linewidth}
      \centering
    \begin{tabular}{|c|c|c|}
        \hline
         Layer & \#Neurons & Act.   \\
         \hline
         FC.     & 128 & LRelu\\ 
         FC.     & 128 & LRelu\\ 
         FC.(Sharp) & 1 & Tanh\\ 
         FC.(Exposure)     & 1 & Tanh\\ 
         FC.(Contrast)     & 1 & Tanh\\ 
         FC.(Tone curve)     & $L$ & Sigmoid\\ 
         FC.(Color curve)     & $3L$ & Sigmoid\\ 
         \hline
    \end{tabular}
    \end{subtable} 
    \caption{(Left): Shared convolutional part of $G_E$. (Right): Specialized densely connected head for predicting foreground and background parameters. `Conv.' is convolutional layer; `FC.' is fully connected layer; `AvgP.' is global average pooling; `InstNorm' is instance normalization; `Act.' is activation function. `LReLU' denotes Leaky ReLU with a factor of 0.2.}
    \label{tab:tab6}
\end{table}

\begin{table}[!htb]
    \begin{subtable}{.5\linewidth}
      \centering
    \begin{tabular}{|l|r|r|c|l|}
        \hline
         Layer & \#Filters/\#Neurons & Size & Stride & Act.   \\
         \hline
         Conv.   & 64  & $4\times4$ & 2  & LReLU \\
         Conv.   & 128 & $4\times4$ & 2  & LReLU \\
         Conv.   & 256 & $4\times4$ & 2 & LReLU \\
         Conv.   & 256  & $4\times4$ & 1  & LReLU \\
         FC. & 128 & - & -  & LReLU\\
         FC. & 1 & - & - & None\\
         
         \hline
    \end{tabular}
    \end{subtable}%
    \begin{subtable}{.5\linewidth}
      \centering
 \begin{tabular}{|c|c|c|}
        \hline
         Layer & \#Neurons & Act.   \\
         \hline
         FC.     & 256 & LRelu\\ 
         FC.     & 256 & LRelu\\ 
         FC.     & 256 & LRelu\\ 
         FC.     & 256 & LRelu\\ 
         FC.     & 10 & None\\ 
        \hline
    \end{tabular}
    \end{subtable} 
\caption{(Left): Architecture of our discriminator. We use a Multi-Scale discriminator~\cite{wang2018high} of scale 3 as it has proven to provide better results with GANs.  `Conv.' is convolutional layer; `FC.' is fully connected layer; `LReLU' denotes Leaky ReLU with a factor of 0.2. (Right): Architecture of the encoder $\text{ENC}$ used to reconstruct the latent vector $z$ in the multi-style setting.}
\label{tab:tab7}
\end{table}

\subsection{Qualitative results}

\begin{figure*}[!ht]
\centering
\setlength{\imwidth}{0.16\linewidth}
\setbox1=\hbox{\includegraphics[width=\imwidth]{Ims/compFig/mask0.jpg}}
\setlength{\tabcolsep}{1pt}
\begin{tabular}{cccccc}
Input & Our & MEC~\cite{mechrez2019saliency} & HAG~\cite{hagiwara2011saliency} & HOR~\cite{mateescu2014attention} & GAT~\cite{gatys2017guiding} 
\\
 \includegraphics[width=\imwidth]{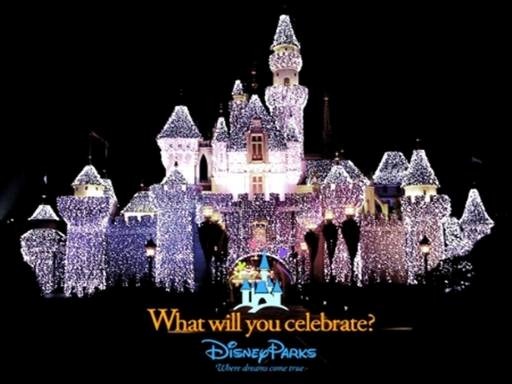}\llap{{\raisebox{0.44\imwidth}{\includegraphics[cfbox=red, width=0.4\imwidth]{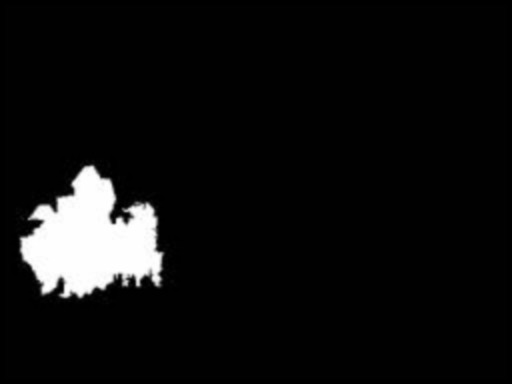}}}}&
 \includegraphics[width=\imwidth]{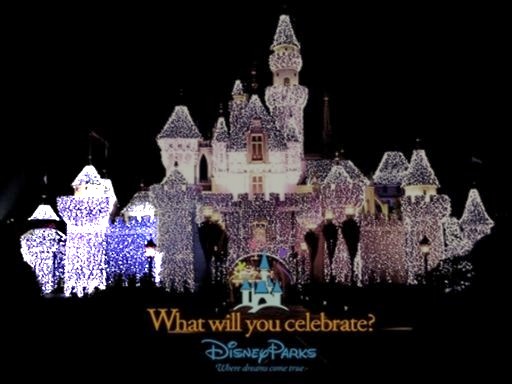} &
 \includegraphics[width=\imwidth]{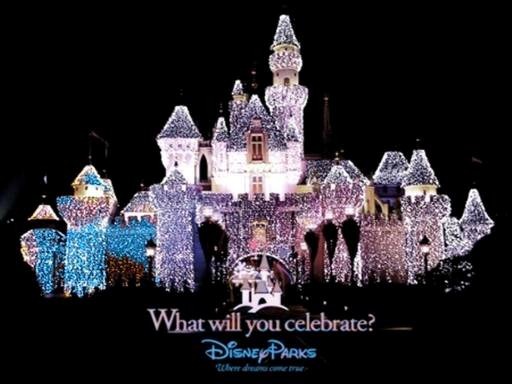} &
 \includegraphics[width=\imwidth]{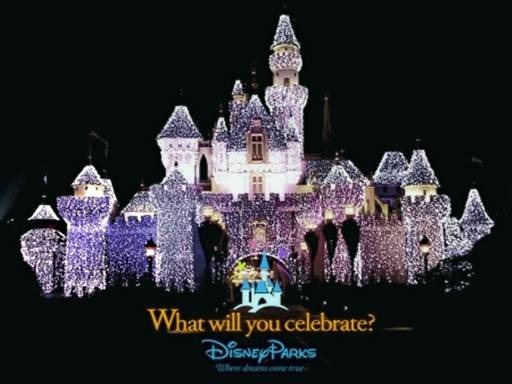} &
 \includegraphics[width=\imwidth]{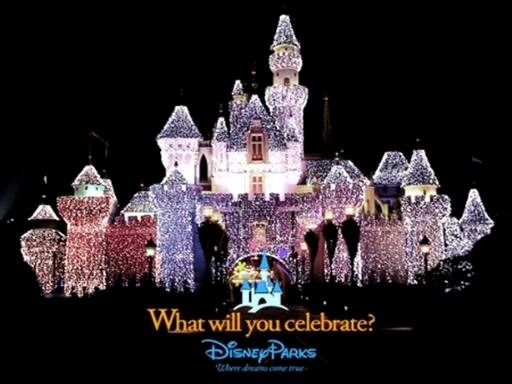} &
 \includegraphics[width=\imwidth]{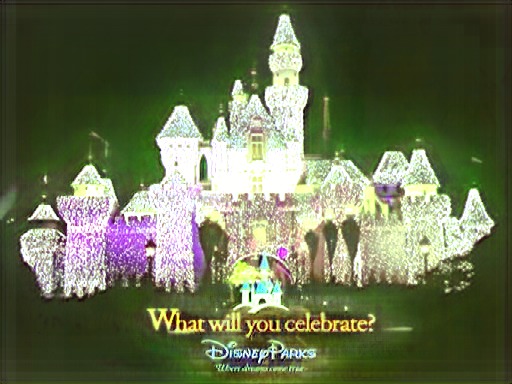}
 \\
 \includegraphics[width=\imwidth]{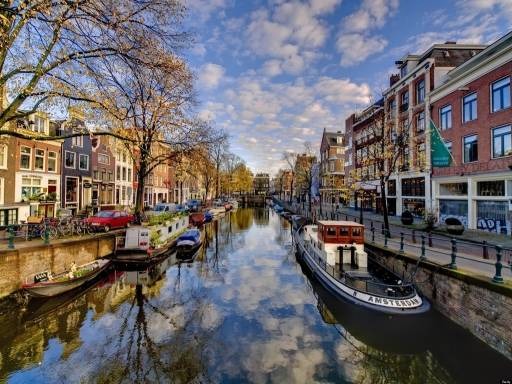}\llap{\makebox[\wd1][l]{\raisebox{0.44\imwidth}{\includegraphics[cfbox=red, width=0.4\imwidth]{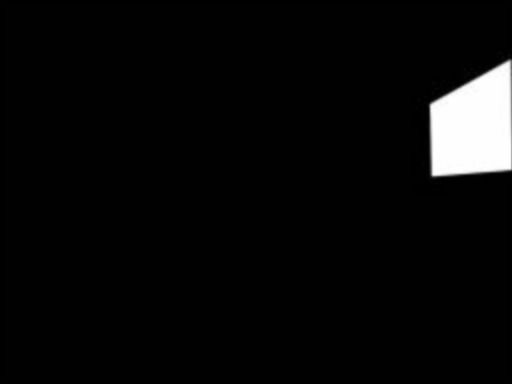}}}}&
 \includegraphics[width=\imwidth]{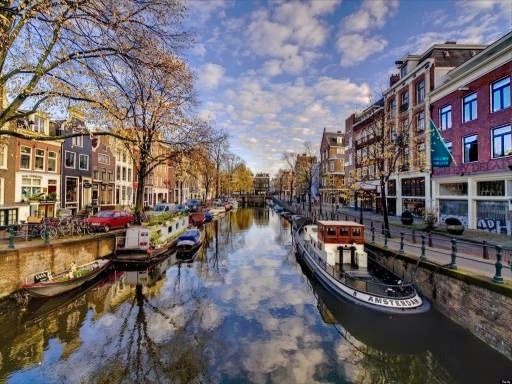}&
 \includegraphics[width=\imwidth]{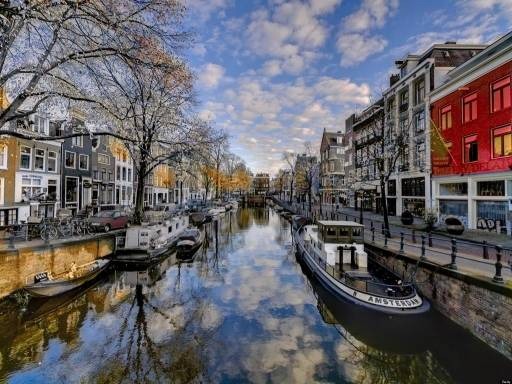} &
 \includegraphics[width=\imwidth]{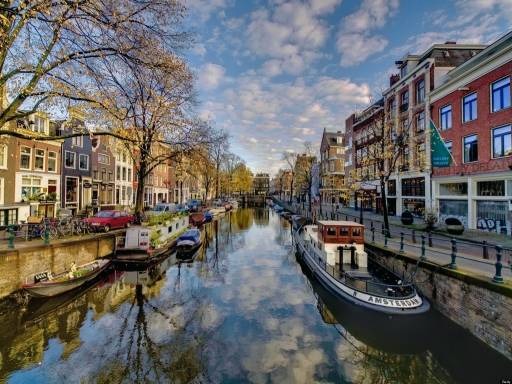} &
 \includegraphics[width=\imwidth]{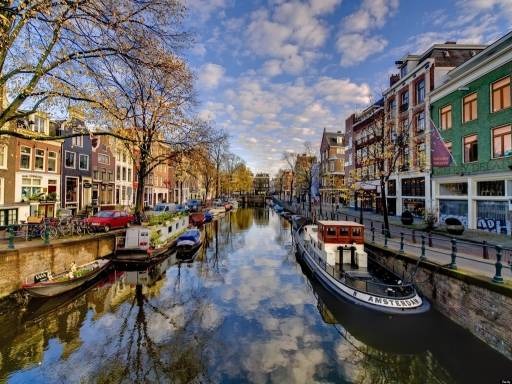} &
 \includegraphics[width=\imwidth]{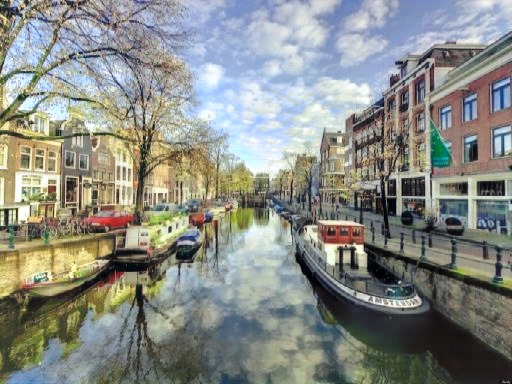}
 \\
 \includegraphics[width=\imwidth]{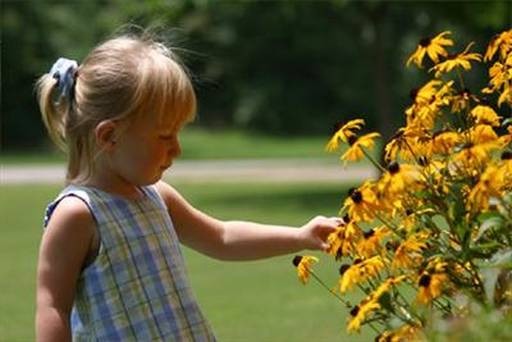}\llap{{\raisebox{0.4\imwidth}{\includegraphics[cfbox=red, width=0.4\imwidth]{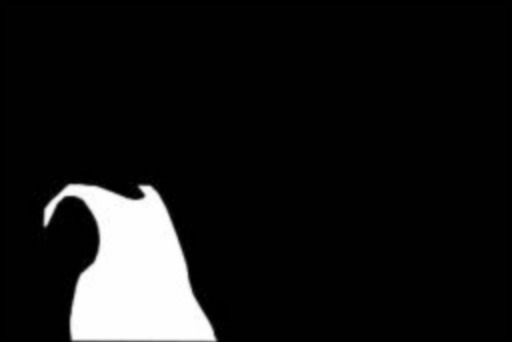}}}}&
 \includegraphics[width=\imwidth]{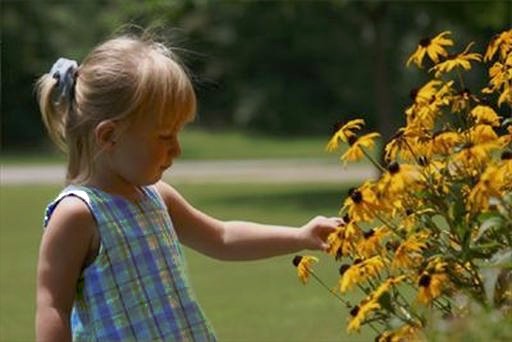}&
 \includegraphics[width=\imwidth]{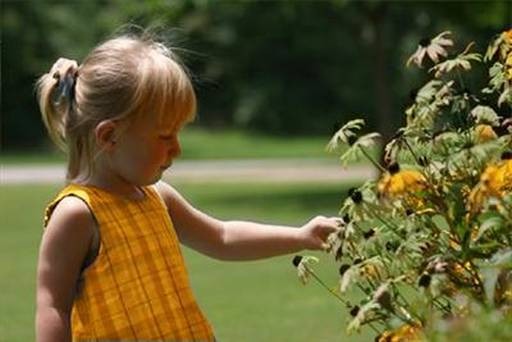} &
 \includegraphics[width=\imwidth]{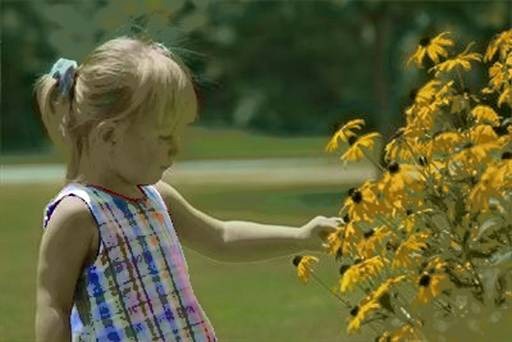} &
 \includegraphics[width=\imwidth]{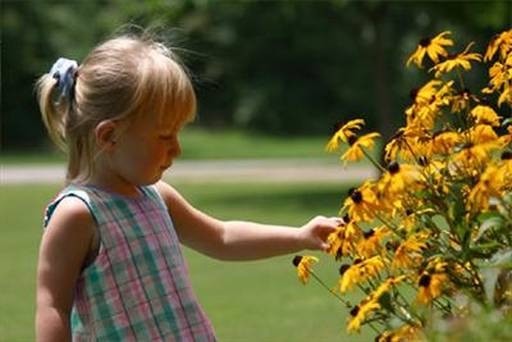} &
 \includegraphics[width=\imwidth]{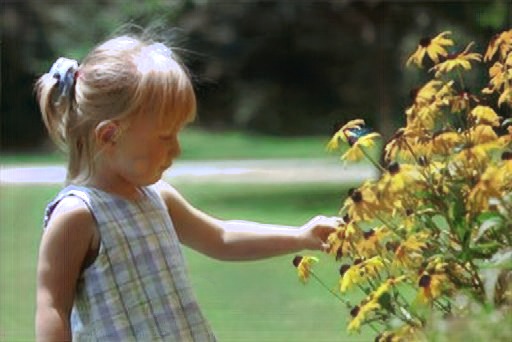}
 \\
 \includegraphics[width=\imwidth]{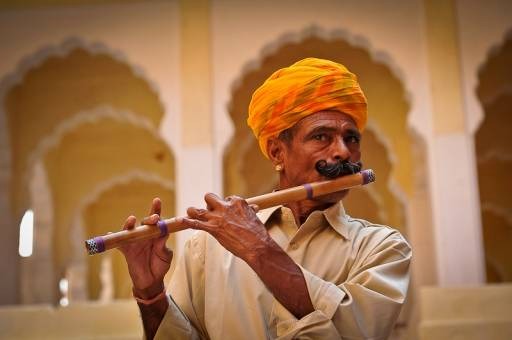}\llap{\makebox[\wd1][l]{\raisebox{0.4\imwidth}{\includegraphics[cfbox=red, width=0.4\imwidth]{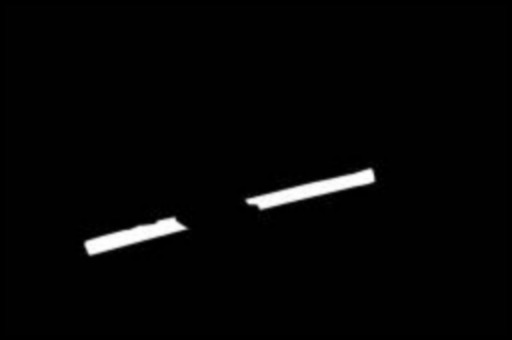}}}} &
 \includegraphics[width=\imwidth]{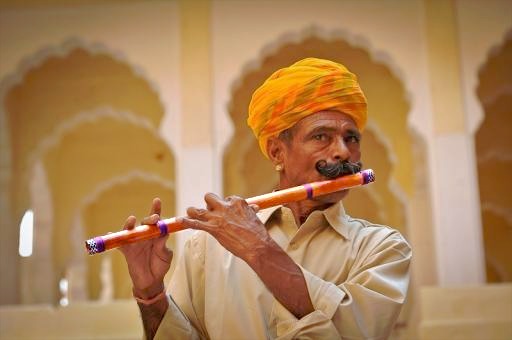}&
 \includegraphics[width=\imwidth]{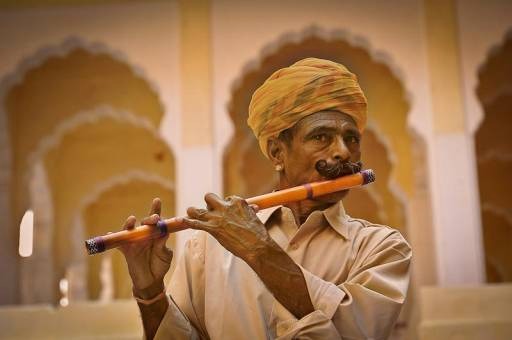} &
 \includegraphics[width=\imwidth]{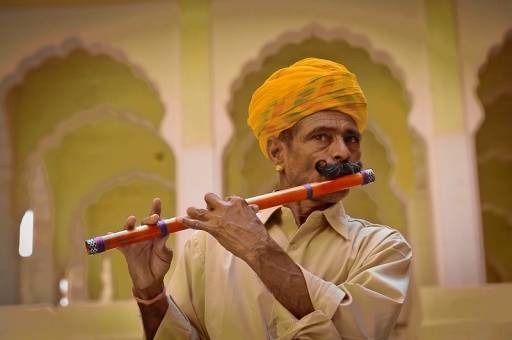} &
 \includegraphics[width=\imwidth]{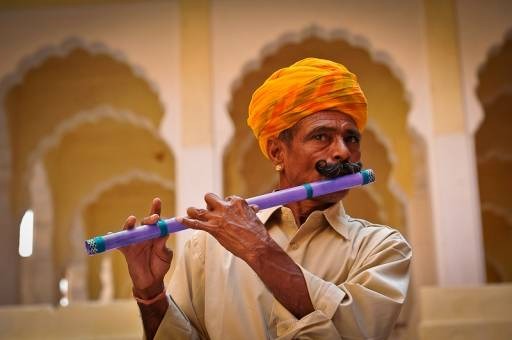} &
 \includegraphics[width=\imwidth]{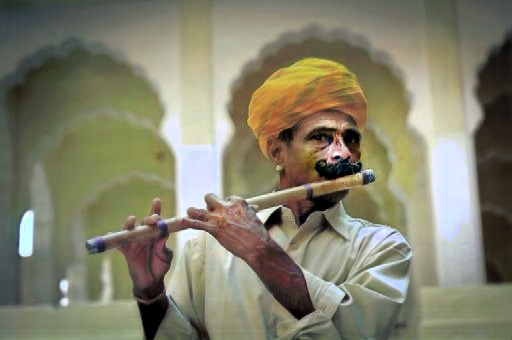}
 \\
 \includegraphics[width=\imwidth]{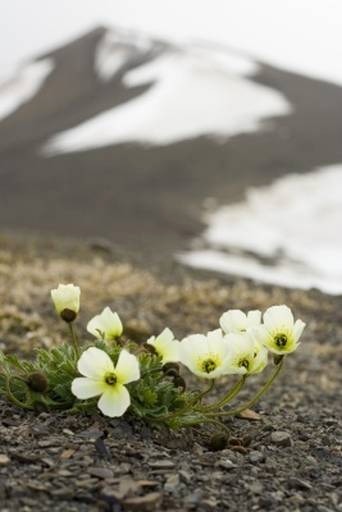}\llap{\makebox[\wd1][l]{\raisebox{0.9\imwidth}{\includegraphics[cfbox=red, width=0.4\imwidth]{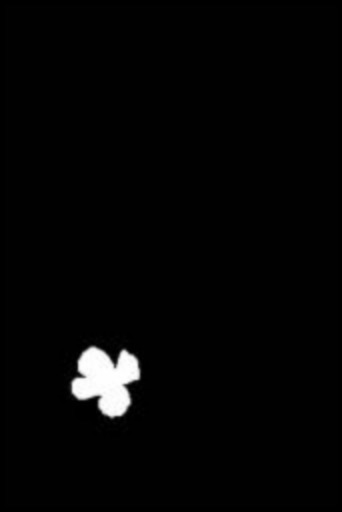}}}}  &
  \includegraphics[width=\imwidth]{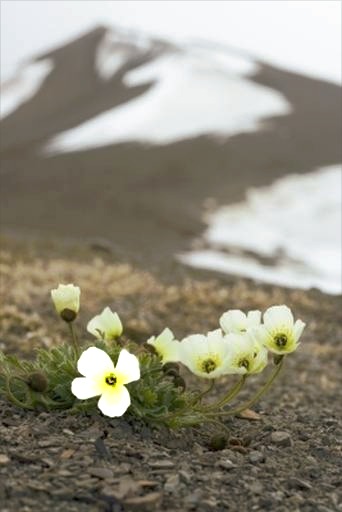} &
 \includegraphics[width=\imwidth]{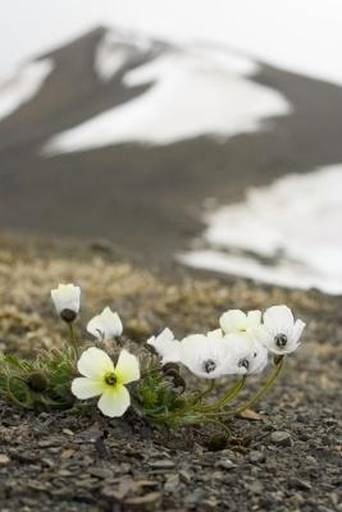} &
 \includegraphics[width=\imwidth]{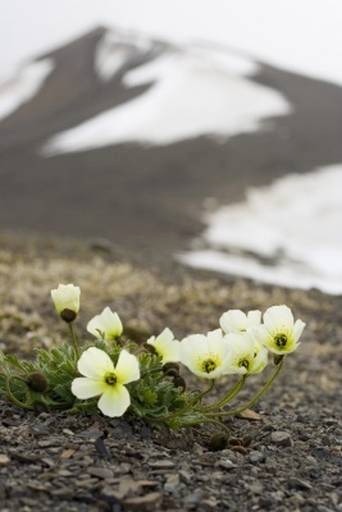} &
 \includegraphics[width=\imwidth]{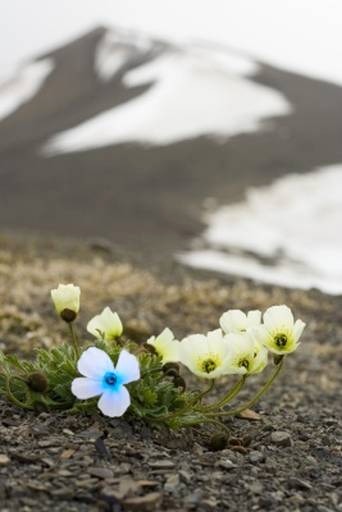} &
 \includegraphics[width=\imwidth]{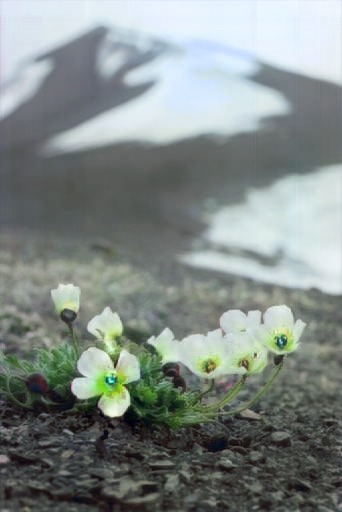}
 \\
 \includegraphics[width=\imwidth]{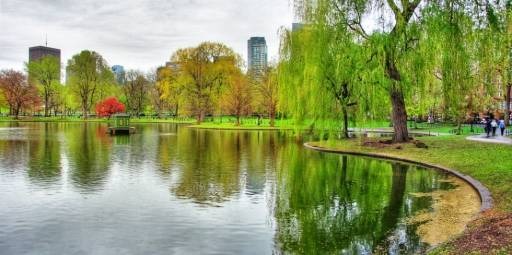}\llap{\makebox[\wd1][l]{{\includegraphics[cfbox=red, width=0.4\imwidth]{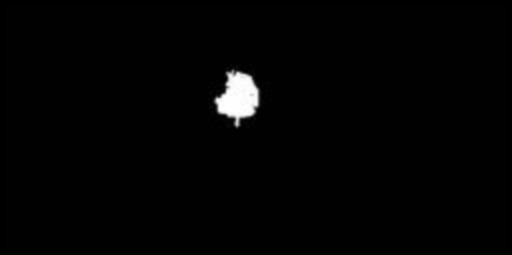}}}} &
 \includegraphics[width=\imwidth]{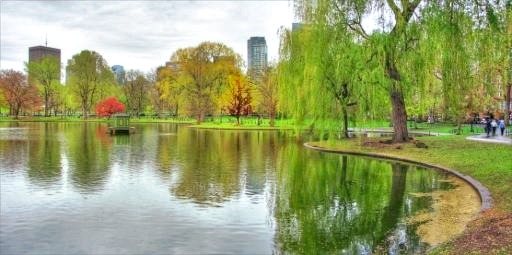}&
 \includegraphics[width=\imwidth]{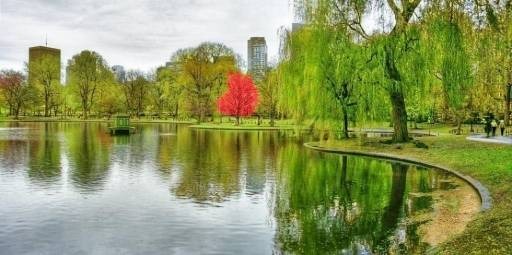} &
 \includegraphics[width=\imwidth]{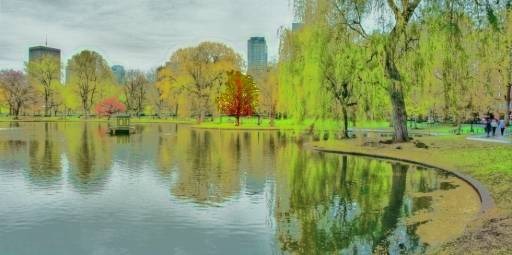} &
 \includegraphics[width=\imwidth]{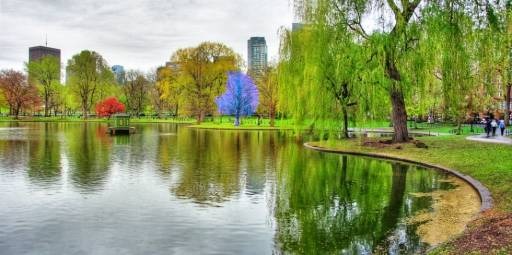} &
 \includegraphics[width=\imwidth]{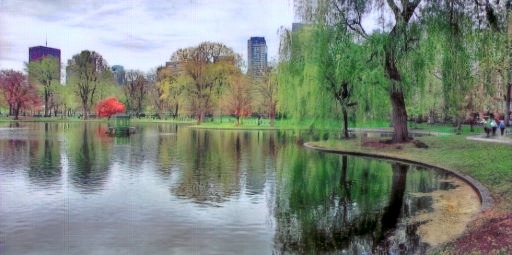}
 \\
 \includegraphics[width=\imwidth]{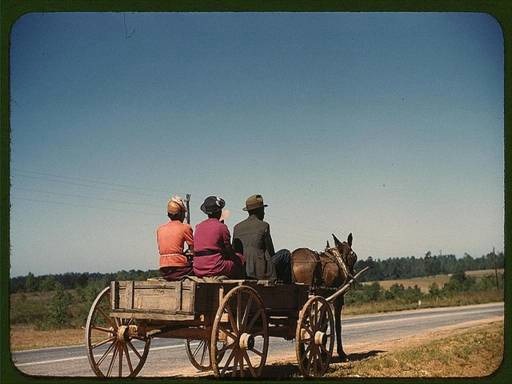}\llap{\makebox[\wd1][l]{\raisebox{0.4\imwidth}{\includegraphics[cfbox=red, width=0.4\imwidth]{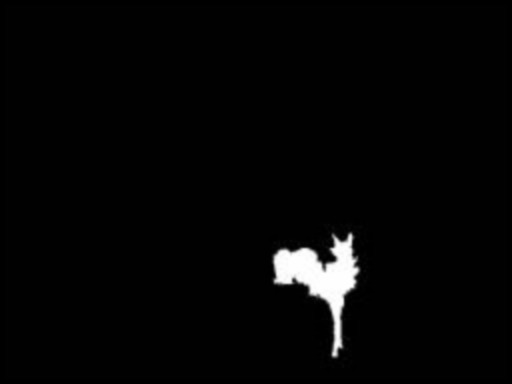}}}}&
  \includegraphics[width=\imwidth]{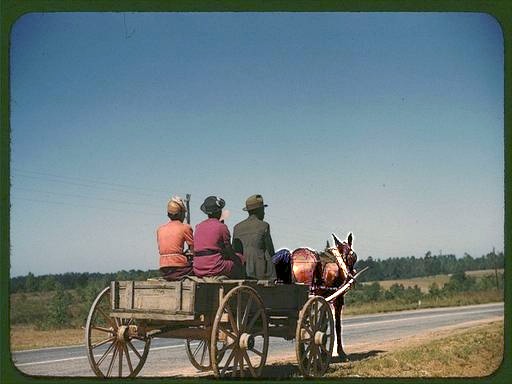} &
 \includegraphics[width=\imwidth]{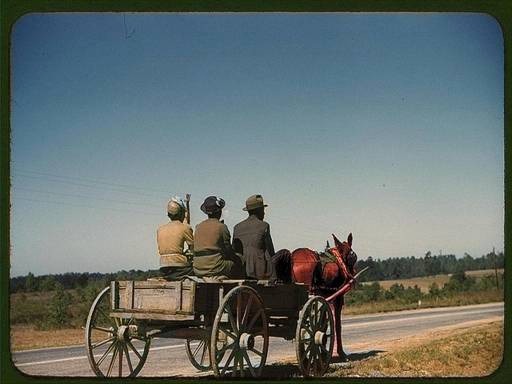} &
 \includegraphics[width=\imwidth]{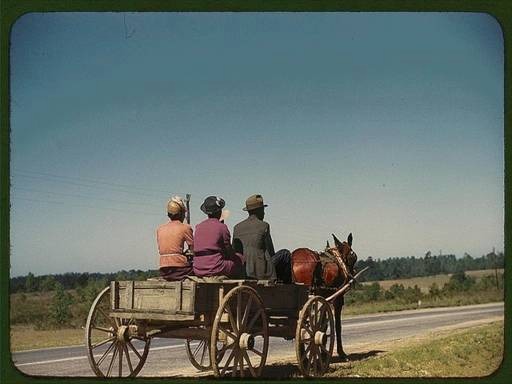} &
 \includegraphics[width=\imwidth]{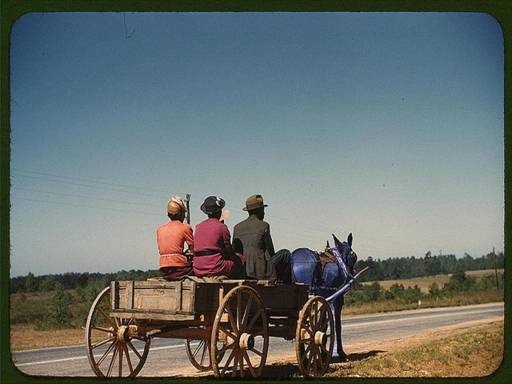} &
 \includegraphics[width=\imwidth]{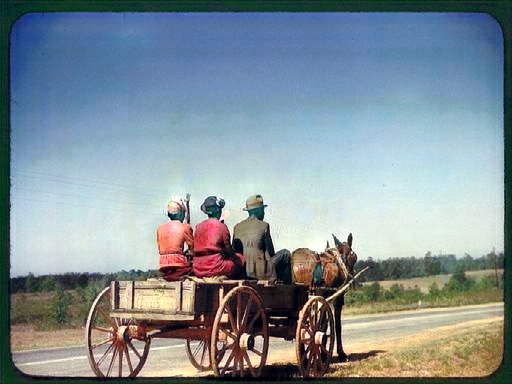}
 \\
 \includegraphics[width=\imwidth]{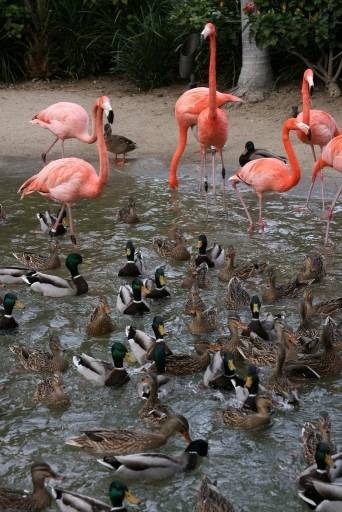}\llap{\makebox[\wd1][l]{{\includegraphics[cfbox=red, width=0.4\imwidth]{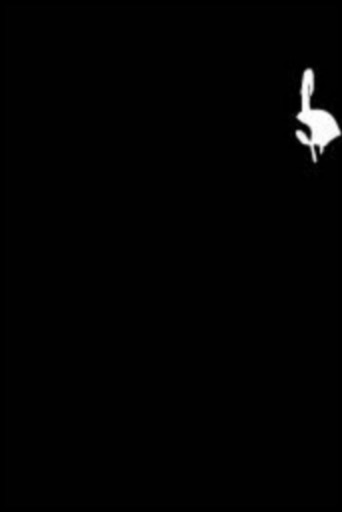}}}} &
 \includegraphics[width=\imwidth]{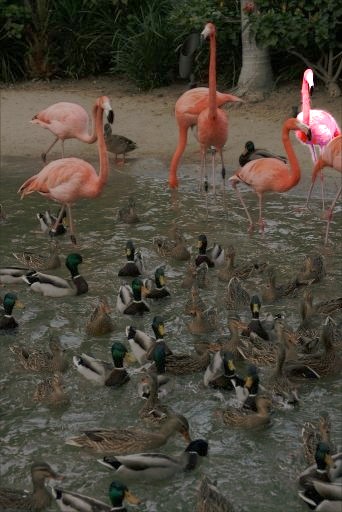}&
 \includegraphics[width=\imwidth]{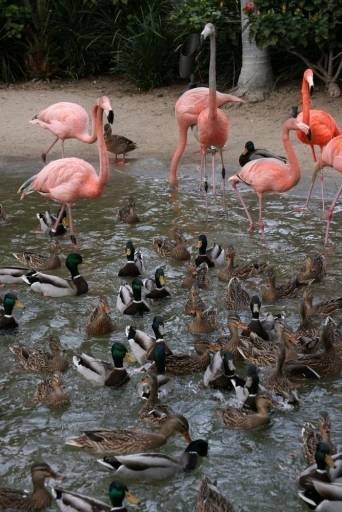} &
 \includegraphics[width=\imwidth]{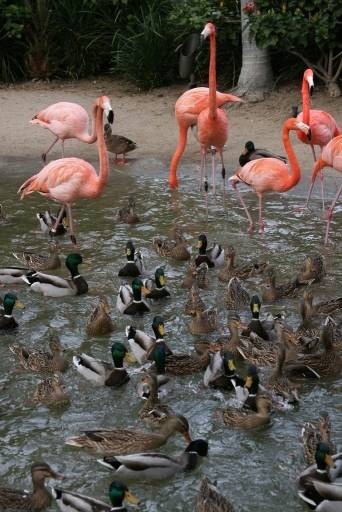} &
 \includegraphics[width=\imwidth]{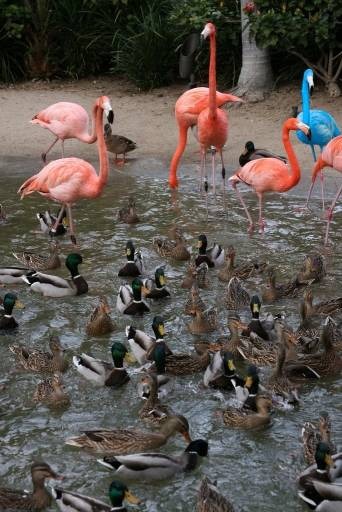} &
 \includegraphics[width=\imwidth]{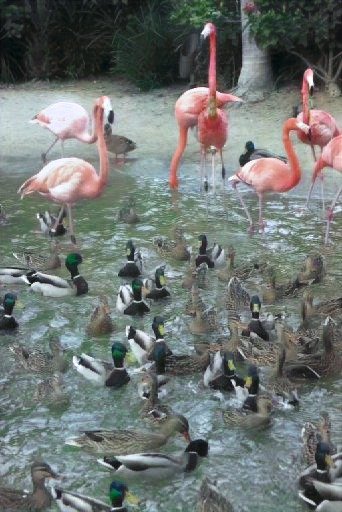}
 \\
 \includegraphics[width=\imwidth]{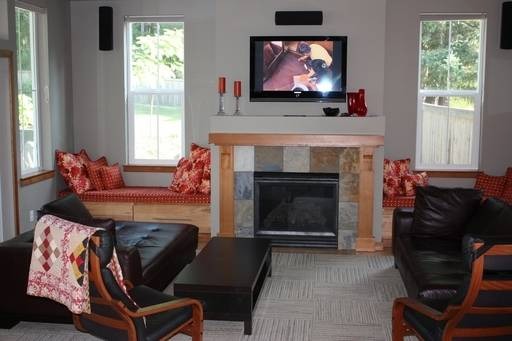}\llap{{\raisebox{0.4\imwidth}{\includegraphics[cfbox=red, width=0.4\imwidth]{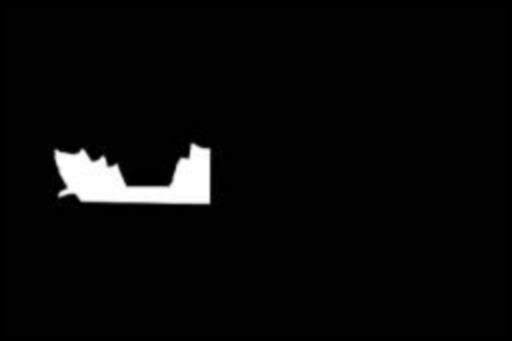}}}}  &
 \includegraphics[width=\imwidth]{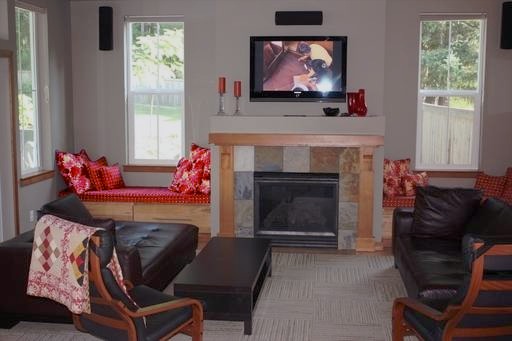} &
 \includegraphics[width=\imwidth]{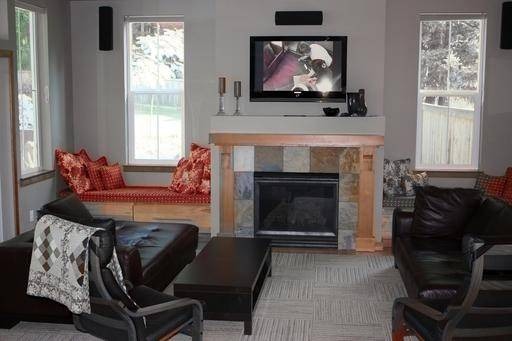} &
 \includegraphics[width=\imwidth]{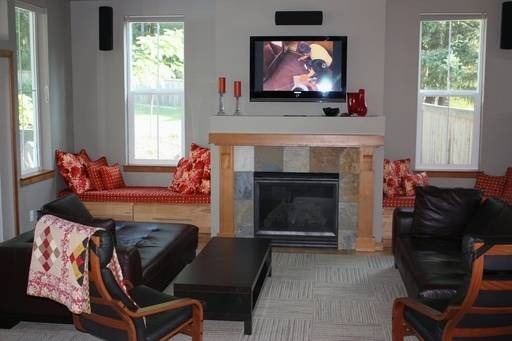} &
 \includegraphics[width=\imwidth]{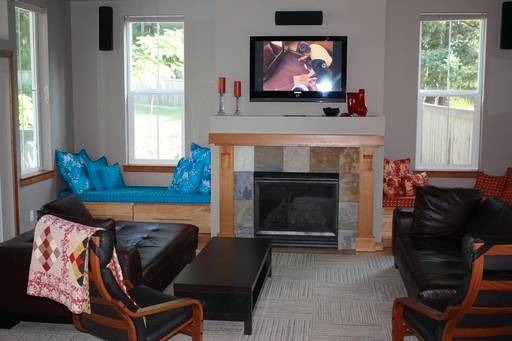} &
 \includegraphics[width=\imwidth]{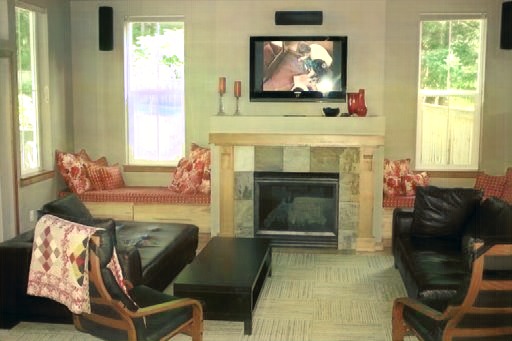}
 \\
 \end{tabular}
\caption{More model comparisons on the Mechrez dataset. }
\label{fig:comparisons_mechrez_1}
\end{figure*}

\begin{figure*}[!ht]
\centering
\setlength{\imwidth}{0.16\linewidth}
\setbox1=\hbox{\includegraphics[width=\imwidth]{Ims/compFig/mask0.jpg}}
\setlength{\tabcolsep}{1pt}
\begin{tabular}{cccccc}
Input & Our & MEC~\cite{mechrez2019saliency} & HAG~\cite{hagiwara2011saliency} & HOR~\cite{mateescu2014attention} & GAT~\cite{gatys2017guiding} 
\\
 \includegraphics[width=\imwidth]{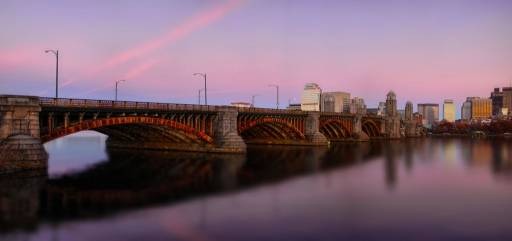}\llap{\makebox[\wd1][l]{\raisebox{0.3\imwidth}{\includegraphics[cfbox=red, width=0.4\imwidth]{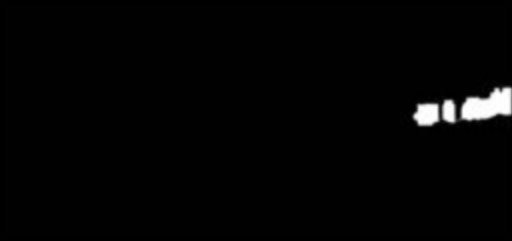}}}}&
 \includegraphics[width=\imwidth]{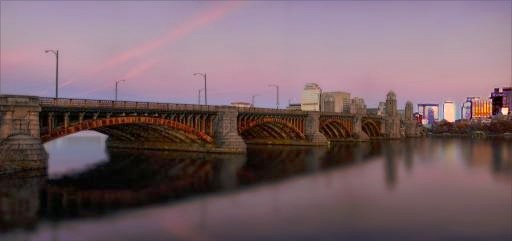} &
 \includegraphics[width=\imwidth]{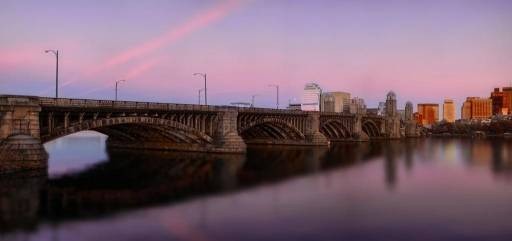} &
 \includegraphics[width=\imwidth]{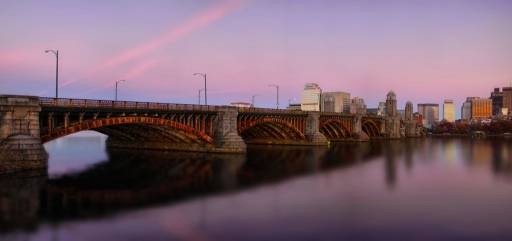} &
 \includegraphics[width=\imwidth]{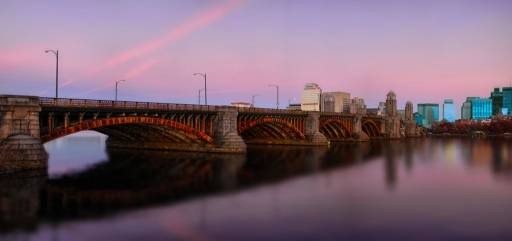} &
 \includegraphics[width=\imwidth]{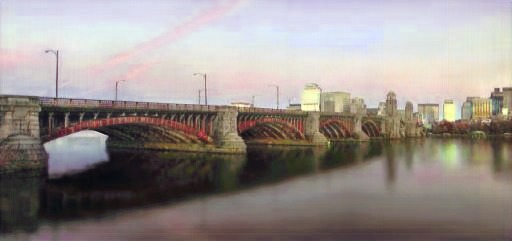}
 \\
  \includegraphics[width=\imwidth]{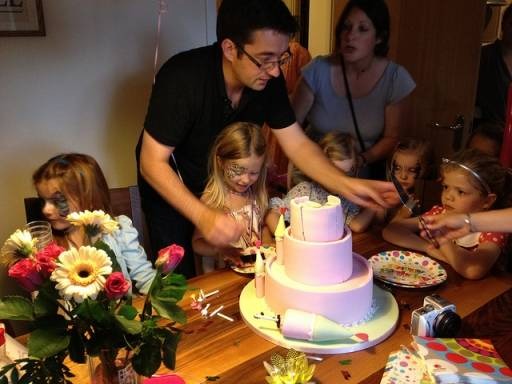}\llap{\makebox[\wd1][l]{\raisebox{0.45\imwidth}{\includegraphics[cfbox=red, width=0.4\imwidth]{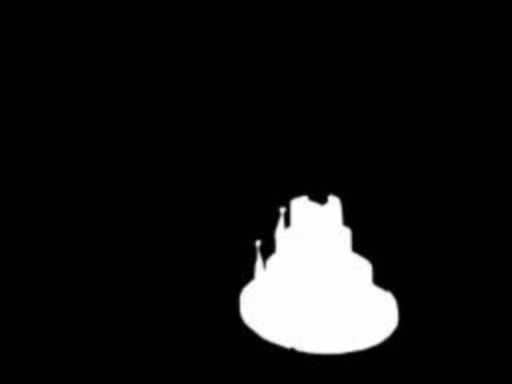}}}}&
 \includegraphics[width=\imwidth]{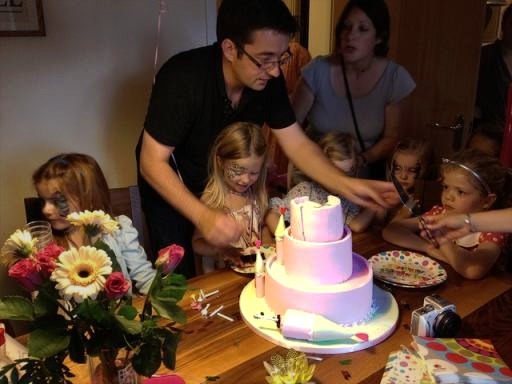} &
 \includegraphics[width=\imwidth]{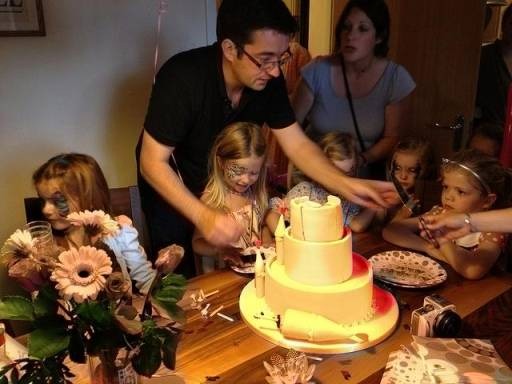} &
 \includegraphics[width=\imwidth]{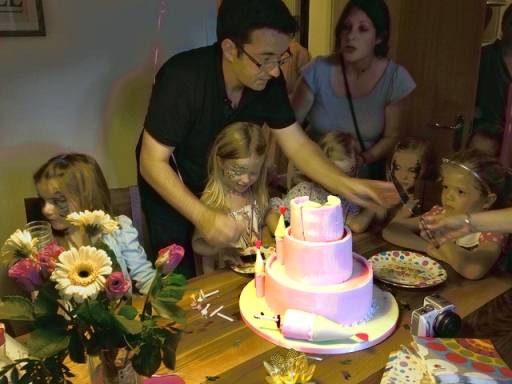} &
 \includegraphics[width=\imwidth]{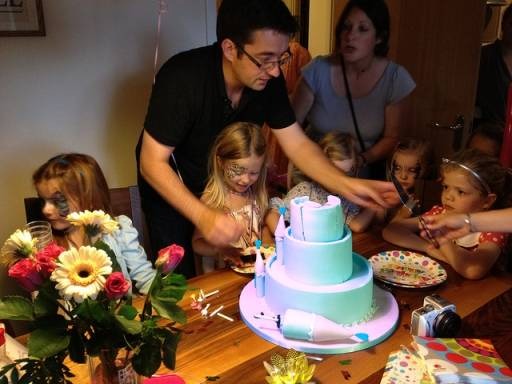} &
 \includegraphics[width=\imwidth]{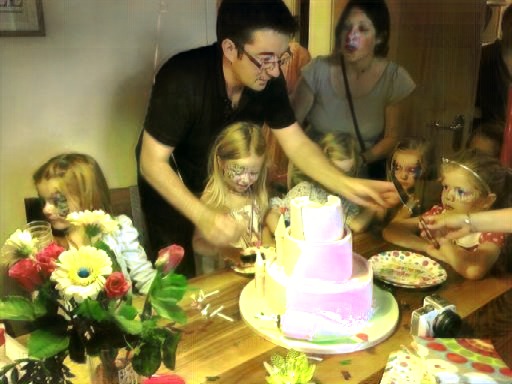}
 \\
  \includegraphics[width=\imwidth]{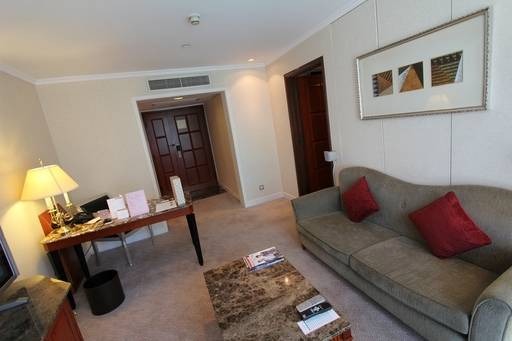}\llap{\makebox[\wd1][l]{\raisebox{0.4\imwidth}{\includegraphics[cfbox=red, width=0.4\imwidth]{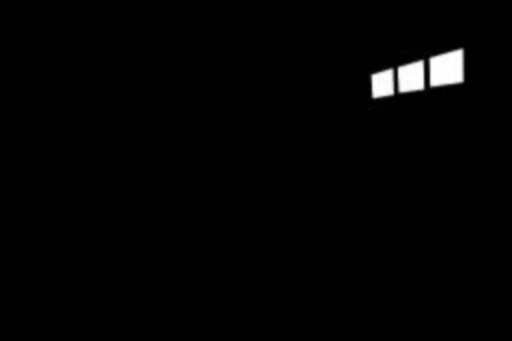}}}}&
 \includegraphics[width=\imwidth]{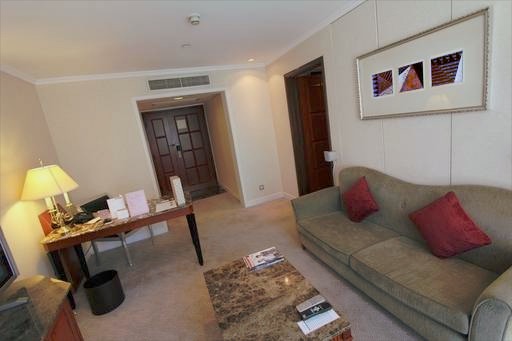} &
 \includegraphics[width=\imwidth]{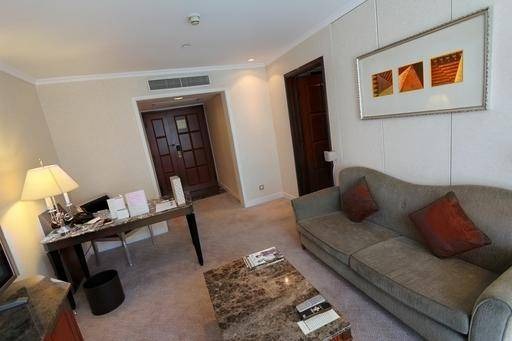} &
 \includegraphics[width=\imwidth]{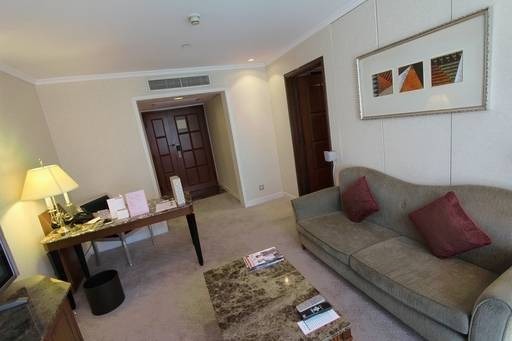} &
 \includegraphics[width=\imwidth]{Ims/compFig_bis/im_hor_inc2_-0029.jpg} &
 \includegraphics[width=\imwidth]{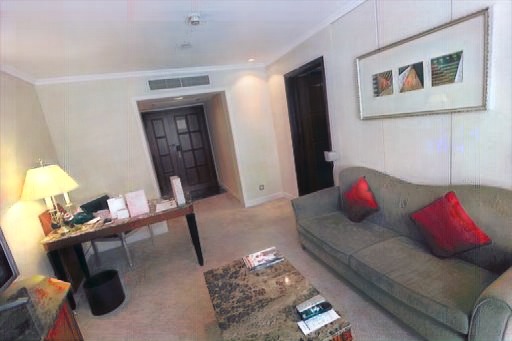}
 \\
  \includegraphics[width=\imwidth]{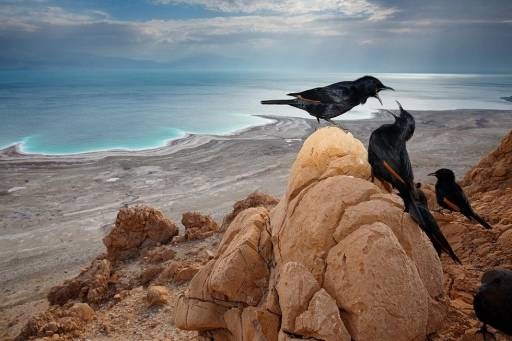}\llap{\makebox[\wd1][l]{\raisebox{0.4\imwidth}{\includegraphics[cfbox=red, width=0.4\imwidth]{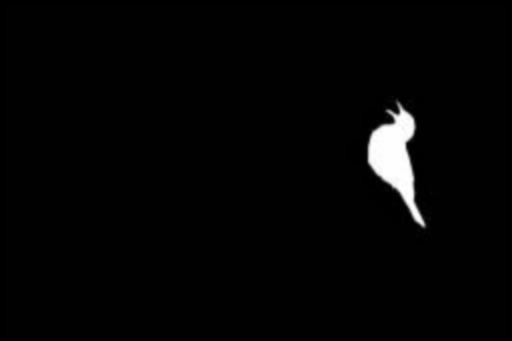}}}}&
 \includegraphics[width=\imwidth]{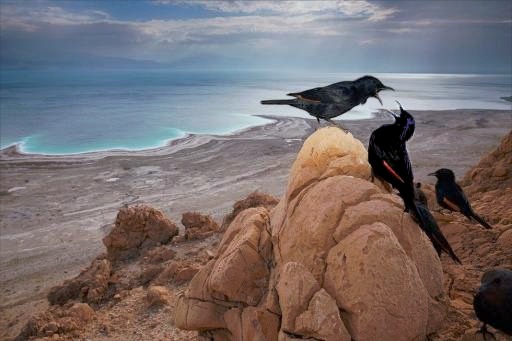} &
 \includegraphics[width=\imwidth]{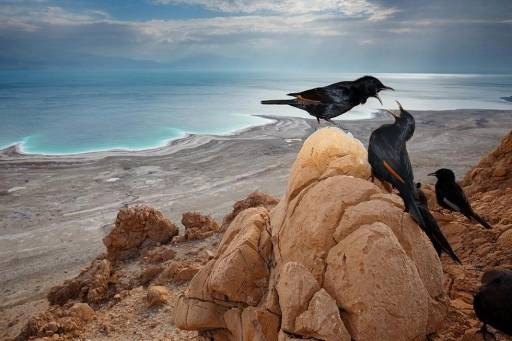} &
 \includegraphics[width=\imwidth]{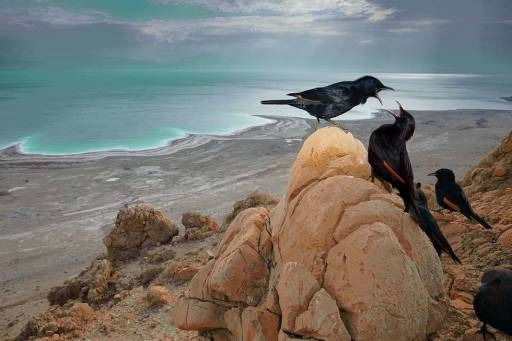} &
 \includegraphics[width=\imwidth]{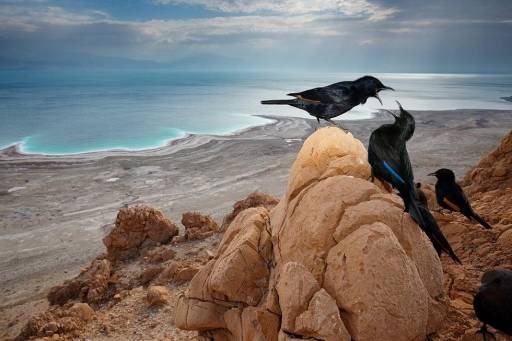} &
 \includegraphics[width=\imwidth]{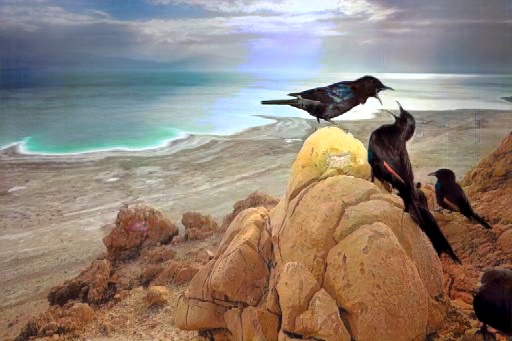}
 \\
  \includegraphics[width=\imwidth]{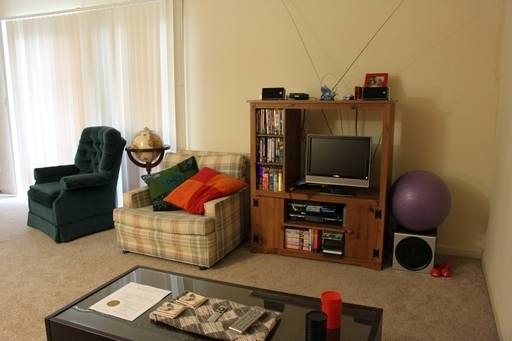}\llap{\makebox[\wd1][l]{\raisebox{0.4\imwidth}{\includegraphics[cfbox=red, width=0.4\imwidth]{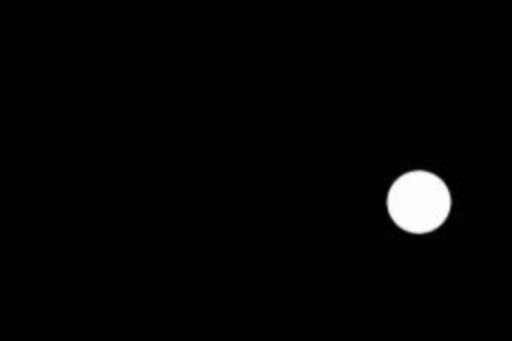}}}}&
 \includegraphics[width=\imwidth]{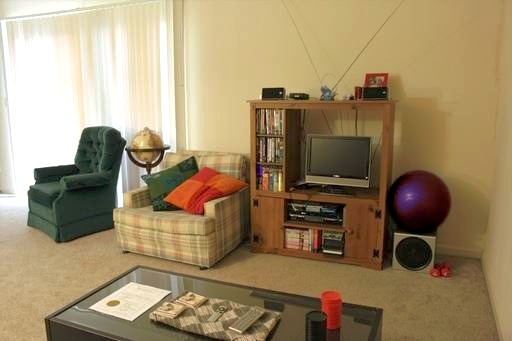} &
 \includegraphics[width=\imwidth]{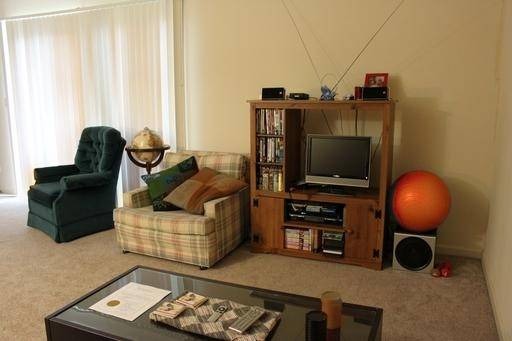} &
 \includegraphics[width=\imwidth]{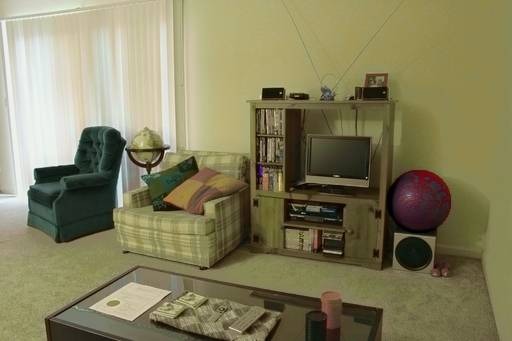} &
 \includegraphics[width=\imwidth]{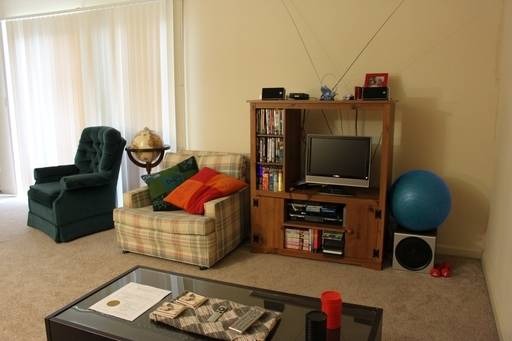} &
 \includegraphics[width=\imwidth]{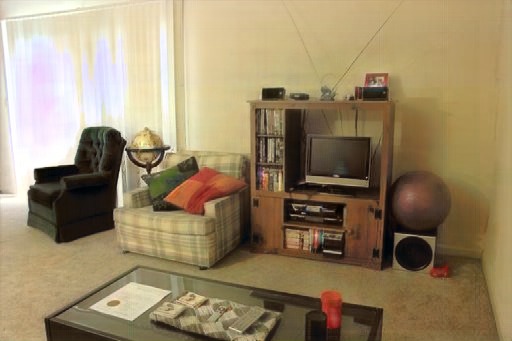}
 \\
   \includegraphics[width=\imwidth]{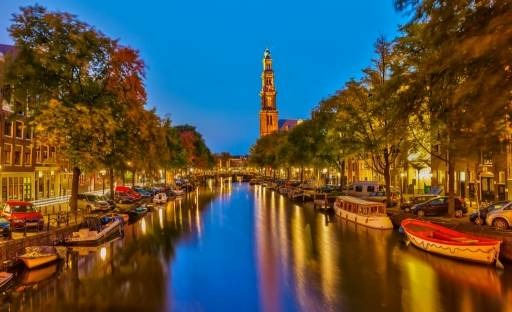}\llap{\makebox[\wd1][l]{\raisebox{0.35\imwidth}{\includegraphics[cfbox=red, width=0.4\imwidth]{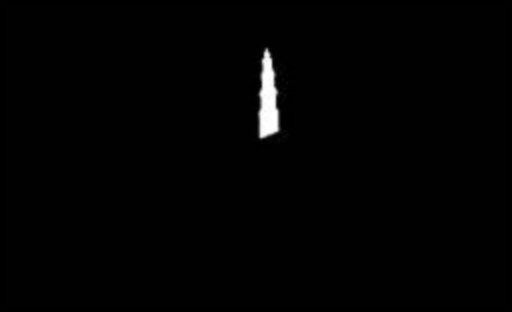}}}}&
 \includegraphics[width=\imwidth]{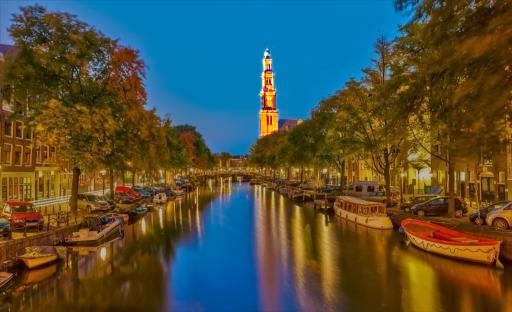} &
 \includegraphics[width=\imwidth]{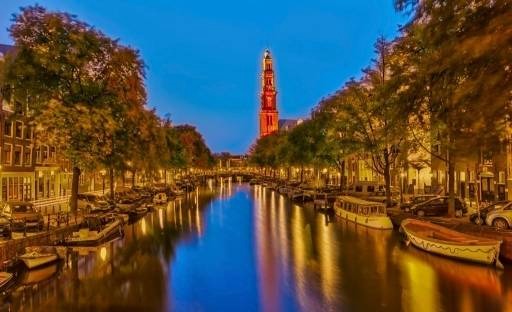} &
 \includegraphics[width=\imwidth]{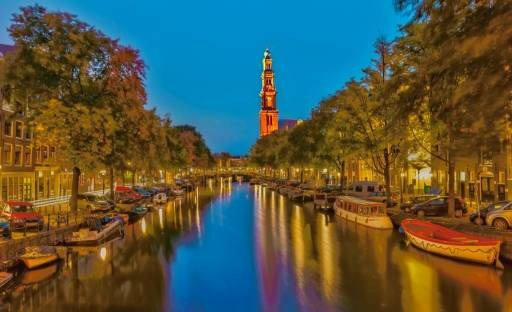} &
 \includegraphics[width=\imwidth]{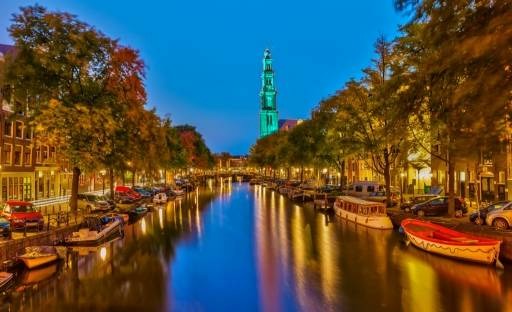} &
 \includegraphics[width=\imwidth]{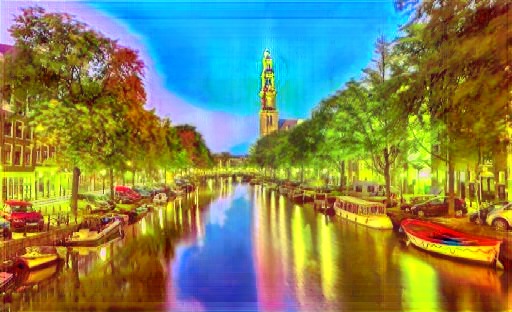}
 \\
   \includegraphics[width=\imwidth]{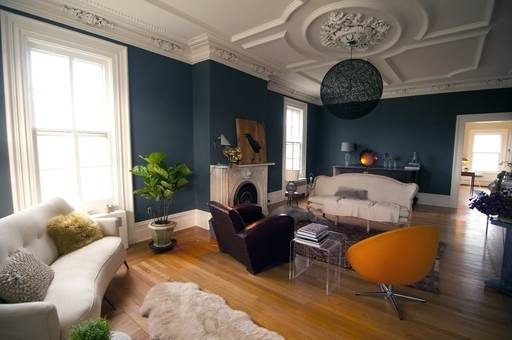}\llap{\makebox[\wd1][l]{\raisebox{0.4\imwidth}{\includegraphics[cfbox=red, width=0.4\imwidth]{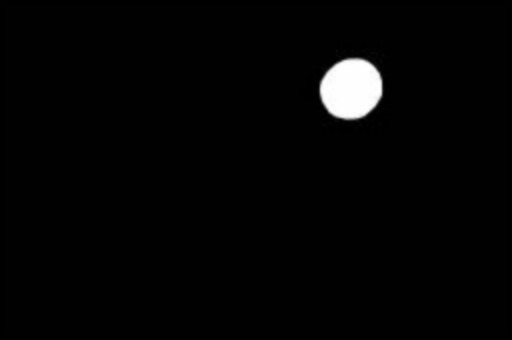}}}}&
 \includegraphics[width=\imwidth]{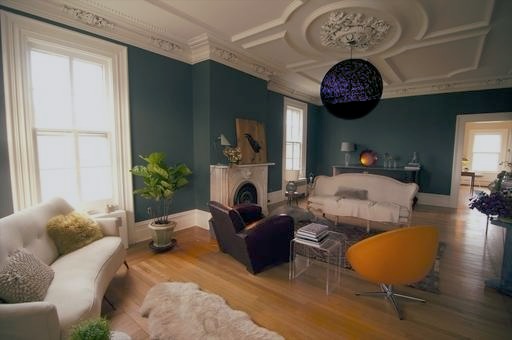} &
 \includegraphics[width=\imwidth]{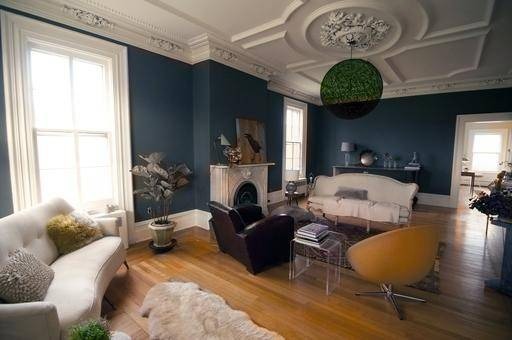} &
 \includegraphics[width=\imwidth]{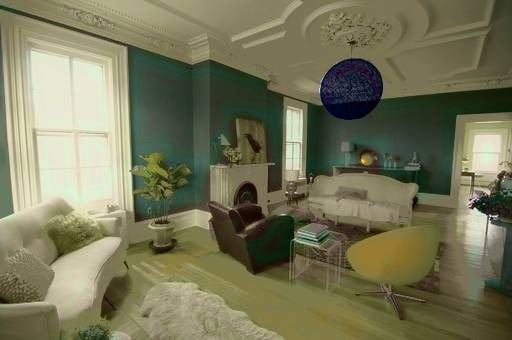} &
 \includegraphics[width=\imwidth]{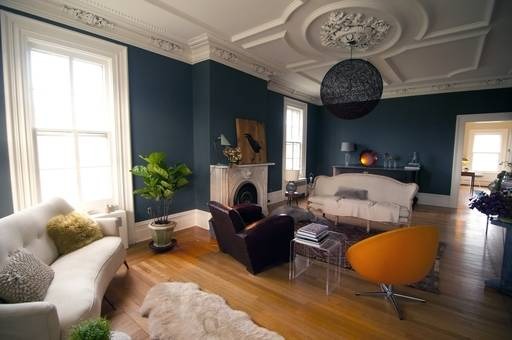} &
 \includegraphics[width=\imwidth]{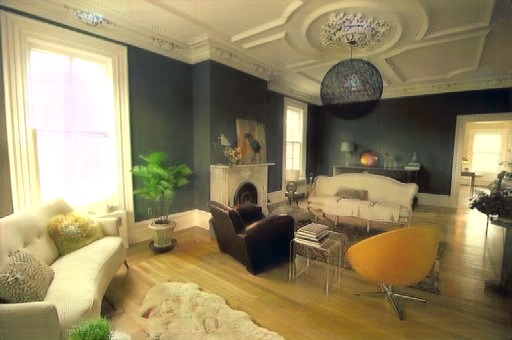}
 \\
   \includegraphics[width=\imwidth]{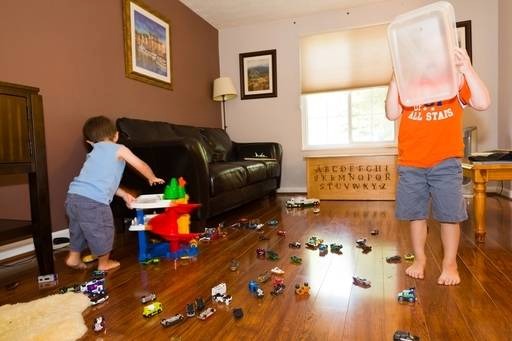}\llap{\makebox[\wd1][l]{\raisebox{0.4\imwidth}{\includegraphics[cfbox=red, width=0.4\imwidth]{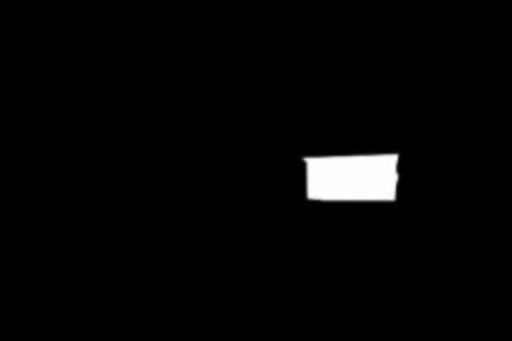}}}}&
 \includegraphics[width=\imwidth]{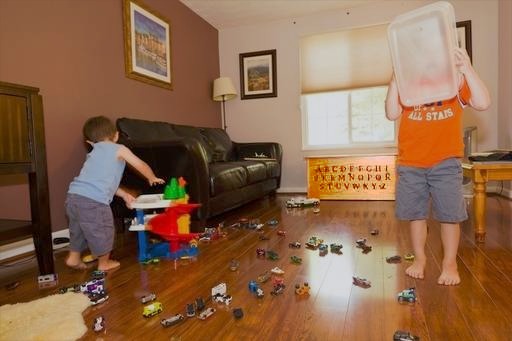} &
 \includegraphics[width=\imwidth]{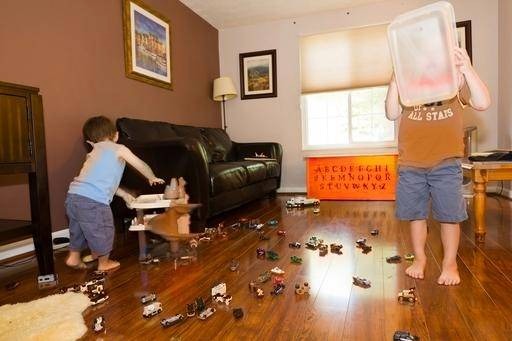} &
 \includegraphics[width=\imwidth]{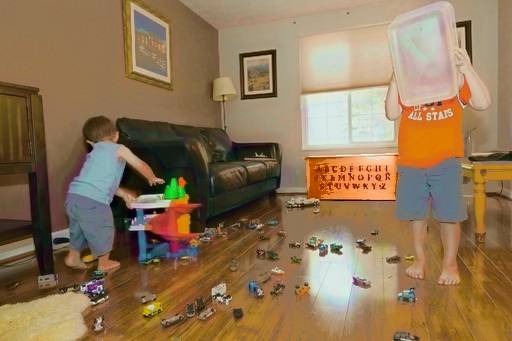} &
 \includegraphics[width=\imwidth]{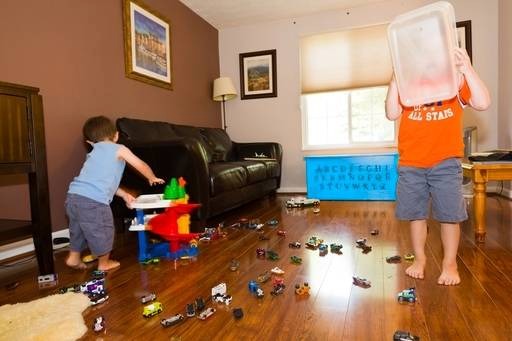} &
 \includegraphics[width=\imwidth]{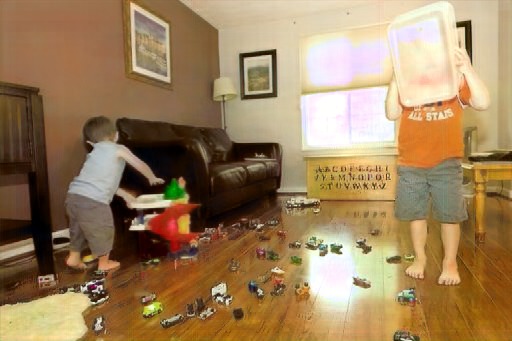}
 \\
   \includegraphics[width=\imwidth]{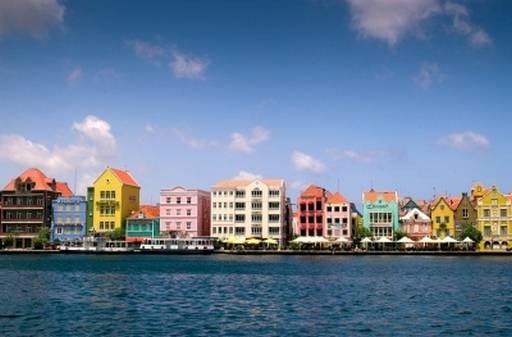}\llap{{\raisebox{0.4\imwidth}{\includegraphics[cfbox=red, width=0.4\imwidth]{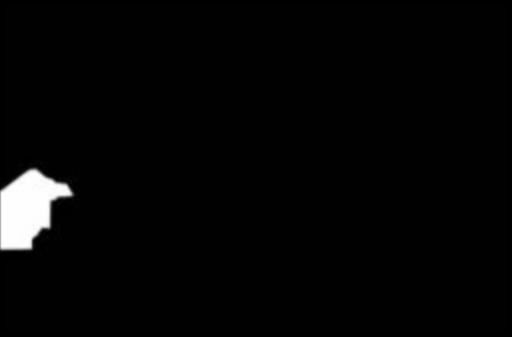}}}}&
 \includegraphics[width=\imwidth]{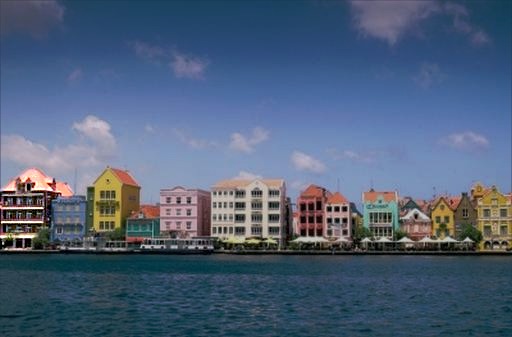} &
 \includegraphics[width=\imwidth]{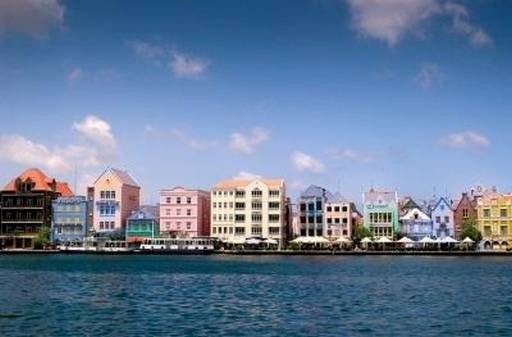} &
 \includegraphics[width=\imwidth]{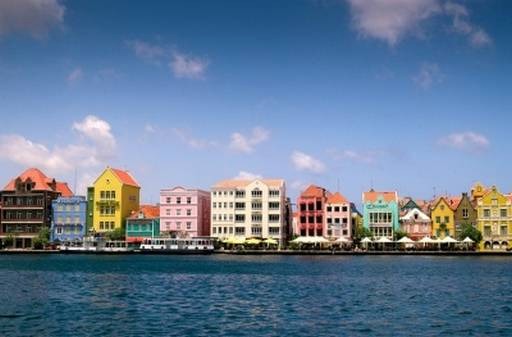} &
 \includegraphics[width=\imwidth]{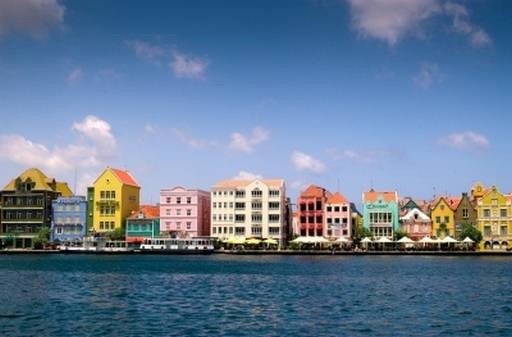} &
 \includegraphics[width=\imwidth]{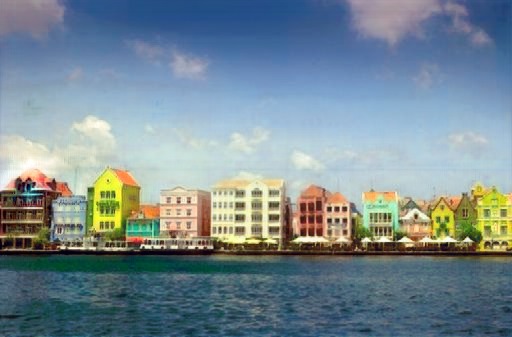}
 \\
    \includegraphics[width=\imwidth]{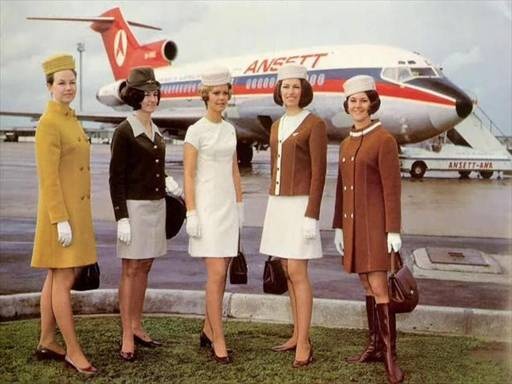}\llap{\makebox[\wd1][l]{\raisebox{0\imwidth}{\includegraphics[cfbox=red, width=0.4\imwidth]{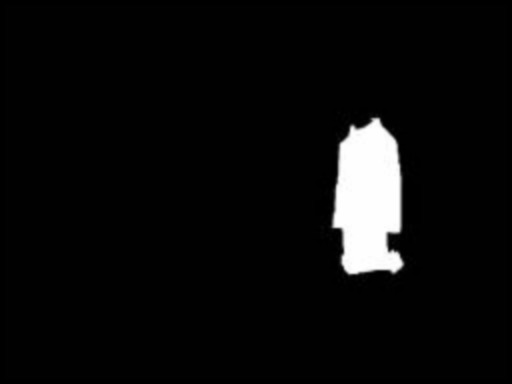}}}}&
 \includegraphics[width=\imwidth]{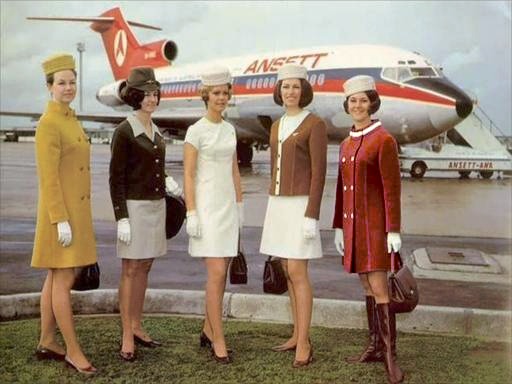} &
 \includegraphics[width=\imwidth]{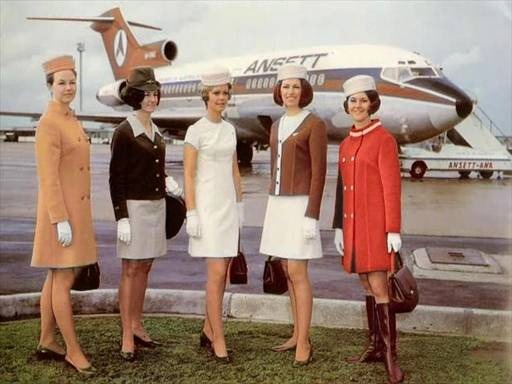} &
 \includegraphics[width=\imwidth]{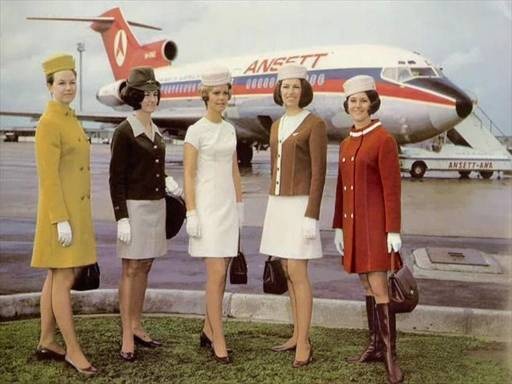} &
 \includegraphics[width=\imwidth]{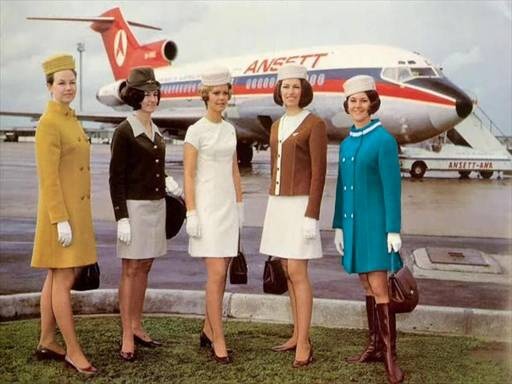} &
 \includegraphics[width=\imwidth]{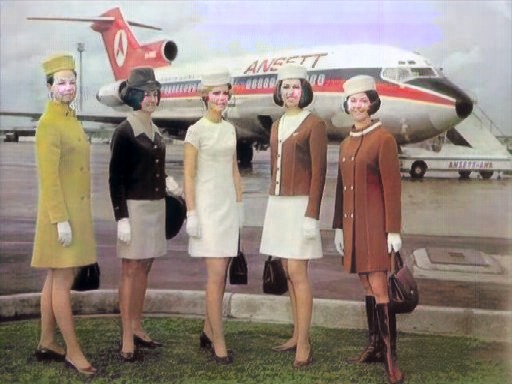}
 \\
     \includegraphics[width=\imwidth]{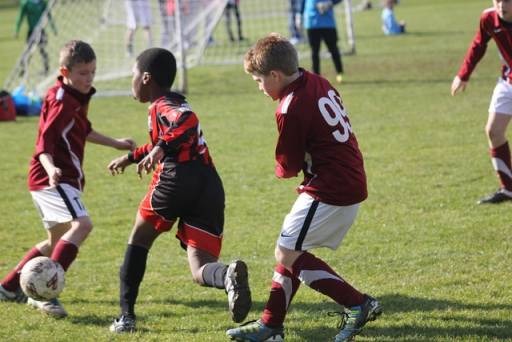}\llap{{\raisebox{0\imwidth}{\includegraphics[cfbox=red, width=0.4\imwidth]{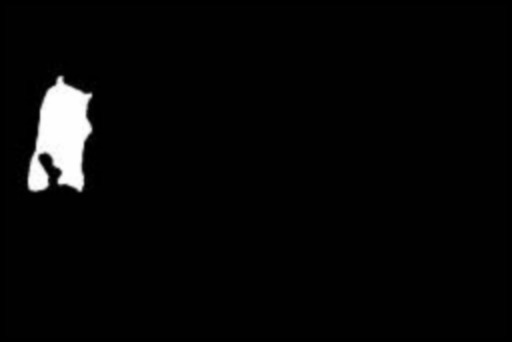}}}}&
 \includegraphics[width=\imwidth]{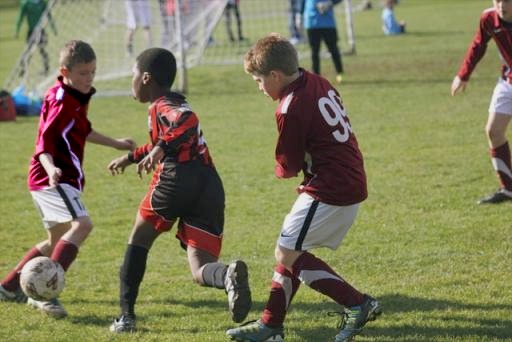} &
 \includegraphics[width=\imwidth]{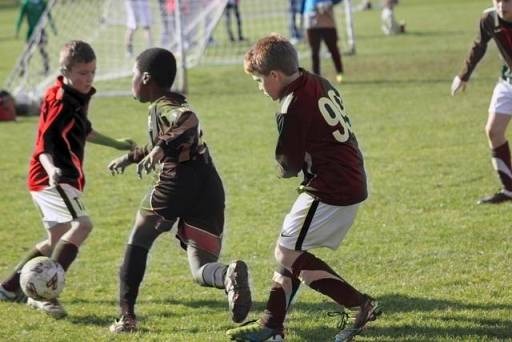} &
 \includegraphics[width=\imwidth]{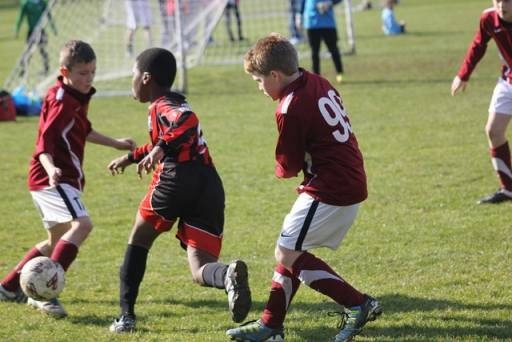} &
 \includegraphics[width=\imwidth]{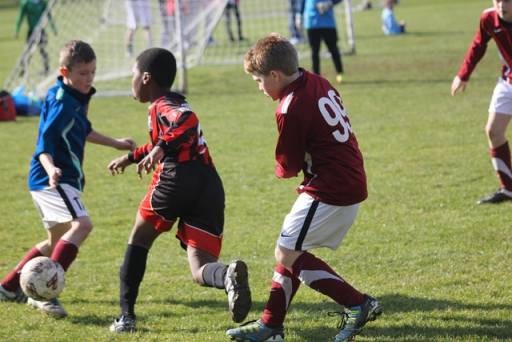} &
 \includegraphics[width=\imwidth]{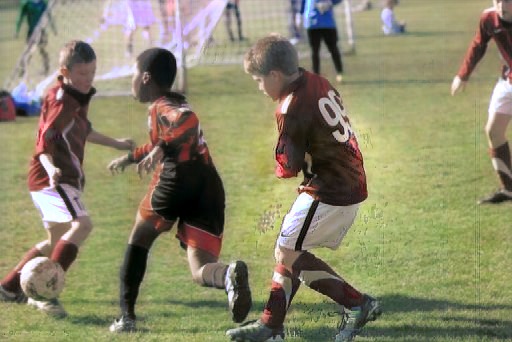}
 \\
      \includegraphics[width=\imwidth]{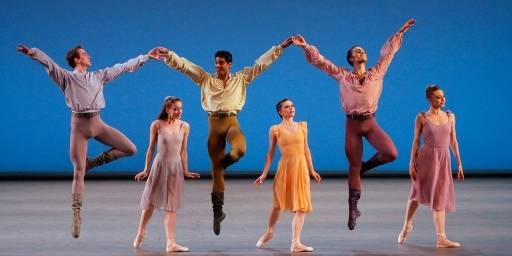}\llap{{\raisebox{0.3\imwidth}{\includegraphics[cfbox=red, width=0.4\imwidth]{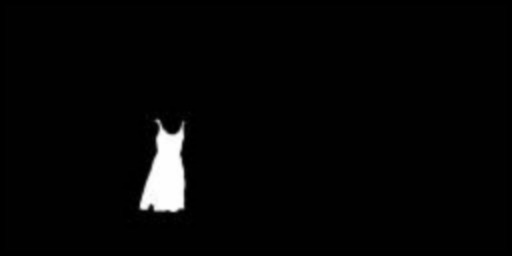}}}}&
 \includegraphics[width=\imwidth]{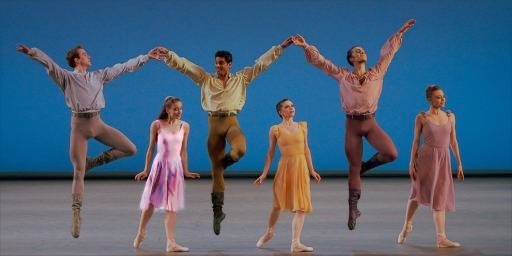} &
 \includegraphics[width=\imwidth]{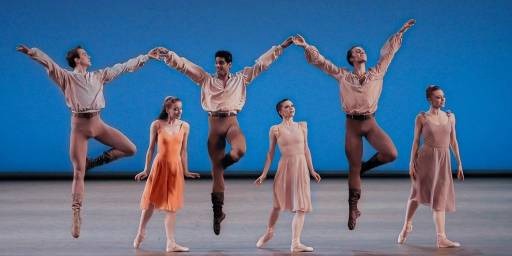} &
 \includegraphics[width=\imwidth]{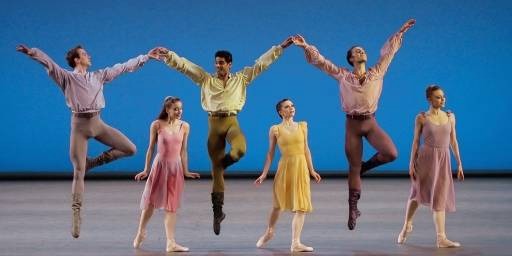} &
 \includegraphics[width=\imwidth]{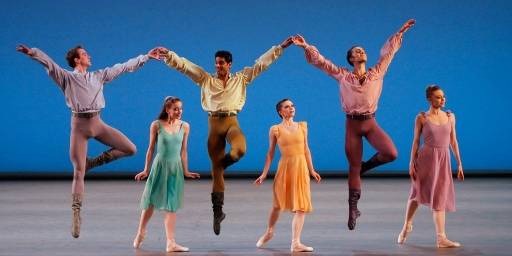} &
 \includegraphics[width=\imwidth]{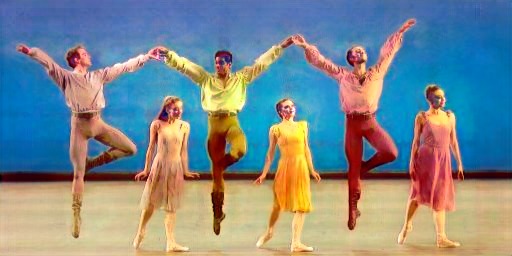}
 \\

 \end{tabular}
\caption{More model comparisons on the Mechrez dataset. }
\label{fig:comparisons_mechrez_2}
\end{figure*}

\begin{figure*}
\centering
\setlength{\imwidth}{0.16\linewidth}
\setbox1=\hbox{\includegraphics[width=\imwidth]{Ims/compFig/mask0.jpg}}
\setlength{\tabcolsep}{1pt}
\begin{tabular}{cccccc}
Input & Mask & Our & HAG~\cite{hagiwara2011saliency} & HOR~\cite{mateescu2014attention} & GAT~\cite{gatys2017guiding} 
\\
 \includegraphics[width=\imwidth]{Ims/compFig_coco/im310.jpg}&
 \includegraphics[width=\imwidth]{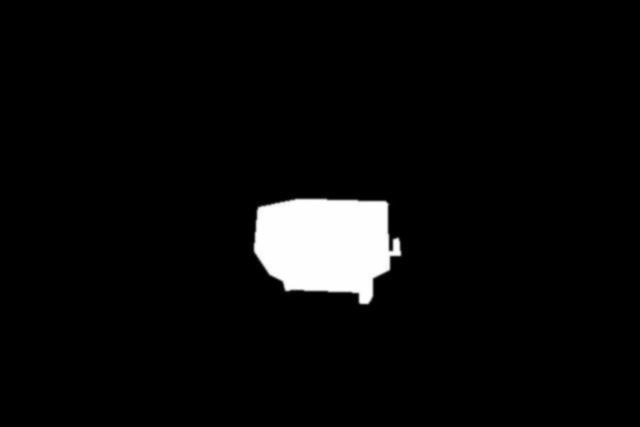}&
 \includegraphics[width=\imwidth]{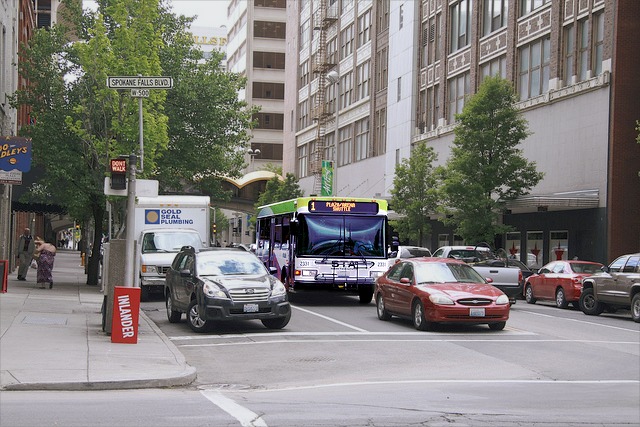}&
 \includegraphics[width=\imwidth]{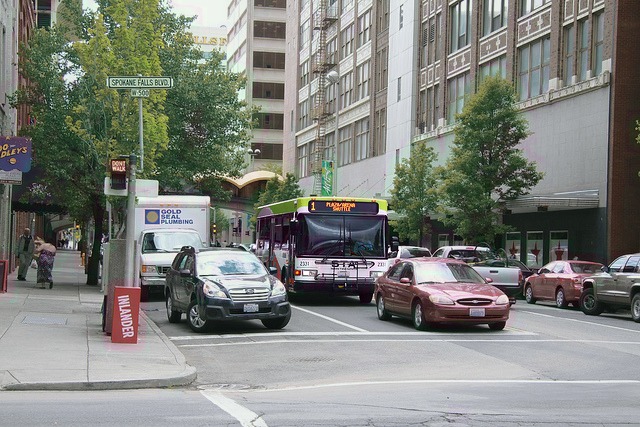}&
 \includegraphics[width=\imwidth]{Ims/compFig_coco/im_hor_inc310_0101.jpg}&
 \includegraphics[width=\imwidth]{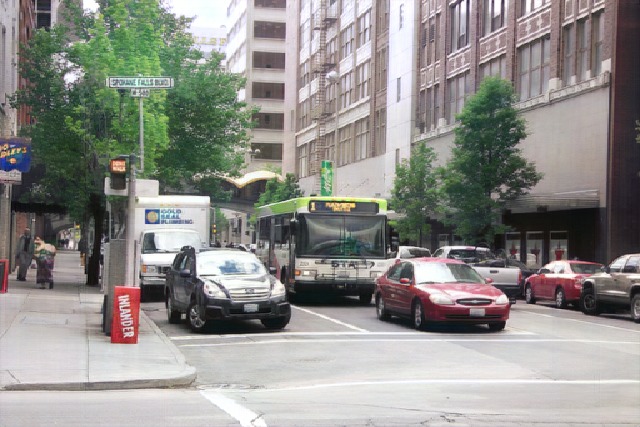}
 \\
 \includegraphics[width=\imwidth]{Ims/compFig_coco/im321.jpg}&
 \includegraphics[width=\imwidth]{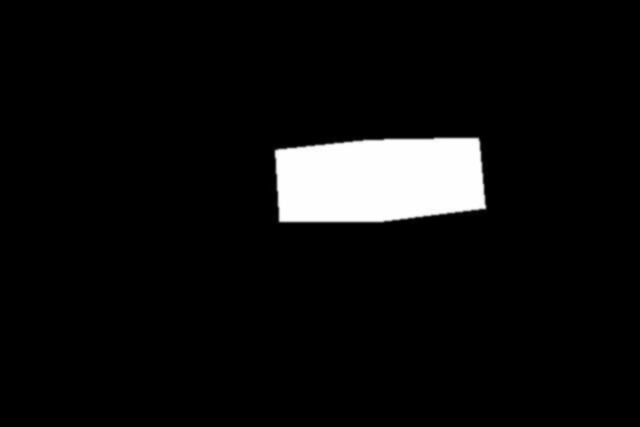}&
 \includegraphics[width=\imwidth]{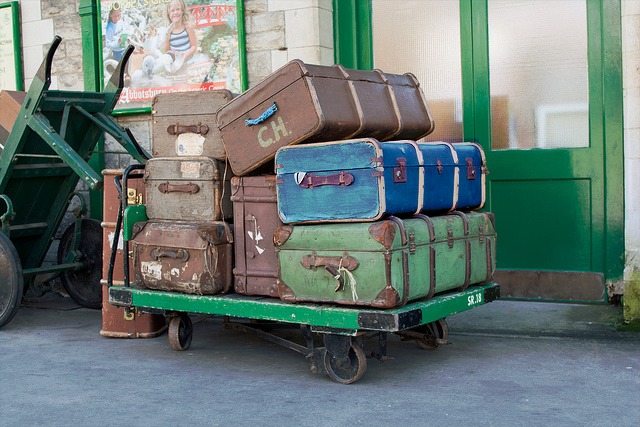}&
 \includegraphics[width=\imwidth]{Ims/compFig_coco/im_hag_inc321_0001.jpg}&
 \includegraphics[width=\imwidth]{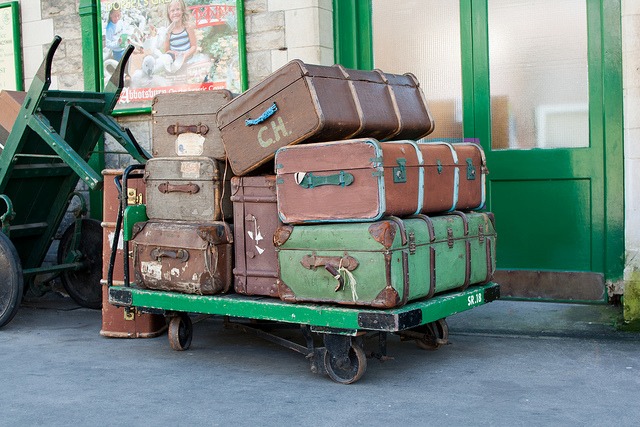}&
 \includegraphics[width=\imwidth]{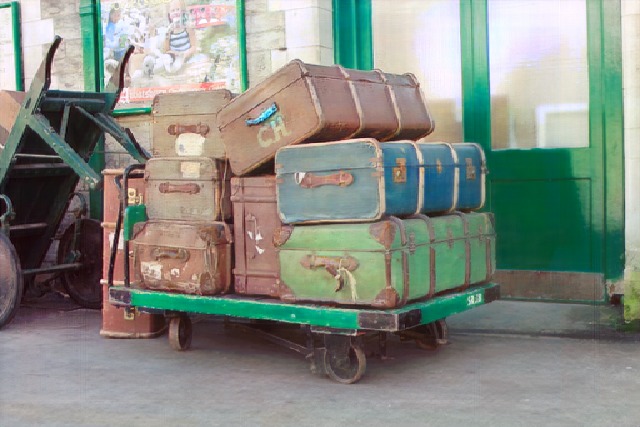}
 \\
 \includegraphics[width=\imwidth]{Ims/compFig_coco/im397.jpg}&
 \includegraphics[width=\imwidth]{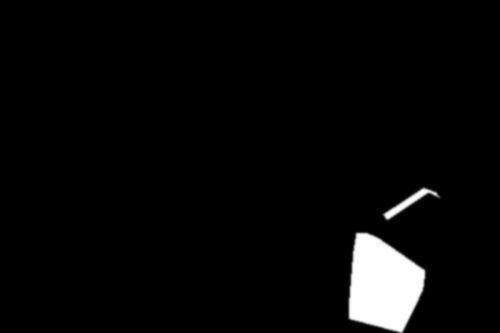}&
 \includegraphics[width=\imwidth]{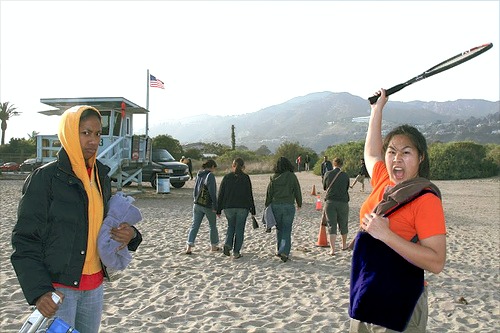}&
 \includegraphics[width=\imwidth]{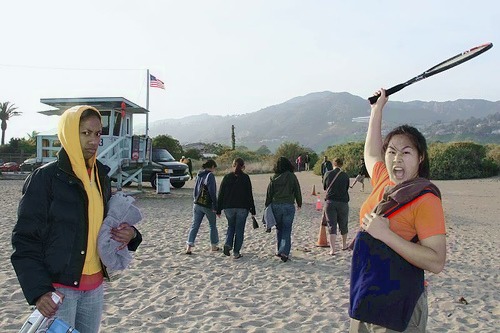}&
 \includegraphics[width=\imwidth]{Ims/compFig_coco/im_hor_inc397_0010.jpg}&
 \includegraphics[width=\imwidth]{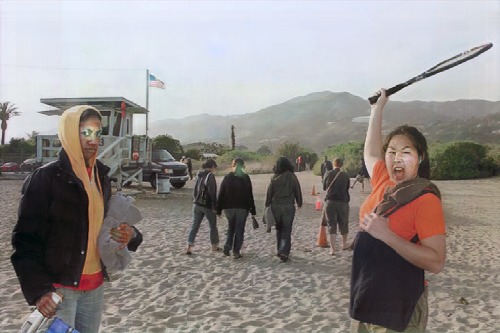}
 \\
 \includegraphics[width=\imwidth]{Ims/compFig_coco/im409.jpg}&
 \includegraphics[width=\imwidth]{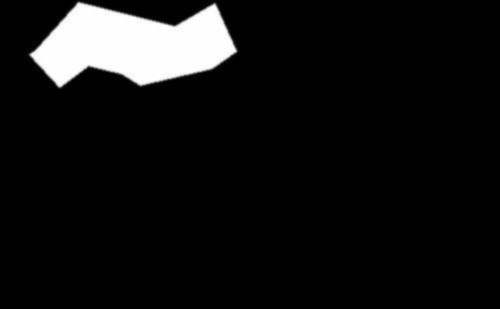}&
 \includegraphics[width=\imwidth]{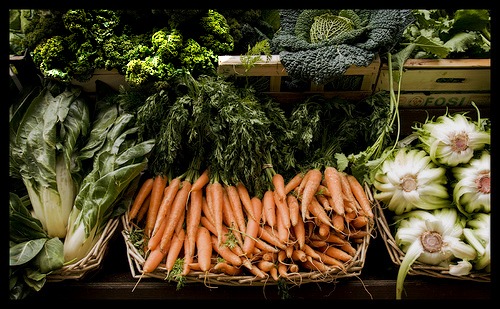}&
 \includegraphics[width=\imwidth]{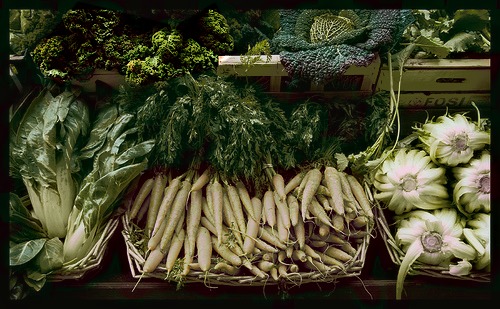}&
 \includegraphics[width=\imwidth]{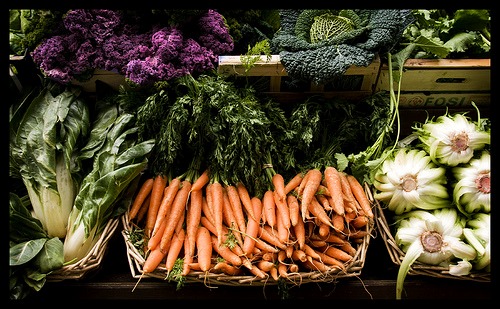}&
 \includegraphics[width=\imwidth]{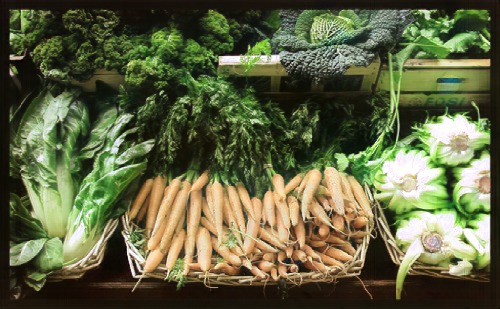}
 \\
 \includegraphics[width=\imwidth]{Ims/compFig_coco/im453.jpg}&
 \includegraphics[width=\imwidth]{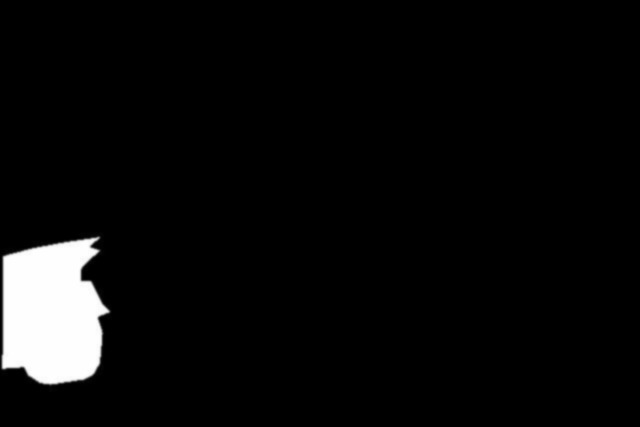}&
 \includegraphics[width=\imwidth]{Ims/compFig_coco/imfinal_inc453_0065.jpg}&
 \includegraphics[width=\imwidth]{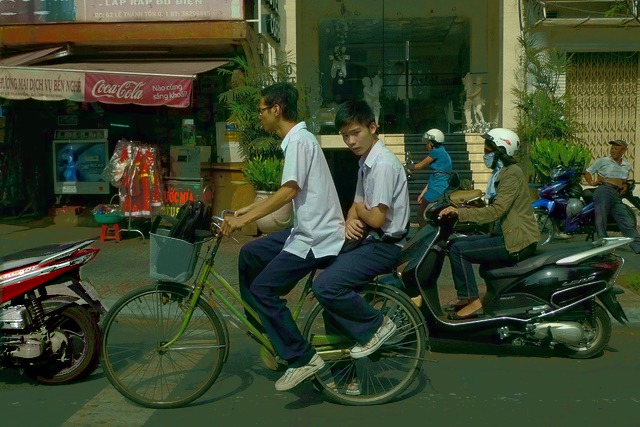}&
 \includegraphics[width=\imwidth]{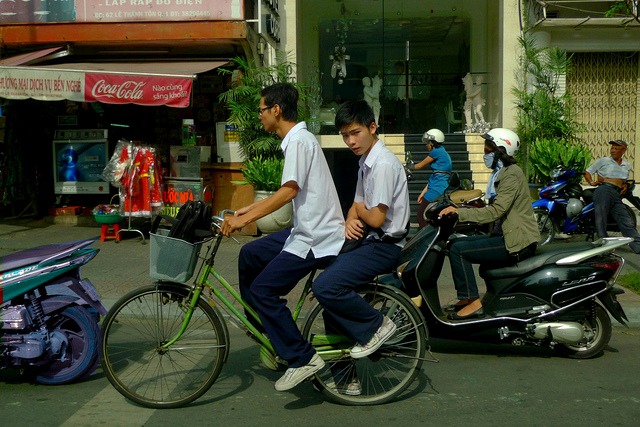}&
 \includegraphics[width=\imwidth]{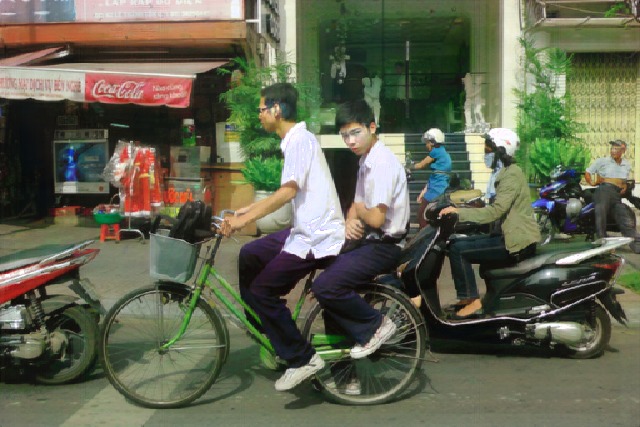}
 \\
 \includegraphics[width=\imwidth]{Ims/compFig_coco/im300.jpg}&
 \includegraphics[width=\imwidth]{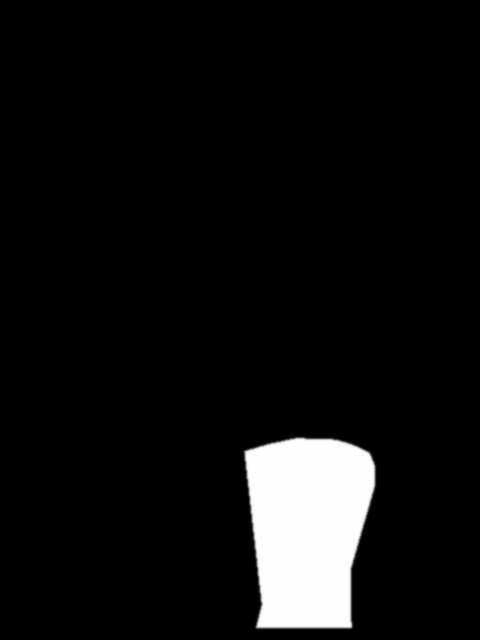}&
 \includegraphics[width=\imwidth]{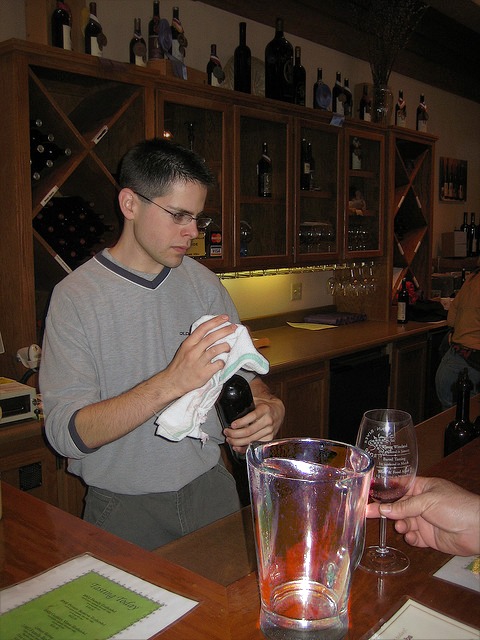}&
 \includegraphics[width=\imwidth]{Ims/compFig_coco/im_hag_inc300_-0000.jpg}&
 \includegraphics[width=\imwidth]{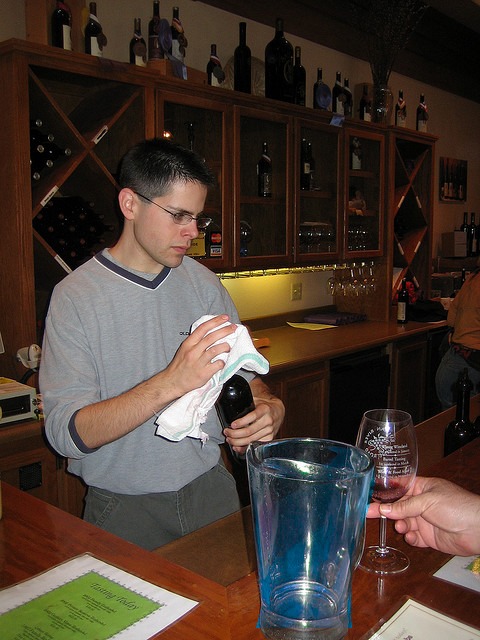}&
 \includegraphics[width=\imwidth]{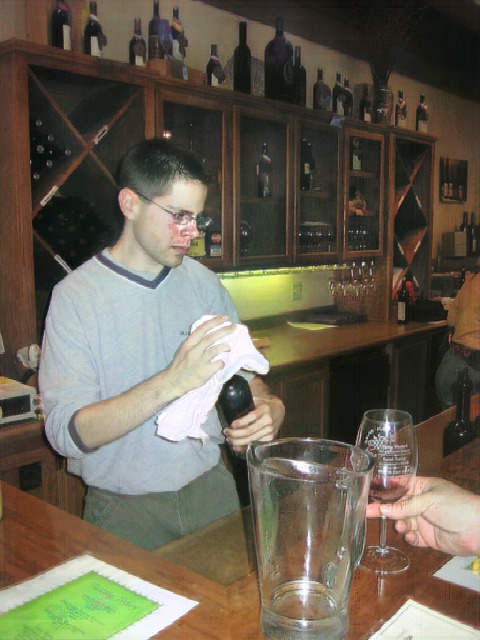}
 \\
  \includegraphics[width=\imwidth]{Ims/compFig_coco/im164.jpg}&
 \includegraphics[width=\imwidth]{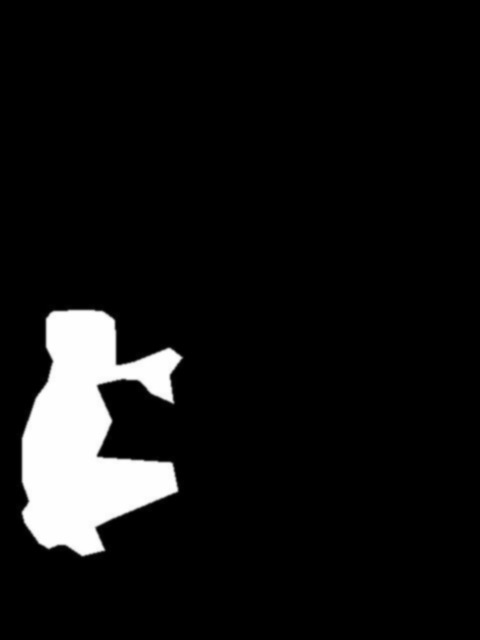}&
 \includegraphics[width=\imwidth]{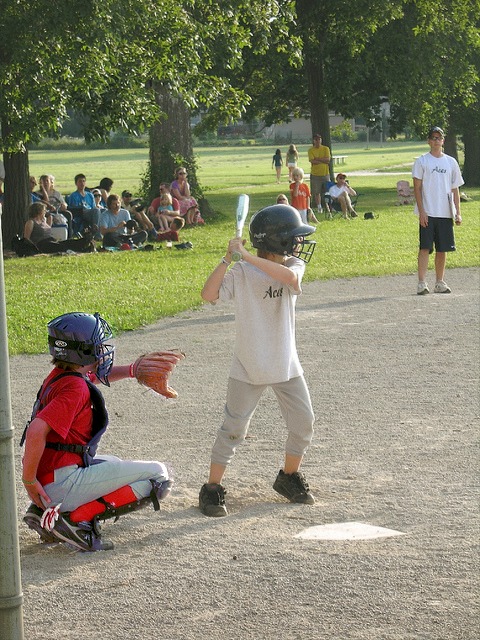}&
 \includegraphics[width=\imwidth]{Ims/compFig_coco/im_hag_inc164_-0000.jpg}&
 \includegraphics[width=\imwidth]{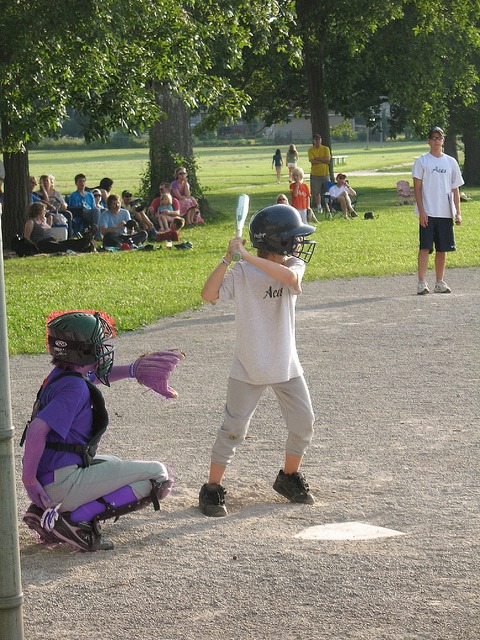}&
 \includegraphics[width=\imwidth]{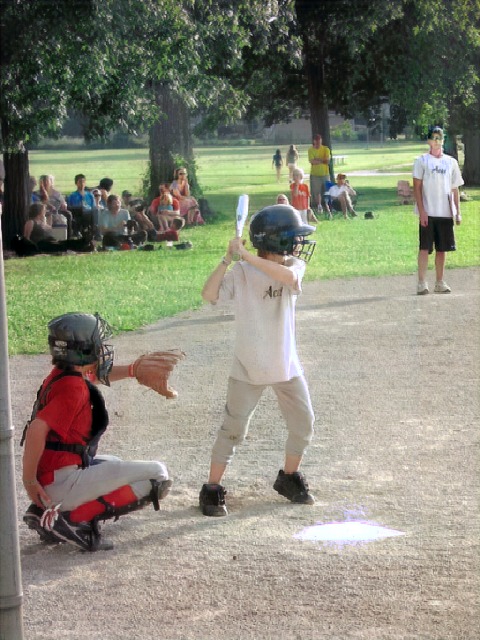}
 \\
   \includegraphics[width=\imwidth]{Ims/compFig_coco/im15.jpg}&
 \includegraphics[width=\imwidth]{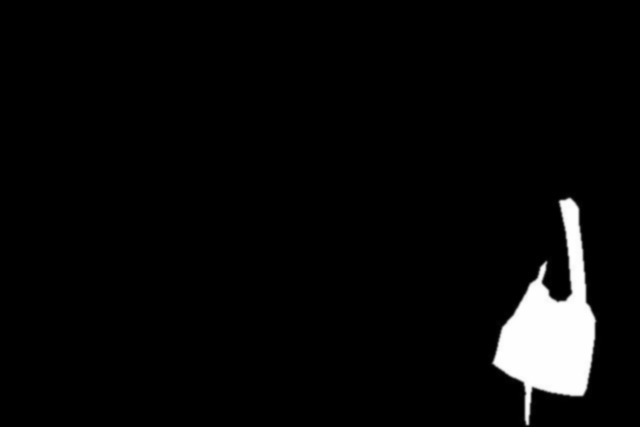}&
 \includegraphics[width=\imwidth]{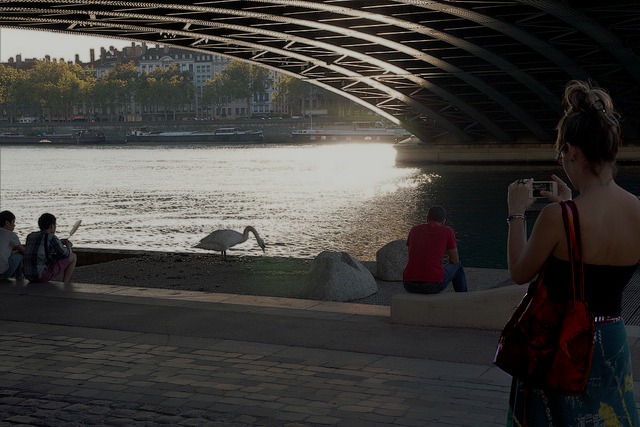}&
 \includegraphics[width=\imwidth]{Ims/compFig_coco/im_hag_inc15_0002.jpg}&
 \includegraphics[width=\imwidth]{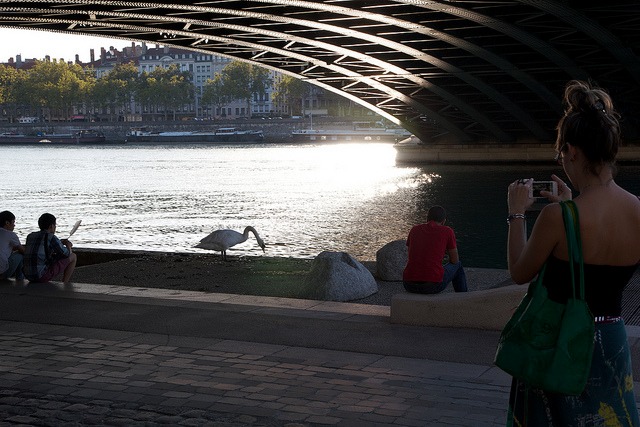}&
 \includegraphics[width=\imwidth]{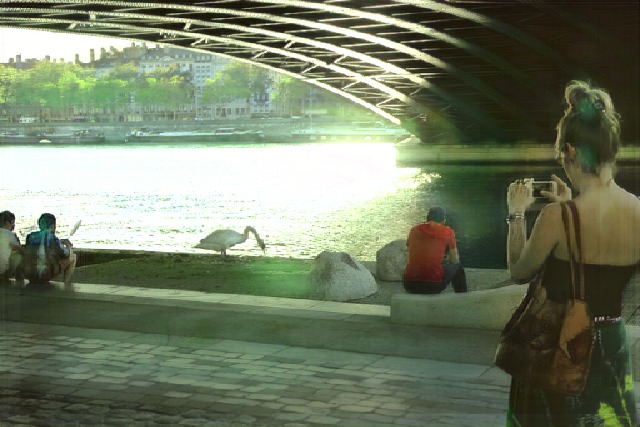}
 \\

 \includegraphics[width=\imwidth]{Ims/compFig_coco/im715.jpg}&
 \includegraphics[width=\imwidth]{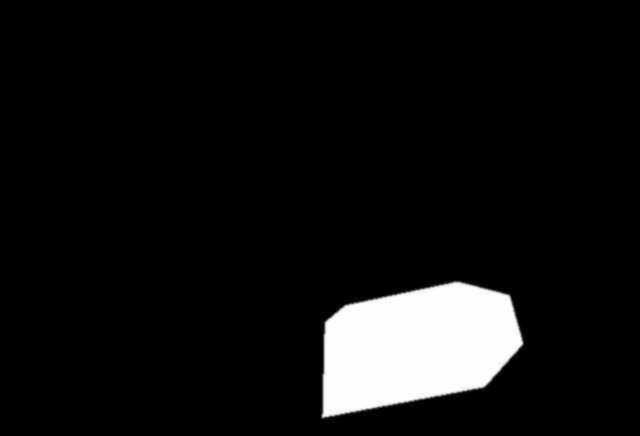}&
 \includegraphics[width=\imwidth]{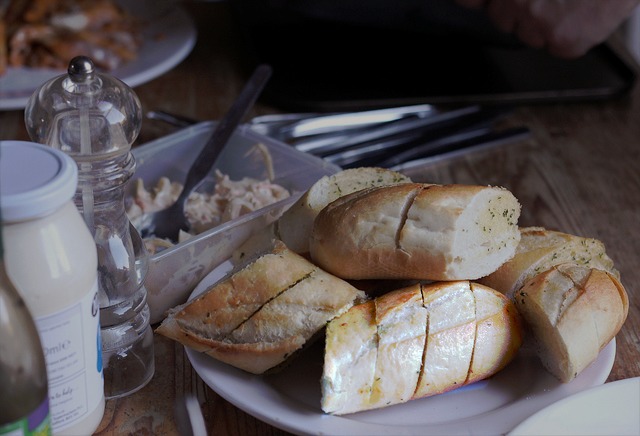}&
 \includegraphics[width=\imwidth]{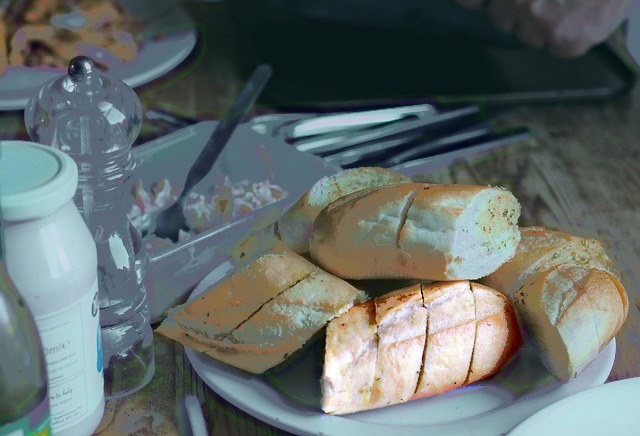}&
 \includegraphics[width=\imwidth]{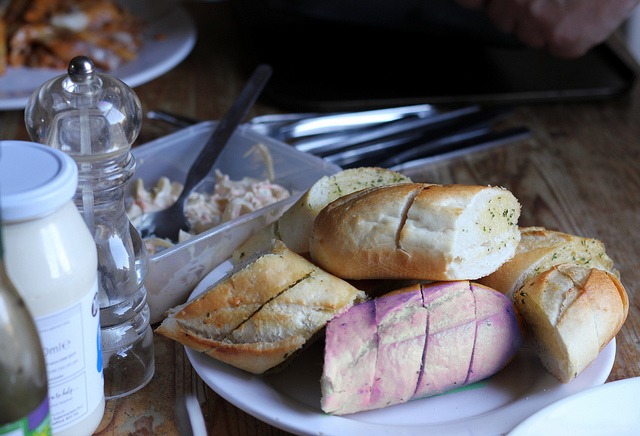}&
 \includegraphics[width=\imwidth]{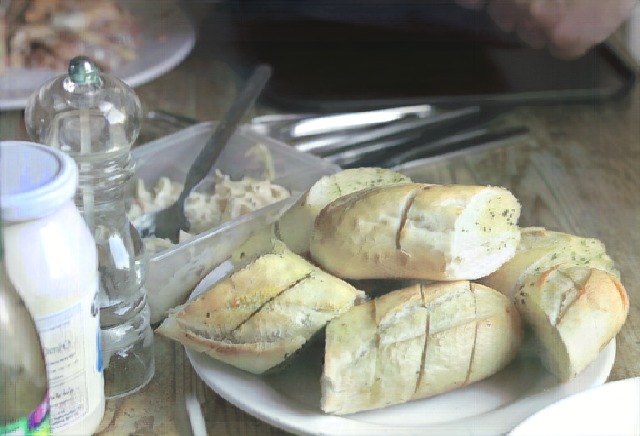}
 \\
    \includegraphics[width=\imwidth]{Ims/CoCoClutter_supplement/im14.jpg}&
 \includegraphics[width=\imwidth]{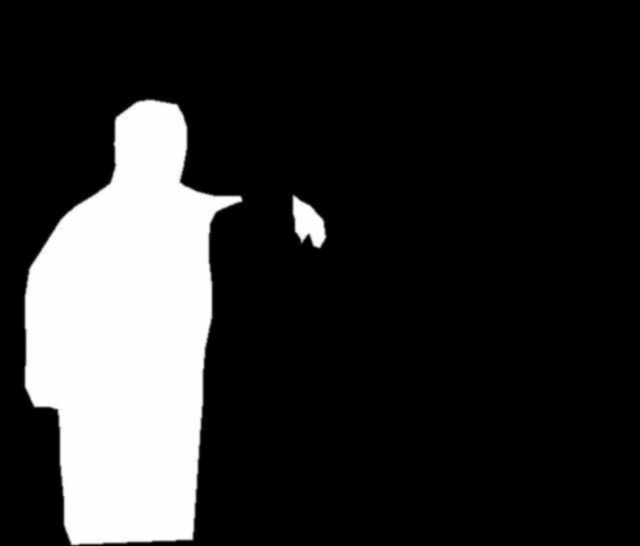}&
 \includegraphics[width=\imwidth]{Ims/CoCoClutter_supplement/imfinal_inc14_-0002.jpg}&
 \includegraphics[width=\imwidth]{Ims/CoCoClutter_supplement/im_hag_inc14_0002.jpg}&
 \includegraphics[width=\imwidth]{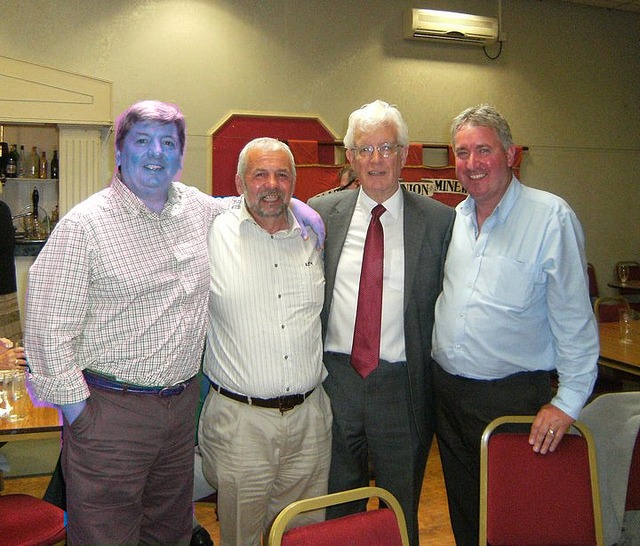}&
 \includegraphics[width=\imwidth]{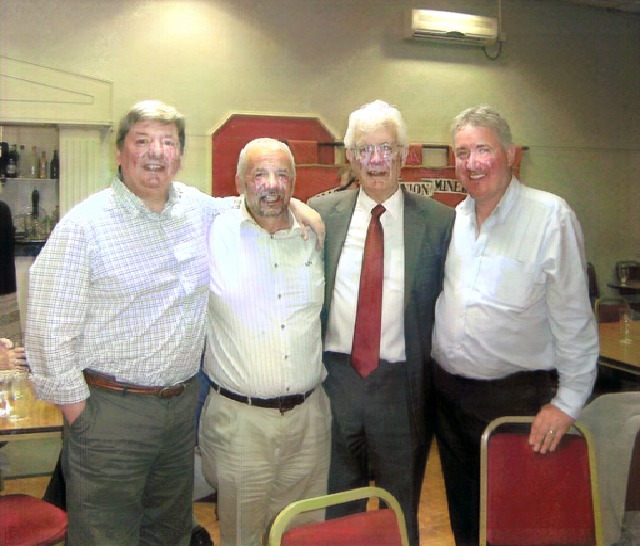}
 \\

 \end{tabular}
\caption{More results on CoCoClutter dataset.}
\label{fig:comparisons_CoCoClutter_1}
\end{figure*}

\begin{figure*}
\centering
\setlength{\imwidth}{0.16\linewidth}
\setbox1=\hbox{\includegraphics[width=\imwidth]{Ims/compFig/mask0.jpg}}
\setlength{\tabcolsep}{1pt}
\begin{tabular}{cccccc}
Input & Mask & Our & HAG~\cite{hagiwara2011saliency} & HOR~\cite{mateescu2014attention} & GAT~\cite{gatys2017guiding} 
\\
 \includegraphics[width=\imwidth]{Ims/CoCoClutter_supplement/im0.jpg}&
 \includegraphics[width=\imwidth]{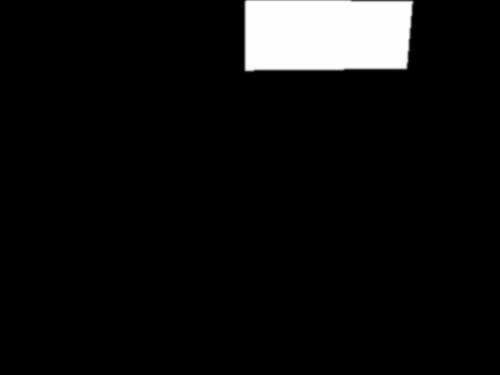}&
 \includegraphics[width=\imwidth]{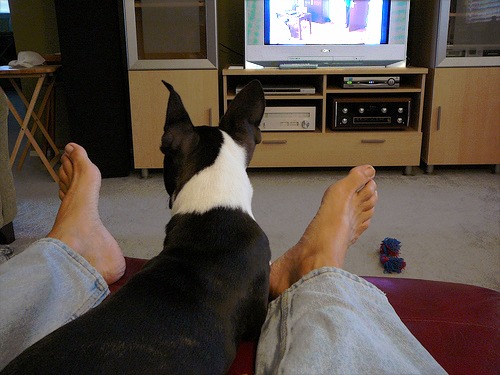}&
 \includegraphics[width=\imwidth]{Ims/CoCoClutter_supplement/im_hag_inc0_-0002.jpg}&
 \includegraphics[width=\imwidth]{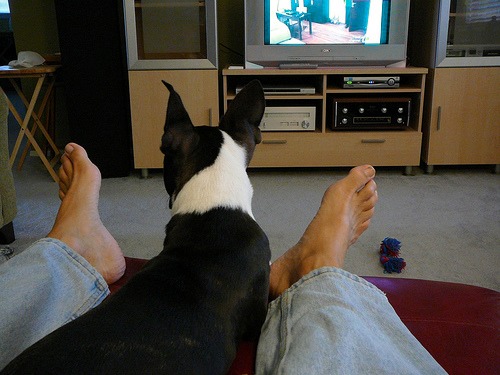}&
 \includegraphics[width=\imwidth]{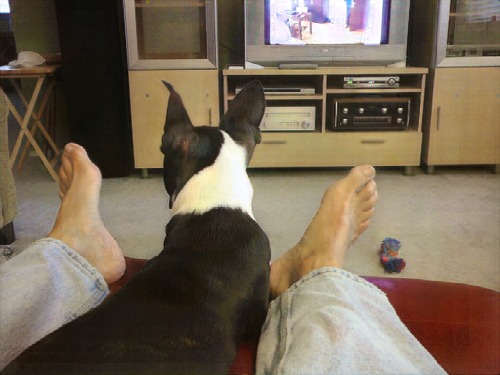}
 \\
  \includegraphics[width=\imwidth]{Ims/CoCoClutter_supplement/im1.jpg}&
 \includegraphics[width=\imwidth]{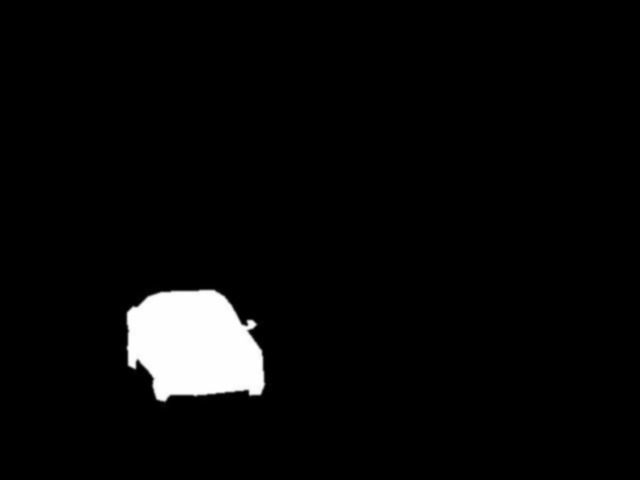}&
 \includegraphics[width=\imwidth]{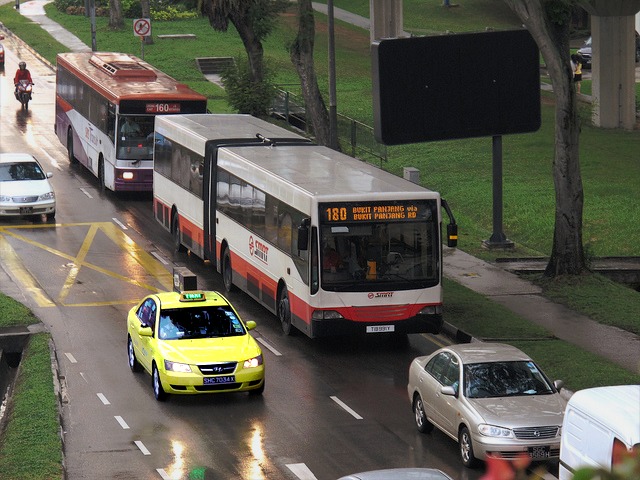}&
 \includegraphics[width=\imwidth]{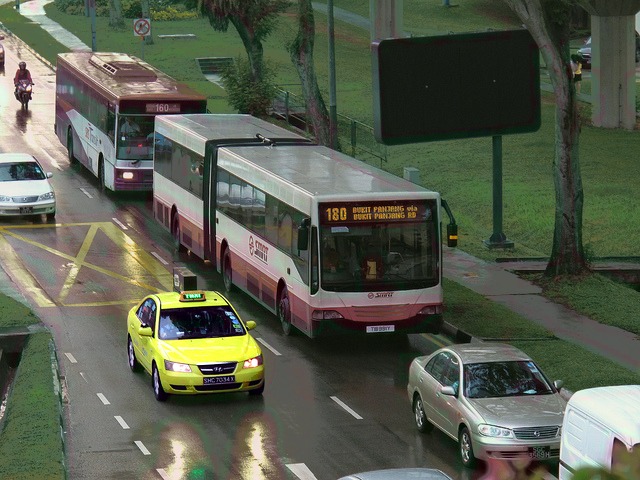}&
 \includegraphics[width=\imwidth]{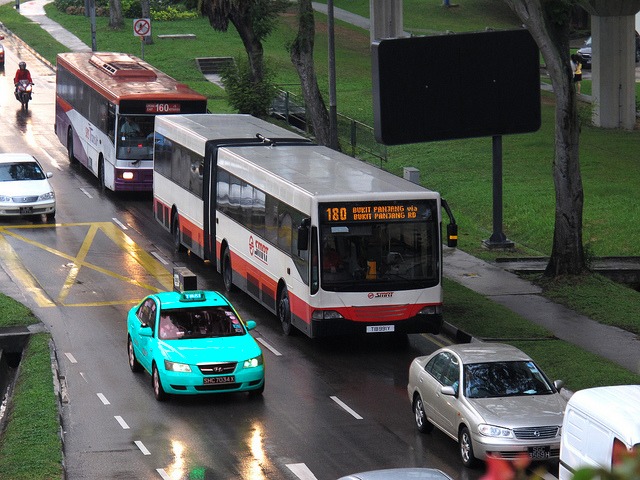}&
 \includegraphics[width=\imwidth]{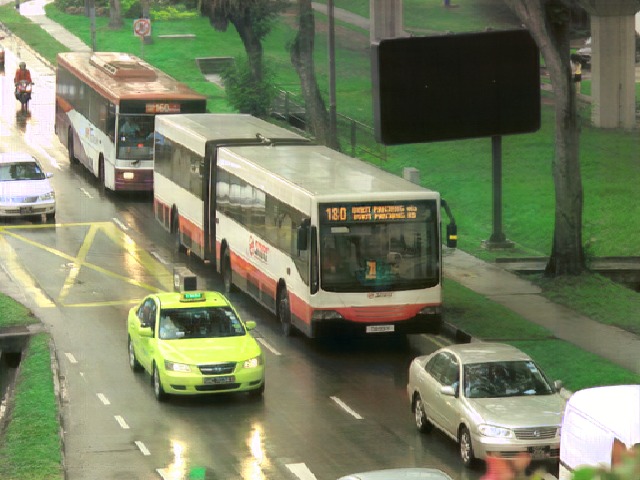}
 \\
  \includegraphics[width=\imwidth]{Ims/CoCoClutter_supplement/im2.jpg}&
 \includegraphics[width=\imwidth]{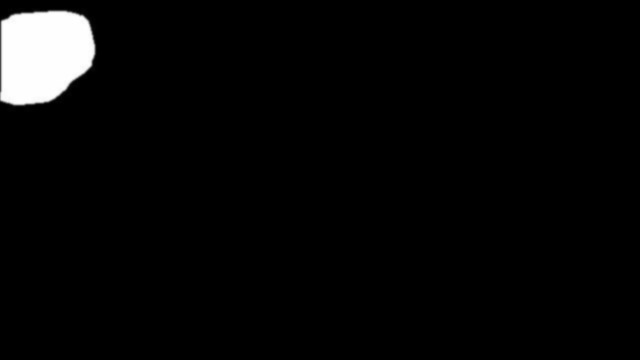}&
 \includegraphics[width=\imwidth]{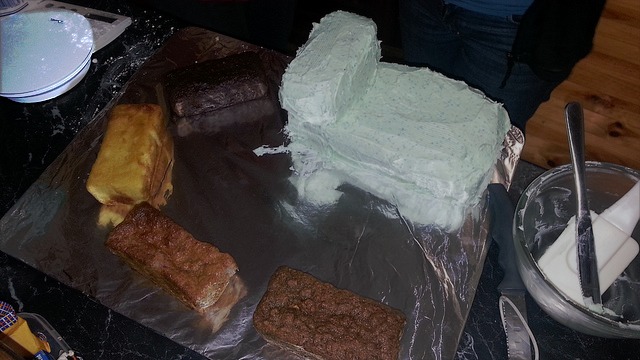}&
 \includegraphics[width=\imwidth]{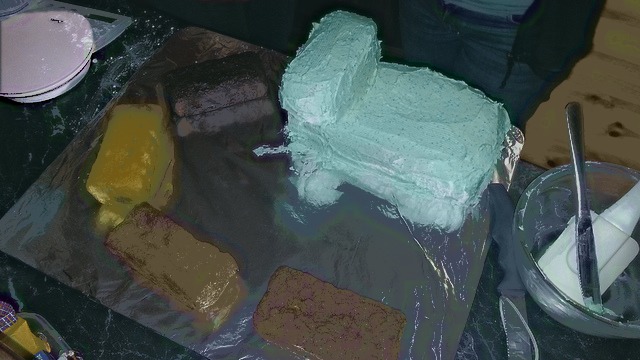}&
 \includegraphics[width=\imwidth]{Ims/CoCoClutter_supplement/im_hor_inc2_-0032.jpg}&
 \includegraphics[width=\imwidth]{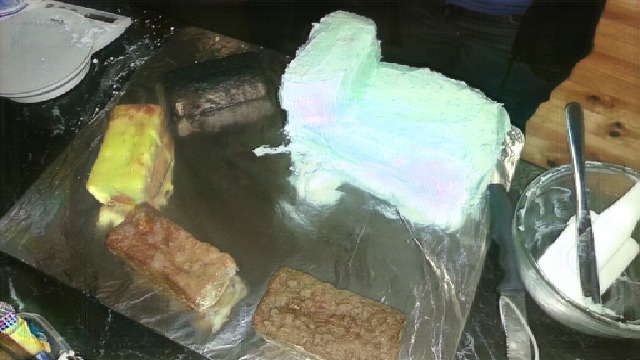}
 \\
  \includegraphics[width=\imwidth]{Ims/CoCoClutter_supplement/im3.jpg}&
 \includegraphics[width=\imwidth]{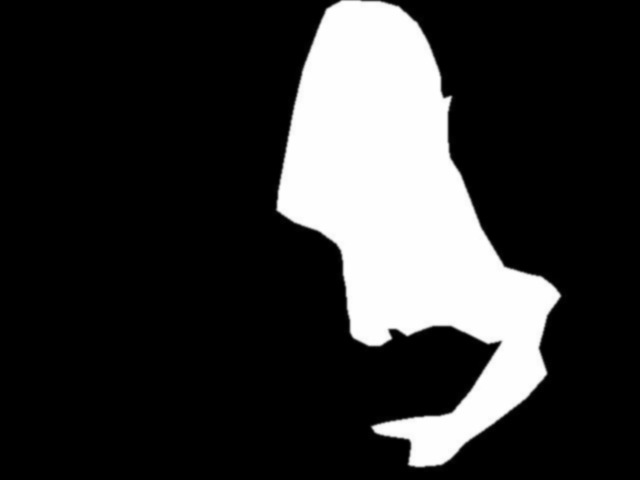}&
 \includegraphics[width=\imwidth]{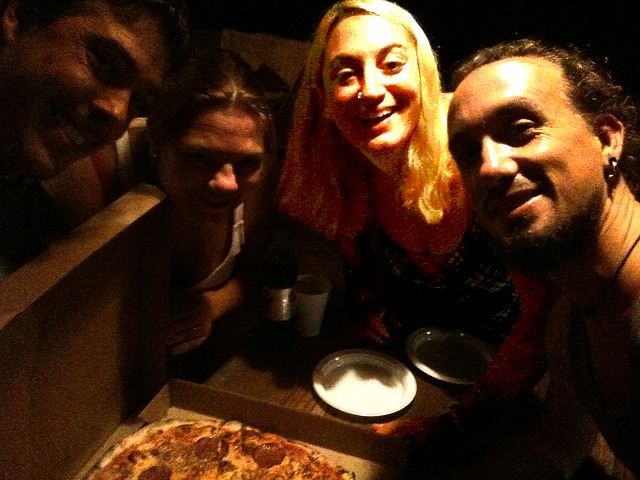}&
 \includegraphics[width=\imwidth]{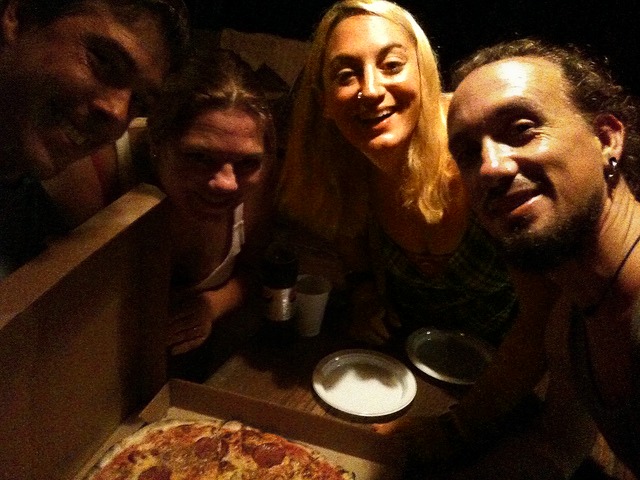}&
 \includegraphics[width=\imwidth]{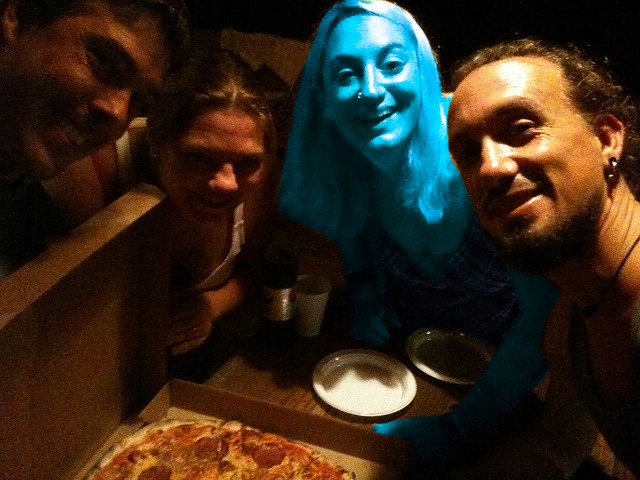}&
 \includegraphics[width=\imwidth]{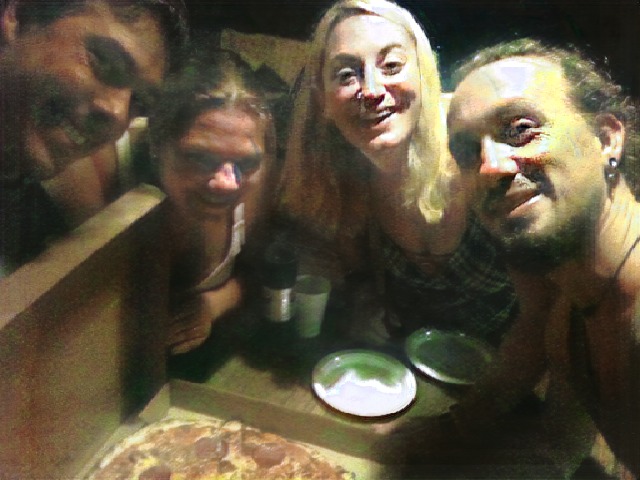}
 \\
  \includegraphics[width=\imwidth]{Ims/CoCoClutter_supplement/im4.jpg}&
 \includegraphics[width=\imwidth]{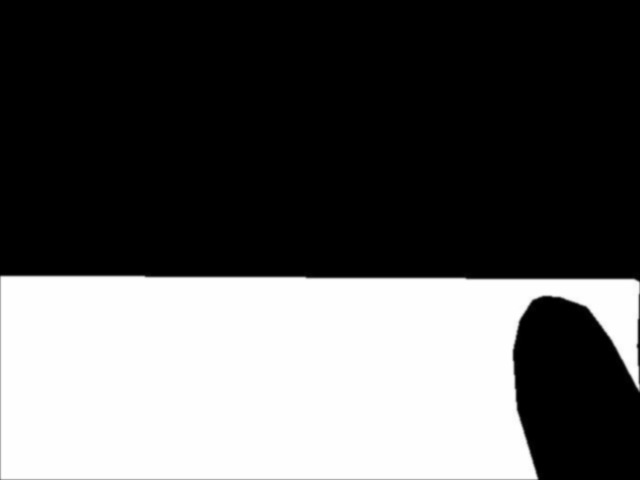}&
 \includegraphics[width=\imwidth]{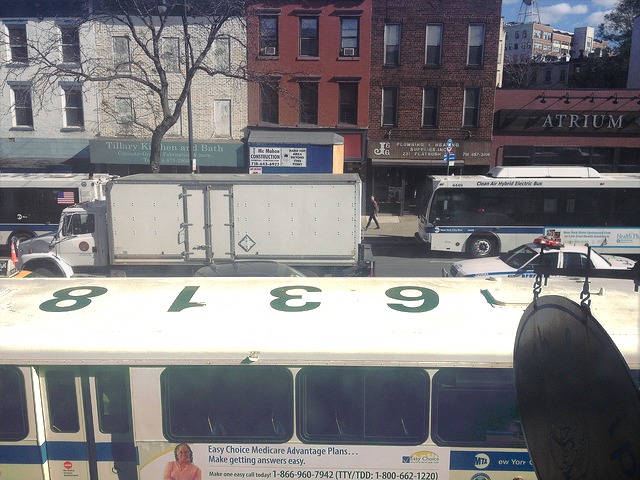}&
 \includegraphics[width=\imwidth]{Ims/CoCoClutter_supplement/im_hag_inc4_-0009.jpg}&
 \includegraphics[width=\imwidth]{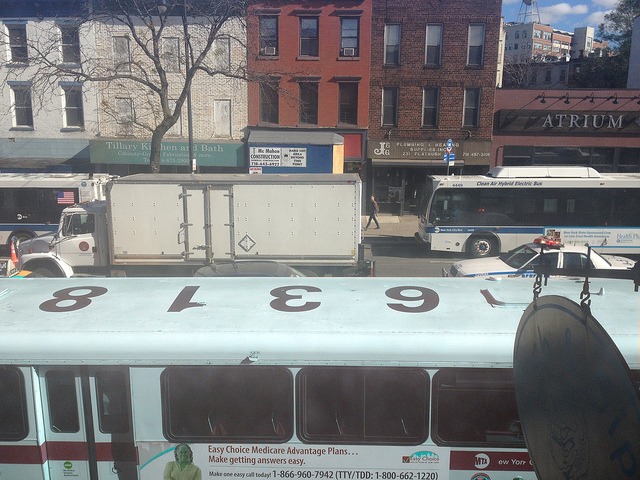}&
 \includegraphics[width=\imwidth]{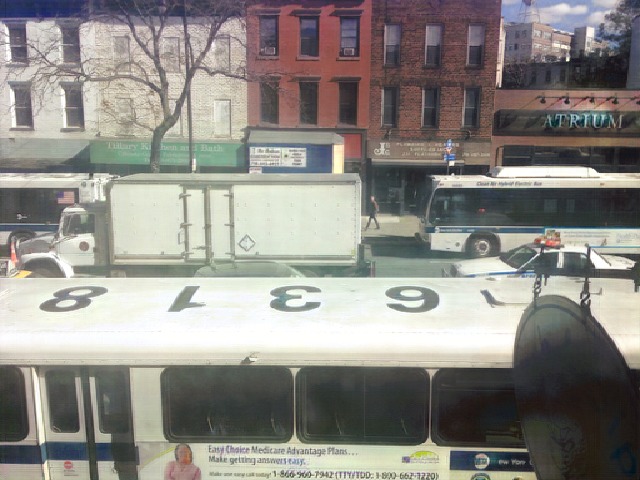}
 \\
  \includegraphics[width=\imwidth]{Ims/CoCoClutter_supplement/im11.jpg}&
 \includegraphics[width=\imwidth]{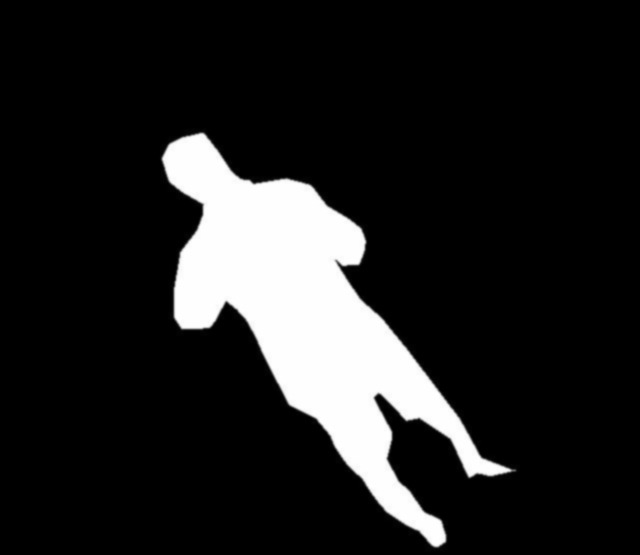}&
 \includegraphics[width=\imwidth]{Ims/CoCoClutter_supplement/imfinal_inc11_0057.jpg}&
 \includegraphics[width=\imwidth]{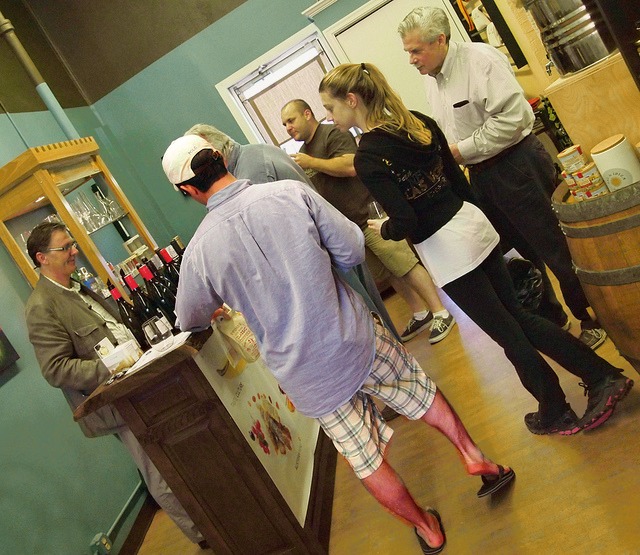}&
 \includegraphics[width=\imwidth]{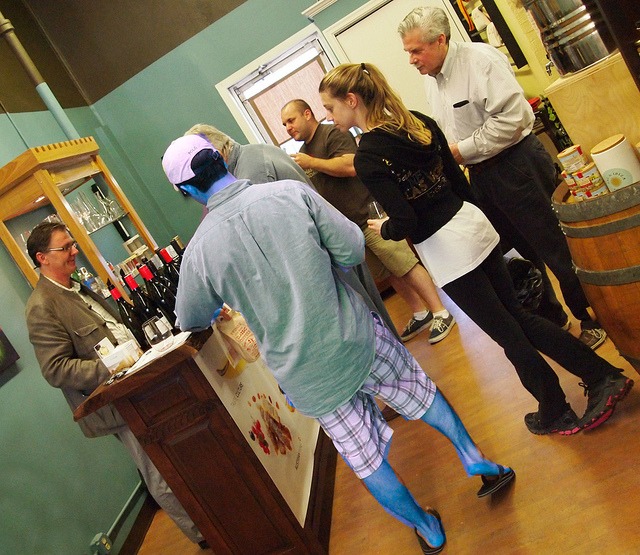}&
 \includegraphics[width=\imwidth]{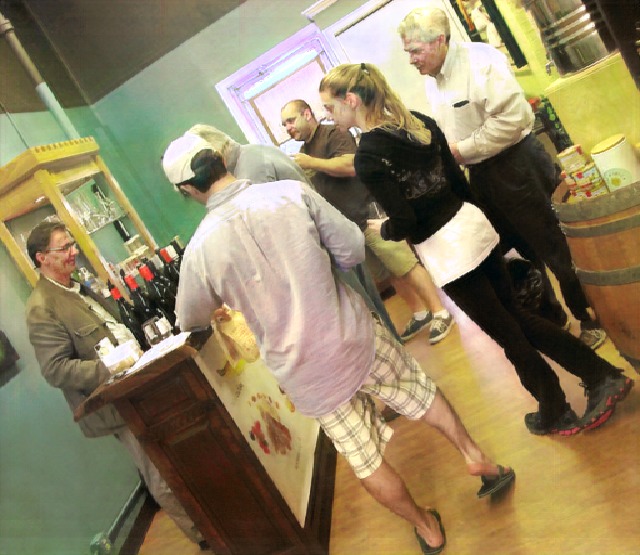}
 \\
  \includegraphics[width=\imwidth]{Ims/CoCoClutter_supplement/im6.jpg}&
 \includegraphics[width=\imwidth]{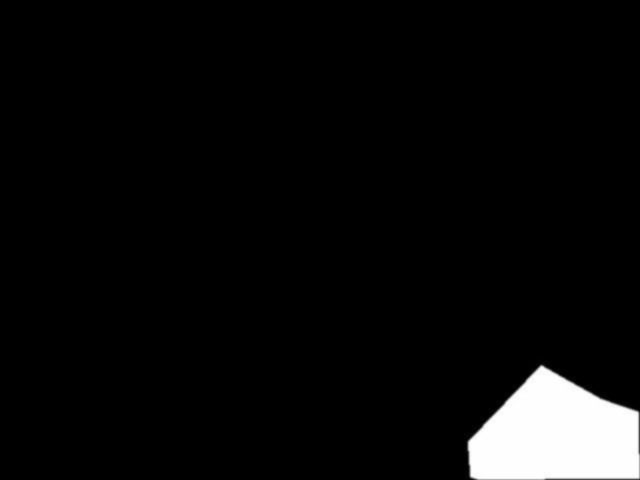}&
 \includegraphics[width=\imwidth]{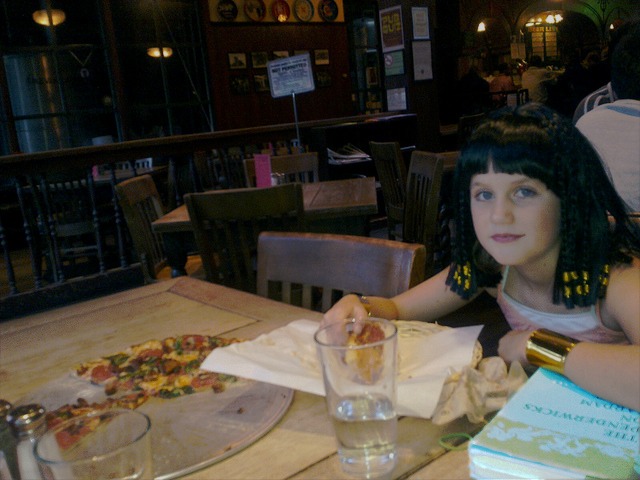}&
 \includegraphics[width=\imwidth]{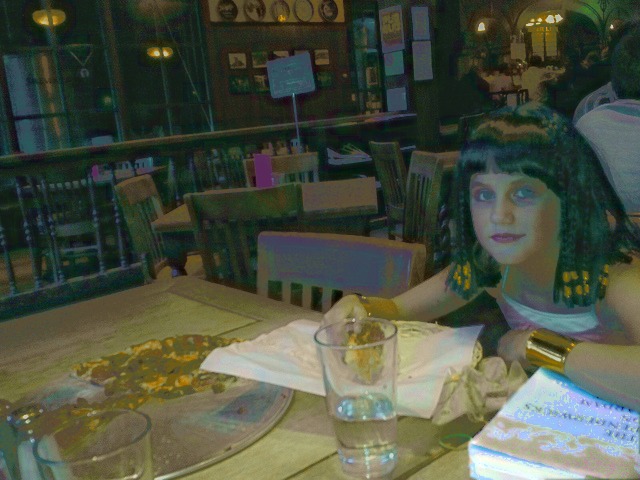}&
 \includegraphics[width=\imwidth]{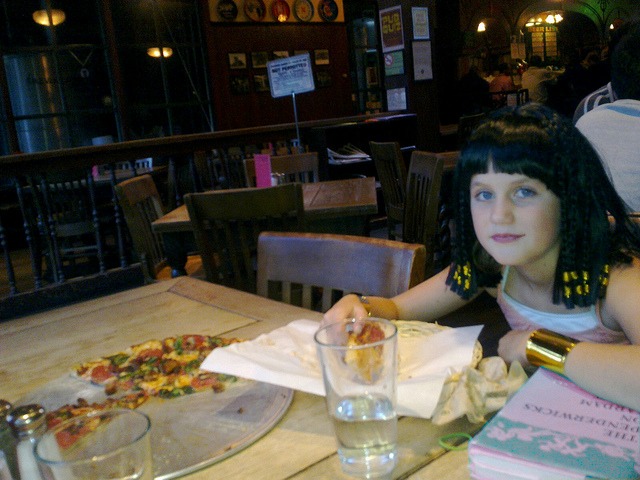}&
 \includegraphics[width=\imwidth]{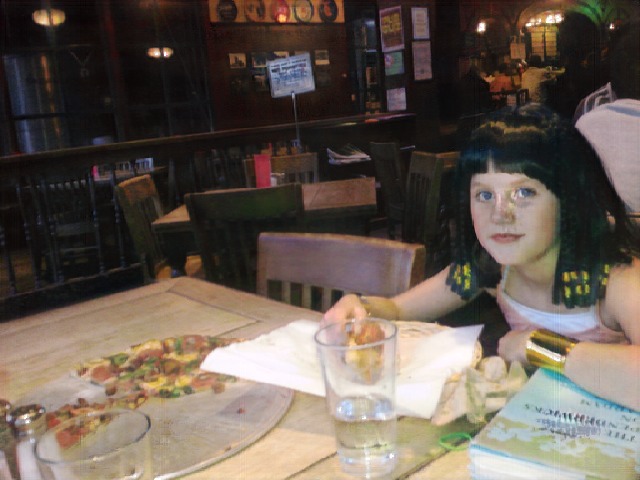}
 \\
  \includegraphics[width=\imwidth]{Ims/CoCoClutter_supplement/im7.jpg}&
 \includegraphics[width=\imwidth]{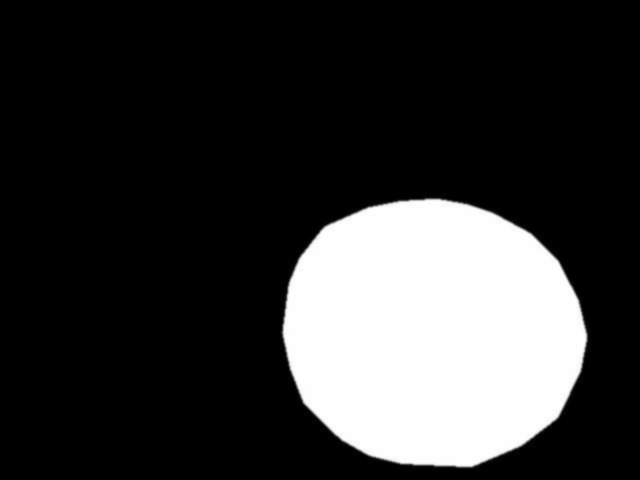}&
 \includegraphics[width=\imwidth]{Ims/CoCoClutter_supplement/imfinal_inc7_0047.jpg}&
 \includegraphics[width=\imwidth]{Ims/CoCoClutter_supplement/im_hag_inc7_-0001.jpg}&
 \includegraphics[width=\imwidth]{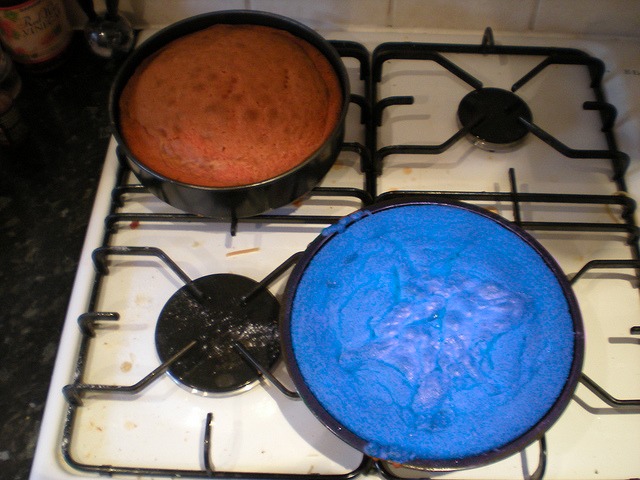}&
 \includegraphics[width=\imwidth]{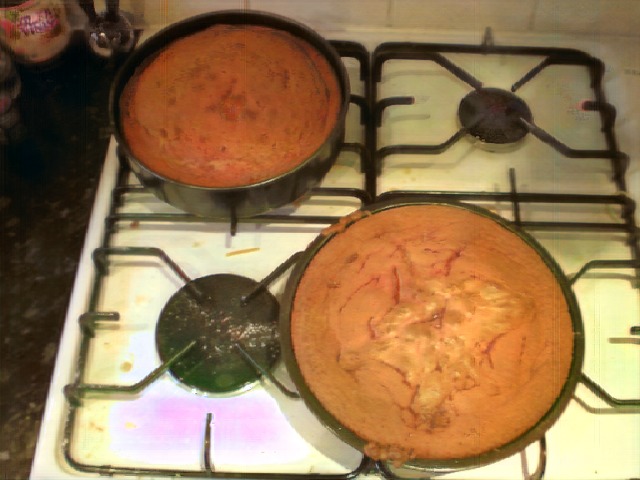}
 \\
   \includegraphics[width=\imwidth]{Ims/CoCoClutter_supplement/im9.jpg}&
 \includegraphics[width=\imwidth]{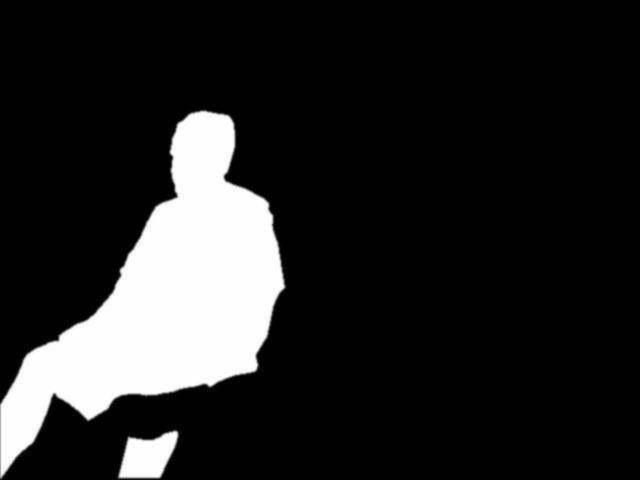}&
 \includegraphics[width=\imwidth]{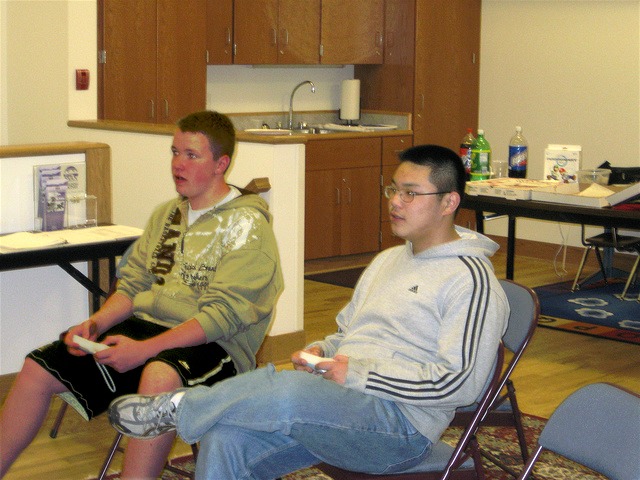}&
 \includegraphics[width=\imwidth]{Ims/CoCoClutter_supplement/im_hag_inc9_0014.jpg}&
 \includegraphics[width=\imwidth]{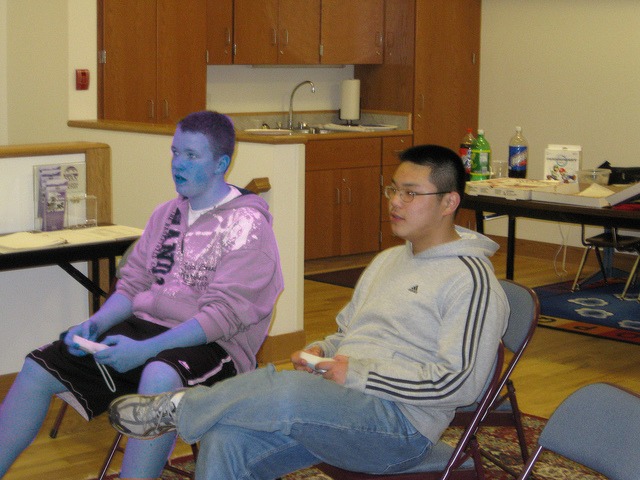}&
 \includegraphics[width=\imwidth]{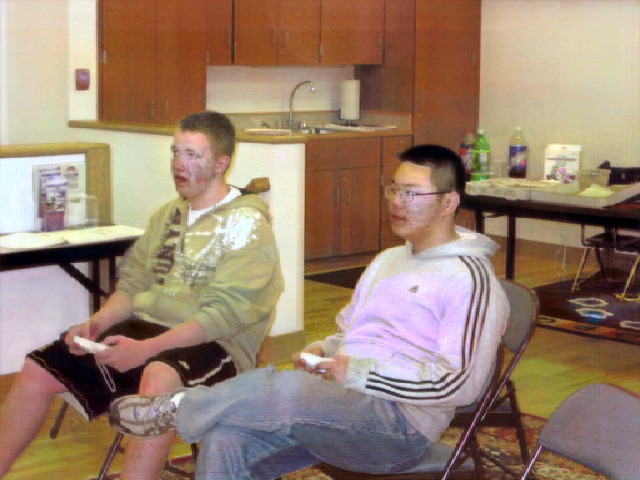}
 \\
   \includegraphics[width=\imwidth]{Ims/CoCoClutter_supplement/im10.jpg}&
 \includegraphics[width=\imwidth]{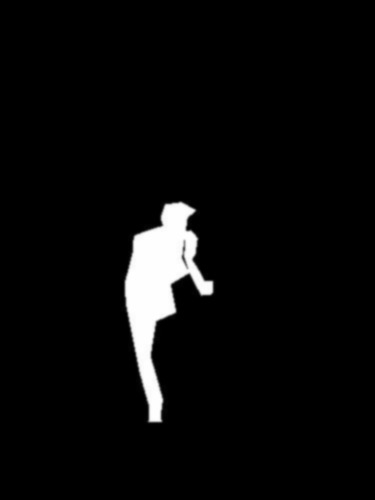}&
 \includegraphics[width=\imwidth]{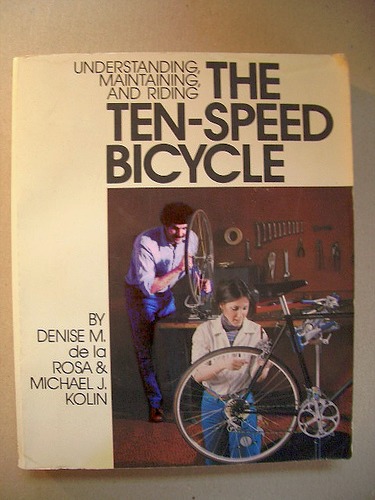}&
 \includegraphics[width=\imwidth]{Ims/CoCoClutter_supplement/im_hag_inc10_0017.jpg}&
 \includegraphics[width=\imwidth]{Ims/CoCoClutter_supplement/im_hor_inc10_0036.jpg}&
 \includegraphics[width=\imwidth]{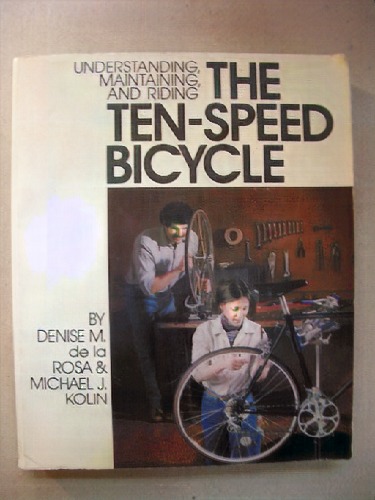}
 \\
    \includegraphics[width=\imwidth]{Ims/CoCoClutter_supplement/im13.jpg}&
 \includegraphics[width=\imwidth]{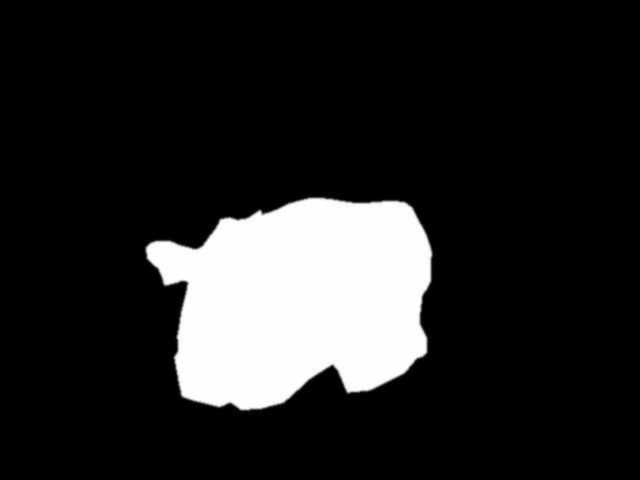}&
 \includegraphics[width=\imwidth]{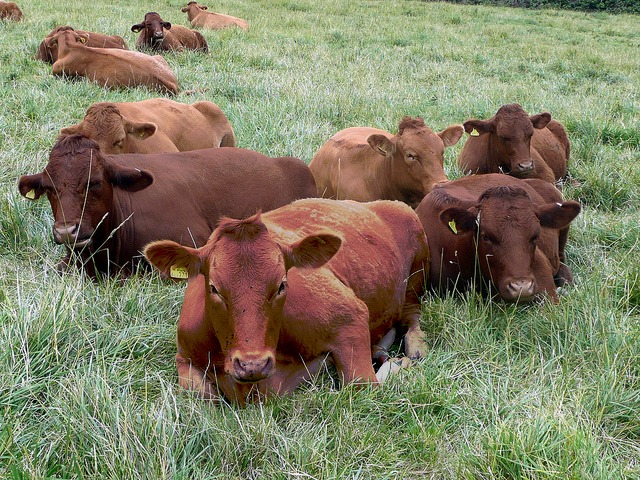}&
 \includegraphics[width=\imwidth]{Ims/CoCoClutter_supplement/im_hag_inc13_0009.jpg}&
 \includegraphics[width=\imwidth]{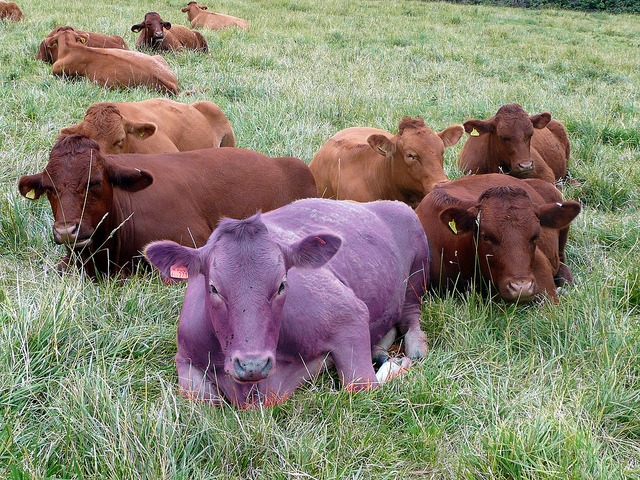}&
 \includegraphics[width=\imwidth]{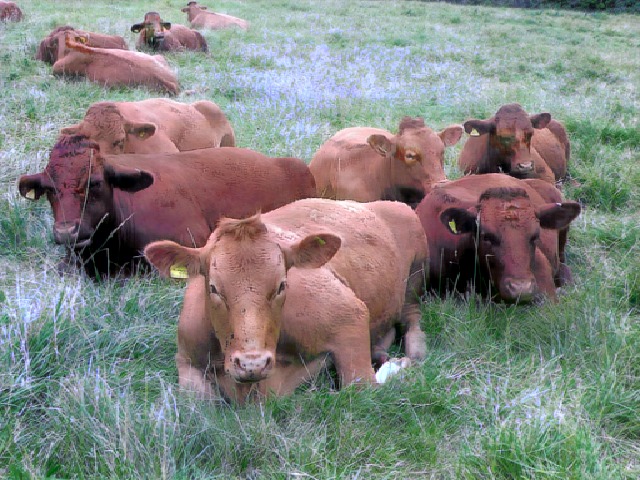}
 \\

 \end{tabular}
\caption{More results on CoCoClutter dataset.}
\label{fig:comparisons_CoCoClutter_2}
\end{figure*}

\begin{figure*}
\centering
\setlength{\imwidth}{0.19\linewidth}
\setlength{\tabcolsep}{1pt}
\begin{tabular}{ccccc}

Input & Mask  & $z_1$ & $z_2$ & $z_3$ 
\\
\includegraphics[width=\imwidth]{Ims/MultiStyle/im262.jpg} &
\includegraphics[width=\imwidth]{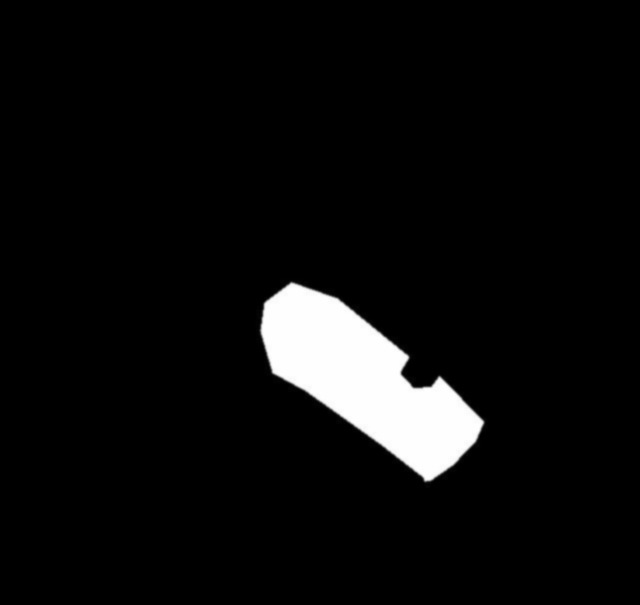}&
\includegraphics[width=\imwidth]{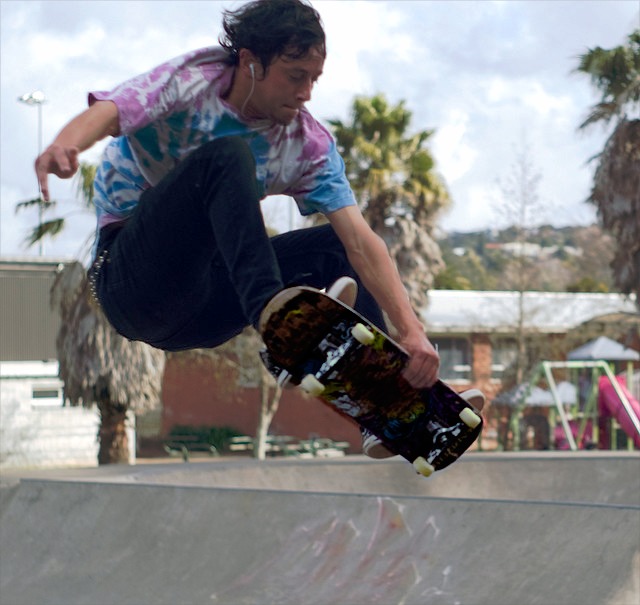}&
\includegraphics[width=\imwidth]{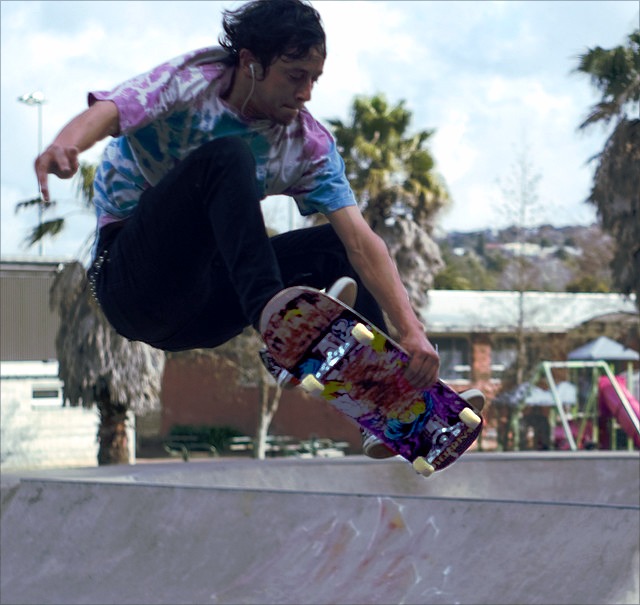}&
\includegraphics[width=\imwidth]{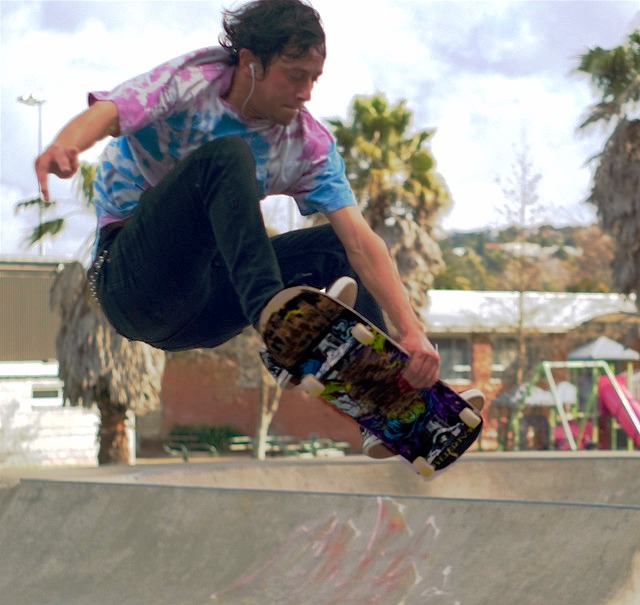}\\

\includegraphics[width=\imwidth]{Ims/MultiStyle/im277.jpg} &
\includegraphics[width=\imwidth]{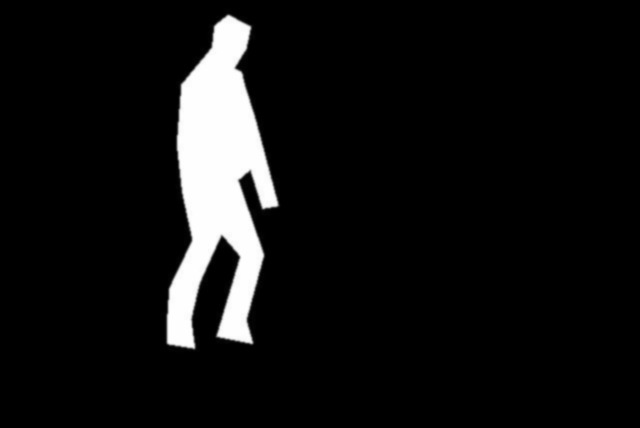}&
\includegraphics[width=\imwidth]{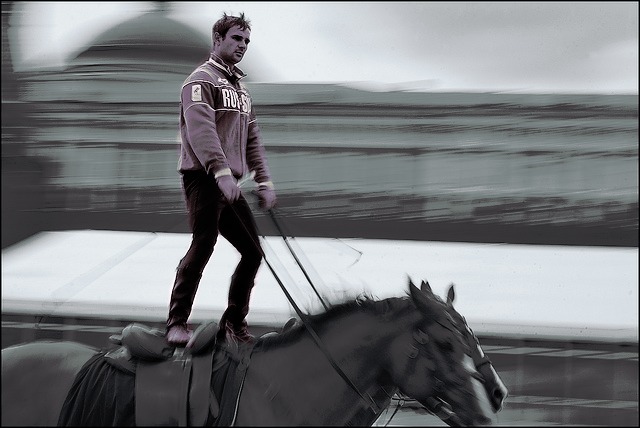}&
\includegraphics[width=\imwidth]{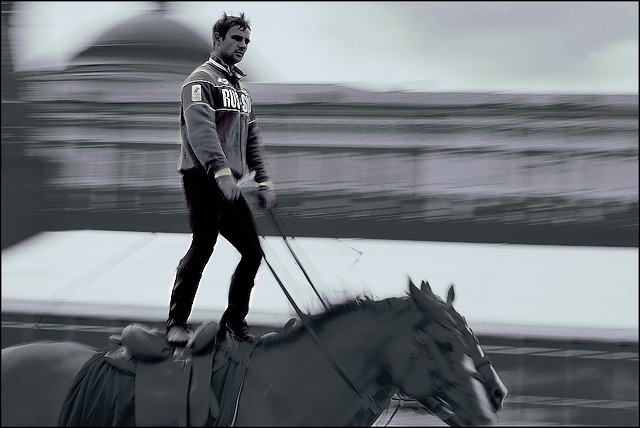}&
\includegraphics[width=\imwidth]{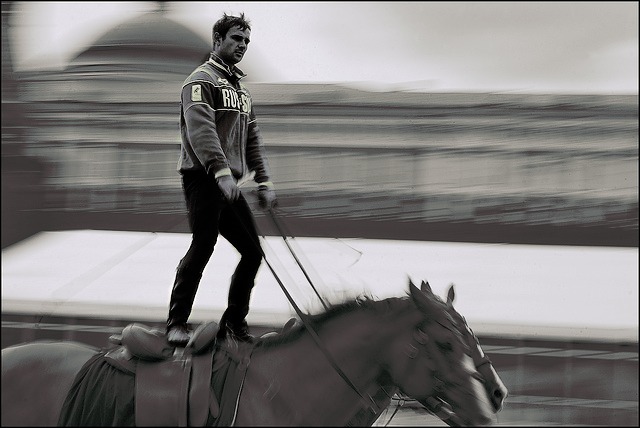}\\

\includegraphics[width=\imwidth]{Ims/MultiStyle/im746.jpg} &
\includegraphics[width=\imwidth]{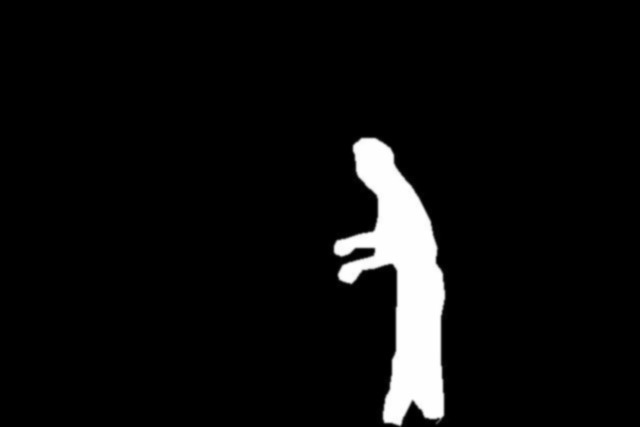}&
\includegraphics[width=\imwidth]{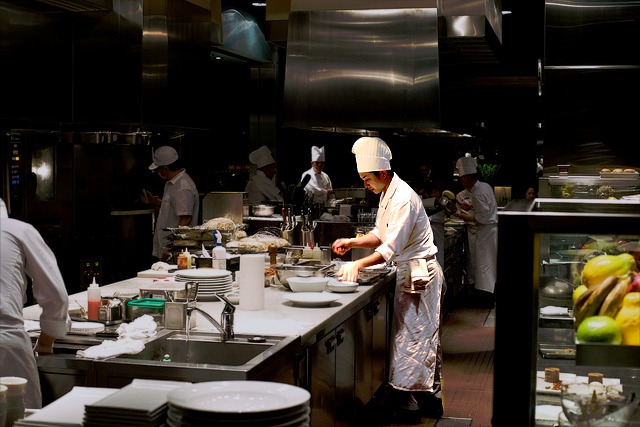}&
\includegraphics[width=\imwidth]{Ims/MultiStyle/imfinal_inc2746_0009.jpg}&
\includegraphics[width=\imwidth]{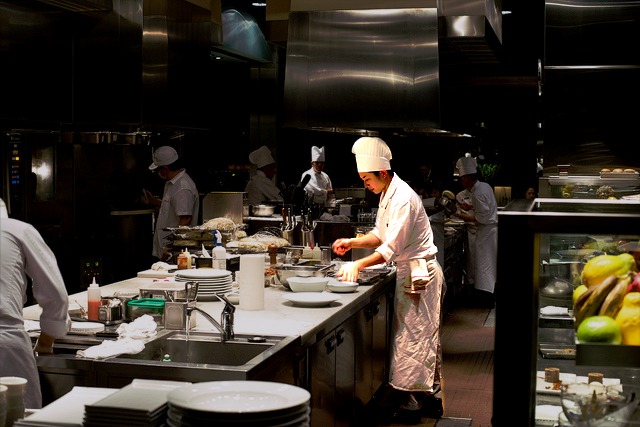}\\

\includegraphics[width=\imwidth]{Ims/MultiStyle/im18.jpg} &
\includegraphics[width=\imwidth]{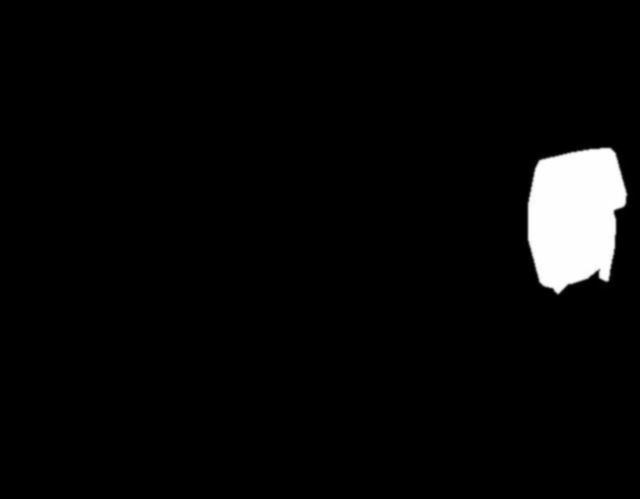}&
\includegraphics[width=\imwidth]{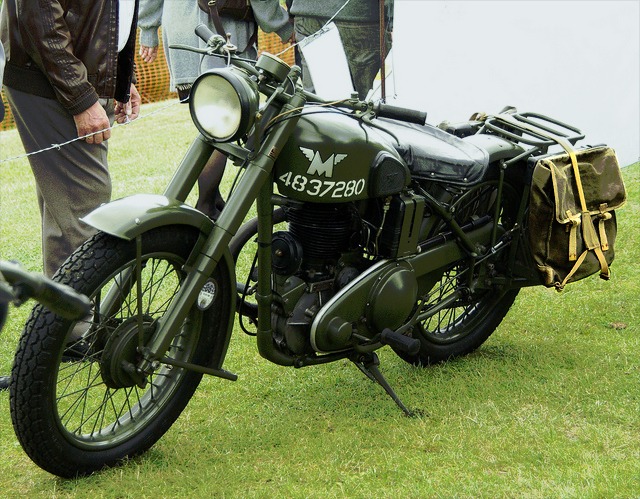}&
\includegraphics[width=\imwidth]{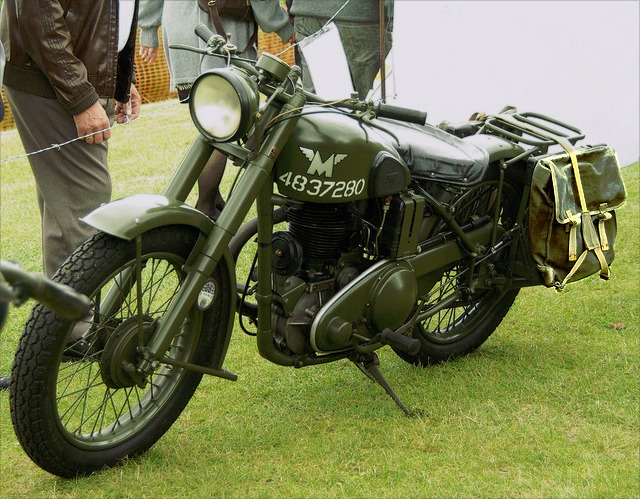}&
\includegraphics[width=\imwidth]{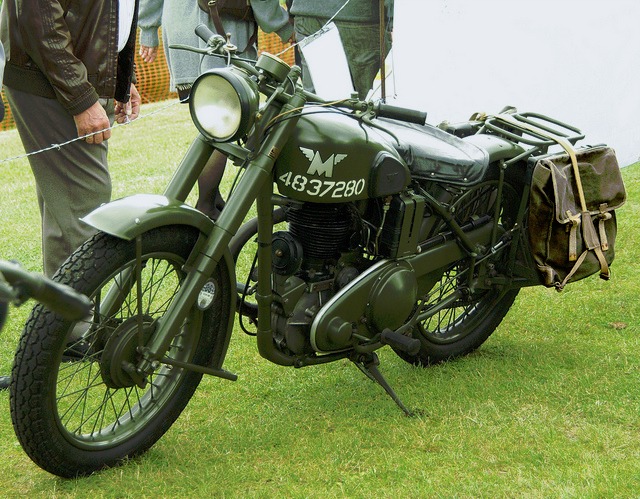}\\

\includegraphics[width=\imwidth]{Ims/MultiStyle/im39.jpg} &
\includegraphics[width=\imwidth]{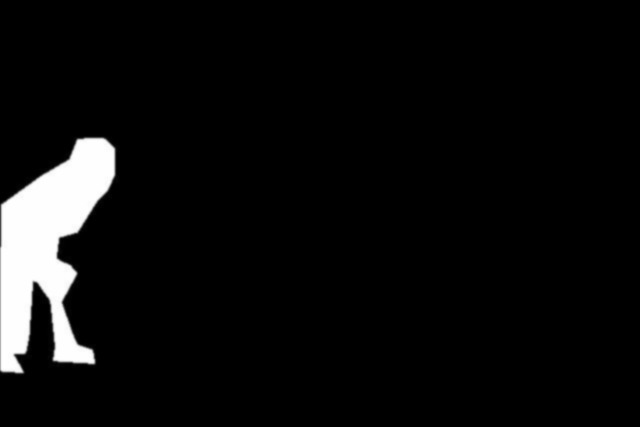}&
\includegraphics[width=\imwidth]{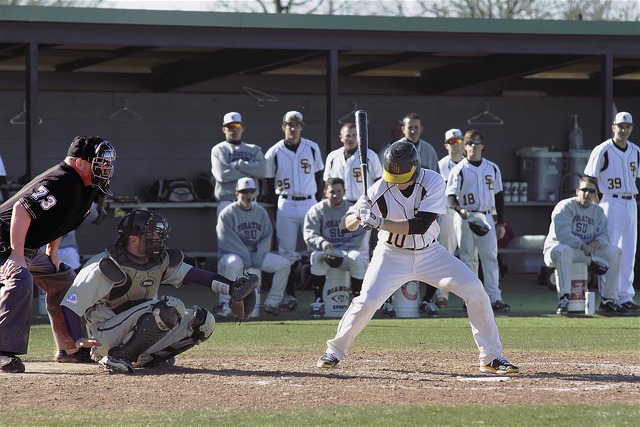}&
\includegraphics[width=\imwidth]{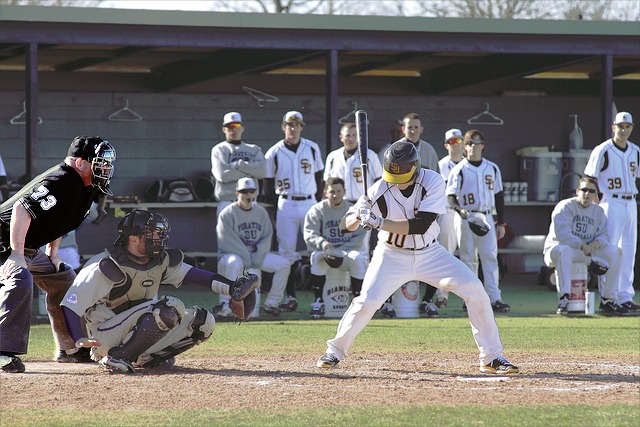}&
\includegraphics[width=\imwidth]{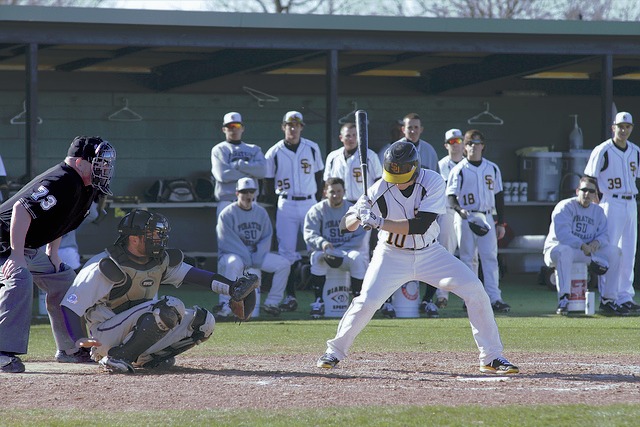}\\

\includegraphics[width=\imwidth]{Ims/MultiStyle/im46.jpg} &
\includegraphics[width=\imwidth]{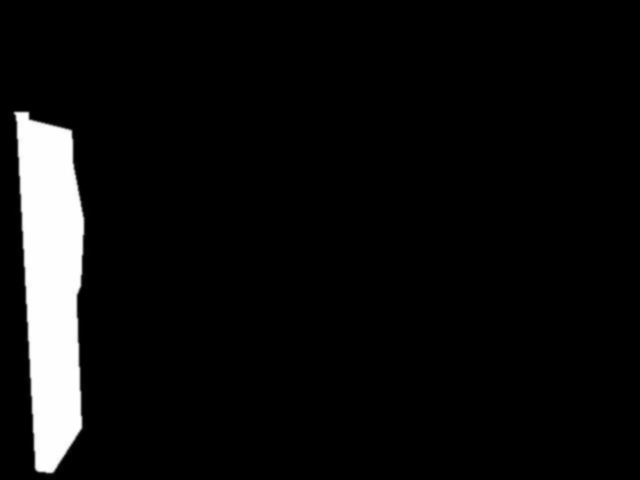}&
\includegraphics[width=\imwidth]{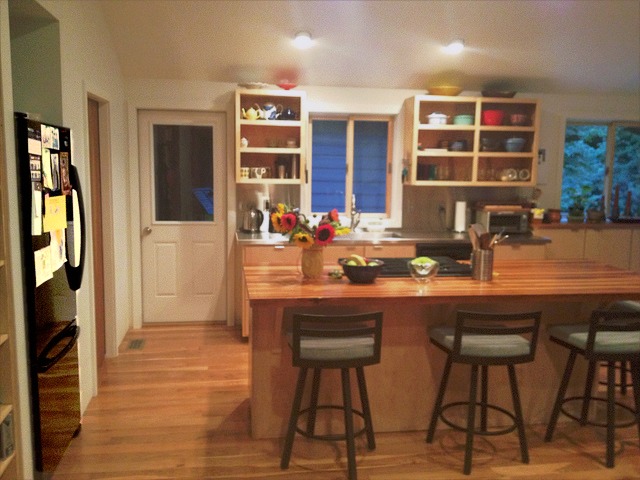}&
\includegraphics[width=\imwidth]{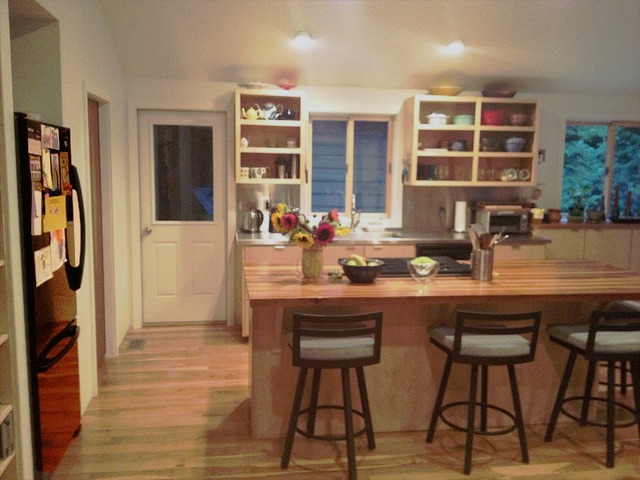}&
\includegraphics[width=\imwidth]{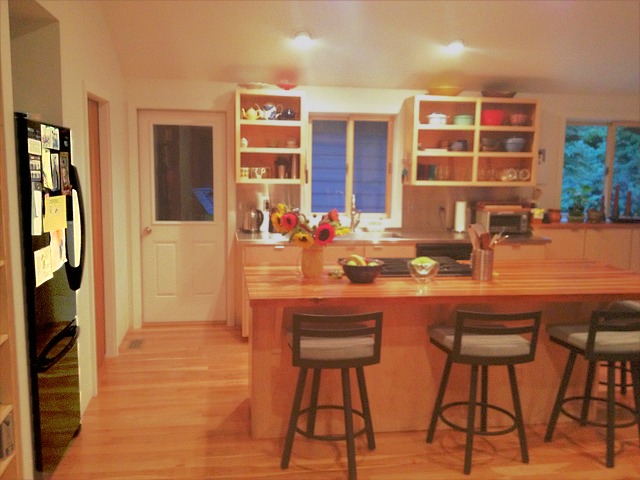}\\

\includegraphics[width=\imwidth]{Ims/MultiStyle/im38.jpg} &
\includegraphics[width=\imwidth]{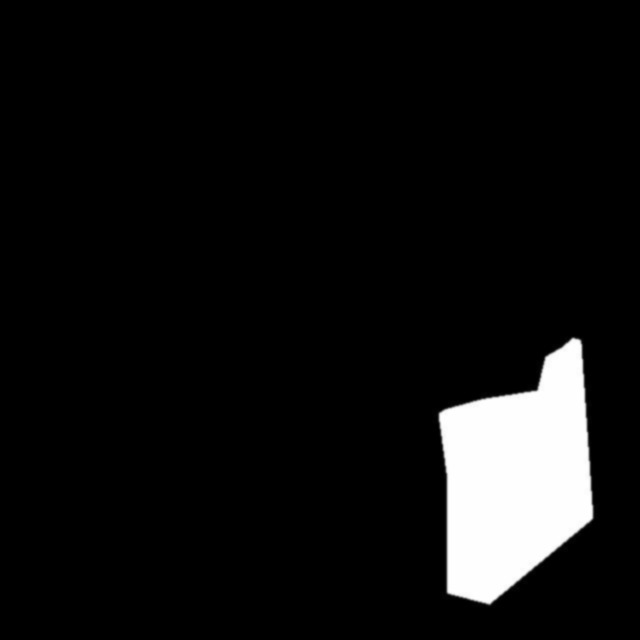}&
\includegraphics[width=\imwidth]{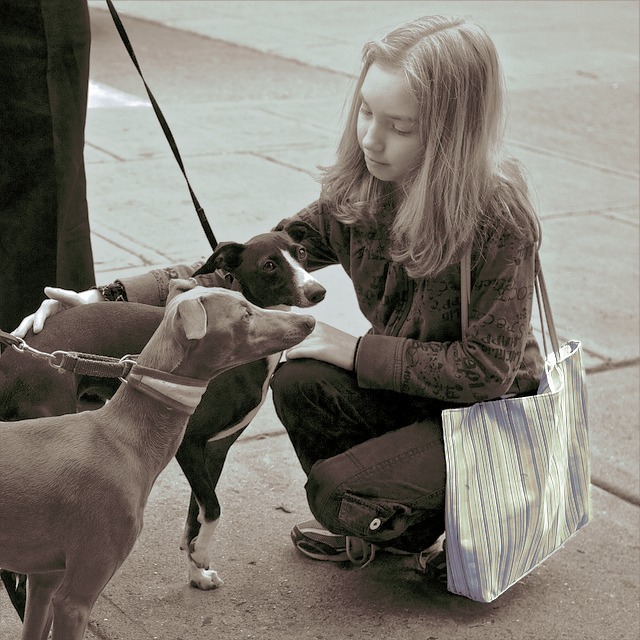}&
\includegraphics[width=\imwidth]{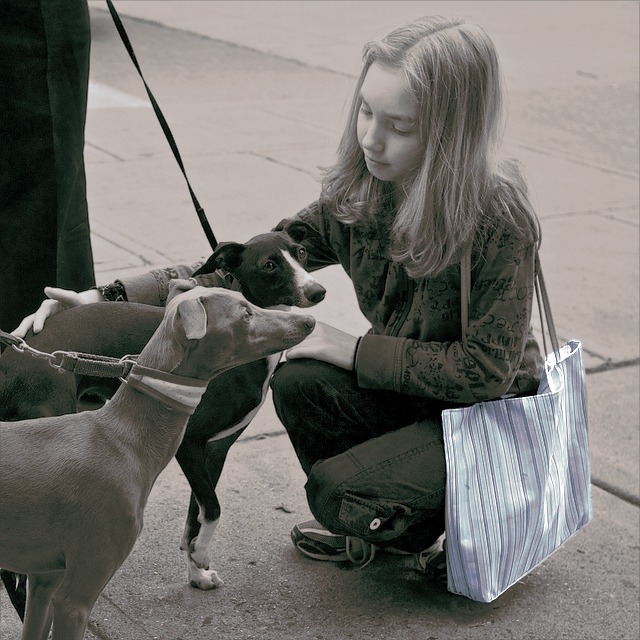}&
\includegraphics[width=\imwidth]{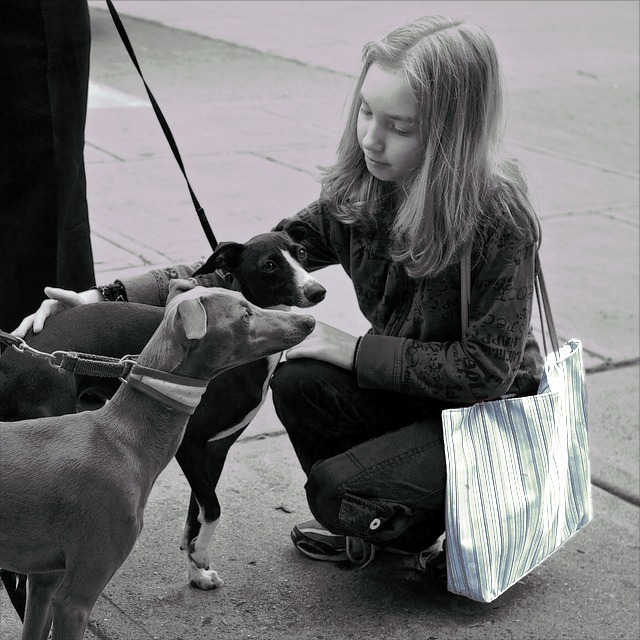}\\

\includegraphics[width=\imwidth]{Ims/MultiStyle/im738.jpg} &
\includegraphics[width=\imwidth]{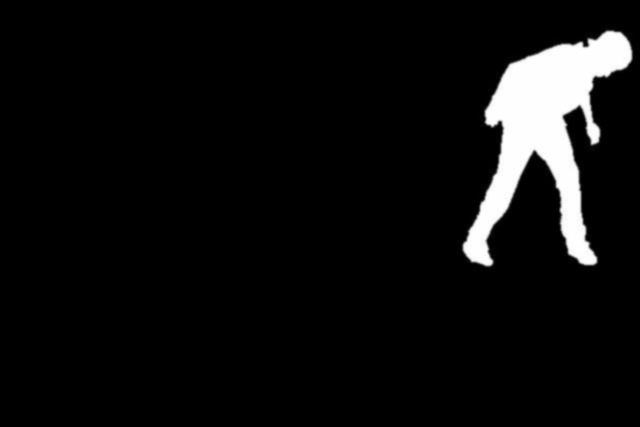}&
\includegraphics[width=\imwidth]{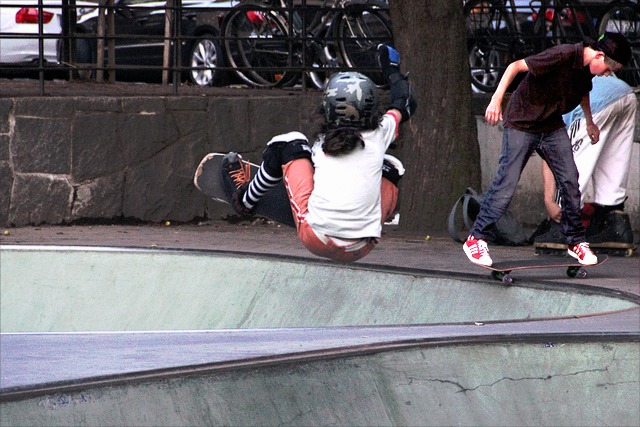}&
\includegraphics[width=\imwidth]{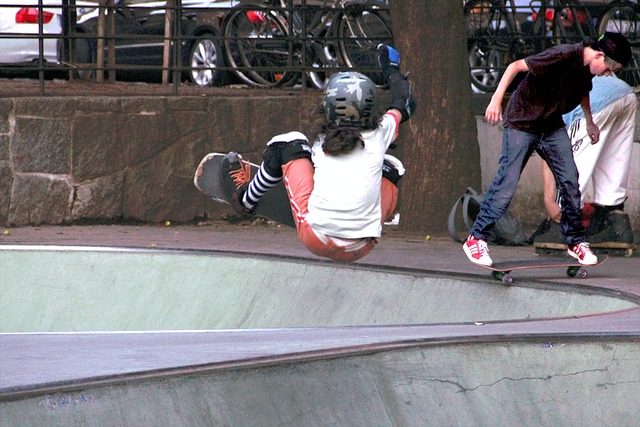}&
\includegraphics[width=\imwidth]{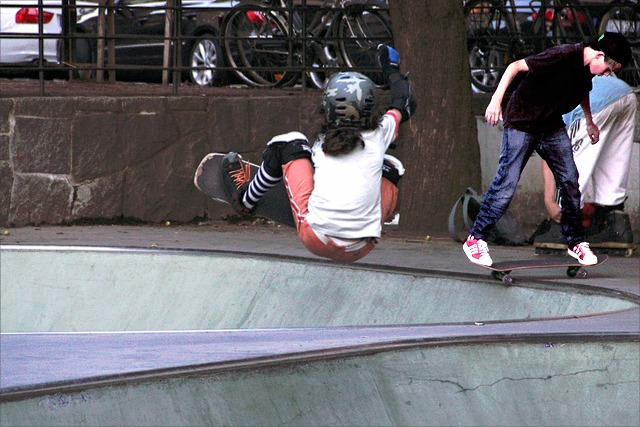}\\

\includegraphics[width=\imwidth]{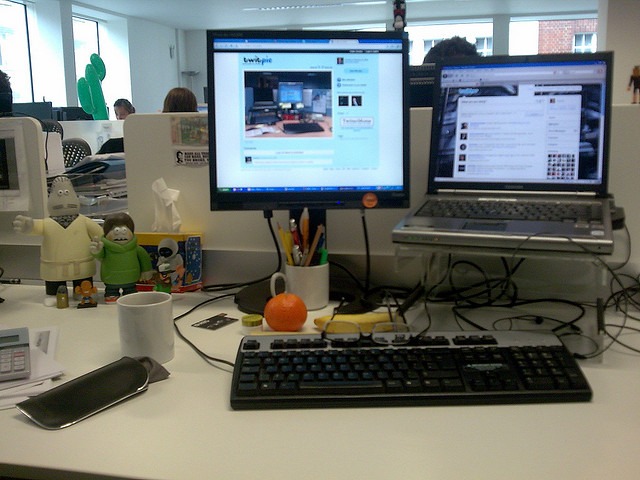} &
\includegraphics[width=\imwidth]{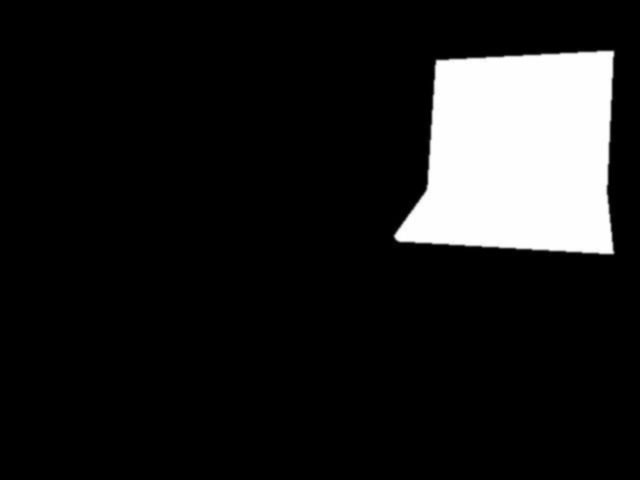}&
\includegraphics[width=\imwidth]{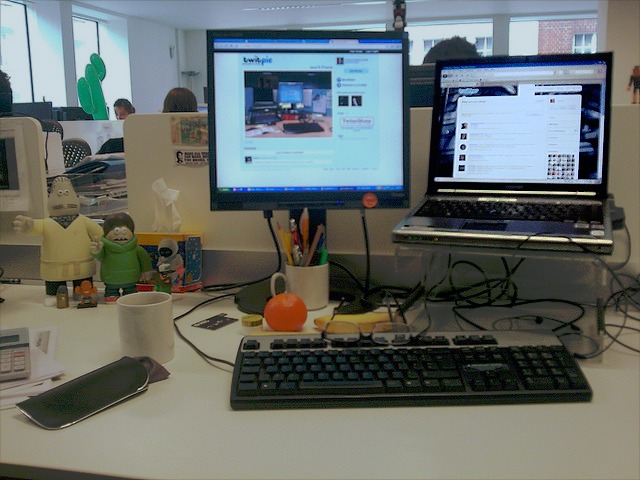}&
\includegraphics[width=\imwidth]{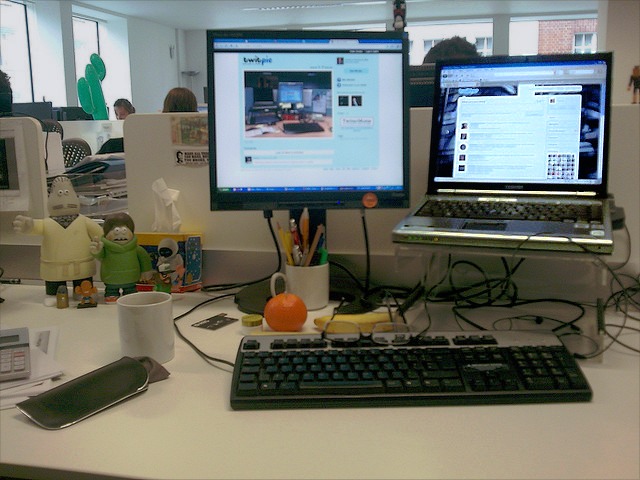}&
\includegraphics[width=\imwidth]{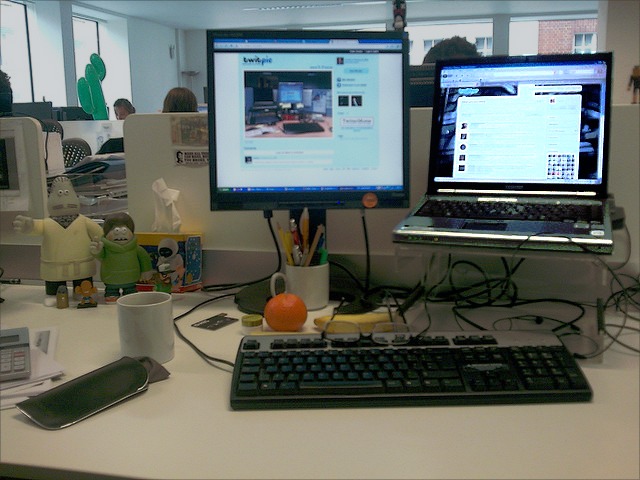}\\

\includegraphics[width=\imwidth]{Ims/MultiStyle/im345.jpg} &
\includegraphics[width=\imwidth]{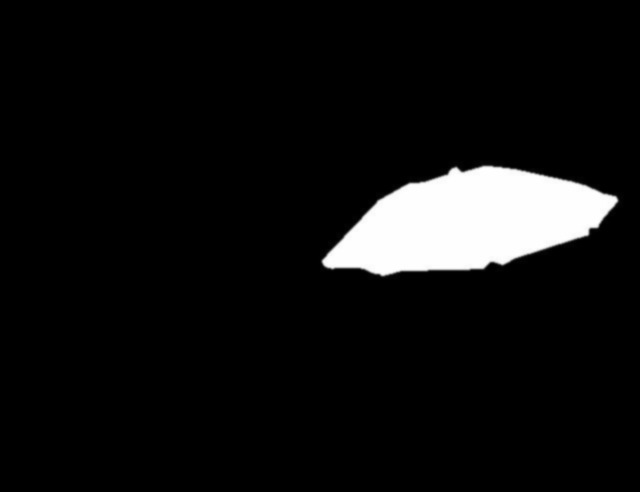}&
\includegraphics[width=\imwidth]{Ims/MultiStyle/imfinal_inc5345_0072.jpg}&
\includegraphics[width=\imwidth]{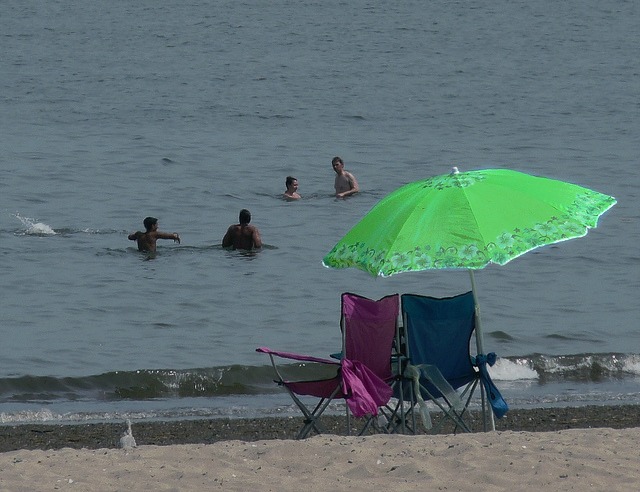}&
\includegraphics[width=\imwidth]{Ims/MultiStyle/imfinal_inc4345_0021.jpg}\\

\end{tabular}
\caption{More results on stochastic image generation.}
\label{fig:multistyle}
\end{figure*}

Additional qualitative results on the Mechrez dataset are provided in Figures~\ref{fig:comparisons_mechrez_1},~\ref{fig:comparisons_mechrez_2} and on the CoCoClutter dataset in Figures~\ref{fig:comparisons_CoCoClutter_1},~\ref{fig:comparisons_CoCoClutter_2}.
Figure~\ref{fig:multistyle} includes additional results using the stochastic image generation model. Figure~\ref{fig:incdec} provide additional results for increasing and decreasing saliency.

\begin{figure*}
\centering
\setlength{\imwidth}{0.16\linewidth}
\setbox1=\hbox{\includegraphics[width=\wc\imwidth]{Ims/incdec_celso/mask15.jpg}}
\setlength{\tabcolsep}{1pt}
\begin{tabular}{cccccc}

Input & Saliency Map & $\uparrow$ Attention & Saliency Map & $\downarrow$ Attention & Saliency Map
\\

\includegraphics[width=\imwidth]{Ims/incdec_supplement/im1730.jpg}\llap{{\raisebox{0\imwidth}{\includegraphics[cfbox=red, width=0.3\imwidth]{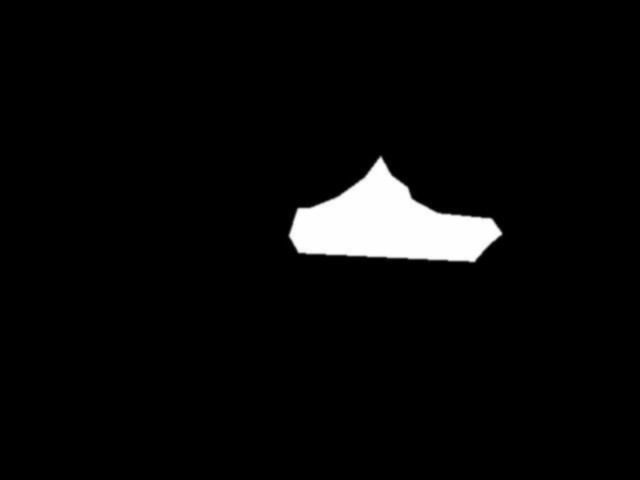}}}}&
\includegraphics[width=\imwidth]{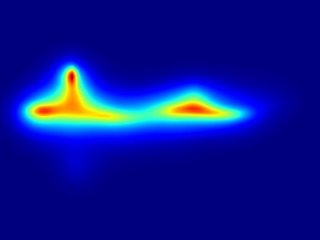} &
\includegraphics[width=\imwidth]{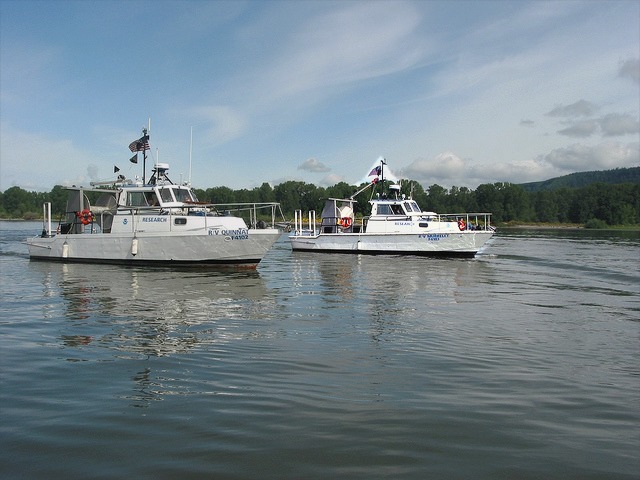} &
\includegraphics[width=\imwidth]{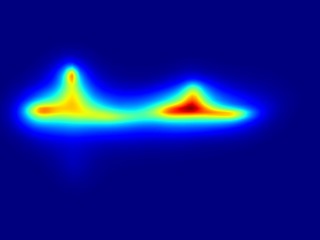} &
\includegraphics[width=\imwidth]{Ims/incdec_supplement/imfinal_inc11730_-0084.jpg}&
\includegraphics[width=\imwidth]{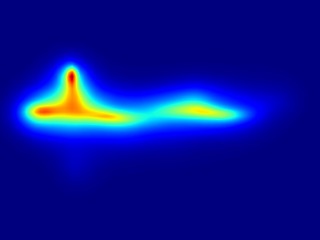}
\\

\includegraphics[width=\imwidth]{Ims/incdec_supplement/im120.jpg}\llap{{\raisebox{0\imwidth}{\includegraphics[cfbox=red, width=0.3\imwidth]{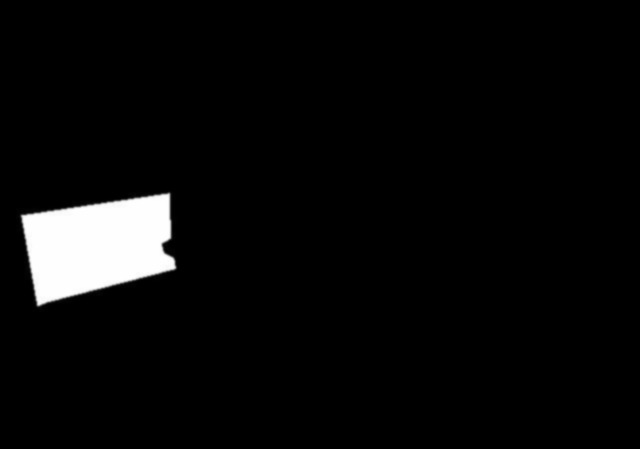}}}}&
\includegraphics[width=\imwidth, height = 1.35cm]{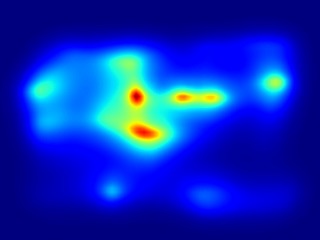} &
\includegraphics[width=\imwidth]{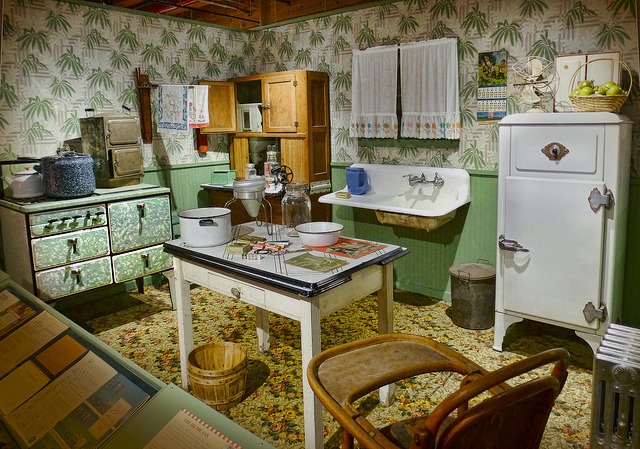} &
\includegraphics[width=\imwidth, height = 1.35cm]{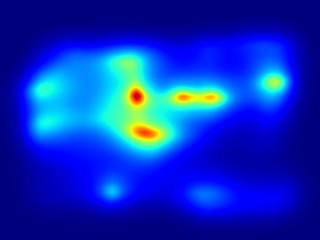} &
\includegraphics[width=\imwidth]{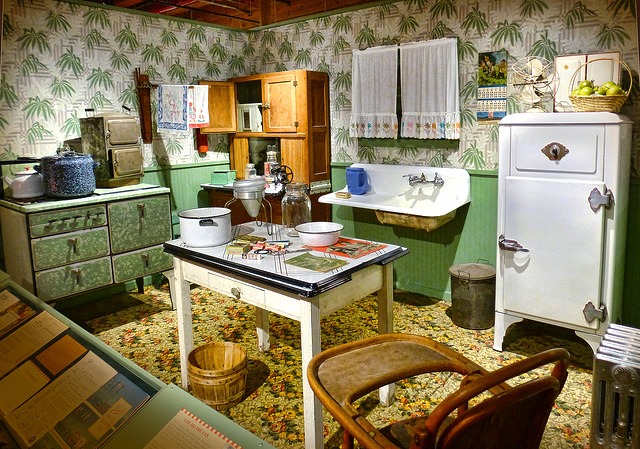}&
\includegraphics[width=\imwidth, height = 1.35cm]{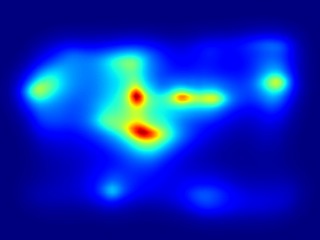}
\\
\includegraphics[width=\imwidth]{Ims/incdec_supplement/im110.jpg}\llap{{\raisebox{0.47\imwidth}{\includegraphics[cfbox=red, width=0.3\imwidth]{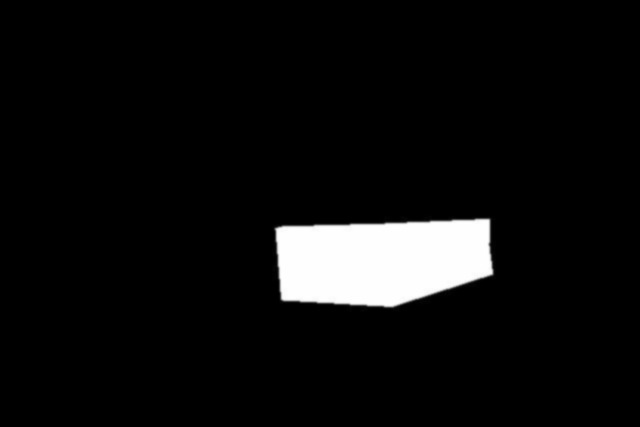}}}}&
\includegraphics[width=\imwidth, height = 1.3cm]{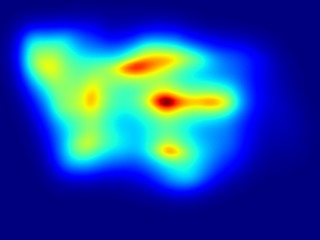} &
\includegraphics[width=\imwidth]{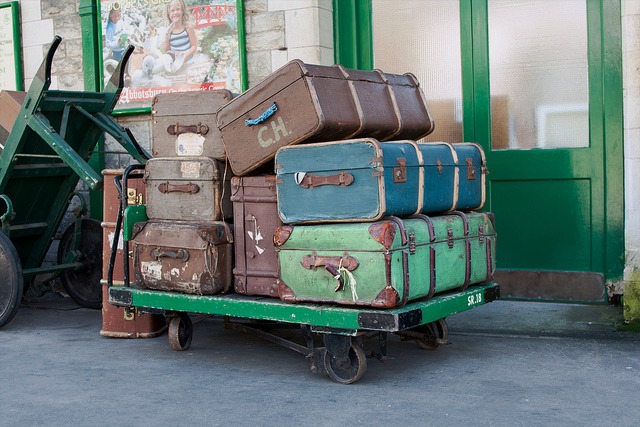} &
\includegraphics[width=\imwidth, height = 1.3cm]{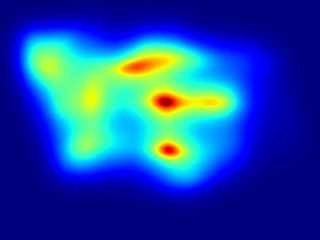} &
\includegraphics[width=\imwidth]{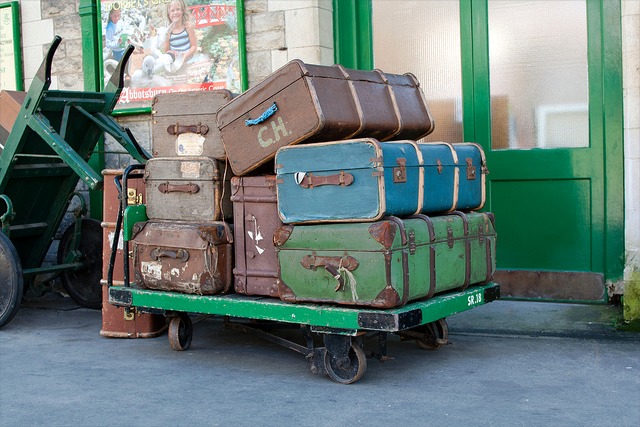}&
\includegraphics[width=\imwidth, height = 1.3cm]{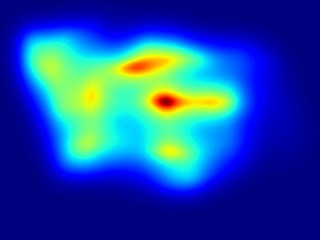}
\\
\includegraphics[width=\imwidth]{Ims/incdec_supplement/im107.jpg}\llap{{\raisebox{0\imwidth}{\includegraphics[cfbox=red, width=0.3\imwidth]{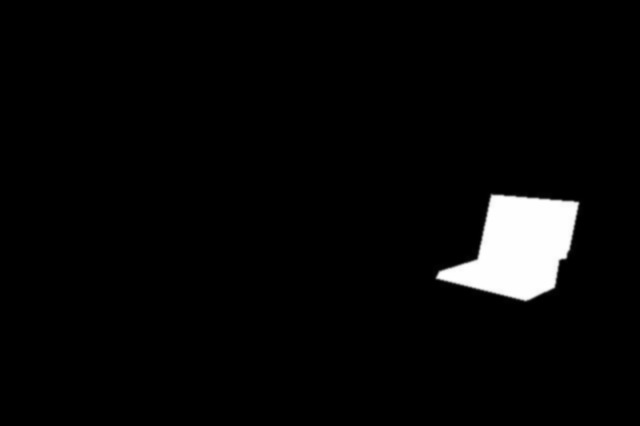}}}}&
\includegraphics[width=\imwidth, height = 1.3cm]{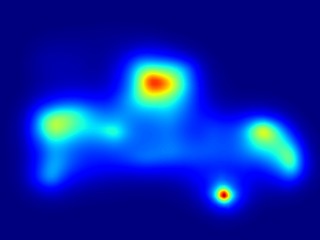} &
\includegraphics[width=\imwidth]{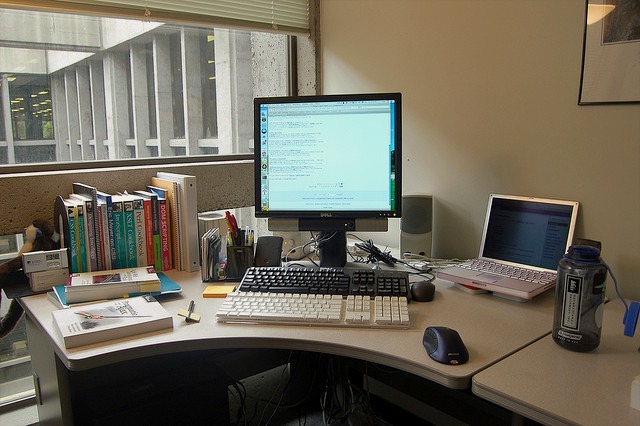} &
\includegraphics[width=\imwidth, height = 1.3cm]{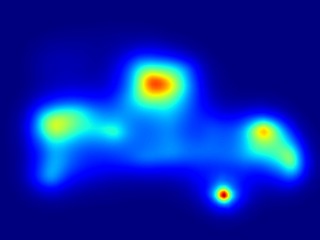} &
\includegraphics[width=\imwidth]{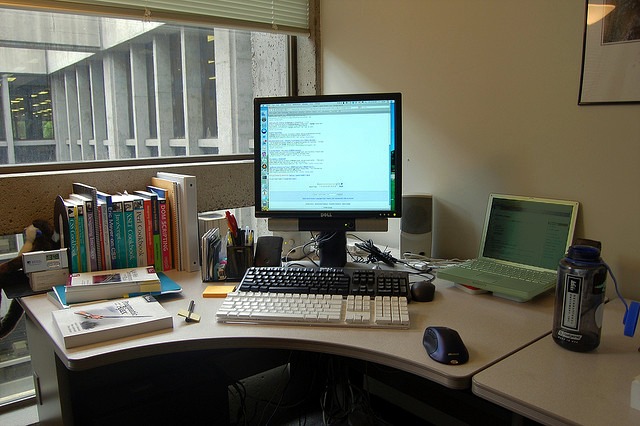}&
\includegraphics[width=\imwidth, height = 1.3cm]{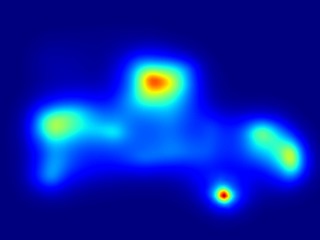}
\\
\includegraphics[width=\imwidth]{Ims/incdec_supplement/im105.jpg}\llap{{\raisebox{0.48\imwidth}{\includegraphics[cfbox=red, width=0.3\imwidth]{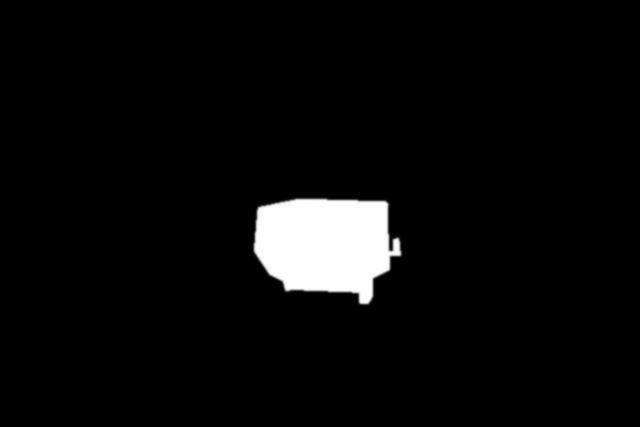}}}}&
\includegraphics[width=\imwidth, height = 1.3cm]{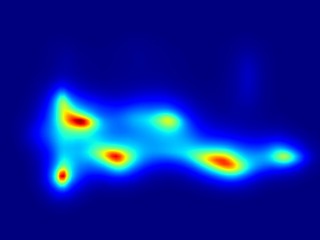} &
\includegraphics[width=\imwidth]{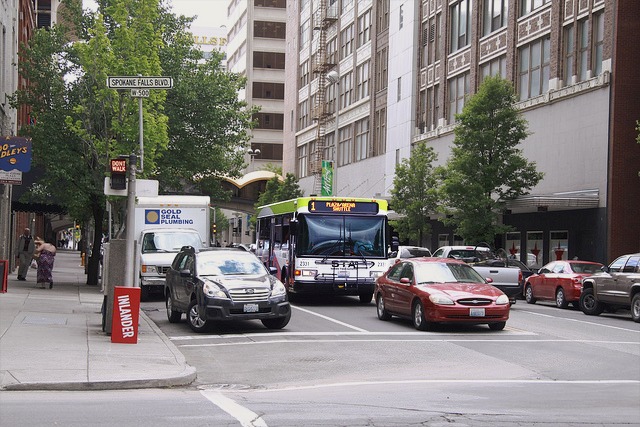} &
\includegraphics[width=\imwidth, height = 1.3cm]{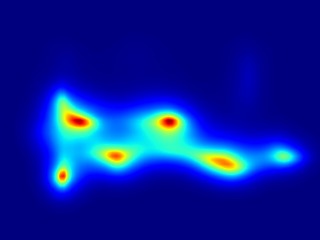} &
\includegraphics[width=\imwidth]{Ims/incdec_supplement/imfinal_inc1105_-0051.jpg}&
\includegraphics[width=\imwidth, height = 1.3cm]{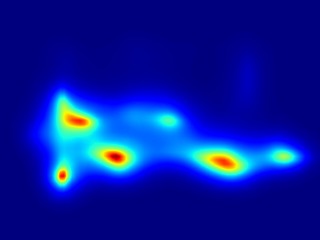}
\\
\includegraphics[width=\imwidth]{Ims/incdec_supplement/im99.jpg}\llap{{\raisebox{0\imwidth}{\includegraphics[cfbox=red, width=0.3\imwidth]{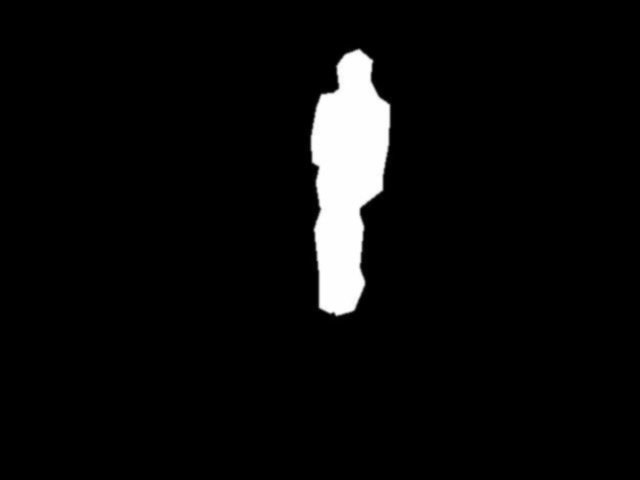}}}}&
\includegraphics[width=\imwidth]{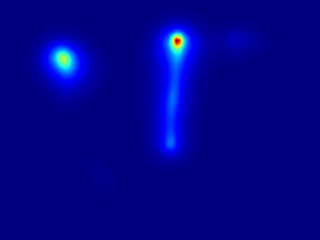} &
\includegraphics[width=\imwidth]{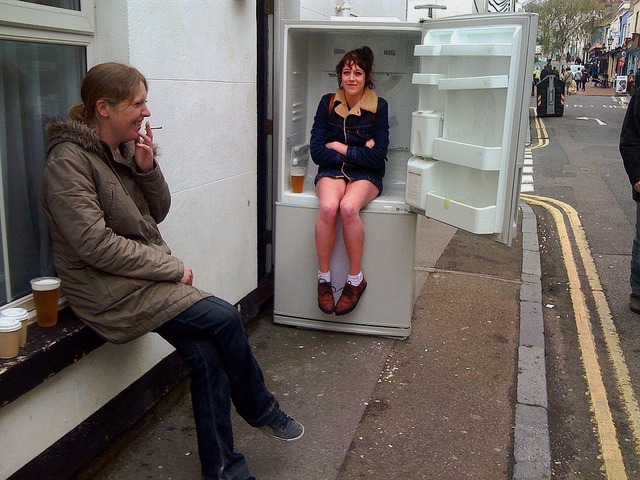} &
\includegraphics[width=\imwidth]{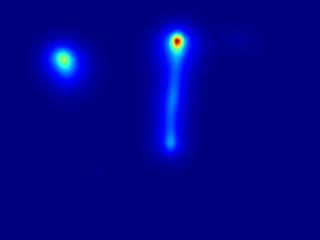} &
\includegraphics[width=\imwidth]{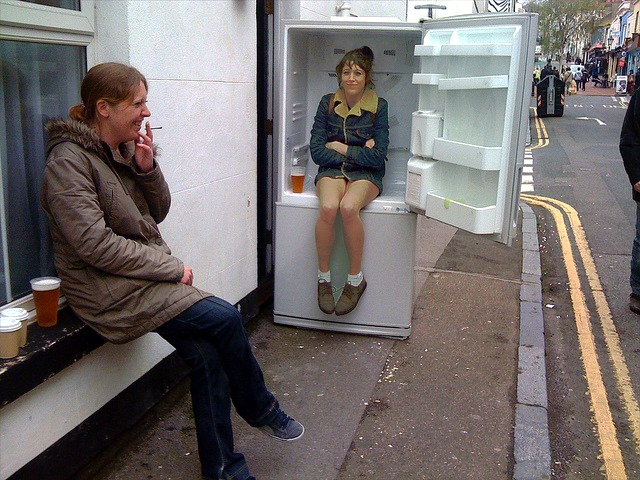}&
\includegraphics[width=\imwidth]{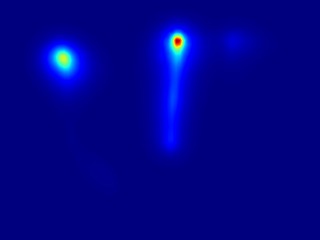}
\\
\includegraphics[width=\imwidth]{Ims/incdec_supplement/im96.jpg}\llap{{\raisebox{0.46\imwidth}{\includegraphics[cfbox=red, width=0.3\imwidth]{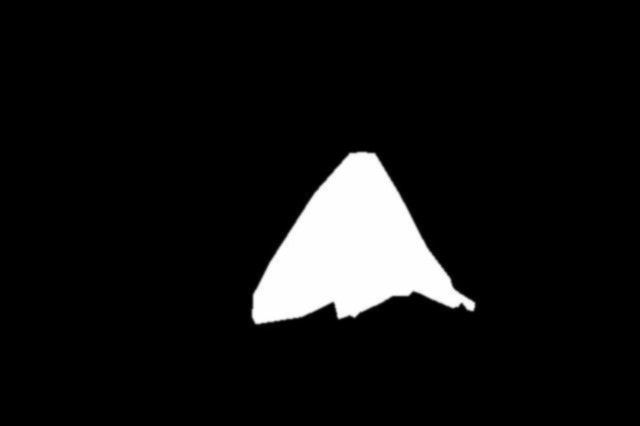}}}}&
\includegraphics[width=\imwidth, height = 1.3cm]{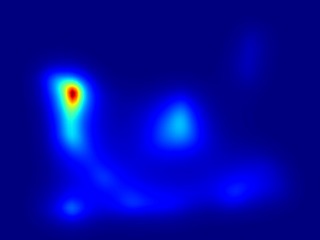} &
\includegraphics[width=\imwidth]{Ims/incdec_supplement/imfinal_inc096_0018.jpg} &
\includegraphics[width=\imwidth, height = 1.3cm]{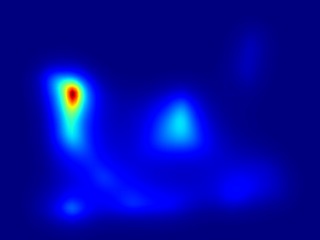} &
\includegraphics[width=\imwidth]{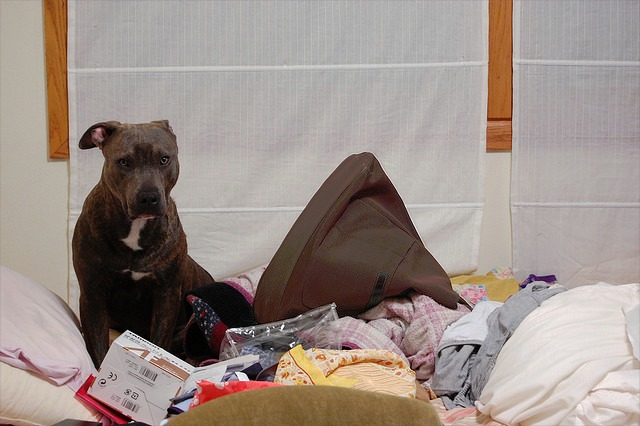}&
\includegraphics[width=\imwidth, height = 1.3cm]{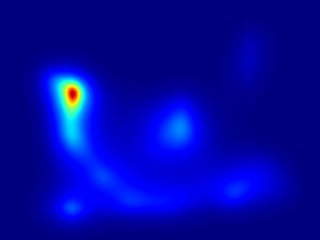}
\\
\includegraphics[width=\imwidth]{Ims/incdec_supplement/im78.jpg}\llap{{\raisebox{0\imwidth}{\includegraphics[cfbox=red, width=0.3\imwidth]{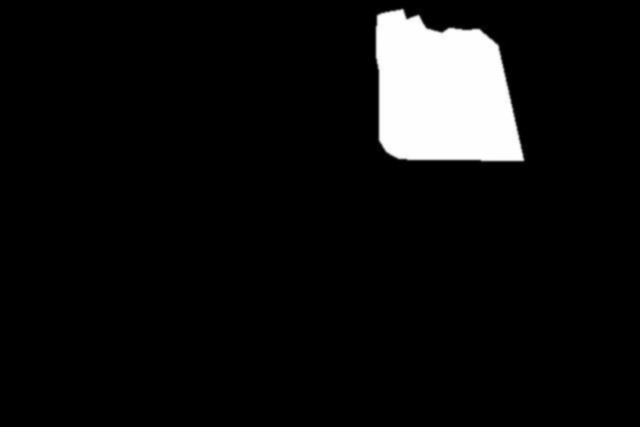}}}}&
\includegraphics[width=\imwidth, height = 1.3cm]{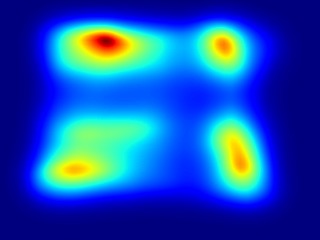} &
\includegraphics[width=\imwidth]{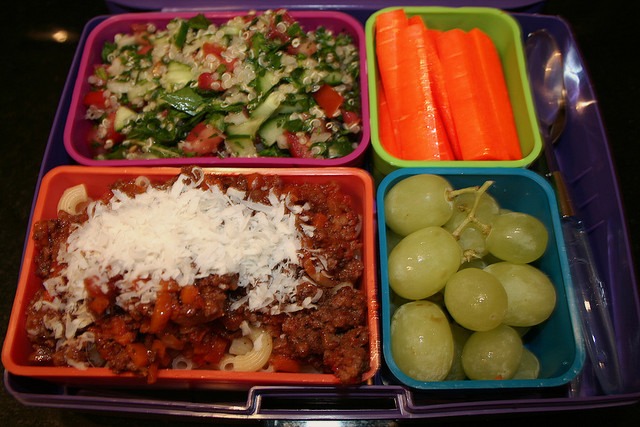} &
\includegraphics[width=\imwidth, height = 1.3cm]{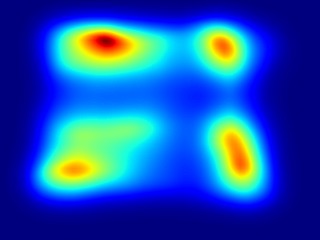} &
\includegraphics[width=\imwidth]{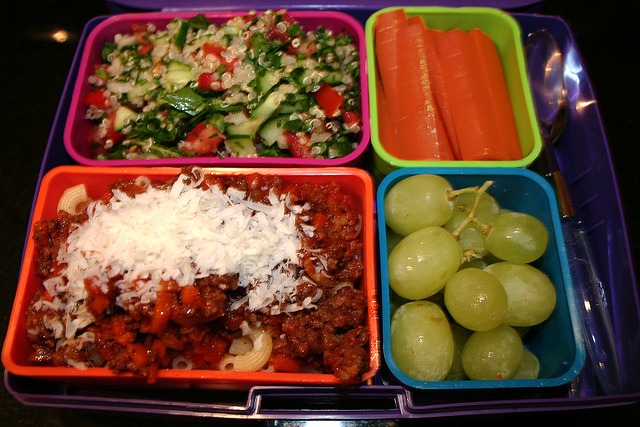}&
\includegraphics[width=\imwidth, height = 1.3cm]{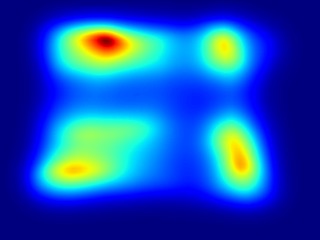}
\\
\includegraphics[width=\imwidth]{Ims/incdec_supplement/im79.jpg}\llap{{\raisebox{0\imwidth}{\includegraphics[cfbox=red, width=0.3\imwidth]{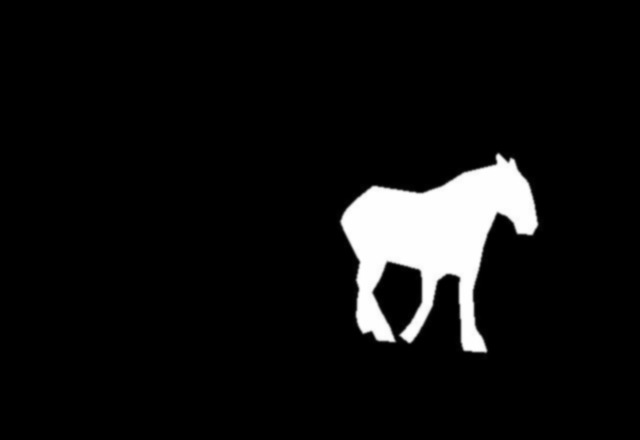}}}}&
\includegraphics[width=\imwidth, height = 1.35cm]{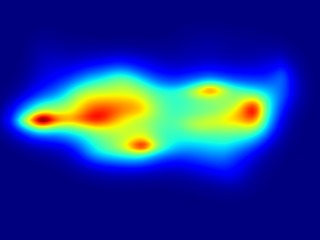} &
\includegraphics[width=\imwidth]{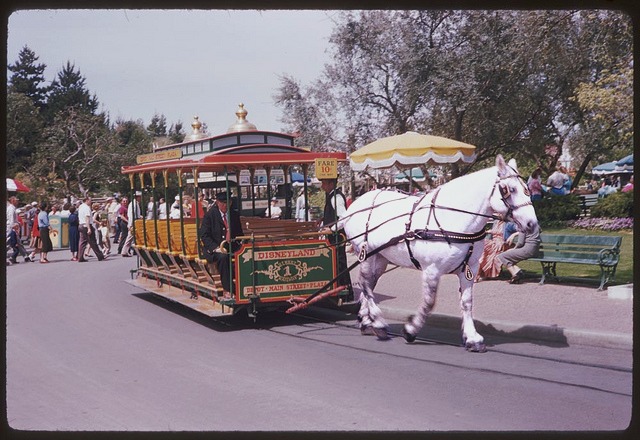} &
\includegraphics[width=\imwidth, height = 1.35cm]{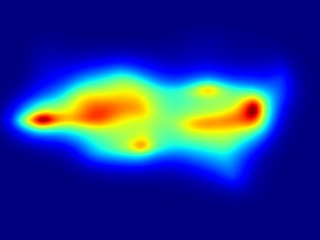} &
\includegraphics[width=\imwidth]{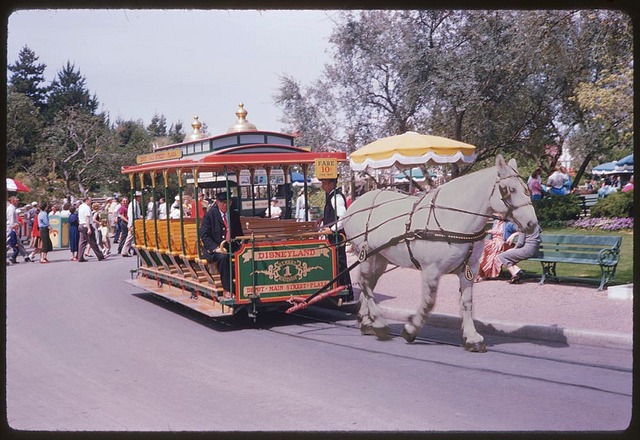}&
\includegraphics[width=\imwidth, height = 1.35cm]{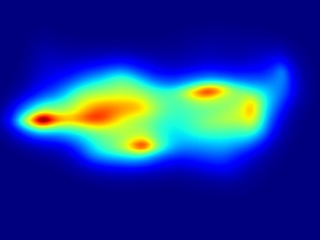}
\\
\includegraphics[width=\imwidth]{Ims/incdec_supplement/im61.jpg}\llap{{\raisebox{0\imwidth}{\includegraphics[cfbox=red, width=0.3\imwidth]{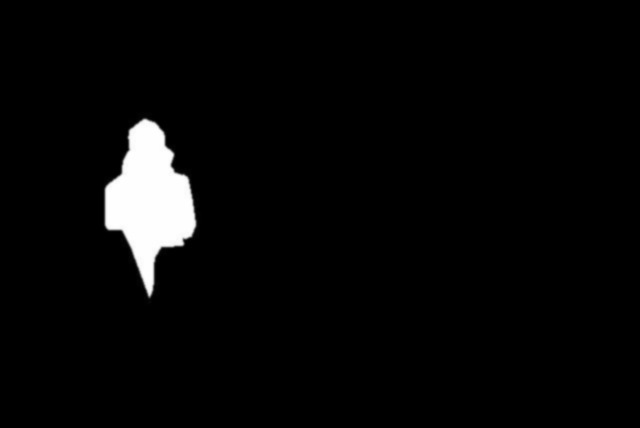}}}}&
\includegraphics[width=\imwidth, height = 1.3cm]{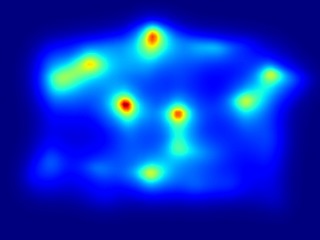} &
\includegraphics[width=\imwidth]{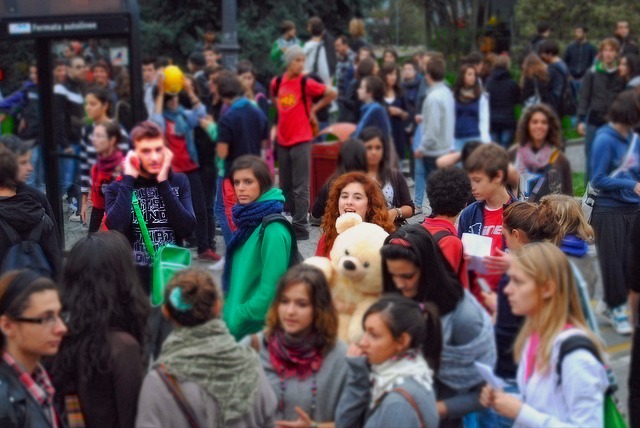} &
\includegraphics[width=\imwidth, height = 1.3cm]{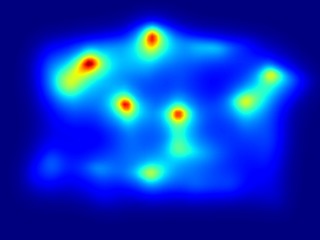} &
\includegraphics[width=\imwidth]{Ims/incdec_supplement/imfinal_inc161_-0056.jpg}&
\includegraphics[width=\imwidth, height = 1.3cm]{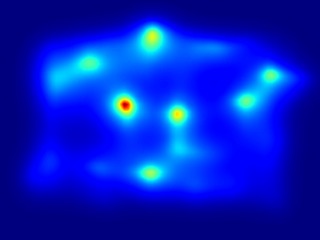}
\\
\includegraphics[width=\imwidth]{Ims/incdec_supplement/im731.jpg}\llap{{\raisebox{0.46\imwidth}{\includegraphics[cfbox=red, width=0.3\imwidth]{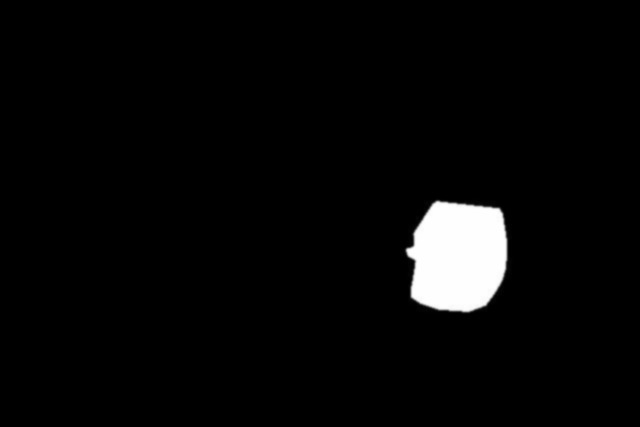}}}}&
\includegraphics[width=\imwidth, height = 1.3cm]{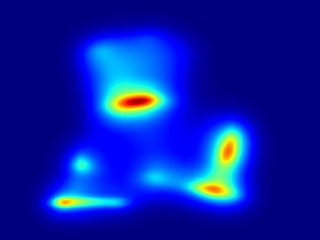} &
\includegraphics[width=\imwidth]{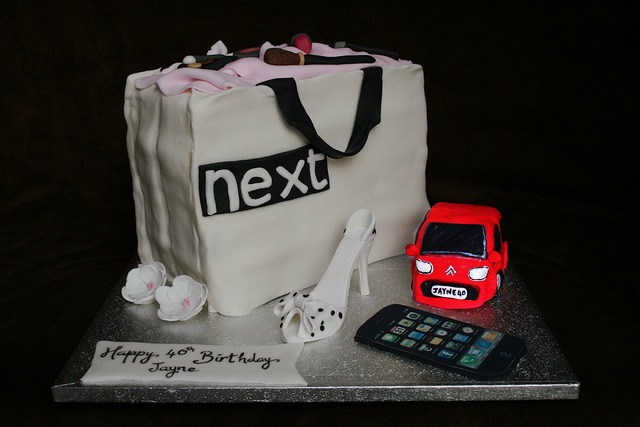} &
\includegraphics[width=\imwidth, height = 1.3cm]{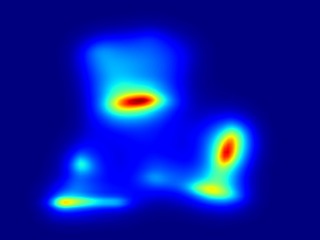} &
\includegraphics[width=\imwidth]{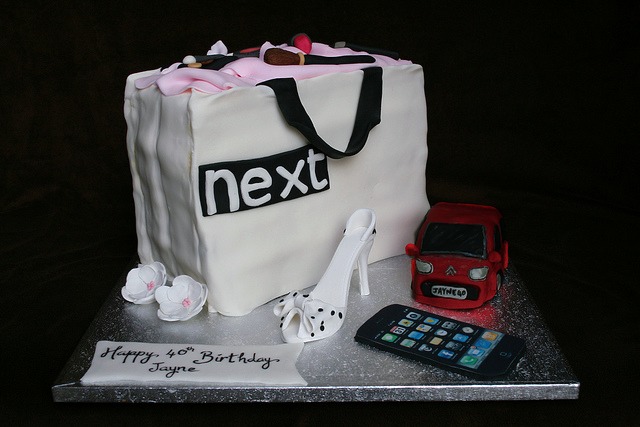}&
\includegraphics[width=\imwidth, height = 1.3cm]{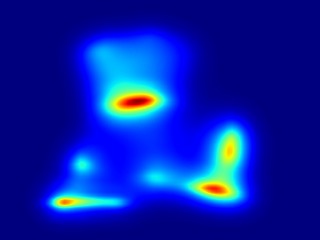}
\\
\includegraphics[width=\imwidth]{Ims/incdec_supplement/im1698.jpg}\llap{{\raisebox{0.46\imwidth}{\includegraphics[cfbox=red, width=0.3\imwidth]{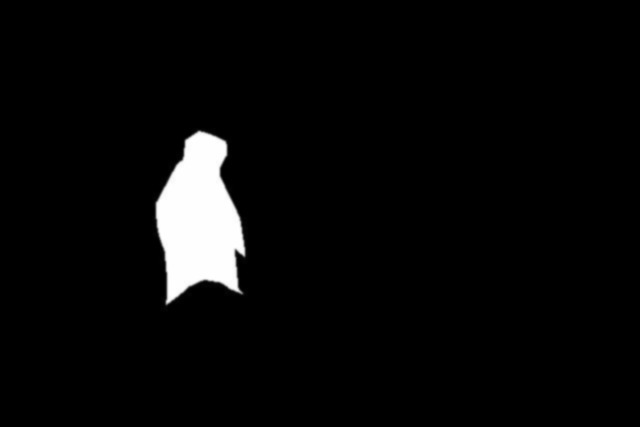}}}}&
\includegraphics[width=\imwidth, height = 1.3cm]{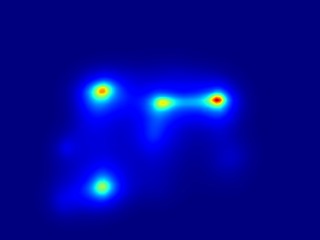} &
\includegraphics[width=\imwidth]{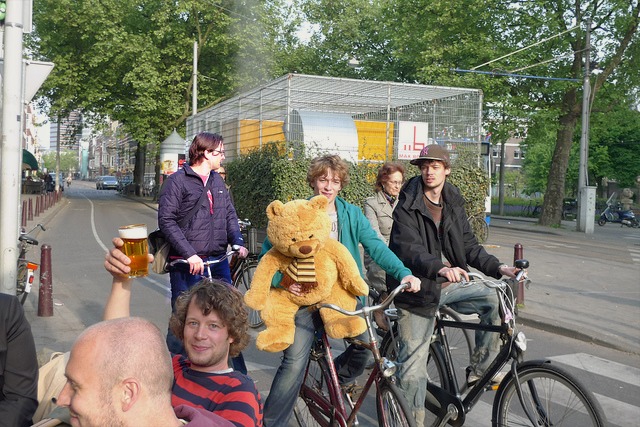} &
\includegraphics[width=\imwidth, height = 1.3cm]{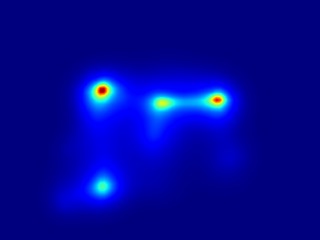} &
\includegraphics[width=\imwidth]{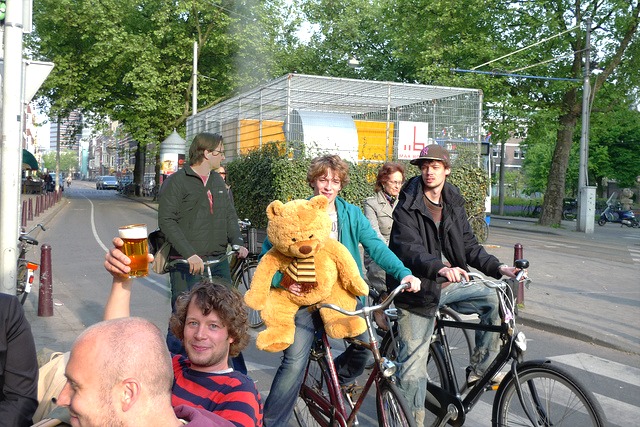}&
\includegraphics[width=\imwidth, height = 1.3cm]{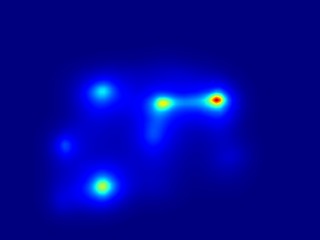}
\\
 \end{tabular}
\caption{Additional results for decreasing human attention.}
\label{fig:incdec}
\end{figure*}%

\end{document}